\PassOptionsToPackage{table,xcdraw}{xcolor}
\documentclass[fleqn,10pt]{wlscirep}

\usepackage[utf8]{inputenc}
\usepackage[T1]{fontenc}
\usepackage{multirow}
\usepackage{amsmath}
\usepackage{algorithm}
\usepackage{algpseudocode}
\usepackage{xurl}
\usepackage{makecell}
\usepackage{booktabs}
\usepackage{rotating}

\hypersetup{
    colorlinks=true,
    linkcolor=blue,
    urlcolor=blue,
    citecolor=blue,
    pdfborder={0 0 0}
}

\makeatletter
\def\url@mystyle{%
  \@ifundefined{selectfont}
    {\def\UrlFont{\sf}}
    {\def\UrlFont{\rmfamily}}}
\makeatother

\title{Generative Active Learning for Label-Efficient Pan-Organ Ultrasound Diagnosis}

\author{
Yaofei Duan$^{1,2,3,\dag}$,
Yuhao Huang$^{4,5,\dag}$,
Yuan Gao$^{2,6}$,
Ruixiang Qi$^{7}$,
Tianyu Zhang$^{2,3}$,\\
Yanlin Geng$^{8}$,
Yue Sun$^{1}$,
Sio-Kei Im$^{1}$,
Dong Ni$^{4,9,10}$,
Dong Xu$^{11,12,*}$,
Ritse Mann$^{2,3,*}$,\\
Tao Tan$^{1,*}$
}

\affil{
$^1$ Macao Polytechnic University, Faculty of Applied Sciences, Macau, China\\
$^2$ The Netherlands Cancer Institute, Department of Radiology, Amsterdam, 1066 CX, The Netherlands\\
$^3$ Radboud University Medical Center, Department of Medical Imaging, Nijmegen, 6525 GA, The Netherlands\\
$^4$ Shenzhen University, Medical Ultrasound Image Computing (MUSIC) Lab, Shenzhen, 518060, China\\
$^5$ Chinese Academy of Sciences, Hong Kong Institute of Science \& Innovation, Centre for Artificial Intelligence and Robotics, Hong Kong, China\\
$^6$ University Medical Center Utrecht, Imaging Division, Utrecht, 3584 CX, The Netherlands\\
$^7$ Westlake University, School of Medicine, Affiliated Hangzhou First People's Hospital, Department of Ultrasound, Hangzhou, 310006, China\\
$^8$ Xidian University, State Key Laboratory of ISN, Xi'an, 710071, China\\
$^9$ Shenzhen University, School of Artificial Intelligence, Shenzhen, 518060, China\\
$^{10}$ Nanjing Medical University, School of Biomedical Engineering and Informatics, Nanjing, 211166, China\\
$^{11}$ Zhejiang Cancer Hospital, Hangzhou Institute of Medicine 
(HIM), Chinese Academy of Sciences, Center of Interventional Oncology, Hangzhou, 310022, China\\
$^{12}$ Zhejiang Cancer Hospital, Hangzhou Institute of Medicine (HIM), Chinese Academy of Sciences, Department of Diagnostic Ultrasound Imaging \& Interventional Therapy, Hangzhou, 310022, China\\
$\dag$ These authors contributed equally to this work.\\
$*$ Corresponding authors: Dong Xu, Ritse Mann, and Tao Tan; contact: taotan@mpu.edu.mo.
}

\begin{abstract}
Medical ultrasound (US) enables real-time, non-invasive diagnosis, yet building US-based AI systems is limited by data scarcity, costly expert annotation, and limited cross-center generalizability. Privacy constraints and instance-level reporting hinder the collection of diverse abnormal cases and frame-level labels, while repeated retraining for new centers increases computational cost. We introduce EvoDiff, a generative active learning framework that integrates diffusion modelling with information-theoretic objectives, evolutionary search, and curriculum-based sampling to select informative samples across heterogeneous US datasets. EvoDiff formulates sample selection as a generate-to-retrieve process, in which diffusion-based optimization in feature space guided by information gain produces synthetic probes, and only the real unlabeled samples retrieved by these probes are annotated and used for downstream training. Across six benchmarks covering fifteen datasets and 16,696 images, we evaluate EvoDiff on convolutional network and transformer-based foundation model. EvoDiff approaches fully supervised performance with substantially fewer labels, outperforms active learning baselines, and improves foundation model generalization under cross-domain external validation on unseen cohorts. A reader study with six radiologists indicates the potential to support clinical annotation workflows, with a model-based estimated 24\% to 84\% reduction in annotation workload at matched downstream performance.
\end{abstract}
\begin{document}

\flushbottom
\maketitle

\thispagestyle{empty}

\section*{Introduction}
Medical ultrasound (US) is one of the most widely adopted imaging modalities in routine clinical care, valued for its non-invasive, real-time, and cost-effective nature across diverse organ systems~\cite{muhanna2022, yoshida2023diagnostic, thies2022}. 
Deep learning has recently accelerated progress in US analysis~\cite{yang2021searching,huang2022online,huang2023fourier,duan2025fetalflex}, achieving strong performance for appendicitis assessment~\cite{akbulut2025artificial}, breast~\cite{huang2025flip,10640221} and ovarian tumor~\cite{xiang2024development} characterization, liver~\cite{lu2025ai} and thyroid~\cite{su2024fully} diagnosis, and a variety of other diagnostic tasks~\cite{huang2024ai}. 
However, translating these advances into reliable clinical deployment remains challenging. 
Daily US workflows rely on instance-level reporting, face strict privacy constraints, and produce highly redundant frame sequences with pronounced class imbalance, all of which hinder the collection of diverse and well-annotated datasets~\cite{won2024sound, cohen2021behind}. 
Frame-level annotation further requires experienced radiologists, making large-scale labels slow and costly to obtain even at high-volume centers. Besides, models trained at a single institution typically require substantial retraining or fine-tuning before they can be safely deployed at new centers, which places an additional computational burden on resource-constrained hospitals. 
These bottlenecks exacerbate data scarcity, inflate expert annotation costs, and ultimately limit cross-center generalizability of US-based AI systems~\cite{wu2025towards, pai2024foundation}. 
These factors lead to pervasive data scarcity and inconsistent label quality, both of which complicate the development of robust deep learning models. 
While recent advances in CNNs, Transformers, and foundation models have improved representation learning, they continue to rely on abundant high-quality labels. 
Na\"{i}vely expanding datasets by aggregating scans from heterogeneous anatomies, devices, and acquisition protocols risks introducing label noise, incorporating uninformative samples, and degrading generalization due to negative transfer effects~\cite{lutnick2019integrated,zhang2025generalist}. This motivates label-efficient strategies capable of systematically prioritizing informative samples, reducing expert workload, and remaining robust to heterogeneous US data distributions and downstream model architectures~\cite{herath2025systematic}.

Active learning (AL) offers a principled way to address annotation bottlenecks by selectively querying the most informative samples for expert review~\cite{ash2019deep,mahapatra2024alfredo,DuaYao_ADAptation_MICCAI2025}. By prioritizing high-value data points, AL can substantially reduce the amount of labeled data needed to achieve strong diagnostic performance. 
However, its effectiveness in medical US remains limited by anatomical complexity, inter-scanner variability, and the multi-organ nature of real-world cohorts~\cite{ovarian2025multicenter,lungus2024generalizability}. 
Existing AL methods are typically developed and evaluated for a single downstream classification task and a single model family, most often CNN classifiers, and perform poorly when scaled to heterogeneous pan-organ settings or when paired with modern Transformer-based foundation models. 
Conceptually, AL strategies fall into two broad paradigms~\cite{ren2023autonomous, li2024survey}. Exploitation-oriented methods estimate sample informativeness from model prediction uncertainties and preferentially select highly uncertain examples~\cite{karamcheti2021mind}. Although intuitive, this reliance on posterior-based uncertainty can skew the training distribution, overlook sample learnability, and produce biased or unstable selections when applied to heterogeneous US cohorts. 
Exploration-oriented methods aim to maximize representativeness, for example via clustering or distance-based sampling~\cite{jin2023label,linmans2023predictive}, yet tend to over-select redundant central samples while missing clinically important outliers. 
Hybrid strategies attempt to balance uncertainty and representativeness, yet they remain constrained by posterior frameworks in which the quality of selection is only revealed after full model retraining and evaluation~\cite{yan2022clustering}. These limitations are magnified in multi-organ US, where speckle noise, variable image quality, and distinct anatomical characteristics create high-dimensional, multi-source feature spaces that challenge conventional AL heuristics~\cite{mittal2019parting,lueth2023pitfalls,coldstart2022imbalanced}.

Recent progress in generative models offers a complementary path for AL. Diffusion models have demonstrated strong capability in capturing complex medical image distributions and generating anatomically coherent samples~\cite{kazerouni2023diffusion,yang2023diffusion}, suggesting that generative priors could guide AL in high-dimensional feature spaces. In principle, AL methods could leverage such priors to design information-theoretic sampling strategies rather than relying on ad hoc posterior heuristics. However, existing generative AL approaches, for example, those based on generative adversarial networks or Bayesian generative models~\cite{zhu2017,bgal}, provide limited theoretical guidance on how to guarantee information gain (IG) for downstream classification tasks and have rarely been explored in pan-organ or foundation model-based US settings.

These observations reveal a critical gap. Current AL frameworks for medical US are predominantly task-specific, region-specific, and model-specific. They do not fully exploit modern generative model capabilities, or integrate seamlessly with foundation models that provide shared, transferable US representations across anatomies and scanners~\cite{awais2025foundation,huang2025foundation}. There is a need for a pan-organ AL framework that can operate in heterogeneous, multi-organ US cohorts, leverage generative priors to explore high-dimensional feature spaces, and prioritizing samples using principled information-theoretic criteria. Such a framework should be compatible with both conventional CNNs and Transformer-based foundation models, while offering improved interpretability by making explicit the information value of selected samples.

% ours
In this work, we introduce EvoDiff, a pan-organ AL framework that integrates \textbf{Evo}lutionary optimization with information gain (IG) guidance during \textbf{Diff}usion sampling. Instead of querying samples from a fixed posterior-based strategy, EvoDiff reformulates sample selection as a diffusion-based feature space optimization problem and establishes a novel theoretical connection between diffusion sampling trajectories and evolutionary search for AL. 
We estimate IG via entropy differentials along DDIM sampling paths, guiding generation toward informative regions without updating the parameters of the pretrained diffusion model or guidance classifier, and use evolutionary perturbations to escape local optima. 
EvoDiff further induces a natural curriculum learning strategy in which generated samples progress from certain to uncertain cases, yielding smoother learning dynamics for downstream classifiers and improved efficiency in limited-data settings.

We conducted comprehensive validation across six representative US benchmarks encompassing fifteen datasets with 16,696 images. 
To demonstrate broad applicability across architectural paradigms, we evaluated EvoDiff with two distinct downstream models, namely a conventional convolutional neural network ResNet-50~\cite{resnet} and a Transformer-based foundation model USFM \cite{jiao2024usfm}. These experiments cover six diagnostic tasks, including appendix diagnosis, breast tumor recognition and diagnosis, follow-up cardiovascular event prediction, fatty liver diagnosis, ovarian tumor assessment, and thyroid nodule diagnosis, covering eight internal cohorts and seven independent external cohorts. 
EvoDiff consistently identifies informative samples in complex, imbalanced multi-organ US feature spaces, approaches fully supervised performance while using substantially fewer labels, and outperforms state-of-the-art AL methods. Moreover, EvoDiff enhances the cross-domain generalization of the foundation model under external validation on unseen datasets. To assess EvoDiff's clinical effectiveness, we conducted a reader study with six radiologists of varying experience levels. EvoDiff yields an estimated 24\% to 84\% reduction in annotation workload relative to random sampling at matched downstream performance. 
Finally, by visualizing curriculum sampling trajectories and monitoring uncertainty reduction across AL iterations, EvoDiff offers enhanced interpretability and clearer insights into the learning process.

%-----------------------------------------------------
\section*{Results}
In the following subsections, we first describe the multi-organ, multi-task dataset and formalize the problem. 
We then introduce the IG-guided EvoDiff framework and outline its pipeline. 
We show that EvoDiff can yield consistent performance improvements on CNN-based and Foundation-based downstream models with distinct architectures. 
Subsequently, we broaden the sample selection strategy from an IG evaluation to explicitly optimizing a foundation model's generalization on external validation cohorts. 
Finally, a comprehensive reader study is conducted in a real-world clinical setting to demonstrate that EvoDiff substantially reduces the annotation time estimated by the Human Annotation Effort (HAE) metric while improving interpretability via intuitive visualizations.

\subsection*{Cross-organ Multi-task Ultrasound Dataset}
% Most existing AL methods in medical imaging focus on single-organ, single-center datasets. These settings implicitly rely on a homogeneous embedding space, where features can be effectively learned using contrastive learning approaches. However, such methods often fail in more complex, pan-organ datasets due to increased heterogeneity across organs, institutions, and imaging quality. 
Our goal is to facilitate informative data selection for pan-organ US image diagnostic tasks by integrating diffusion models with semi-supervised active learning (SSAL). 
In this study, we constructed an extensive US imaging dataset from 12 public repositories and three private cohorts, comprising 16,696 images across 6 major organs. As shown in Table \ref{tdataset}, the extensive downstream diagnostic tasks included: pediatric appendicitis diagnosis \cite{app}, breast tumor recognition and malignancy classification \cite{BUSI,BUS-BRA,UDIAT,ALN-Ultra,QAMEBI}, carotid artery US for cardiovascular event prediction \cite{CUBS}, fatty liver assessment \cite{liver}, ovarian health and pathology evaluation \cite{mmotu}, and thyroid nodule diagnosis \cite{TN3K, tn5000}. The ovary dataset (MMOTU) originally comprised seven distinct ovarian disease categories and one normal category. We manually combined these seven ovarian disease categories into a single abnormality class, thereby developing a binary classification task that distinguishes between normal and abnormal samples. 
Eight datasets were partitioned using a fixed 8:1:1 ratio and merged to construct the internal training set, internal test set, and a limited labeled subset for model initialization followed by supervised fine-tuning. The 8:1:1 split is fixed throughout the study. At each cumulative annotation budget the expanded labeled training set is subjected to five-fold cross-validation, in which four folds are used for training and one fold for validation and model selection, and the five resulting fold-specific models are evaluated on the same fixed internal test set. Strict case-level partitioning was ensured for seven of the eight public internal datasets. BUSI was the only dataset for which patient-level partitioning could not be verified, because its public release contains 780 images from 600 women without patient identifiers. For BUSI we performed image-level partitioning and additionally applied image-similarity matching to keep highly similar images within the same split wherever possible, and we note that strict patient-level separation for this dataset cannot be formally verified. The internal validation cohorts were employed to assess EvoDiff's capability in selecting informative samples that enhance the performance of two downstream architectures, e.g., ResNet-50 and USFM. Additionally, seven independent datasets were used in their entirety as external validation cohorts to rigorously evaluate the cross-domain generalization ability of the USFM model fine-tuned with EvoDiff-selected samples. Four of these are public repositories and three are private cohorts collected at Zhejiang Cancer Hospital. The three private external cohorts were assembled retrospectively under ethics approval by the Institutional Review Board of Zhejiang Cancer Hospital (No.~2026ZN004-1). 
Notably, the pan-organ US dataset introduces additional challenges arising from substantial inter-organ heterogeneity. The cohort exhibits pronounced long-tailed class imbalance and marked structural variability across anatomical regions, disease phenotypes, imaging devices, and patient populations. For example, thyroid US images constitute 34.31\% of the internal training set, whereas appendix samples account for only 4.66\%, illustrating the extreme disparity in sample availability. This heterogeneity, coupled with imbalanced data distribution, poses stringent challenges for developing unified and robust representations that generalize across organ systems.

\subsection*{Problem Formulation}
Consider a heterogeneous pan-organ US dataset spanning $K$ anatomical regions, denoted as $\mathcal{O} = \{o_k\}_{k=1}^{K}$, collected from $M$ independent medical centers. The full dataset is represented as $\mathcal{D} = \{\mathcal{D}_i\}_{i=1}^{M}$, where each $\mathcal{D}_i$ corresponds to data from center $i$.  
The distribution of imaging sources is typically uneven across centers and organs, introducing inter-center variations in acquisition protocols and scanner characteristics, where each organ category may have data collected from either multiple centers or a single center. 
Each center-specific dataset is defined as $\mathcal{D}_i = \{(x_{j}^{i}, y_{j}^{i})\}_{j=1}^{N_i}$, where $x_{j}^{i} \in \mathbb{R}^{h \times w}$ represents a US image and $y_{j}^{i}$ is the corresponding diagnostic label, which varies according to the clinical task associated with each organ. 

According to our AL formulation, the dataset from each center is partitioned in a fixed 8:1:1 ratio and subsequently integrated to construct a unified pan-organ US learning benchmark. This partitioning serves three purposes.
\begin{enumerate}[label=(\arabic*)]
    \item an \textbf{unlabeled candidate pool $\mathcal{D}_i^{\text{U}}$} (80\%), containing only organ-level metadata and used as the candidate pool from which EvoDiff selects informative samples for annotation;
    \item a \textbf{labeled initialization set $\mathcal{D}_i^{\text{L}}$} (10\%), comprising fully annotated samples with both organ and diagnostic labels for warm-start model training;
    \item a \textbf{test set $\mathcal{D}_i^{\text{T}}$} (10\%), reserved for internal evaluation of downstream performance in each AL iteration.
\end{enumerate}
The 8:1:1 partition is fixed for the whole study, and all reported performance is obtained by five-fold cross-validation within the expanded labeled training set at each cumulative annotation budget, with every fold-specific model evaluated on the same fixed test set $\mathcal{D}_i^{\text{T}}$. We implement $K=6$ anatomical regions collected from $M=8$ independent medical centers. Specifically, breast US images are contributed by three centers, while each of the remaining five organ cohorts originates from a single institution.

Unlike prior AL studies that typically target homogeneous, single-organ datasets~\cite{unisal}
% (e.g., pathology-only methods) 
or multi-modal settings where separate models are trained per dataset~\cite{ma2024adaptive}, 
% (e.g., Ma et al. \cite{ma2024adaptive} for brain MRI and histopathology), 
our task presents unique challenges to existing AL methods. 
These challenges arise from a high-dimensional, complex representation space in our pan-organ US dataset, characterized by class boundary ambiguity and feature space entanglement across different anatomical structures. 
The dataset requires AL frameworks to learn simultaneously across multiple organs and diagnostic tasks in the presence of extreme class imbalance, inter-class overlap, and subtle feature variations. This multi-domain heterogeneity introduces distribution shifts that exacerbate the representation collapse problem in limited annotation scenarios. 
CNN-based AL strategies implicitly assume homogeneous representation spaces and uniform data distributions, failing to generalize under severe cross-domain heterogeneity. Their inability to learn discriminative representations from limited labeled data becomes particularly pronounced when facing intra-class variability across anatomical domains. 
Consequently, our problem formulation necessitates a more effective sample selection strategy that can identify structurally informative examples while progressively organizing a coherent embedding space suitable for pan-organ US analysis.

%---------------------------------------------------------------------------
\subsection*{Overview of EvoDiff}
\begin{figure*}[!ht]
	\centering
	\includegraphics[width=1\linewidth]{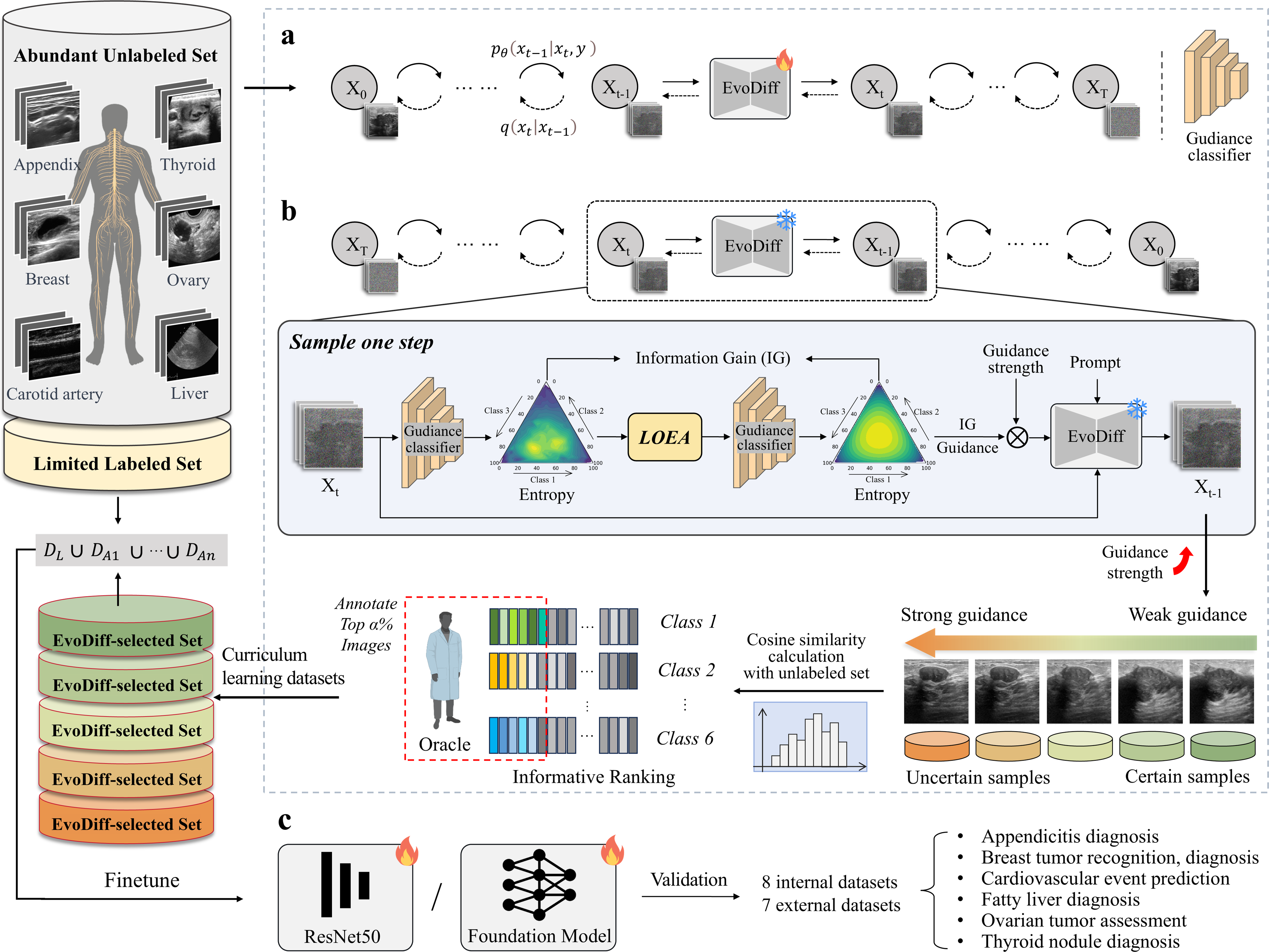}
    \caption{\textbf{Overview of the EvoDiff pipeline.} \textbf{a)} Organ-conditioned coarse pretraining of the diffusion model and the guidance classifier on multi-organ, organ-identified but pathology-unlabeled ultrasound images, followed by pathology-level fine-tuning on the limited labeled subset. A flame symbol indicates that the parameters of a network are updated in that stage, and a snowflake symbol indicates that the network is frozen. \textbf{b)} Information-gain (IG)-guided sampling combined with the Local Optimization Evolutionary Algorithm (LOEA), which performs gradient-free evolutionary refinement along the frozen diffusion trajectory and uses entropy-based IG signals to steer latents toward informative regions. The triangular maps show the predictive-entropy landscape of the guidance classifier before and after LOEA refinement, coloured from dark blue for low entropy to yellow for high entropy. As the guidance strength increases, generation moves from certain samples, drawn as green disks, to uncertain samples, drawn as orange disks, as indicated by the green-to-orange arrow. The generated probes and the unlabeled candidate pool then undergo feature extraction and cosine-similarity ranking, and the expert oracle, shown in the red dashed box, annotates the top $\alpha$\% most informative real samples. The generated probes are never annotated and are never used as downstream training samples. \textbf{c)} Newly annotated data are merged with the existing labeled data to fine-tune the downstream models, a ResNet-50 and the ultrasound foundation model (USFM), and each subsequent active learning (AL) iteration increases the uncertainty of the training set. After all AL iterations, the downstream models are validated on 15 datasets, 8 internal and 7 external, spanning 6 diagnostic tasks.}
    \label{archi}
\end{figure*}

\begin{figure*}[!ht]
	\centering
	\includegraphics[width=1\linewidth]{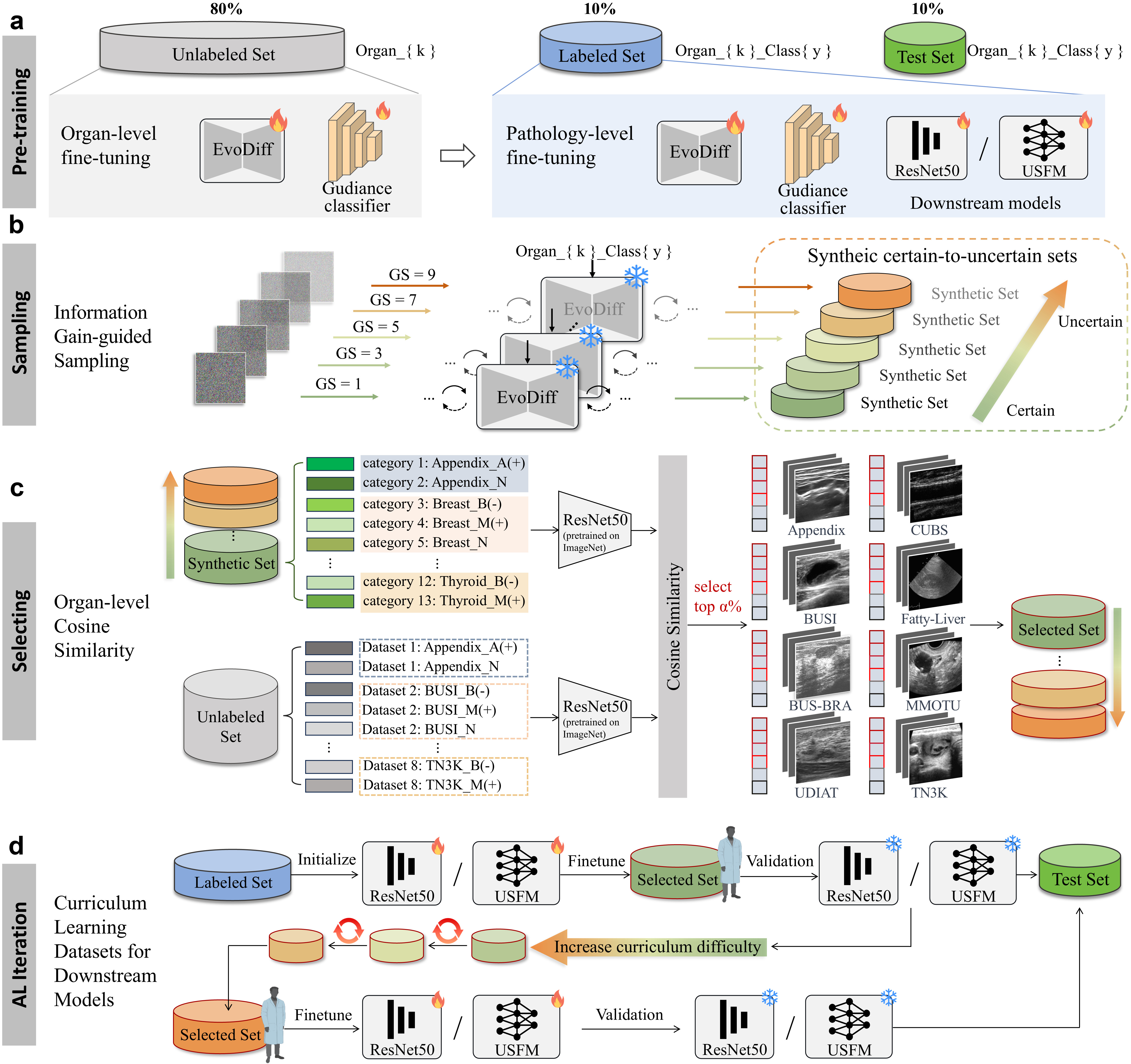}
    \caption{\textbf{Technical pipeline and implementation details of the EvoDiff framework.} \textbf{a)} Pre-training phase. Organ-conditioned coarse pretraining of the diffusion model and the guidance classifier uses the 80\% pathology-unlabeled pool, followed by pathology-level fine-tuning on the 10\% pathologically labeled subset, with the remaining 10\% reserved as the test set for final validation. A flame symbol indicates that a network is updated in that step, and a snowflake symbol indicates that it is frozen. \textbf{b)} Sampling phase. Information gain-guided sampling with the frozen diffusion model generates synthetic sample sets under guidance strengths (GS) from 1 to 9, forming the certain-to-uncertain curriculum sets depicted by the green-to-orange cylinders and arrow. \textbf{c)} Selecting phase. Features of the synthetic probes and of the unlabeled candidate pool are extracted by a ResNet-50 pretrained on natural images, cosine similarity between the two feature sets is computed within each organ-level category, and the top $\alpha$\% most similar real images, outlined in red in the ranking columns, are selected for expert annotation. \textbf{d)} Active learning (AL) iteration phase. The newly annotated samples are merged into the training set to fine-tune the downstream models, a ResNet-50 and the ultrasound foundation model (USFM), which are then validated on the frozen test set, and successive cycles progressively increase the curriculum difficulty, as shown by the green-to-orange arrow.}
    \label{alround}
\end{figure*}

To address these challenges in heterogeneous pan-organ US learning, 
we propose EvoDiff, a generative AL framework that couples diffusion modeling with information-theoretic sampling to enable label-efficient learning from multi-domain US data. As shown in Fig. \ref{archi} and Fig. \ref{alround}, a diffusion model and a guidance classifier are first jointly pretrained on abundant organ-identified but pathology-unlabeled US images and a limited labeled subset, through organ-conditioned coarse pretraining followed by pathology-level fine-tuning, to learn anatomical representations. Second, during generation, we introduce an IG-guided sampling strategy that computes entropy-based information gradients along the DDIM denoising trajectory, steering sample synthesis toward regions of maximal epistemic value while the parameters of the pretrained diffusion model and guidance classifier remain frozen. Meanwhile, Local Optimization Evolutionary Algorithm (LOEA) refines intermediate diffusion latents by maximizing entropy under category consistency constraints, yielding highly informative synthetic candidates that better explore diagnostically relevant regions of the data manifold. Finally, EvoDiff performs iterative AL by ranking generated samples via cosine similarity against the unlabeled dataset, selecting the top informative US images for expert annotation, and scaling the learning difficulty of downstream models through curriculum learning subset selection from certain to uncertain cases. The enriched labeled pool is then used to finetune downstream diagnostic models (e.g., ResNet-50 or USFM), achieving label-efficient generalization across 6 clinical tasks spanning 6 organs. Together, EvoDiff establishes a principled integration of generative modeling, information theory, and evolutionary optimization to reduce annotation cost in real-world medical imaging.

\subsection*{Comparative Performance Analysis on CNN-based Downstream Model}
% To ensure a comprehensive evaluation, random sample selection was established as the baseline for each querying round.
We benchmarked EvoDiff against 6 representative baselines: (1) FSL (Fully Supervised Learning), which uses the full labeled dataset and defines an empirical upper bound on performance;
(2) Random sampling, which uniformly selects samples without prioritization; 
(3) VAAL \cite{vaal}, a variational adversarial AL method that aligns labeled and unlabeled distributions via latent-space discrimination;
(4) Core-Set \cite{core}, which selects samples that minimize the maximum distance to the labeled set in feature space;
(5) Max-Entropy \cite{entro}, which queries samples with high predictive uncertainty; and (6) BALD \cite{bald}, which measures mutual information between predictions and model parameters under Bayesian inference. 
All methods were evaluated under identical annotation budgets of 5\%, 10\%, 20\%, 30\%, 40\%, 50\%, and 100\% of labeled data. 
For consistency, all methods employed ResNet-50 as the downstream architecture, initialized with ImageNet pre-trained weights and further fine-tuned on 10\% of the labeled dataset to ensure a fair comparison.

\begin{figure*}[!ht]
	\centering
	\includegraphics[width=1\linewidth]{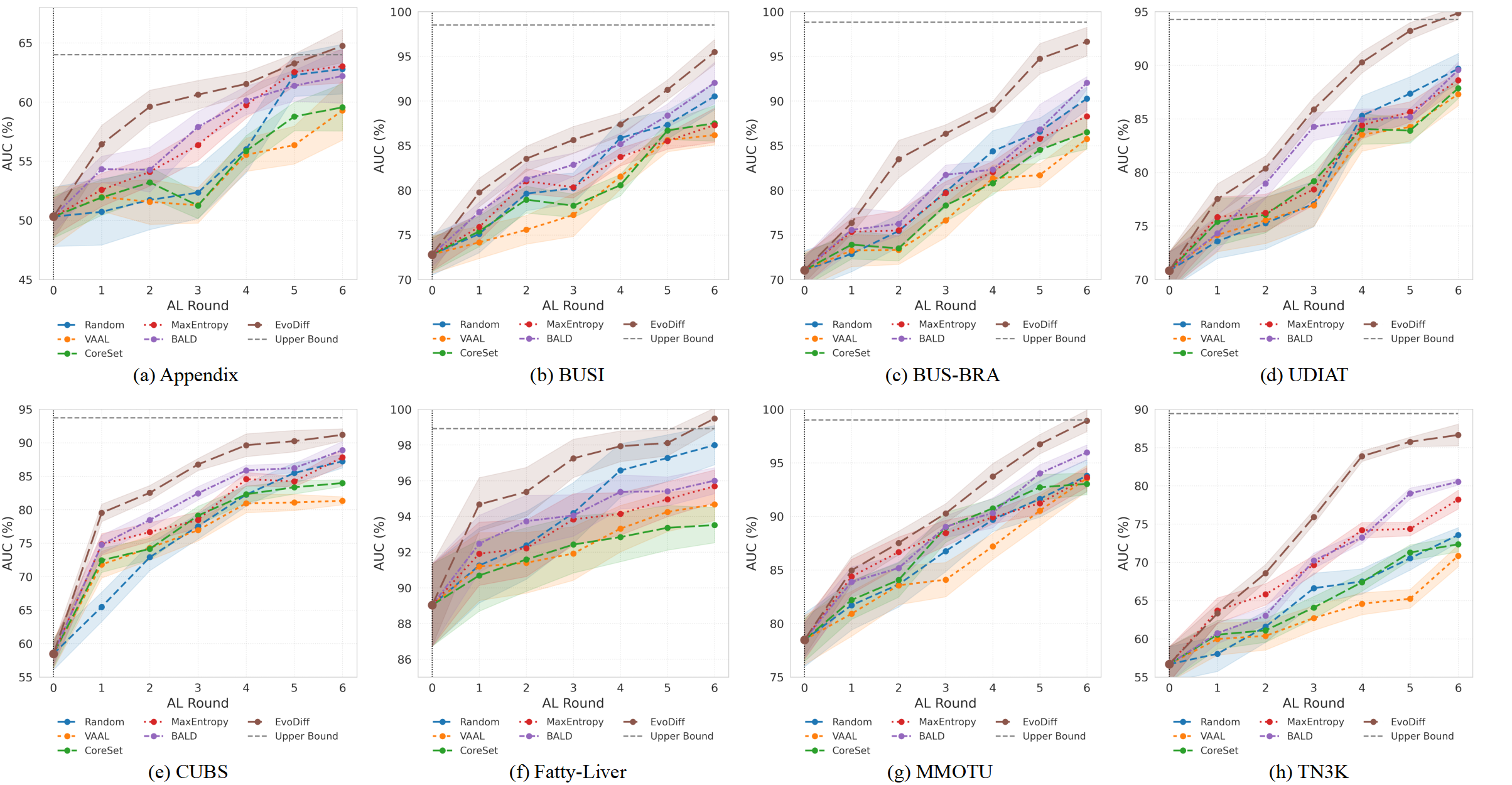}
	\caption{\textbf{Performance comparison of six active learning methods across annotation rounds on eight ultrasound image datasets.} Results are reported as the area under the receiver operating characteristic curve (AUC, \%) of the downstream ResNet-50 classifier. Round 0 represents the baseline performance of the initialized ResNet-50, and rounds 1 to 6 correspond to cumulative annotation budgets of 5\%, 10\%, 20\%, 30\%, 40\%, and 50\% of the unlabeled candidate pool. \textbf{a)} Appendicitis diagnosis on Appendix. \textbf{b)} Breast tumor diagnosis on BUSI. \textbf{c)} Breast tumor diagnosis on BUS-BRA. \textbf{d)} Breast tumor diagnosis on UDIAT. \textbf{e)} Cardiovascular event prediction on CUBS. \textbf{f)} Fatty liver diagnosis on Fatty-Liver. \textbf{g)} Ovarian tumor assessment on MMOTU. \textbf{h)} Thyroid nodule diagnosis on TN3K. In every panel, the blue dashed line denotes Random, the orange dashed line denotes VAAL, the green dashed line denotes Core-Set, the red dotted line denotes Max-Entropy, the purple dash-dotted line denotes BALD, and the brown dashed line denotes EvoDiff, each with circular markers at the evaluated rounds. The grey horizontal dashed line marks the fully supervised upper bound trained on the complete labeled set. The shaded band around each curve, drawn in the same colour as the curve, indicates the s.d. (standard deviation) across the five cross-validation folds.}
    \label{zhexian}
\end{figure*}

Experiments were conducted across eight heterogeneous US datasets spanning 6 anatomical organs: Appendix \cite{app}, BUSI \cite{BUSI}, BUS-BRA \cite{BUS-BRA}, UDIAT \cite{UDIAT}, CUBS \cite{CUBS}, Fatty-Liver \cite{liver}, MMOTU \cite{mmotu}, and TN3K \cite{TN3K}. 
As summarized in Table \ref{tdataset}, this benchmark exhibits substantial inter-organ heterogeneity and class imbalance. For instance, the Appendix dataset comprises only 4.66\% of the total images, while several organ datasets show pronounced intra-class imbalance between benign and malignant cases. This design enables a rigorous and realistic evaluation of EvoDiff under clinically relevant data distributions. 
Across the 48 evaluated dataset and budget settings, EvoDiff ranked first in accuracy in 43 and second in the remaining five, where the largest gap from the best-performing method was 0.88 percentage points (Table \ref{c1}). To further examine performance trends over iterative AL rounds, we reported the area under the ROC curve (AUC) across AL iterations in Fig. \ref{zhexian}. EvoDiff not only achieves the highest AUC across all datasets but also exhibits a consistently steeper improvement trajectory compared to uncertainty-based and representation-based baselines. Notably, EvoDiff demonstrates stable convergence behavior with lower variance across AL rounds, indicating enhanced robustness in sample acquisition.

AL algorithms face great challenge at low cumulative annotation budgets (5\%, 10\%), where the precise identification of informative samples from extreme labeling budgets becomes critical for downstream classifier optimization. 
Under the low-label conditions, EvoDiff consistently outperformed established methods across multiple datasets (Table \ref{c1} and Fig. \ref{zhexian}). 
At 5\% annotation budget, EvoDiff achieved substantial performance improvements, with the most pronounced gain observed on the Appendix dataset (13.63\%{\color{red}$\uparrow$} over random sampling). Consistent gains were also observed on other datasets, ranging from 3.42\% {\color{red}$\uparrow$} to 12.13\% {\color{red}$\uparrow$}, demonstrating EvoDiff's ability to acquire high-value samples even from scarce data. 
At 10\% annotation, EvoDiff maintained strong advantages. For example, on the BUSI dataset, EvoDiff achieved 0.8075 accuracy, significantly exceeding Core-Set (0.7599, 6.26\%{\color{red}$\uparrow$}) and BALD (0.7449, 8.40\%{\color{red}$\uparrow$}). Similarly, comparable performance advantages were observed on MMOTU, where EvoDiff reached 0.8095 accuracy, surpassing Core-Set (0.7630) and BALD (0.7630) by 4.65\% {\color{red}$\uparrow$}. These results confirm EvoDiff's superior sample selection ability in early AL stages.

Class imbalance is common in real-world medical datasets, and accuracy alone conceals how an acquisition strategy behaves under it. The CUBS dataset (Table \ref{tdataset}) exemplifies this challenge with its pronounced distributional bias, comprising 81.57\% normal versus 18.43\% abnormal samples. 
Supplementary Table~S1 reports accuracy, balanced accuracy, sensitivity, specificity, PPV, NPV, F1-score, and AUC for every method, dataset, and budget, and the CUBS entries at the 10\% cumulative annotation budget separate the two behaviours. Max-Entropy attained the second highest accuracy at 0.7462 while recording the lowest balanced accuracy at 0.5097 and a minority-class sensitivity of 0.1769, which falls below the 0.1923 obtained by random sampling. EvoDiff instead reached a balanced accuracy of 0.6165, a sensitivity of 0.4154, an F1-score of 0.3769, and an AUC of 0.8242, exceeding Max-Entropy by 10.7, 23.9, 18.7 and 5.7 percentage points respectively, with $q=0.0001$ against random sampling after Benjamini--Hochberg correction. The gain on the minority class costs 6.0 percentage points of specificity relative to random sampling, so the improvement reflects a shift of the operating point toward the underrepresented class rather than a uniform gain across both classes. 
This behaviour follows from the information-guided latent exploration of EvoDiff, which balances exploration of underrepresented classes against exploitation of decision boundaries. Uncertainty-based methods concentrate on the majority class, where predictive entropy is dense enough to dominate the ranking, and diversity-based methods may miss rare but critical cases, whereas EvoDiff selects samples by their contribution to model improvement rather than by their position in the class distribution.

A notable trend in our results reveals that at higher cumulative annotation budgets (20\%-50\%), most baseline methods exhibit diminishing returns, where each successive round of annotation yields progressively smaller improvements to downstream classifier performance. Under these conditions, EvoDiff maintained consistent performance gains. 
For example, on the TN3K dataset at 50\% cumulative annotation budget, EvoDiff achieved 0.8656 accuracy, substantially outperforming BALD (5.33\%{\color{red}$\uparrow$}) and Core-Set (7.10\%{\color{red}$\uparrow$}).
Such a trend is also reflected in the AUC progression shown in Fig. \ref{zhexian}. Unlike baseline methods that plateau after the third AL round, EvoDiff demonstrates a monotonic increase in AUC throughout all six AL rounds across all datasets, indicating that EvoDiff continues to acquire informative and discriminative samples even in later acquisition steps. Furthermore, EvoDiff exhibits smaller variance across runs, particularly in BUSI, BUS-BRA, and MMOTU, suggesting more stable learning behavior and resilience to sampling bias. The steeper upward trajectory of EvoDiff in Fig. \ref{zhexian} reinforces its advantage in sustained knowledge acquisition, while other methods tend to over-exploit boundary uncertainty or suffer from sample redundancy.

The sustained performance improvement stems from a curriculum learning strategy within the EvoDiff framework that progressively transitions from generating easily learnable samples to challenging, informative samples as the classifier becomes more robust, thereby effectively preventing the performance plateaus commonly observed in conventional AL methods. 
In contrast, traditional uncertainty-based sampling often suffers from redundant selection of similar boundary samples after initial iterations, while diversity-based methods may favor outliers with limited learning utility. EvoDiff overcomes these limitations by maintaining a dynamic sampling policy that adapts to the model's evolving knowledge state. 
Strategy-versus-Random comparisons with Benjamini--Hochberg correction are reported in Supplementary Table~S1. EvoDiff achieved a higher AUC than random sampling in all 24 evaluated dataset and budget settings, and the difference reached $q<0.05$ in 10 of them. At the 50\% cumulative annotation budget, EvoDiff comes closest to the fully supervised learning (FSL) upper bound on Fatty-Liver and Appendix, where the remaining accuracy gaps are 0.17 and 0.31 percentage points respectively. In terms of accuracy, EvoDiff remains below the FSL upper bound on all eight datasets, which shows that principled data acquisition can approach dense labeling in medical imaging pipelines at a fraction of the annotation cost.

EvoDiff obtains these gains at a higher acquisition-side computational cost than uncertainty-based selection. For a representative acquisition round on BUS-BRA at the 10\% cumulative annotation budget on a single NVIDIA A40, EvoDiff requires 6.8 GPU-hours, compared with 19.4 seconds for Max-Entropy and 86.5 seconds for BALD. Nearly all of this cost arises from information-gain-guided sampling, whereas synthetic-to-real retrieval takes approximately 34 seconds. Pretraining of the diffusion model and guidance classifier is a one-time cost shared across organs, datasets, and acquisition rounds. The acquisition step runs offline and requires no radiologist time. The stage-wise costs of all compared methods are reported in Supplementary Table~S10.

\subsection*{Validation Performance on Foundation-based Downstream Model}
To evaluate the generalizability and scalability of EvoDiff beyond CNN architectures, we conducted comprehensive validation using USFM, a state-of-the-art Transformer-based US foundation model pre-trained on 2,187,915 US images of 12 common organs. This evaluation addresses a critical gap in current AL research, whether sample selection strategies optimized for traditional architectures can effectively transfer to large-scale pre-trained models with different inductive biases.

\begin{figure*}[!ht]
	\centering
	\includegraphics[width=1\linewidth]{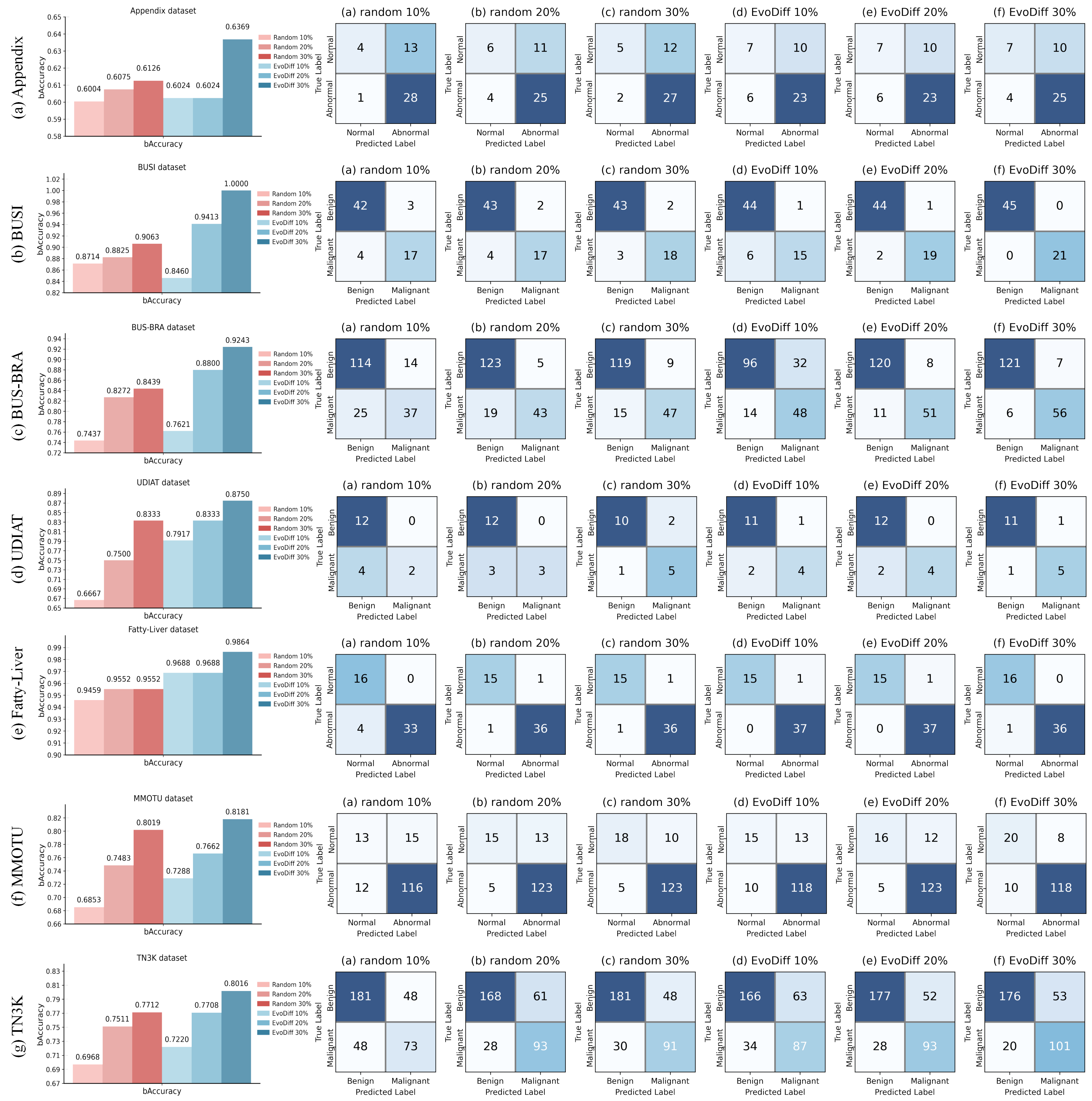}
	\caption{\textbf{Validation of USFM performance comparing random sampling with EvoDiff IG-guided curriculum learning across multiple ultrasound datasets.} Each row corresponds to one internal dataset. \textbf{a)} Appendix. \textbf{b)} BUSI. \textbf{c)} BUS-BRA. \textbf{d)} UDIAT. \textbf{e)} Fatty-Liver. \textbf{f)} MMOTU. \textbf{g)} TN3K. Within each row, the leftmost bar chart reports the balanced accuracy (bAccuracy) of the ultrasound foundation model (USFM) fine-tuned on samples selected by random sampling at cumulative annotation budgets of 10\%, 20\%, and 30\%, drawn as light-to-dark red bars, and by EvoDiff at the same three budgets, drawn as light-to-dark blue bars, with the value printed above each bar. The six confusion matrices to the right of the bar chart, titled from random 10\% to EvoDiff 30\%, report the corresponding predictions on the fixed internal test set, with rows giving the true label and columns the predicted label for the two task-specific classes, normal versus abnormal or benign versus malignant. Matrix cells are shaded in blue, with darker shading indicating a larger number of samples, and the sample count is printed in each cell. IG, information gain.}
    \label{hunxiao}
\end{figure*}

As demonstrated in Fig. \ref{hunxiao} and Supplementary Table~S2, which reports accuracy, balanced accuracy, sensitivity, specificity, PPV, NPV, F1-score, and AUC for every method, dataset, and budget under USFM, EvoDiff consistently outperforms random sampling across most datasets under three cumulative annotation budgets (10\%, 20\%, 30\%), establishing its effectiveness in fine-tuning foundation models under label-scarce conditions. 
The performance gains are particularly pronounced in balanced accuracy metric (bAcc), with the most substantial improvements observed in breast imaging tasks (Fig.~\ref{hunxiao} (b)-(d)), where USFM possesses strong prior domain knowledge derived from its pre-training corpus comprising > 90\% breast US images. 
Under a 30\% annotation budget, EvoDiff improves bAcc by 9.37\%{\color{red}$\uparrow$} on BUSI and 8.04\% {\color{red}$\uparrow$} on BUS-BRA compared to random sampling, indicating that EvoDiff sample selection effectively guides foundation models toward diagnostically meaningful training signals early in fine-tuning. 
Confusion matrices in Fig.~\ref{hunxiao} show that EvoDiff shifts the operating point toward higher sensitivity on several datasets. Pooling the eight datasets and five folds into 40 paired observations at each budget, the mean sensitivity difference between EvoDiff and random sampling was $+15.04$, $+11.17$, and $-0.13$ percentage points at the 10\%, 20\%, and 30\% budgets for ResNet-50, and $+6.33$, $+8.41$, and $+2.29$ percentage points for USFM. 
For example, on BUS-BRA at 10\% cumulative annotation budget (Fig.~\ref{hunxiao} (c)), the number of missed malignancies decreased from 25 cases with random sampling to 14 cases with EvoDiff, corresponding to a 44\% reduction. 
A similar pattern is observed on TN3K, where sensitivity increases with the annotation budget. 
Because the sensitivity gain is not uniform across budgets and is not accompanied by a matching specificity gain, we report these differences as performance observations under the evaluated settings and do not interpret them as evidence of clinical safety. 
The method's robustness extends to challenging imbalanced datasets, where conventional sampling strategies typically struggle with minority class recognition. In UDIAT and MMOTU, EvoDiff improves minority class recognition without sacrificing specificity. Notably, in UDIAT at 30\% sampling, EvoDiff achieves perfect specificity while increasing sensitivity, showing that the curriculum learning strategy mitigates class imbalance bias during active selection.

These findings suggest a synergistic interaction between EvoDiff and USFM. First, the EvoDiff methodology's sample selection through diffusion sampling demonstrates exceptional precision in identifying samples with maximal IG. While the USFM has been pre-trained on an extensive dataset, it provides strong anatomical priors that accelerate convergence. Together, they form a label-efficient training paradigm that generalizes across organs and imaging distributions. 
This intelligent sample curation process allows the foundation model to efficiently leverage its pre-learned representations while focusing computational resources on the most informative examples, resulting in rapid adaptation capabilities that facilitate seamless transfer to diagnostic tasks across different medical centers with varying imaging protocols and patient populations. 
Overall, EvoDiff delivers consistent and clinically meaningful performance gains over random sampling, reinforcing its value as a practical and scalable AL strategy for real-world foundation model deployment in medical US analysis.

\subsection*{Informative Sample Selection for Improved Foundation Model Generalizability}
\begin{figure*}[!ht]
	\centering
	\includegraphics[width=1\linewidth]{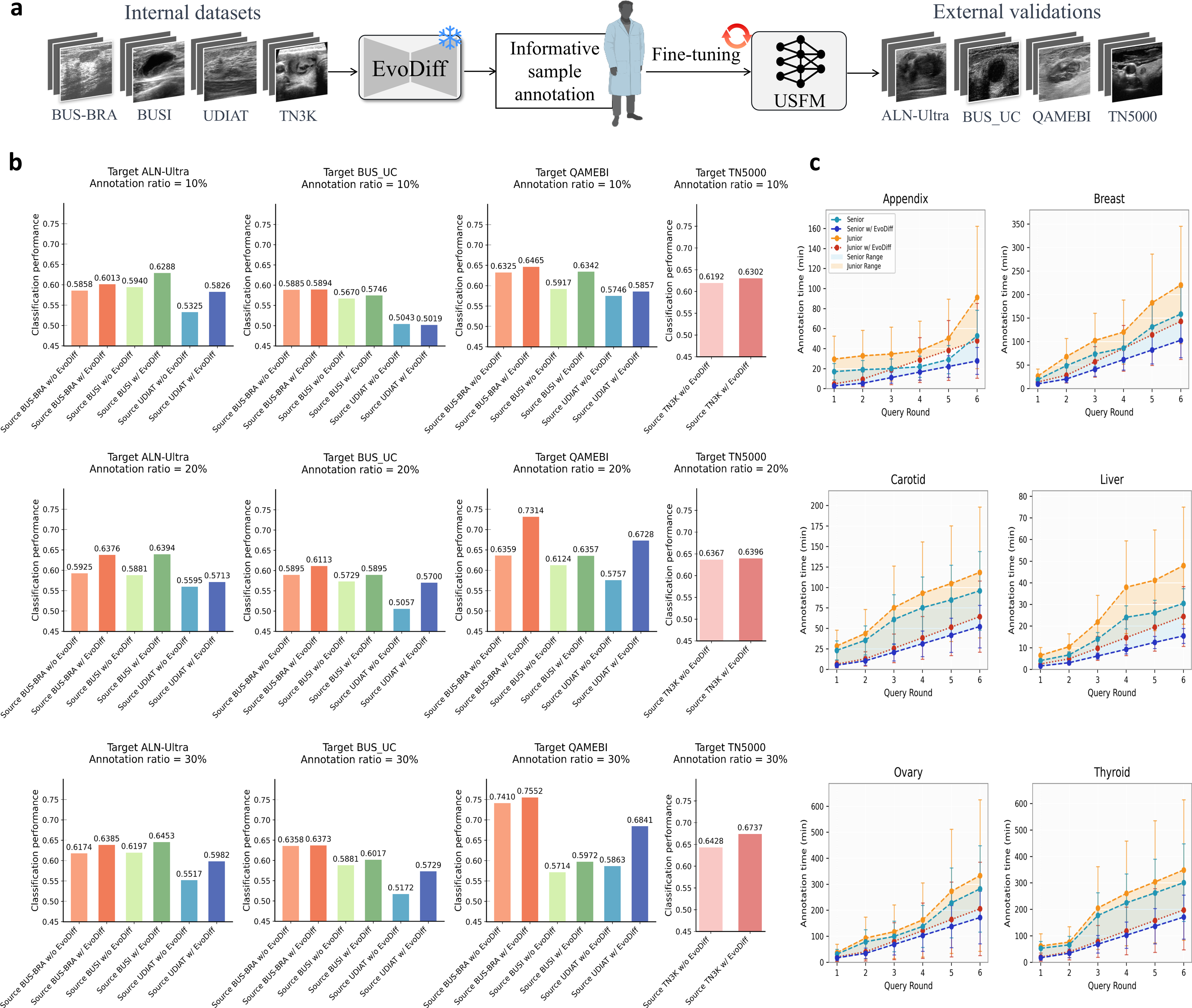}
	\caption{\textbf{Foundation model performance in cross-domain validation and annotation efficiency analysis.} \textbf{a)} External validation pipeline. The frozen EvoDiff model, marked by a snowflake symbol, selects informative samples from the internal source datasets (BUS-BRA, BUSI, UDIAT, and TN3K) for expert annotation, the ultrasound foundation model (USFM) is fine-tuned on the annotated samples, and the fine-tuned model is evaluated on four previously unseen external datasets (ALN-Ultra, BUS\_UC, QAMEBI, and TN5000). \textbf{b)} Balanced accuracy of USFM on each external target dataset at annotation ratios of 10\%, 20\%, and 30\%, arranged from the top row to the bottom row. Bars are grouped in pairs by source dataset, where the lighter bar of each pair gives the result without EvoDiff, that is with randomly sampled annotation, and the darker bar gives the result with EvoDiff-selected annotation. Orange bars denote BUS-BRA as the source dataset, green bars denote BUSI, blue bars denote UDIAT, and red bars denote TN3K, with the value printed above each bar. \textbf{c)} Model-based estimates of annotation cost across query rounds for the six organ tasks, computed using the mean unit annotation time within each experience stratum of the six-reader panel. The teal dashed line with circular markers shows the estimated annotation time for random sampling in the senior stratum, the dark blue dashed line shows the estimate for EvoDiff selection in the senior stratum, the orange dashed line shows the estimate for random sampling in the junior stratum, and the red dotted line shows the estimate for EvoDiff selection in the junior stratum. Error bars indicate the between-reader s.d. (standard deviation) within each stratum. The pale blue band and the pale orange band between the paired curves mark the estimated reduction in annotation workload with EvoDiff for the senior and junior strata, respectively.}
    \label{reader}
\end{figure*}

Recent studies suggest that foundation models can learn transferable representations through large-scale pretraining~\cite{moor2023foundation,zhang2024generalist,lu2024visual}. 
However, their generalizability to new clinical environments remains constrained by annotation quality and sample informativeness rather than dataset size alone. 
To investigate whether informative sample selection can unlock latent transferable capacity, we conducted a cross-domain external validation of USFM using the training data selected by EvoDiff and a randomly sampled baseline.
As shown in Fig. \ref{reader} and Supplementary Table~S3, EvoDiff improves cross-domain performance in 42 of the 45 comparisons spanning fifteen source-to-target transfer settings and three annotation budgets, without any fine-tuning.
The overall \textit{generalization gains} demonstrate EvoDiff's strong ability to support the rapid migration of the foundation model (USFM) to new domains.

Specifically, across the public external breast and thyroid targets, USFMs fine-tuned with EvoDiff-selected samples outperformed their randomly-trained counterparts in 29 of the 30 comparisons. The relative gains in bAcc differ by source and target rather than forming a single range. For the three source-to-BUS\_UC breast evaluations they span 2.9\% to 12.7\%, for the source-to-ALN-Ultra and source-to-QAMEBI breast evaluations they reach 16.9\%, and for TN3K$\rightarrow$TN5000 the gain is 4.8\%. 
At a 10\% cumulative annotation budget, EvoDiff can improve transfer performance in most cases, indicating its effective generalization ability in the early stage. 
These gains typically amplify at the 20\% and 30\% annotation budgets, reflecting EvoDiff's curriculum strategy of gradually exploring more informative samples over AL iterations. 

Cross-domain transfer validation on the external breast targets showed the largest relative improvement for UDIAT$\rightarrow$QAMEBI at 16.9\%. 
Notably, EvoDiff also shows cross-domain generalization ability in the thyroid task (i.e., TN3K$\rightarrow$TN5000), improving the bAcc from 0.6428 to 0.6737 at 30\% annotation budget (4.8\% {\color{red}$\uparrow$}). 
This further demonstrates EvoDiff's applicability across different anatomies. 
The three private cohorts collected at Zhejiang Cancer Hospital were not accessible to the USFM pretraining corpus and therefore provide an overlap-free test of cross-domain transfer. On these cohorts EvoDiff improved balanced accuracy over random sampling in 13 of the 15 comparisons, with relative gains reaching 25.0\% on PrivAppendix, 18.4\% on PrivBreast and 18.3\% on PrivLiver. Both exceptions arose in the UDIAT to PrivBreast setting, where the two strategies differed by 0.02 percentage points at the 10\% annotation budget and random sampling was higher at the 30\% budget. Because these cohorts were never exposed to the foundation model during pretraining, the observed generalization gains cannot be attributed to prior familiarity with the evaluation data.

The results above highlight two key conclusions. 
First, random sampling fails to unlock the inherent transferability of USFM because it cannot ensure the selection of high-quality and informative samples. 
Second, EvoDiff can act as a quality-controlled supervision mechanism, selecting domain-bridging and information-dense samples that promote cross-domain representation alignment. 
Consequently, EvoDiff addresses the critical bottleneck of sample informativeness and annotation efficiency, enabling foundation models to generalize without massive data scaling or costly target-domain fine-tuning. 
These findings further confirm that generalization is driven more by supervision quality than data quantity, positioning EvoDiff as a scalable and annotation-efficient solution for deployment across heterogeneous clinical sites.

\subsection*{Reader Study of Annotation Cost Reduction Across Multi-Organ Ultrasound Datasets}
To evaluate EvoDiff's clinical utility in reducing radiologist annotation burden, we conducted a reader study to estimate the Human Annotation Effort (HAE) and annotation workload savings ($\triangle$HAE) of EvoDiff-selected annotation relative to random sampling at matched downstream performance. Both HAE and $\triangle$HAE are model-based estimates that combine annotation counts at matched downstream performance with unit annotation times measured in the reader study, and the same unit annotation time is applied to both strategies. 
The study used randomized unlabeled subsets from our collected multi-organ US dataset and involved six board-certified radiologists spanning two experience strata, namely three senior readers with 7, 8, and 13 years of clinical experience and three junior readers with 2, 3, and 3 years of experience. Each reader independently reviewed US images and provided clinical diagnoses for the assigned cases within a fixed reading session. We recorded the number of correctly diagnosed samples and calculated the unit annotation cost for every reader, establishing baseline annotation performance at both experience levels.

Results revealed substantial variability in both accuracy and efficiency across readers and organ tasks, and the per-reader values are reported in Supplementary Table~S6. Averaged within each experience stratum, senior readers reached 70.0\% accuracy and junior readers 67.1\% across the six tasks. Both strata performed best on liver, thyroid, breast, and carotid, with stratum means between 74.7\% and 86.5\%, and worst on ovary, where the stratum means were 36.0\% for senior and 38.4\% for junior readers. 
These findings expose critical realities in clinical annotation workflows. First, annotation quality depends more on the difficulty of the organ task than on reader experience, with both strata falling below 40\% accuracy on ovary while exceeding 74\% on liver, thyroid, breast, and carotid. Second, error-induced waste represents a hidden cost burden rarely acknowledged in deep learning literature.

Using Table \ref{c1} as performance benchmarks, we evaluated EvoDiff's HAE effectiveness across multiple cumulative annotation budgets (5\%, 10\%, 20\%, 30\%, 40\%, and 50\%). For each cumulative annotation budget, we calculated the HAE required for random sampling to achieve equivalent performance to EvoDiff-optimized selection. This analysis yields model-based estimates of the total annotation workload and the corresponding workload savings ($\triangle$HAE) for achieving comparable diagnostic accuracy with random sampling and with EvoDiff selection. 
We plotted the estimated annotation costs separately for the senior and junior strata. Fig. \ref{reader}.c shows the estimated HAE across AL rounds for random sampling and for EvoDiff selection in each stratum, in which the shaded regions represent the estimated workload saving ($\triangle$HAE) for each stratum. And Supplementary Table~S4 shows the corresponding numerical results. 
The estimated annotation workload is lower with EvoDiff than with random sampling in both experience strata, in all six organ tasks and at every cumulative annotation budget. Under the common unit-time assumption, the estimated relative reduction is identical in the senior and junior strata. The estimated absolute time saving is larger in the junior stratum, reflecting its longer measured time per correctly annotated image in this panel. 
At the 5\% cumulative annotation budget, the estimated workload reduction on the carotid artery task reaches 77\% for both the junior and the senior stratum. 
Under the 10\% cumulative annotation budget, the estimated workload for breast tumor diagnosis is 58\% lower with EvoDiff for both senior and junior readers. 
Moreover, the estimated workload reduction is present at every cumulative annotation budget in both strata, including later AL stages when numerous diverse samples had been labeled. 
These estimates indicate that EvoDiff's sample selection strategy reduces the annotation workload required to reach a given diagnostic accuracy. The consistent pattern of estimated HAE reduction across all six anatomical regions supports the applicability of EvoDiff's label efficiency across diverse US applications.

\subsection*{Ablation Study of EvoDiff Framework}
To assess the contribution of each proposed component in the EvoDiff framework, we conduct ablation experiments on two representative datasets, i.e., \textit{Appendix} with the fewest training samples, and \textit{TN3K} contains the largest. 
Therefore, the Appendix dataset reflects a highly data-scarce scenario, while TN3K provides a data-rich environment. 
This specific design allows us to rigorously evaluate both the effectiveness and scalability of EvoDiff under varying annotation budgets. 
As shown in Table \ref{ab1} and Table \ref{ab2}, the framework is progressively assembled from three key components, including IG Guidance, Local Optimization Evolutionary Algorithm (LOEA), and Curriculum Learning (CL). 
All experiments are conducted under 10\%, 30\%, and 50\% cumulative annotation budgets, using Accuracy (ACC), F1-score, and AUC as evaluation metrics.

On the Appendix dataset (see Table \ref{ab1}), the baseline configuration, which includes only IG Guidance, achieved modest performance with F1-score of 0.4938, 0.5265, and 0.5294 at 10\%, 30\%, and 50\% cumulative annotation budgets, respectively. 
Incorporating LOEA can yield minimal improvement at lower cumulative annotation budgets (i.e., 0.4938 v.s. 0.4956 at 10\% annotation), but showed better performance at higher cumulative annotation budgets (i.e., 0.5294 v.s. 0.5319 at 50\% annotation). 
The incorporation of CL produced notable performance improvements across all metrics. The configuration without CL differed significantly from the complete EvoDiff framework ($p<0.01$, two-sided paired $t$-test with the five cross-validation folds used as matched pairs). This validates that progressive sampling strategies effectively stabilize learning dynamics in low-data conditions. 
The full EvoDiff model (the last row in Table \ref{ab1}) consistently achieves the highest performance (e.g., 0.5652 ACC and 0.5960 AUC at 10\% annotation), with improvements over the IG-only baseline reaching statistical significance ($p<0.01$, two-sided paired $t$-test with the five cross-validation folds used as matched pairs), demonstrating its robust capability under limited supervision. 
Similar improvement pattern is observed on the TN3K dataset (see Table \ref{ab2}), where LOEA alone offers incremental gains, while CL significantly enhances performance. 
With 50\% annotations, EvoDiff reaches satisfactory ACC (0.8656), F1-score (0.7849), and AUC (0.8519), substantially closing the gap toward the fully supervised performance. 
These findings demonstrate that EvoDiff maintains consistent advantages in both label-scarce and data-rich settings.

\subsection*{Information Gain-Guided Synthesis Results}
\begin{figure*}[!ht]
	\centering
	\includegraphics[width=1\linewidth]{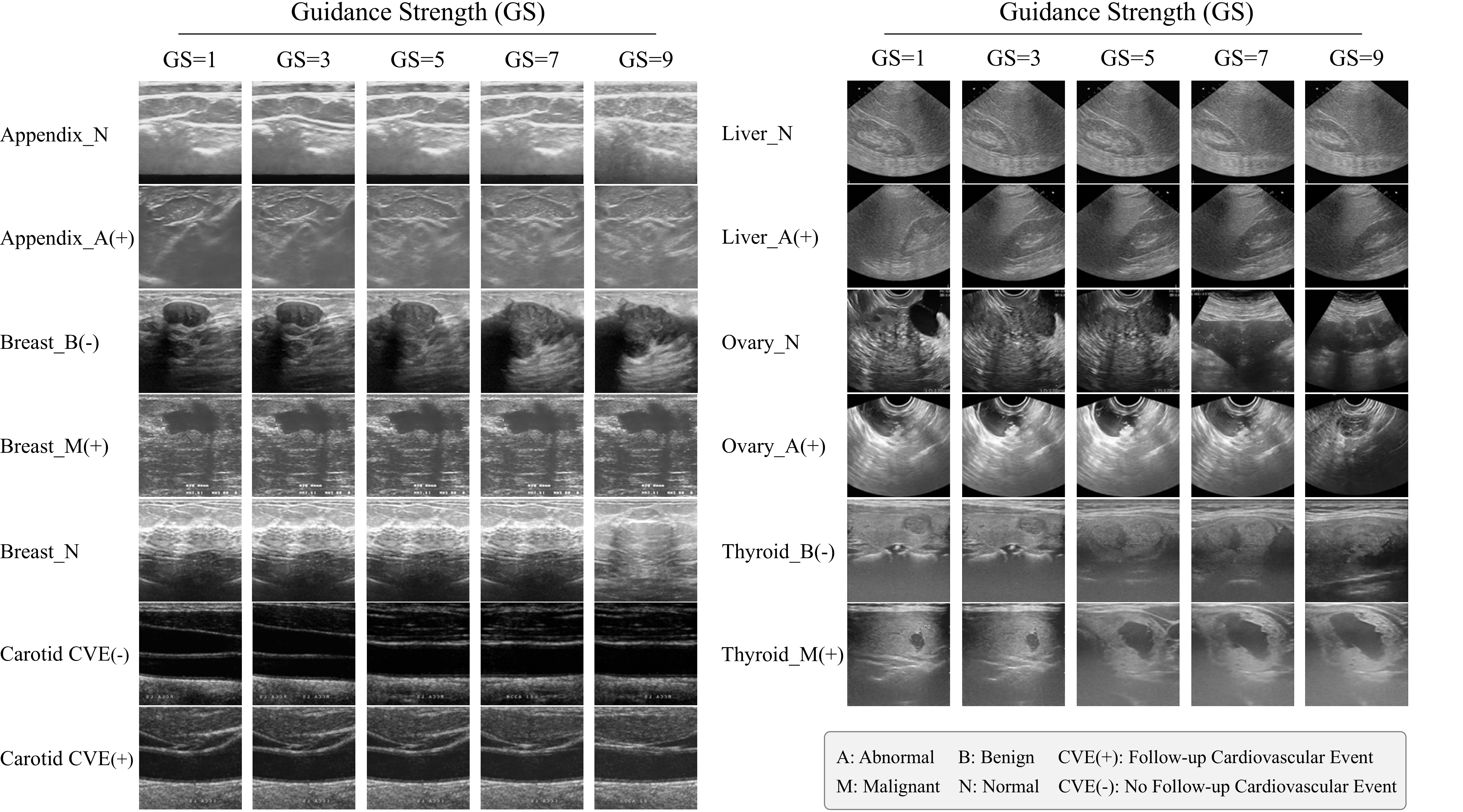}
	\caption{\textbf{EvoDiff-generated synthetic probes across varying guidance strengths.} Each row shows synthetic ultrasound images generated for one organ and class, and the five columns correspond to guidance strengths (GS) of 1, 3, 5, 7, and 9. In the row labels, N denotes normal, A denotes abnormal, B denotes benign, M denotes malignant, and CVE(+) and CVE(-) denote cases with and without a follow-up cardiovascular event. Progressive intensification of the information-gain (IG) guidance across active learning (AL) iterations creates a curriculum learning framework in which sample difficulty systematically increases, moving generation from typical appearances toward diagnostically ambiguous appearances.}
    \label{gs}
\end{figure*}

To gain deeper insight into the IG-guided generative behavior of EvoDiff and its role in facilitating progressive downstream classifier learning under limited annotations, we further visualize US images generated under varying guidance strengths. This experiment not only examines feature fidelity across anatomical categories but also reveals how EvoDiff constructs a curriculum learning mechanism at the data level through controllable difficulty modulation. As illustrated in Fig. \ref{gs}, representative images were synthesized from multiple organs, including appendix, breast, carotid artery, liver, ovary and thyroid, under five guidance strength (GS) settings ranging from 1 to 9.

The sampling process in EvoDiff is regulated by IG gradient guidance, which directs diffusion trajectories toward semantically informative regions of the feature space, while controlled stochastic perturbations preserve sample diversity. In parallel, LOEA performs non-backpropagation-based refinements, enabling EvoDiff to maintain stability during iterative AL rounds. At lower guidance strengths (GS = 1 or 3), synthesized images exhibit clear intra-class coherence and strong expression of discriminative pathology-related attributes. For example, benign and malignant lesions in breast US remain visually separable, and vessel wall continuity in carotid cases is preserved, demonstrating that EvoDiff can accurately model early-stage class structure with high semantic clarity.

As guidance strength increases, however, we observe a progressive transformation in synthesized samples, that intra-class distinctions gradually become more ambiguous, while inter-organ anatomical characteristics remain identifiable. This controlled blurring of class-specific pathological cues can be attributed to the fact that stronger guidance steers generation toward higher-entropy regions in the classifier's decision space, effectively emphasizing samples near classification boundaries. Notably, this transition from semantically clear to increasingly ambiguous examples is not a degradation of generative quality but rather a deliberate mechanism that enables curriculum learning within the AL pipeline. Specifically, low-GS samples act as ``easy (certain)'' synthetic queries that support initial classifier convergence, whereas higher-GS samples introduce ``hard (uncertain)'' cases that stimulate discriminative refinement in subsequent learning rounds.

This visualization confirms two critical properties of EvoDiff. First, the framework has the potential to control the certainty of synthesized data, enabling the construction of a continuous curriculum where sample informativeness gradually increases alongside model capability. Second, the integration of IG guidance and LOEA ensures that as annotation rounds progress and additional supervision becomes available, EvoDiff continues to produce high-information synthetic probes that remain informative for sample retrieval rather than collapsing into trivial noise. Consequently, the progressive increase in guidance strength aligns naturally with the iterative structure of AL, contributing to a stable and monotonic improvement in model performance as more annotations are introduced. These results reinforce the advantage of EvoDiff as a controllable, information-aware synthesis framework that enhances AL efficiency in medical US imaging scenarios constrained by annotation scarcity.

The synthetic images serve exclusively as retrieval probes and are neither annotated nor used for downstream training. We nevertheless assessed whether they preserve pathology-related information and a realistic US appearance. Diagnostic ResNet-50 models trained exclusively on real images recognized the synthetic probes with a pooled AUC between 0.733 and 0.966 across the six organ tasks (Supplementary Table~S7). In a blinded assessment by four radiologists, the pooled accuracy of real-versus-synthetic identification was 45\%, close to the 50\% chance level (Supplementary Table~S8), and synthetic diseased images received higher mean diagnostic scores than synthetic non-diseased images in all six organ tasks (Supplementary Table~S9 and Supplementary Figures~S9 and~S10).

\subsection*{Guidance-Driven Curriculum Learning Sample Analysis and Uncertainty Interpretation}
\begin{figure*}[!ht]
	\centering
	\includegraphics[width=0.95\linewidth]{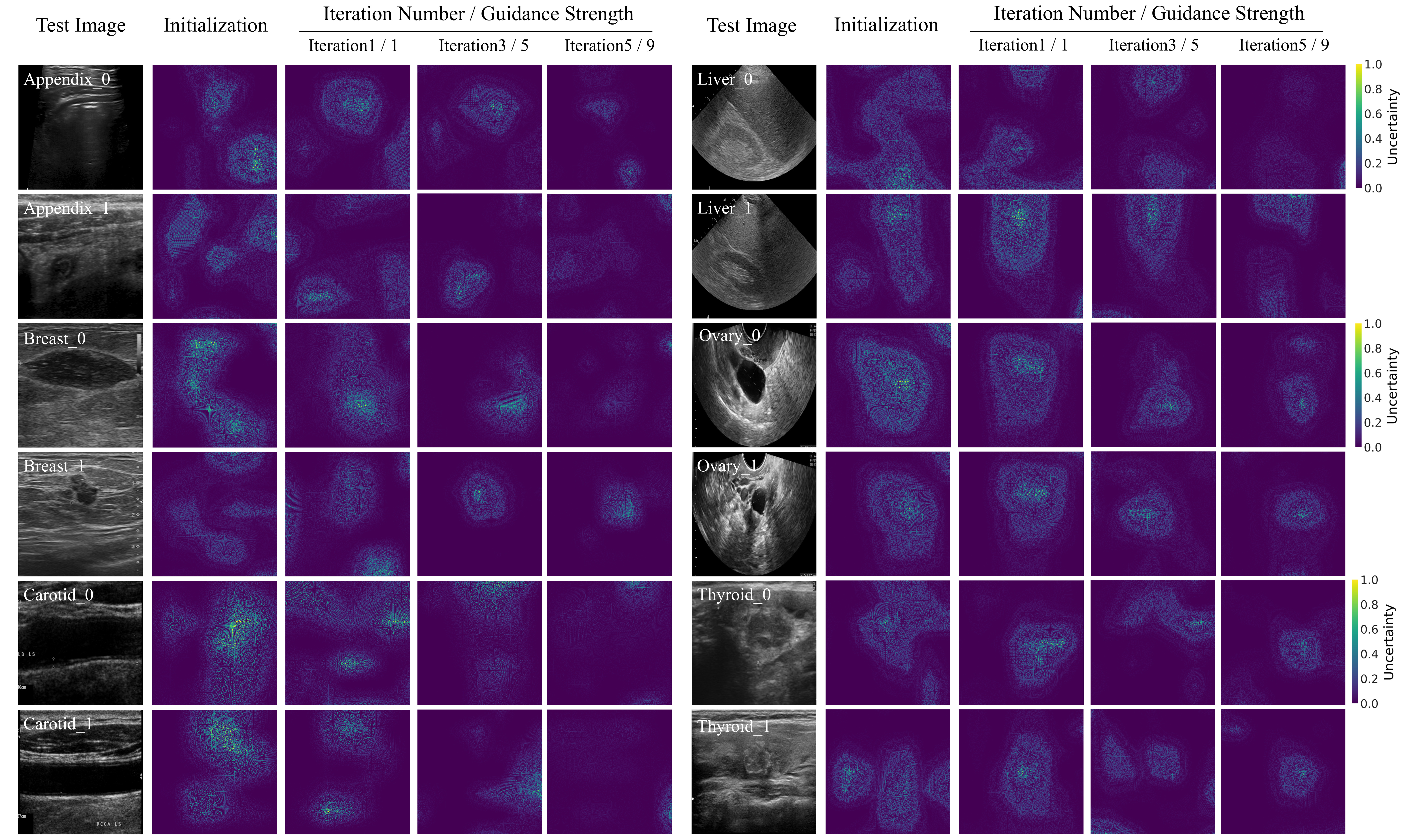}
	\caption{\textbf{Systematic uncertainty reduction in downstream model predictions across multi-organ ultrasound test sets.} Uncertainty maps are shown for six anatomical sites (appendix, breast, carotid, liver, ovary, and thyroid) with two representative cases per organ, where the suffix 0 or 1 in each row label gives the binary ground-truth label, benign 0 and malignant 1, or normal 0 and abnormal 1. Within each half of the figure, the first column shows the original test image, the second column shows the uncertainty map of the initialized model, and the following three columns show the maps after active learning (AL) rounds 1, 3, and 5, generated with guidance strengths of 1, 5, and 9, respectively. Each map is coloured from dark purple for an uncertainty of 0, through blue and green, to yellow for an uncertainty of 1, as indicated by the colour bars, and the shrinking bright regions from left to right illustrate decreasing predictive uncertainty.}
    \label{uncertain}
    % \vspace{3em}
\end{figure*}

Fig. \ref{uncertain} provides a comprehensive visualization of the EvoDiff framework selecting information-rich samples from a diverse multi-organ US dataset for downstream model training, and the evolutionary dynamics of classification uncertainty in multi-organ classification tasks. 
The visualization results show that as EvoDiff constructs a curriculum learning dataset through IG-guided sampling and increasing guided strength values, it optimizes the feature learning of the downstream model, resulting in a systematic reduction in the prediction uncertainty of the downstream model as the number of AL rounds increases.

The uncertainty maps reveal distinct organ-specific patterns of information extraction. In the appendix cases, we observe initially diffuse uncertainty distributions that progressively concentrate toward anatomically relevant regions. Notably, the abnormal appendix case shows more rapid uncertainty reduction compared to normal label, suggesting that malignant/abnormal cases may contain more distinctive features that facilitate faster model convergence. 
Similar differential learning patterns are evident in breast tissue samples, where the uncertainty concentrations progressively shift from globally distributed to locally focused regions, likely corresponding to diagnostically relevant anatomical structures. The carotid artery uncertainty maps exhibit a particularly interesting transformation, with initial broad uncertainty that gradually localizes to vessel wall boundaries, which are critical regions for pathological assessment.

The color-coded uncertainty scale (0.0-1.0) illustrates how the guidance strength parameter (increasing from 1 to 9) systematically modulates the uncertainty landscape. At iteration 1 with minimal guidance (strength=1), uncertainty patterns appear diffuse and relatively uniform across organ types. By iteration 3 (guidance strength=5), we observe a critical transition phase where uncertainty begins to concentrate in diagnostically informative regions, with noticeable differentiation between benign/normal and malignant/abnormal cases. 
By iteration 5 (guidance strength=9), uncertainty has converged to minimal, highly localized patterns, particularly evident in thyroid and ovary samples where anatomical complexity typically presents diagnostic challenges. This convergence pattern validates our hypothesis that the EvoDiff framework effectively navigates the solution space toward optimal annotation subsets through iterative refinement.

The visualization reveals that different organ systems exhibit varying rates of uncertainty reduction, reflecting their inherent complexity and feature distinctiveness. Liver and thyroid samples demonstrate relatively slower uncertainty convergence, likely due to their heterogeneous parenchymal textures that present more complex feature spaces for model interpretation. In contrast, ovarian and carotid samples show more efficient uncertainty reduction, suggesting greater feature discriminability in these anatomical contexts. This visualization empirically validates our theoretical framework by demonstrating how the integration of evolutionary algorithms with diffusion sampling progressively identifies the most informative regions for annotation across diverse anatomical contexts, thus enabling truly adaptive and organ-agnostic AL for US image classification.

\section*{Discussion} 
Progress in clinical deep learning continues to be constrained by the scarcity of high-quality annotations and the absence of principled strategies for prioritizing informative training samples~\cite{candemir2021training,ching2018opportunities}. Conventional CNN-based medical AI models rely heavily on dense supervision yet lack mechanisms to identify which images contribute most to model refinement. Although recent foundation models offer transferable representations across tasks and modalities, their adaptation to new clinical settings still demands high-information data, limiting scalability in real-world practice~\cite{pai2024foundation,huang2025foundation}. 
These fundamental challenges are particularly pronounced in US imaging, where technical limitations intersect with unique workflow characteristics to create compounding barriers for supervised learning. From a technical perspective, operator-dependent acquisition, heterogeneous scanning protocols, and pervasive speckle noise hinder reliable model training. However, the most critical challenge stems from US's distinct clinical workflow. The temporal separation between image acquisition and diagnostic labeling creates an ``image-label separation" phenomenon, where individual frames often lack direct diagnostic associations at the time of storage~\cite{drukker2020introduction, prevedello2019challenges}. 
This labeling challenge is exacerbated by US's inherent data redundancy and imbalance. Sequential scanning generates overlapping visual content, while normal anatomical variations vastly outnumber pathological cases. Consequently, traditional full-dataset training approaches become suboptimal, as models may overfit to redundant normal samples while failing to adequately learn from rare but clinically critical pathological cases.

To address these limitations, we introduce EvoDiff, a generative AL framework designed to improve label efficiency by integrating data quality awareness with information-driven sampling~\cite{hu2025addressing}. 
By systematically identifying high-information samples and reducing reliance on labeled data, EvoDiff addresses both the technical constraints of US imaging and the workflow-specific challenges of clinical practice. Importantly, EvoDiff remains compatible with both classical CNN workflows and foundation model architectures, enabling its adoption across diverse clinical pipelines without requiring fundamental infrastructure changes. 
EvoDiff reconceptualizes AL as a diffusion-based feature space optimization process rather than a heuristic query selection problem. By embedding IG directly into the generative process, the framework steers sampling along entropy-increasing gradients with respect to the sampling state, without any parameter update to the pretrained networks. In contrast to uncertainty-based strategies that passively depend on posterior variance, EvoDiff actively explores the latent feature manifold to identify samples expected to provide maximal information. An evolutionary algorithm mitigates premature convergence to suboptimal regions of the data distribution, while a curriculum learning-inspired progression from low- to high-uncertainty samples promotes stable model optimization under sparse supervision. Together, these components establish a principled formulation of AL as a controllable generative search process. 
We next evaluated how this theoretical advantage translates to empirical performance across established AL strategies such as margin sampling and entropy-based querying, EvoDiff alleviates the instability that arises from unreliable posterior probability estimates, a problem that is particularly acute in the US due to artifacts and class imbalance~\cite{rastogi2011opportunities}. In contrast to clustering and diversity-based selection methods, which may inadvertently select redundant samples, EvoDiff synthesizes semantic probes that retrieve previously underrepresented yet informative regions of the data distribution.

Beyond these methodological advantages, EvoDiff exhibited broad architectural compatibility, consistently enhancing performance across both CNN-based models and Transformer-based foundation models. 
 Experimental validation on ResNet-50 revealed superior performance compared to four baseline methods across 6 benchmark tasks encompassing 8 datasets. Under low annotation resource conditions (10\% annotation), the data selected by EvoDiff achieved an accuracy of 79.51\% on the downstream cardiovascular event prediction task, while the comparative method BALD only reached 74.03\%. Under higher annotation resource scenarios, such as 50\% cumulative annotation budget, the curriculum learning design of EvoDiff ensures consistent performance improvements for downstream models. 
Furthermore, in the fatty liver diagnosis task, data selected by EvoDiff achieved an accuracy of 97.24\% with 50\% cumulative annotation budget, approaching the upper bound of FSL results at 97.41\% accuracy. Similar findings were observed in the appendicitis diagnosis task, where EvoDiff achieved 63.73\% accuracy with 50\% cumulative annotation budget, exhibiting only a 0.31 percentage point accuracy degradation compared to the fully supervised results. 
To further validate the generalization capabilities, external experimental evaluation demonstrated that EvoDiff substantially enhanced USFM's cross-domain performance, particularly in cross-center evaluations, indicating robust resilience to domain shift and reliable sample selection across heterogeneous clinical cohorts. 
On three unseen external breast validation sets, fine-tuning USFM with EvoDiff-selected samples yielded generalization gains ranging from 2.9\% to 16.9\%. For the thyroid dataset, when generalizing from TN3K to the TN5000 dataset, EvoDiff achieved a 4.8\% improvement. This demonstrates that EvoDiff can serve as an effective approach in the medical US imaging domain to reduce annotation costs while automatically excluding noisy annotations and enhancing the generalization capability of foundation models on unseen domains. 
Collectively, these findings establish EvoDiff as a versatile and transferable AL paradigm that is not constrained by model architecture or training paradigm. 
In addition to improving the diagnostic accuracy of downstream models, EvoDiff offers practical value for real-world annotation workflows. EvoDiff employs a curriculum guided sampling strategy that aligns with clinical reasoning by first selecting high-confidence cases and then progressively introducing diagnostically challenging cases. This learning trajectory stabilizes early optimization and improves annotation usefulness. At matched downstream performance, EvoDiff is estimated to require 24\% to 84\% fewer annotations than random sampling. The corresponding workloads in minutes are estimated using unit annotation times measured in the six-reader panel (Supplementary Table~S4). This estimated reduction reflects the smaller number of annotations required, as EvoDiff preferentially selects samples that maximize model learning rather than those that are merely ambiguous.

Beyond accuracy, a practical motivation for label-efficient learning in ultrasound is what we term \textbf{image-label separation}. Unlike static modalities such as CT or MRI, ultrasound examination proceeds through real-time scanning, storage of key frames, and case-level report generation only after all stored images have been reviewed~\cite{drukker2020introduction}. Individual frames therefore carry no frame-level diagnostic label at the time of storage. In a thyroid examination a radiologist may store 15 to 35 images across bilateral transverse and longitudinal views, yet the report records a single case-level conclusion, so only 3 to 5 images directly demonstrate the finding while the remainder cannot be associated with that label without expert annotation. Hospital information systems compound the problem, because imaging archives and diagnostic reports are typically held in separate systems under different access permissions.

Active learning also remains useful when labels already exist, because ultrasound data are highly redundant. Consecutive frames in dynamic acquisition overlap substantially in content, standard views of the same organ repeat structurally, and normal studies vary little across patients, which leaves datasets in long-tailed distributions~\cite{prevedello2019challenges}. Training on the full archive therefore spends computation on repeated patterns, amplifies class imbalance, and lowers annotation consistency when large batches are labeled at once. EvoDiff addresses both settings by selecting the informative minority of an archive, which reduces expert labeling burden while preserving downstream performance.

Despite its advantages, EvoDiff has limitations. First, the framework relies on a frozen feature encoder during the hybrid generation-selection process. Although this encoder is used solely for representation embedding rather than for universal uncertainty estimation across downstream classifiers, potential embedding shifts may still influence sample prioritization. Second, EvoDiff is currently validated on 2D B-mode US images; future work is required to extend the framework to MRI and CT, where spatiotemporal coherence may alter AL dynamics. Third, EvoDiff involves a trade-off between downstream performance gains and acquisition time. In the representative BUS-BRA benchmark reported in Supplementary Table~S10, one EvoDiff acquisition round requires 6.8 GPU-hours on a single NVIDIA A40, compared with 19.4 seconds for Max-Entropy and 86.5 seconds for BALD. This acquisition step runs offline and requires no radiologist time. The present implementation incorporates DDIM acceleration. Future work could investigate reduced-step sampling, diffusion distillation, or lightweight generative adapters to reduce acquisition time and improve scalability. Fourth, the 24\% to 84\% reduction in annotation workload is estimated with the HAE metric, which combines annotation counts at matched downstream performance with unit annotation times measured in the six-reader panel. For each reader and organ task, the same unit annotation time is applied to both strategies, an assumption that follows from diagnostic classification annotation, in which every image receives one categorical label. Future work could develop more fine-grained measures of human annotation effort that account for sample-level variation in annotation difficulty, confidence, and decision time.

Looking forward, EvoDiff opens promising directions for generative AL. Future extensions may include learning adaptive guidance schedules instead of a fixed curriculum, enabling the sampler to dynamically adjust difficulty based on downstream classifiers' uncertainty. Another promising avenue is integrating EvoDiff with segmentation tasks to reduce annotation time. Although the present study focuses on the US, the framework is intrinsically modality agnostic and may be extended to other medical imaging domains, such as MRI and CT, where acquisition variability and annotation scarcity continue to limit scalability. 
In summary, EvoDiff reframes AL as a process of guided generative exploration rather than passive sample selection. By integrating IG with diffusion priors and curriculum sampling, the framework offers a principled strategy for label-efficient model development. Its consistent benefits across CNN and foundation model architectures, together with its demonstrated annotation efficiency, indicate strong potential for enabling data-centric clinical AI under real-world constraints.

\section*{Methods}

\subsection*{Overview}
As illustrated in Fig.~\ref{alround}, our EvoDiff framework follows a two-phase training scheme:
(1) Organ-conditioned coarse pretraining. EvoDiff and a guidance classifier are pretrained on $\mathcal{D}_i^{\text{U}}$ with organ identity as the only conditioning signal, so that anatomical representations are learned without any pathology-level annotation. The diffusion denoising objective itself remains self-supervised, because its regression target is the injected noise. 
(2) Full supervised fine-tuning. The pretrained guidance classifier will be incorporated as part of the classifier-guided EvoDiff architecture for joint fine-tuning on $\mathcal{D}_i^{\text{L}}$ using class-level pathology annotations to capture diagnostic features. 
During sampling, we apply varying guidance strengths for generating the Synthetic set $\mathcal{D}_i^{\text{G}} = \{(x_{j}^{i}, p_{j}^{i})\}_{j=1}^{G}$, where $p_{j}^{i}$ denotes pseudo-labels, and $G$ is the synthetic sample count (set to 100 to speed up the denoising process). 
Subsequently, we compute the cosine similarity between the generated samples and unlabeled data in $\mathcal{D}_i^{\text{U}}$, and select the most informative samples for expert annotation. This forms the Actively annotated set $\mathcal{D}_i^{\text{A}} = \{(x_k^i, y_k^i)\}_{k=1}^{Z}$, where $Z$ denotes the constrained number of annotated samples. EvoDiff therefore follows a \textbf{generate-to-retrieve} paradigm. The synthetic set $\mathcal{D}_i^{\text{G}}$ serves only as a set of retrieval probes, and its members are neither annotated by experts nor added to the downstream training data. Only the retrieved real samples in $\mathcal{D}_i^{\text{A}}$ are annotated and used for downstream fine-tuning. 
This process is iteratively applied to each dataset. Downstream models are then finetuned on the EvoDiff-selected labeled data $\mathcal{D}_i^{\text{A}}$. The downstream CNN experiments use one independently trained ResNet-50 per dataset rather than a unified pan-organ model or one model per organ, and no model parameters are shared across datasets or organs. For each dataset the ResNet-50 is initialized from the same ImageNet-pretrained weights, so that the acquisition strategy is the only factor that differs across the compared methods.

\subsection*{Model Pre-training}
In order to fully utilize the large amount of data without diagnostic labels $\mathcal{D}_i^{\text{U}}$ and the small amount of data with diagnostic labels $\mathcal{D}_i^{\text{L}}$, the training process was divided into two phases, namely (1) an organ-conditioned coarse pretraining phase and (2) a pathology-level supervised fine-tuning phase. However, both training objectives are learning discriminative US image features.

Inspired by \cite{nichol2021improved}, we optimize guidance classifier's loss and then used a loss function to train the diffusion model as follow:
\begin{equation}
    \begin{aligned}
L_{\text {diff}}= \sum_{t}\mathbb{E}_{x_{0}, \epsilon}\left[\left\|\epsilon-\epsilon_{\theta}\left(\sqrt{\bar{\alpha}_{t}} x_{0}+\sqrt{1-\bar{\alpha}_{t}} \epsilon, y, t\right)\right\|^{2}\right] + 0.001L_{\mathrm{vlb}}.
\end{aligned}
\end{equation}

To facilitate category-based feature distinguish, which serves as the foundation for IG-based gradient sampling, we adopted the approach proposed by Luo et al.~\cite{luo2024measurement} to train a classifier $f_{\phi}\left(x_{t}, t\right)$. This guidance classifier architecture is based on the downsampling trunk of the U-Net model augmented with an attention pooling mechanism. The training loss function is formulated as follows:
\begin{equation}
    L_{\text {guidance\_classifier }}=\sum_{t}\mathbb{E}_{x_{0}, \epsilon}\left[\left\|y-f_{\phi}\left(x_{t}, t\right)\right\|_{c}\right],
\end{equation}
where $\|\cdot\|_{c}$ represents the cross-entropy loss function.

\subsection*{Information-Guided Diffusion Sampling}
% bald
Houlsby et al. \cite{houlsby2011bayesian} proposed an information-theoretic AL approach that approximates mutual information scores through the inconsistency of Bayesian network parameter posterior distributions, seeking data points that maximize IG. 
{
Inspired by this information-theoretic principle, we distinguish the ideal Bayesian information-gain objective from a tractable trajectory-local information-gain surrogate used during classifier-guided Denoising Diffusion Implicit Model (DDIM) sampling. At each reverse-diffusion step, the trajectory-local surrogate evaluates the predictive-entropy gain of an LOEA-refined candidate relative to the current diffusion state at the same timestep. This criterion provides a tractable mechanism for steering continuous latent-space sampling toward more informative regions while keeping the diffusion model and guidance-classifier parameters fixed during sampling.
}
%最终
{Ultimately, the generated images serve as information-guided synthetic probes that direct retrieval toward corresponding real samples in the unlabeled candidate pool; the probes themselves are not annotated or used for downstream training.}

% DDPM
According to the Denoising Diffusion Probabilistic Model (DDPM), the forward diffusion process progressively adds random Gaussian noise to data distribution $q(x_{0})$ through a fixed Markov chain:
    \begin{equation}
        q\left(\boldsymbol{x}_{1: T} \mid \boldsymbol{x}_{0}\right)=\prod_{t=1}^{T} q\left(\boldsymbol{x}_{t} \mid \boldsymbol{x}_{t-1}\right),
    \end{equation}
    
    \begin{equation}
    q\left(x_{t} \mid x_{t-1}\right)=\mathcal{N}\left(x_{t} ; \sqrt{\alpha_{t}} x_{t-1},\left(1-\alpha_{t}\right) I\right),
    \label{e1}
    \end{equation}
where ${x_t}$ is original data $x$ after $t$ timesteps noise added, $\alpha_t=1-\beta_{t}$ is the noise variance schedule factor, $\beta_t$ is the hyperparameter that controls the noise variance increment. At each step, the noise follows $ \ \mathcal{N}(0, \beta_t I)$.

Equation \ref{e1} indicates that at each timestep, $x_t$ is sampled from a Gaussian distribution, which is scaled of the previous timestep $x_{t-1}$, with added noise. Therefore, we can directly compute $x_t$ from the initial state $x_0$ at any arbitrary timestep $t$ using the following closed-form expression:
\begin{equation}
    x_t = \sqrt{\bar{\alpha}_t}x_0 + \sqrt{1-\bar{\alpha}_t}\epsilon, \quad \epsilon \sim \mathcal{N}(0,I),
\end{equation}
where $\epsilon$ represents noise. 
And the reverse denoising process occurs through network $\epsilon_{\theta}\left(x_{t}, t\right)$ predicting added noise $\epsilon$ to estimated $x_0$.

{We use deterministic DDIM sampling with $\sigma_t=0$ and the cumulative schedule $\bar{\alpha}_t=\prod_{s=1}^{t}\alpha_s$, with $\bar{\alpha}_0=1$. The raw diffusion-model prediction and IG-guided noise estimate are distinguished as
\begin{equation}
\epsilon_t^{\mathrm{raw}}=\epsilon_\theta(x_t,t,y),
\qquad
\widetilde{\epsilon}_t
=\epsilon_t^{\mathrm{raw}}-\sqrt{1-\bar{\alpha}_t}\,\boldsymbol{g}_t,
\end{equation}
where $\boldsymbol{g}_t$ is the sampling-state guidance vector defined below. The guided estimate $\widetilde{\epsilon}_t$ is used consistently in both parts of the deterministic transition:
\begin{equation}
\widehat{x}_0
=\frac{x_t-\sqrt{1-\bar{\alpha}_t}\,\widetilde{\epsilon}_t}{\sqrt{\bar{\alpha}_t}},
\qquad
x_{t-1}
=\sqrt{\bar{\alpha}_{t-1}}\,\widehat{x}_0
+\sqrt{1-\bar{\alpha}_{t-1}}\,\widetilde{\epsilon}_t.
\end{equation}
No additional Gaussian-noise term is introduced during the reverse update.}

Previous BALD method demonstrates that the core objective of information theory-based AL is to reduce the number of possible hypotheses as rapidly as possible. 
{
Accordingly, we formulate an ideal Bayesian information-gain objective to express the information-seeking principle underlying EvoDiff. Let $\Phi$ denote the random guidance-classifier parameters under the Bayesian posterior, $\phi$ the fixed guidance-classifier parameters used during practical sampling, and $\mathcal{D}_{\mathrm{L}}$ the labeled data available at the current AL stage. For an intermediate diffusion state $x_t$, the ideal Bayesian IG objective is defined as
    \begin{equation}
    \begin{aligned}
        \mathrm{IG}^{*}(x_t)
        &= I\!\left(Y;\Phi\mid x_t,t,\mathcal{D}_{\mathrm{L}}\right) \\[6pt]
        &= H\!\left(Y\mid x_t,t,\mathcal{D}_{\mathrm{L}}\right)
        - \mathbb{E}_{\Phi\sim p(\Phi\mid\mathcal{D}_{\mathrm{L}})}
        \left[H\!\left(Y\mid x_t,t,\Phi\right)\right].
    \end{aligned}
    \label{e2}
    \end{equation}
This quantity represents the ideal information-theoretic objective. Its exact evaluation requires integration over the posterior distribution of the model parameters and is therefore not computed directly at each diffusion-sampling step.
}

\begin{algorithm}[t]

\caption{Information Gain-Guided Diffusion Sampling}
\label{alg1}

\textbf{Input:}
diffusion model $\epsilon_{\theta}$;
guidance classifier $f_{\phi}$;
guidance strength $\lambda$;
LOEA;
number of diffusion timesteps $T$;
cumulative DDIM noise schedule $\{\bar{\alpha}_t\}_{t=0}^{T}$
with $\bar{\alpha}_0=1$
\\
\textbf{Output:}
$x_0$: generated synthetic probe

\begin{algorithmic}[1]

\State Sample $x_T \sim \mathcal{N}(0,\mathbf{I})$

\For{$t=T,\dots,1$}

    \Statex \textit{/* Step 1: Trajectory-local information-gain evaluation */}

    \State
    $p_t \gets
    \operatorname{softmax}\!\left(f_{\phi}(x_t,t)\right)$

    \State
    $H_{\phi}(Y\mid x_t,t)
    \gets
    -\sum_{c=1}^{C}
    p_{t,c}\log p_{t,c}$

    \State
    $z_t^{*}
    \gets
    \operatorname{LOEA}(x_t,t)$

    \State
    $p_t^{*}
    \gets
    \operatorname{softmax}\!\left(f_{\phi}(z_t^{*},t)\right)$

    \State
    $H_{\phi}(Y\mid z_t^{*},t)
    \gets
    -\sum_{c=1}^{C}
    p_{t,c}^{*}\log p_{t,c}^{*}$

    \State
    $\mathrm{IG}_{t}^{\mathrm{local}}
    (z_t^{*};x_t)
    \gets
    H_{\phi}(Y\mid z_t^{*},t)
    -
    H_{\phi}(Y\mid x_t,t)$

    \Statex \textit{/* Step 2: Sampling-state guidance */}

    \State
    $\mathcal{L}_{\mathrm{guide}}(x_t)
    \gets
    \lambda\,
    \mathrm{IG}_{t}^{\mathrm{local}}
    (z_t^{*};x_t)$

    \State
    $\boldsymbol{g}_t
    \gets
    \nabla_{x_t}
    \mathcal{L}_{\mathrm{guide}}(x_t)$

    \State
    $\epsilon_{t}^{\mathrm{raw}}
    \gets
    \epsilon_{\theta}(x_t,t,y)$

    \State
    $\widetilde{\epsilon}_t
    \gets
    \epsilon_{t}^{\mathrm{raw}}
    -
    \sqrt{1-\bar{\alpha}_t}\,
    \boldsymbol{g}_t$

    \Statex \textit{/* Step 3: Deterministic DDIM update */}

    \State
    $\widehat{x}_0
    \gets
    \dfrac{
        x_t
        -
        \sqrt{1-\bar{\alpha}_t}\,
        \widetilde{\epsilon}_t
    }{
        \sqrt{\bar{\alpha}_t}
    }$

    \State
    $x_{t-1}
    \gets
    \sqrt{\bar{\alpha}_{t-1}}\,
    \widehat{x}_0
    +
    \sqrt{1-\bar{\alpha}_{t-1}}\,
    \widetilde{\epsilon}_t$

\EndFor

\State \Return $x_0$

\end{algorithmic}
\end{algorithm}

{
Exact evaluation of the Bayesian objective in Equation~\ref{e2} is intractable during iterative diffusion sampling because it requires integration over the model-parameter posterior. Rather than numerically estimating this posterior mutual information at every reverse-diffusion step, EvoDiff operationalizes the same information-seeking principle using a tractable \textbf{trajectory-local information-gain surrogate}. For the current diffusion state $x_t$ and an LOEA candidate $z$ evaluated at the same noise level $t$, we define
    \begin{equation}
    \mathrm{IG}_{t}^{\mathrm{local}}(z;x_t)
    = H_{\phi}\!\left(Y\mid z,t\right)
    - H_{\phi}\!\left(Y\mid x_t,t\right).
    \label{e4}
    \end{equation}
A positive value indicates that the local candidate induces greater predictive uncertainty than the current diffusion state. Importantly, Equation~\ref{e4} compares $x_t$ and its locally refined candidate at the \textbf{same timestep}; it does not compute an entropy difference between adjacent DDIM timesteps.

For any state $u$ evaluated at timestep $t$, predictive entropy is computed from the frozen guidance classifier as
    \begin{equation}
        H_{\phi}[Y \mid u,t]
        =-\sum_{i=1}^{C} p_{\phi}(y_i \mid u,t) \log p_{\phi}(y_i \mid u,t),
    \label{eq7}
    \end{equation}
where $p_{\phi}(y_i \mid u,t)$ denotes the softmax probability for category $y_i$ predicted by $f_{\phi}$, and $C$ is the number of categories.

To incorporate the trajectory-local IG surrogate into diffusion sampling, we define the scalar guidance objective
    \begin{equation}
        \mathcal{L}_{\mathrm{guide}}
        \gets
        \lambda \cdot
        \mathrm{IG}_{t}^{\mathrm{local}}(z_t^{*};x_t)
        \label{eq11}
    \end{equation}
where $z_t^{*}$ is the LOEA-refined candidate. The corresponding sampling-state guidance vector is
    \begin{equation}
        \boldsymbol{g}_t
        =\nabla_{x_t}\mathcal{L}_{\mathrm{guide}}(x_t).
    \end{equation}
Thus, $\mathcal{L}_{\mathrm{guide}}$ is a scalar objective, whereas $\boldsymbol{g}_t$ is the vector used to correct the DDIM noise estimate. The gradient is computed once with respect to $x_t$ while the diffusion-model and guidance-classifier parameters remain frozen.

Thus, Equation~\ref{e2} specifies the ideal Bayesian information-gain principle, whereas Equation~\ref{e4} provides the tractable trajectory-local surrogate used during sampling. As illustrated in Algorithm~\ref{alg1}, this surrogate constructs an IG-based implicit guidance field that redirects the reverse-diffusion trajectory toward regions of higher predictive uncertainty without updating model parameters.
}

\subsection*{Local Optimization Evolutionary Algorithm}
Evolutionary algorithms are optimization algorithms that simulate the process of natural selection by iteratively optimizing solutions through operations such as selection, crossover, and mutation. 
Zhang et al. \cite{zhang2024diffusion} were the first to mathematically establish a link between diffusion models and evolutionary algorithms. Leveraging the noise-addition and denoising characteristics of diffusion models, evolutionary algorithms can precisely refine solutions within the parameter space.

{
We propose a LOEA with a category-similarity constraint to refine candidate states locally at each reverse-DDIM timestep. As detailed in Algorithm~\ref{loea}, LOEA searches a candidate population $\mathcal{P}_t$ and favors candidates with high predictive entropy while penalizing candidates that violate the feature-similarity constraint.

This entropy term is consistent with the trajectory-local IG surrogate in Equation~\ref{e4}. Within a single LOEA call, $H_{\phi}(Y\mid x_t,t)$ is constant for all candidates $z\in\mathcal{P}_t$; therefore, ranking feasible candidates by $H_{\phi}(Y\mid z,t)$ yields the same information-term ranking as $\mathrm{IG}_{t}^{\mathrm{local}}(z;x_t)$. The similarity term acts as an additional constraint that preserves category consistency rather than defining a different information-gain quantity. Specifically, ResNet-50 features from labeled US images are used to construct the feature bank $B$, and cosine similarity is used to penalize candidates that deviate from the corresponding category feature space. 
Selection retains the top-performing half of the population according to the constrained fitness score, followed by uniform crossover and Gaussian mutation. After $G$ generations, LOEA returns the candidate with the highest constrained fitness. Thus, Algorithm~\ref{loea} provides the local candidate optimization used by Algorithm~\ref{alg1}, while the information-seeking component remains consistent with the trajectory-local IG surrogate defined in Equation~\ref{e4}.
}

\begin{algorithm}[!ht]

\caption{Local Optimization Evolutionary Algorithm}
\label{loea}
{\textbf{Input:} current diffusion state $x_t$; population size $n$; timestep $t$; feature extractor $\mathcal{F}$; frozen guidance classifier $f_\phi$;} \\
\hspace*{4.2em} feature bank $\mathcal{B}$; similarity threshold $\tau$ \\
{\textbf{Output:} $z_t^{*}$: locally refined candidate with the highest constrained fitness}
\begin{algorithmic}[1]
\State $P \gets \operatorname{InitializePopulation}(x_t,n)$ \hfill \textit{// Local perturbations around $x_t$}
\State $G \gets 40$, $p_{\text{mut}} \gets 0.1$, $\sigma_{\text{mut}} \gets 0.1$
\For{$g = 1$ to $G$}
    \State {$p \gets \operatorname{softmax}(f_\phi(P, t))$, \quad $\log p \gets \log \operatorname{softmax}(f_\phi(P, t))$}
    \State {$\mathcal{H} \gets -\sum_i p_i \log p_i$ \hfill \textit{// Entropy term; equivalent local-IG ranking before similarity penalty}}
    \State $\mathbf{V} \gets \mathcal{F}(P)$, \quad $\pi \gets \mathbf{0}$
    \For{$i = 1$ to $n$}
        \State $f_i \gets \mathbf{V}[i]$
        \State $s_c \gets \cos(f_i, \mathcal{B}[c]), \forall c$
        \If{$\max_c(s_c) < \tau$}
            \State $\pi[i] \gets 1$
        \EndIf
    \EndFor
    \State {$\text{fitness} \gets \mathcal{H} - 10 \cdot \pi$ \hfill \textit{// Local information term with similarity penalty}}
    \State $P_{\text{sel}} \gets \operatorname{TopK}(P,\ \text{fitness},\ k = \lfloor n/2 \rfloor)$

    \Statex \textit{/* Crossover */}
    \For{$i = 1$ to $n$}
        \State Sample $p_1, p_2 \sim P_{\text{sel}}$
        \State Sample $m \sim \{0,1\}^d$
        \State $c_i \gets m \cdot p_1 + (1 - m) \cdot p_2$
    \EndFor
    \State $P \gets \{c_i\}_{i=1}^n$

    \Statex \textit{/* Mutation */}
    \State $\epsilon \sim \mathcal{N}(0,\ \sigma_{\text{mut}}^2)$
    \State $M \sim \operatorname{Bernoulli}(p_{\text{mut}})$
    \State $P \gets P + M \cdot \epsilon$
\EndFor

\State {$p \gets \operatorname{softmax}(f_\phi(P, t))$, \quad $\log p \gets \log \operatorname{softmax}(f_\phi(P, t))$}
\State $\mathcal{H} \gets -\sum_i p_i \log p_i$
\State Recompute $\pi$ using same procedure
\State {$\text{fitness} \gets \mathcal{H} - 10 \cdot \pi$ \hfill \textit{// Local information term with similarity penalty}}
\State {$z_t^{*} \gets P[\arg\max(\text{fitness})]$}
\State \Return $z_t^{*}$
\end{algorithmic}
\end{algorithm}

\subsection*{Curriculum Learning Strategy for Progressive Sample Selection}
Directly prioritizing high-uncertainty samples in early AL stages can introduce unstable optimization and slow convergence to downstream classifiers due to noisy gradients and unreliable supervision signals. To address this challenge, EvoDiff incorporates a Curriculum Learning (CL) strategy for progressive sample selection, enabling a controlled transition from certain (easy) to uncertain (hard) samples over AL iterations. This design ensures that the downstream classifier first establishes robust decision primitives before being progressively exposed to ambiguous, boundary-level cases that refine decision surfaces.

We modulate sample difficulty using a discrete scheduling of the GS parameter $\lambda$ applied during IG-guided diffusion sampling. Let $i$ denote the AL iteration index, the curriculum schedule is defined as:
\begin{equation}
    \lambda^{(i)} \in\{1,3,5,7,9\},
\end{equation}
{where lower $\lambda$ emphasizes more certain regions and larger $\lambda$ guides sampling toward higher predictive entropy. Round~0 uses only the fixed labeled initialization set. Rounds~1--6 correspond to cumulative candidate-pool budgets of 5\%, 10\%, 20\%, 30\%, 40\%, and 50\%, respectively. The corresponding newly annotated batches are therefore 5\%, 5\%, 10\%, 10\%, 10\%, and 10\% of the original candidate-pool size. The lowest guidance strength $\lambda{=}1$ spans the first two rounds, and the remaining guidance strengths 3, 5, 7, and 9 are each used for one subsequent round.}

{
Following Equation~\ref{eq11}, the curriculum-stage guidance objective becomes
\begin{equation}
    \mathcal{L}_{\mathrm{guide}}^{(i)}(x_t)
    = \lambda^{(i)} \cdot \mathrm{IG}_{t}^{\mathrm{local}}(z_t^{*};x_t),
\end{equation}
which modifies the DDIM denoising through the sampling-state gradient
\begin{equation}
    \boldsymbol{g}_t^{(i)}
    =\nabla_{x_t}\mathcal{L}_{\mathrm{guide}}^{(i)}(x_t),
    \qquad
    \widetilde{\epsilon}_t^{(i)}
    =\epsilon_t^{\mathrm{raw}}-\sqrt{1-\bar{\alpha}_t}\,\boldsymbol{g}_t^{(i)}.
\end{equation}
This formulation provides explicit control over sample difficulty while keeping the model parameters frozen during sampling.
}

{EvoDiff uses curriculum-guided synthetic images as probes to retrieve real samples from the remaining candidate pool $\mathcal{D}_{i}^{\mathrm{U}}$. At stage $i$, the frozen retrieval encoder $\mathcal{F}$ maps a generated probe $g\in\mathcal{D}_{i}^{\mathrm{G}}$ and a real candidate $u\in\mathcal{D}_{i}^{\mathrm{U}}$ to $\boldsymbol{\phi}_g=\mathcal{F}(g)$ and $\boldsymbol{\phi}_u=\mathcal{F}(u)$. Their cosine similarity is
\begin{equation}
\operatorname{cos}\!\left(\boldsymbol{\phi}_g,\boldsymbol{\phi}_u\right)
=\frac{\boldsymbol{\phi}_g^{\top}\boldsymbol{\phi}_u}
{\lVert\boldsymbol{\phi}_g\rVert_2\lVert\boldsymbol{\phi}_u\rVert_2}.
\end{equation}
The $K_i$ highest-ranked real candidates are selected, where $K_i$ equals the stage-specific batch size above (equivalently, the round-specific $\alpha_i\%$ selection ratio). Previously selected images are removed before ranking, so acquisition is without replacement. Only the retrieved real images are annotated and used for downstream training.}

{EvoDiff adopts a cumulative curriculum rather than resetting the labeled set at each stage. Every selected image is retained in the expanded labeled training set, and subsequent acquisition operates only on the remaining candidate pool.}

\subsection*{Implementation details}
During the pre-training phase, all US images were resized to $224\times224$ pixels. For pre-training with the large-scale unlabeled data pool, the diffusion model was optimized using a total of 1,000 timesteps with a linear noise schedule, where the noise coefficient $\beta_\mathrm{t}$ ranged from $\beta_{1}=1 \times 10^{-4} \text { to } \beta_\mathrm{T}=0.02.$ The training was conducted for 20,000 iterations with a batch size of 4. During the fine-tuning stage with pathologically labeled data, the noise coefficient range was adjusted to $\beta_{1}=1 \times 10^{-5} \text { to } \beta_\mathrm{T}=0.02.$ For classifier pre-training, we employed the same number of training iterations with a batch size of 2 and a learning rate of $2 \times 10^{-4}$. The learning rate was subsequently reduced to $1 \times 10^{-4}$ during fine-tuning with pathologically labeled data.

CNN-based diagnostic model ResNet-50~\cite{resnet}: During the image preprocessing stage, all images were resized to $224\times224$ pixels and subjected to a series of data augmentation techniques, including random horizontal flipping, Color Jitter (brightness=0.2, contrast=0.2, saturation=0.2, hue=0.1), and random rotation (Random Rotation with degrees ranging from $-10^{\circ}$ to $10^{\circ}$). The experimental procedure was as follows: we first loaded ResNet-50 weights pre-trained on ImageNet, then fine-tuned the model using pathologically labeled data to obtain the initialized downstream model. The training hyperparameters were set as follows: batch size of 2, learning rate of $1 \times 10^{-4}$, with a cosine annealing strategy to facilitate convergence, and a total of 40 epochs. During the AL iterations, the initialized weights were loaded at the beginning of each round while maintaining consistent experimental settings, and model weights were saved after each AL iteration. 
Transformer-based foundation model USFM~\cite{jiao2024usfm} The data preprocessing pipeline followed the original paper's configuration, with images resized to $256\times256$ pixels and augmented with random rotations and flips. We first loaded the pre-trained weights released by USFM~\cite{jiao2024usfm}, then fine-tuned the model using pathologically labeled data and saved the weights before proceeding with incremental learning across AL iterations. Training was performed using the Adam optimizer with a cosine learning rate adjustment strategy. To ensure convergence during USFM fine-tuning, we configured dataset-specific hyperparameters, as detailed in Supplementary Table~S5.

All experiments were conducted on four NVIDIA GeForce A40 GPUs. All models were developed using the PyTorch framework. 
The global random seed was 42 for PyTorch, CUDA, NumPy, and the Python \texttt{random} module, and deterministic cuDNN execution was enabled. Each internal dataset was partitioned once using a fixed 8:1:1 split into an unlabeled candidate pool (80\%), a labeled initialization set (10\%), and an internal test set (10\%), and the case-level separation of the internal datasets, including the BUSI-specific limitation, is reported in the \textit{Cross-organ Multi-task Ultrasound Dataset} subsection. The cumulative candidate-pool budgets were 5\%, 10\%, 20\%, 30\%, 40\%, and 50\%. The corresponding newly annotated batches, and therefore the round-specific $\alpha_i$ selection ratios, were 5\%, 5\%, 10\%, 10\%, 10\%, and 10\% of the original candidate-pool size, drawn without replacement. EvoDiff generated 100 synthetic probes per requested class condition at each curriculum stage. The diffusion sampler followed the DDIM configuration described above. LOEA used a population size of 50, 40 generations, top-50\% selection, a mutation probability of 0.1, and a Gaussian mutation s.d. of 0.1. 
Random used uniform sampling without replacement. Max-Entropy ranked candidates by predictive entropy. BALD used 25 Monte Carlo dropout passes. Core-Set used greedy $k$-center selection. VAAL used its published VAE and discriminator acquisition procedure. Within each downstream backbone setting, all comparison methods used the same labeled initialization set, unlabeled candidate pool, stage-specific batch sizes, downstream architecture, initialization weights, optimizer, learning-rate schedule, number of training epochs, and fixed internal test set, so that the acquisition strategy is the only varying factor.

\subsection*{Evaluation metrics and statistical analysis}
Downstream model classification tasks were evaluated using different metrics tailored to the specific characteristics and capabilities of each model architecture. For CNN-based models, we primarily use Accuracy (ACC) as the evaluation metric, while for foundation models, we employ balanced accuracy (bAcc) as the primary metric. Additionally, Area Under the Curve (AUC) is used as a supplementary metric for comprehensive performance assessment.

The ACC metric measures the overall proportion of correctly classified samples across all classes, providing a general assessment of model performance. The ACC is defined as follows:
\begin{equation}
    \text { Accuracy }=\frac{T P+T N}{T P+F N+F P+T N},
\end{equation}
where TP refers to true positives, FP refers to false positives, FN refers to false negatives, TN refers to true negatives. 
For CNN-based downstream models (e.g. ResNet-50), ACC serves as an appropriate metric due to their training on relatively balanced subsets during the AL process, where the iterative sample selection helps mitigate severe class imbalance issues. 

However, foundation models, which are pre-trained on large-scale datasets and fine-tuned for our tasks, demonstrate different sensitivity to class distribution variations. To address this and provide a more robust evaluation that accounts for the inherent class imbalance in medical imaging datasets, we employ bAcc for foundation model evaluation. The bAcc metric provides equal weight to each class regardless of sample size: 
\begin{equation}
\begin{aligned}
\text{sensitivity} &= \frac{TP}{TP + FN} \\
\text{specificity} &= \frac{TN}{TN + FP} \\
\text{bAcc} &= \frac{\text{sensitivity} + \text{specificity}}{2}
\end{aligned}
\end{equation}

The AUC metric evaluates the model's ability to distinguish between classes by measuring the area under the Receiver Operating Characteristic (ROC) curve. 
%%%%%%%%%%%%%%%%%%%%%%%

All reported performance is obtained by five-fold cross-validation within the expanded labeled training set at each cumulative annotation budget, and results are presented as mean values with s.d. across the five fold-specific models evaluated on the fixed internal test set. Each sample-selection strategy is compared with random sampling using two-sided paired $t$-tests, and multiplicity is controlled with the Benjamini--Hochberg procedure applied across the five strategy-versus-Random comparisons that share a given dataset and annotation budget. Both unadjusted $p$-values and Benjamini--Hochberg-calibrated $q$-values are reported. Supplementary Figures~S1 to~S8 present these comparisons as heat maps, one metric per figure, reporting the unadjusted $p$-values and the mean differences relative to random sampling for both downstream architectures across the eight internal datasets and the three cumulative annotation budgets, with Benjamini--Hochberg-corrected significance marked in each map. For the ablation study, each reduced configuration was compared with the complete EvoDiff framework using two-sided paired $t$-tests, with the five cross-validation folds used as matched pairs. 
The five cross-validation runs constitute the repeated experiments referred to throughout, rather than independently repeated dataset splits.

\begin{algorithm}[!ht]
\caption{HAE Metric Computation}
\label{HAE}
\textbf{Input:} $D^U$: unlabeled candidate pool; $S_{AL}$: active learning strategy; $\Theta = \{\theta_1, \theta_2, \ldots, \theta_m\}$: performance thresholds; \\
\hspace*{4.2em} $R = \{r_1, r_2, \ldots, r_n\}$: cumulative annotation budgets; $T = 30$: time window; $\epsilon = 0.01$: precision parameter \\
\textbf{Output:} $HAE_{list}: HAE$: human annotation effort, $\triangle HAE$: human annotation effort saved for each threshold

\begin{algorithmic}[1]
\State $W_{unit} \gets \text{calibrate\_annotation\_time}(T)$ \hfill \textit{// Measure unit annotation cost}
\State $A_{random}, A_{AL} \gets \text{evaluate\_strategies}(D^U, S_{AL}, R)$ \hfill \textit{// Generate performance curves}
\State $HAE_{list} \gets []$ \hfill \textit{// Initialize HAE results}

\For{$k = 1$ to $|\Theta|$}
    \State $\theta_k \gets \Theta[k]$
    \State $r_{AL}^{\theta_k} \gets \min\{r : A_{AL}(r) \geq \theta_k\}$ \hfill \textit{// Eq. (1)}
    \State $r_i, r_{i+1} \gets \text{find\_adjacent\_ratios}(A_{random}, \theta_k)$ 
    
    \If{$A_{random}(r_i) \leq \theta_k \leq A_{random}(r_{i+1})$}
        \State $\rho \gets \frac{A_{random}(r_{i+1}) - A_{random}(r_i)}{(r_{i+1} - r_i) \times |D^U|}$ \hfill \textit{// Performance improvement rate}
        \State $r_{equiv}^{\theta_k} \gets r_i + \frac{\theta_k - A_{random}(r_i)}{\rho \times |D^U|}$ \hfill \textit{// Equivalent ratio interpolation}
    % \Else
    %     \State $r_{equiv}^{\theta_k} \gets \text{interpolate}(A_{random}, \theta_k)$ 
    \EndIf
    
    \State $N_{AL}^{\theta_k} \gets r_{AL}^{\theta_k} \times |D^U|$ \hfill \textit{// AL sampling requirement}
    \State $N_{random}^{\theta_k} \gets r_{equiv}^{\theta_k} \times |D^U|$ \hfill \textit{// Random sampling requirement}
    \State $HAE_{AL}^{\theta_k} \gets N_{AL}^{\theta_k} \times W_{unit}$ \hfill \textit{// AL sampling workload}
    \State $HAE_{random}^{\theta_k} \gets N_{random}^{\theta_k} \times W_{unit}$ \hfill \textit{// Random sampling workload}
    \State $\triangle HAE \gets (N_{random}^{\theta_k} - N_{AL}^{\theta_k}) \times W_{unit}$ \hfill \textit{// Workload savings}
    \State $HAE_{list}.\text{append}(HAE_{random}^{\theta_k}, HAE_{AL}^{\theta_k},\triangle HAE^{\theta_k})$
\EndFor

% \State $HAE_{avg} \gets \frac{1}{|\Theta|} \times \sum_{k=1}^{|\Theta|} HAE_{list}[k]$ \hfill \textit{// Average across thresholds}
% \State \Return $HAE_{avg}$
\end{algorithmic}
\end{algorithm}

\subsection*{Human Annotation Effort estimation}
Given a selected and annotated dataset $D$ from unlabeled candidate pool $D^{U}$ with $|D^{U}|$ samples and corresponding downstream diagnostic model performance metrics (e.g., accuracy), existing evaluation approaches for AL typically focus on measuring improvements in model performance or convergence rates. However, these metrics fail to quantify the practical annotation costs that radiologists encounter in real-world clinical scenarios, where evaluating the efficiency gains of AL over random sampling is crucial for AL algorithm deployment. 
Therefore, we propose a novel evaluation metric, \textbf{Human Annotation Effort (HAE)}, which quantifies the annotation workload required by different sample selection strategies when reaching equivalent downstream diagnostic performance. 
Based on HAE, we further define \textbf{Human Annotation Effort Saved ($\triangle$HAE)} to estimate the workload reduction achieved by AL algorithms. 
Diagnostic classification annotation primarily involves cognitive operations; the annotation process consists of: (1) medical image interpretation through visual examination, (2) differential diagnosis reasoning based on clinical knowledge, and (3) diagnostic decision confirmation through categorical labeling. Classification annotation focuses on diagnostic decision-making, quantifiable through time-based measurements.

Let $A_{\text{AL}}(r)$ and $A_{\text{random}}(r)$ represent the downstream classification accuracy achieved by AL and random sampling strategies after annotating samples at ratio $r$ (where $r \in [0,1]$ represents the fraction of unlabeled candidate pool $D^{U}$ annotated), respectively. Given a target performance threshold $\theta$ (e.g., 85\% accuracy), we define:
\begin{align}
r_{\text{AL}}^{\theta} &= \min\{r : A_{\text{AL}}(r) \geq \theta\} \label{eq:r_al}, \\
r_{\text{random}}^{\theta} &= \min\{r : A_{\text{random}}(r) \geq \theta\} \label{eq:r_random},
\end{align} 
where $r_{\text{AL}}^{\theta}$ represents the minimum cumulative annotation budget required by AL to achieve performance threshold $\theta$, and $r_{\text{random}}^{\theta}$ represents the minimum cumulative annotation budget required by random sampling to achieve the same performance threshold.

The unit annotation time used by the HAE metric is obtained from empirical time measurements. We conduct controlled experiments where experienced radiologists perform diagnostic classification tasks within fixed time windows $T$ (e.g., 30 minutes). The unit annotation cost is calculated as $W_{\text{unit}} = T / C_{\text{correct}}$, where $C_{\text{correct}}$ represents the number of correctly annotated samples within the time window. Reader responses were verified against the ground-truth labels, so that only correctly annotated samples contribute to $C_{\text{correct}}$. The diagnostic workload encompasses multiple measurable components, including visual examination duration, differential diagnosis consideration time, and clinical reasoning complexity.

To establish equivalent performance benchmarks, we systematically evaluate both random sampling and AL strategies across multiple cumulative annotation budgets $r_1, r_2, \ldots, r_n$ and measure their downstream diagnostic accuracy $A_{\text{random}}(r_i)$ and $A_{\text{AL}}(r_j)$. For cases where the AL performance $A_{\text{AL}}(r_j)$ falls between two consecutive random sampling performance points, we employ interpolation to determine the equivalent random sampling ratio. Specifically, we identify adjacent random sampling ratios $r_i$ and $r_{i+1}$ such that $A_{\text{random}}(r_i) \leq A_{\text{AL}}(r_j) \leq A_{\text{random}}(r_{i+1})$. The performance improvement rate within this interval is calculated as:
\begin{equation}
\rho = \frac{A_{\text{random}}(r_{i+1}) - A_{\text{random}}(r_i)}{(r_{i+1} - r_i) \times |D|},
\end{equation}
representing the accuracy gain per additional annotated sample. The equivalent random sampling ratio is then determined as: 
\begin{equation}
r_{\text{equiv}} = r_i + \frac{A_{\text{AL}}(r_j) - A_{\text{random}}(r_i)}{\rho \times |D|}.
\end{equation}

The core HAE computation follows:
\begin{align}
\text{HAE}_{\text{random}} &= N_{\text{random}} \times W_{\text{unit}}, \\
\text{HAE}_{\text{AL}} &= N_{\text{AL}} \times W_{\text{unit}}, \\
\triangle \mathrm{HAE} &= (\text{HAE}_{\text{random}} - \text{HAE}_{\text{AL}}),
\end{align}
where $N_{\text{random}} = r_{\text{equiv}} \times |D|$ and $N_{\text{AL}} = r_j \times |D|$ represent the sample quantities required by random sampling and AL, respectively, to achieve equivalent downstream performance. The performance equivalence is established through the interpolation described above: $A_{\text{random}}(r_{\text{equiv}}) = A_{\text{AL}}(r_j)$. For each reader and organ task, the same unit annotation time $W_{\text{unit}}$ is applied to both strategies, assuming comparable average time per correctly annotated image. Under this assumption, the relative saving equals $1-N_{\text{AL}}/N_{\text{random}}$ and does not depend on $W_{\text{unit}}$, whereas the absolute workload estimates depend on the measured unit annotation times. 
Overall, the computation procedure is systematically outlined in Algorithm~\ref{HAE}, encompassing empirical annotation time calibration through controlled radiologist sessions, performance curve generation across varying cumulative annotation budgets, equivalent performance sample requirement estimation, and workload savings quantification across multiple benchmarks.

\subsection*{Ethics approval}
This study was conducted in accordance with the Declaration of Helsinki. The retrospective collection and analysis of the three private external cohorts were approved by the Institutional Review Board of Zhejiang Cancer Hospital (approval No.~2026ZN004-1). Owing to the retrospective design of the study and the use of de-identified imaging data, the requirement for individual informed consent was waived by the same board. All public datasets used in this study are fully anonymized, openly available collections that were released under the ethical oversight of their original studies and were used in accordance with their respective licenses and access conditions.

\section*{Data availability}
The public medical datasets used in this study are accessible from the following sources: Appendix, \url{https://zenodo.org/records/7669442}; BUSI, \url{https://www.kaggle.com/datasets/aryashah2k/breast-ultrasound-images-dataset}; BUS-BRA, \url{https://zenodo.org/records/8231412}; UDIAT, \url{https://ieeexplore.ieee.org/abstract/document/8003418}; ALN-Ultra, \url{https://zenodo.org/records/15003119}; QAMEBI, \url{https://qamebi.com/breast-ultrasound-images-database}; CUBS, \url{https://data.mendeley.com/datasets/fpv535fss7/1}; Fatty-Liver, \url{https://zenodo.org/records/1009146}; MMOTU, \url{https://figshare.com/articles/dataset/_zip/25058690?file=44222642}; TN3K, \url{https://github.com/haifangong/TRFE-Net-for-thyroid-nodule-segmentation}; TN5000, \url{https://figshare.com/s/cb6a67f17c04b29e7edd}. The datasets generated and/or analyzed during the current study from the three private external cohorts collected at Zhejiang Cancer Hospital are not publicly available due to institutional ethics approval and patient privacy restrictions that do not permit open redistribution of the imaging data, but are available from the corresponding author on reasonable request. Shared data will be provided in de-identified form, and requests are subject to approval by the institutional review board.

\section*{Code availability}
The source code for EvoDiff is publicly available without access restrictions at \url{https://github.com/IsBaSO4/EvoDiff}. All experiments were implemented in Python using PyTorch 1.12.1 and torchvision 0.13.1, together with scikit-learn 1.3.2, SciPy 1.9.1, pandas 2.0.3, timm 0.6.12, OpenCV 4.10.0 and albumentations 1.3.0; the complete pinned software environment is listed in the \texttt{requirements.txt} file of the repository. The specific hyperparameters, random seed, and acquisition settings used to generate and analyze the current datasets are reported in the Methods (\textit{Implementation details}) and are provided as default configurations in the repository. Pre-trained weights and implementation code for the downstream models are accessible through the following repositories: USFM at \url{https://github.com/openmedlab/USFM} and ResNet-50 at \url{https://github.com/pytorch/vision/blob/main/torchvision/models/resnet.py}.

\section*{Acknowledgements}
Not applicable.

\section*{Funding}
This work was supported by Shenzhen Medical Research Fund (D2501013), the Science and Technology Development Fund of Macao (0041/2023/RIB2), the National Natural Science Foundation of China (12326619), the Frontier Technology Development Program of Jiangsu Province (BF2024078), the Science and Technology Planning Project of Guangdong Province (2023A0505020002), the National Natural Science Foundation of China (62601018), and the Macao Polytechnic University Grant (RP/FCA-25/2025). The funders had no role in study design, data collection and analysis, decision to publish, or preparation of the manuscript.

\section*{Author Contributions}
Y.D., Y.H., Y.G., YL.G. and T.T. conceived the conception and design of the study; Y.D. and Y.H. developed the deep learning model; T.T. and D.N. contributed to device support for model training and inference; Y.D. and D.X. contributed to acquisition and annotation of the data; Y.D., Y.H., R.Q., Y.S., R.M. and T.T. contributed to analysis and interpretation of the data; T.Z., D.X. and R.M. conducted the reader study. Y.D. and Y.H. wrote the initial manuscript; S.I., R.M., and T.T. supervised the work. All authors contributed to the revision of the manuscript.

\section*{Competing Interests}
The authors declare no competing financial or non-financial interests.

\bibliography{sample}

\clearpage

%%%%%%%%%%%%%%%%  Figures  %%%%%%%%%%%%%%%%

\clearpage

\clearpage

\clearpage

\clearpage

\clearpage

\clearpage

\clearpage

%%%%%%%%%%%%%%%%  Tables  %%%%%%%%%%%%%%%%

\begin{table*}[!ht]
\centering
\caption{Overview of 6 clinical applications and dataset distributions across multi-organ US datasets. Fifteen datasets comprising 16,696 images across six anatomical regions were used, with 10,182 images for model training and internal validation, and 6,514 images from ALN-Ultra, BUS\_UC, QAMEBI, TN5000, and the three private cohorts for external validation.}
\label{tdataset}

\footnotesize

\resizebox{\textwidth}{!}{

\begin{tabular}{l|l|l|cccc|l|l|l|c|l|l}

\toprule

\multirow{2}{*}{\textbf{Organ}} & 

\multirow{2}{*}{\textbf{Dataset}} & 

\multirow{2}{*}{\textbf{Challenge}} &

\multicolumn{4}{c|}{\textbf{Sample Distribution}} & 

\multirow{2}{*}{\textbf{\begin{tabular}[c]{@{}l@{}}Country\end{tabular}}} & 

\multirow{2}{*}{\textbf{\begin{tabular}[c]{@{}l@{}}Collection\\Period\end{tabular}}} & 

\multirow{2}{*}{\textbf{\begin{tabular}[c]{@{}l@{}}Ultrasound\\Equipment\end{tabular}}} & 

\multirow{2}{*}{\textbf{\begin{tabular}[c]{@{}c@{}}Patient\\Count\end{tabular}}} & 

\multirow{2}{*}{\textbf{\begin{tabular}[c]{@{}l@{}}Age Range\\(Mean±SD)\textsuperscript{a}\end{tabular}}} & 

\multirow{2}{*}{\textbf{\begin{tabular}[c]{@{}l@{}}Dataset\\Type\textsuperscript{b}\end{tabular}}} \\

\cmidrule{4-7}

& & & \textbf{Normal} & \textbf{Benign} & \textbf{Abnormal} & \textbf{Total} & & & & & & \\

\midrule

\multirow{2}{*}{Appendix} & 

Appendix~\cite{app} & 

Appendicitis diagnosis &

174 & -- & 300 & 474 & 

Germany & 

2016--2021 & 

Toshiba Xario, Aplio XG machines & 

579 & 

0--18 & 

\cellcolor{gray!20}Internal \\

& 

PrivAppendix & 

Appendicitis diagnosis &

21 & -- & 24 & 45 & 

China & 

-- & 

-- & 

-- & 

-- & 

\cellcolor{orange!30}External \\

\cmidrule[0.5pt](lr){1-2} \cmidrule[0.5pt](lr){13-13}

\multirow{7}{*}{Breast} & 

BUSI~\cite{BUSI} & 

Breast tumor recognition &

133 & 437 & 210 & 780 & 

Egypt & 

2018 & 

GE Logiq E9 & 

-- & 

25--75 & 

\cellcolor{gray!20}Internal \\

& 

BUS-BRA~\cite{BUS-BRA} & 

Breast tumor diagnosis &

-- & 1{,}268 & 607 & 1{,}875 & 

Brazil & 

-- & 

GE Logiq 5 \& 7, Aplio 300, U-Systems &

1{,}642 & 

16--89 & 

\cellcolor{gray!20}Internal \\

& 

UDIAT~\cite{UDIAT} & 

Breast tumor diagnosis &

-- & 109 & 54 & 163 & 

Spain & 

2012 & 

Siemens C512 &

163 & 

-- & 

\cellcolor{gray!20}Internal \\

& 

ALN-Ultra~\cite{ALN-Ultra} & 

Breast tumor diagnosis &

-- & 121 & 136 & 257 & 

China & 

2019-2021 & 

-- & 

257 & 

30--85 & 

\cellcolor{orange!30}External \\

& 

BUS\_UC & 

Breast tumor diagnosis &

-- & 358 & 453 & 811 & 

China & 

-- & 

SonoSkills, Hitachi Medical Systems & 

-- & 

-- & 

\cellcolor{orange!30}External \\

&

QAMEBI~\cite{QAMEBI} & 

Breast tumor diagnosis &

-- & 109 & 123 & 232 & 

Iran & 

-- & 

AirPlorer Ultimate & 

232 & 

above 18 & 

\cellcolor{orange!30}External \\ 

& 

PrivBreast & 

Breast tumor diagnosis &

-- & 59 & 41 & 100 & 

China & 

-- & 

-- & 

-- & 

-- & 

\cellcolor{orange!30}External \\

\cmidrule[0.5pt](lr){1-2} \cmidrule[0.5pt](lr){13-13}

Carotid & 

CUBS~\cite{CUBS} & 

Cardiovascular event prediction &

1{,}124 & -- & 254 & 1{,}378 & 

Toronto & 

-- & 

Epiq 7G, MyLab25, ATL HDI500, FAMUS & 

-- & 

-- & 

\cellcolor{gray!20}Internal \\

\cmidrule[0.5pt](lr){1-2} \cmidrule[0.5pt](lr){13-13}

\multirow{2}{*}{Liver} & 

Fatty-Liver~\cite{liver} & 

Fatty liver diagnosis &

170 & -- & 380 & 550 & 

Poland & 

-- & 

GE Vivid E9 & 

55 & 

-- (40.1±9.1) & 

\cellcolor{gray!20}Internal \\

& 

PrivLiver & 

Fatty liver diagnosis &

24 & -- & 45 & 69 & 

China & 

-- & 

-- & 

-- & 

-- & 

\cellcolor{orange!30}External \\

\cmidrule[0.5pt](lr){1-2} \cmidrule[0.5pt](lr){13-13}

Ovary & 

MMOTU~\cite{mmotu} & 

Ovarian tumor assessment &

267 & -- & 1{,}202 & 1{,}469 & 

China & 

-- & 

Mindray Resona8 & 

294 & 

-- & 

\cellcolor{gray!20}Internal \\

\cmidrule[0.5pt](lr){1-2} \cmidrule[0.5pt](lr){13-13}

\multirow{2}{*}{Thyroid} & 

TN3K~\cite{TN3K} & 

Thyroid nodule diagnosis &

-- & 2{,}283 & 1{,}210 & 3{,}493 & 

-- & 

2016--2020 & 

-- & 

2{,}421 & 

-- & 

\cellcolor{gray!20}Internal \\

& 

TN5000~\cite{tn5000} & 

Thyroid nodule diagnosis &

-- & 1{,}426 & 3{,}574 & 5{,}000 & 

China & 

-- & 

GE Logiq E9 \&  S7& 

-- & 

-- & 

\cellcolor{orange!30}External \\

\midrule

\multicolumn{3}{l|}{\textbf{Total (Internal)}} & 

\textbf{1{,}868} & \textbf{4{,}097} & \textbf{4{,}217} & \textbf{10{,}182} & 

-- & -- & -- & 

-- & -- & -- \\

\multicolumn{3}{l|}{\textbf{Total (External)}} & 

\textbf{45} & \textbf{2,073} & \textbf{4,396} & \textbf{6,514} & 

-- & -- & -- & 

-- & -- & -- \\

\multicolumn{3}{l|}{\textbf{Grand Total}} & 

\textbf{1,913} & \textbf{6,170} & \textbf{8,613} & \textbf{16,696} & 

-- & -- & -- & 

-- & -- & -- \\

\bottomrule

\multicolumn{13}{l}{\textsuperscript{a}Age presented as range (mean±s.d.) in years; ``--'' indicates data not available} \\

\multicolumn{13}{l}{\textsuperscript{b}Internal datasets used for training/cross-validation; External datasets reserved for independent validation} \\

\multicolumn{13}{l}{\textsuperscript{c}Class definitions by task. Appendix, normal versus appendicitis. Breast, benign versus malignant, with the BUSI cohort additionally containing a normal class. Carotid, no event} \\

\multicolumn{13}{l}{\phantom{\textsuperscript{c}}versus event. Liver, normal versus fatty liver. Ovary, normal versus abnormal, where the seven original ovarian disease categories were merged into a} \\

\multicolumn{13}{l}{\phantom{\textsuperscript{c}}single abnormality class. Thyroid, benign versus malignant} \\

\end{tabular}

}

\end{table*}

\clearpage

\begin{table*}[!ht]
\small
\centering
\renewcommand{\arraystretch}{1}
\setlength{\tabcolsep}{5pt}
\caption{Comparison of classification accuracy between EvoDiff and other AL methods on the test sets. Values are reported as mean $\pm$ s.d. across the five fold-specific models obtained by five-fold cross-validation. One ResNet-50 was trained independently for each dataset, with no parameter sharing across datasets or organs. Strategy-versus-Random statistical comparisons, including the two-sided paired $t$-tests and the Benjamini--Hochberg multiplicity correction applied across the five strategy-versus-Random comparisons, are reported separately in the Supplementary Information. (\textbf{Bold} represents the best result, \underline{Underline} represents the second best result).}
\resizebox{\textwidth}{!}{
\begin{tabular}{l|l|l|cccccc|c}
\toprule
& & & \multicolumn{7}{c}{\textbf{Cumulative Annotation Budget}} \\
\cmidrule(lr){4-10}
\multirow{-2}{*}{\textbf{Organ}} & \multirow{-2}{*}{\textbf{Dataset}} & \multirow{-2}{*}{\textbf{Method}} & 5\% & 10\% & 20\% & 30\% & 40\% & 50\% & 100\% \\
\midrule
& & \cellcolor[HTML]{F8F8D1}FSL (upper bound) & \cellcolor[HTML]{F8F8D1}\textbf{-} & \cellcolor[HTML]{F8F8D1}\textbf{-} & \cellcolor[HTML]{F8F8D1}\textbf{-} & \cellcolor[HTML]{F8F8D1}\textbf{-} & \cellcolor[HTML]{F8F8D1}\textbf{-} & \cellcolor[HTML]{F8F8D1}\textbf{-} & \cellcolor[HTML]{F8F8D1}0.6404±{\scriptsize 0.0104} \\
& & \cellcolor[HTML]{ECF4FF}Random & \cellcolor[HTML]{ECF4FF}0.4783±{\scriptsize 0.0583} & \cellcolor[HTML]{ECF4FF}0.4870±{\scriptsize 0.1163} & \cellcolor[HTML]{ECF4FF}0.4913±{\scriptsize 0.0668} & \cellcolor[HTML]{ECF4FF}0.5391±{\scriptsize 0.0422} & \cellcolor[HTML]{ECF4FF}0.6000±{\scriptsize 0.0379} & \cellcolor[HTML]{ECF4FF}0.6087±{\scriptsize 0.0550} & \cellcolor[HTML]{ECF4FF}\textbf{-} \\
& & VAAL \cite{vaal} & 0.4635±{\scriptsize 0.0967} & 0.4891±{\scriptsize 0.0763} & 0.4936±{\scriptsize 0.0869} & 0.5279±{\scriptsize 0.0448} & 0.5435±{\scriptsize 0.0967} & 0.5832±{\scriptsize 0.1383} & \textbf{-} \\
& & Core-Set \cite{core} & 0.4845±{\scriptsize 0.0692} & 0.5096±{\scriptsize 0.0149} & 0.4858±{\scriptsize 0.0918} & 0.5392±{\scriptsize 0.0573} & 0.5502±{\scriptsize 0.0270} & 0.5745±{\scriptsize 0.0854} & \textbf{-} \\
& & Max-Entropy \cite{entro} & 0.5016±{\scriptsize 0.0708} & 0.5277±{\scriptsize 0.0645}  & 0.5556±{\scriptsize 0.0794} & 0.5804±{\scriptsize 0.0940} & 0.5932±{\scriptsize 0.0809} & 0.6217±{\scriptsize 0.0927} & \textbf{-} \\
\multirow{-6}{*}{Appendix} & \multirow{-6}{*}{Appendix~\cite{app}} & BALD \cite{bald} & 0.5286±{\scriptsize 0.1077} & 0.5145±{\scriptsize 0.0692} & 0.5769±{\scriptsize 0.0870} & 0.5932±{\scriptsize 0.1383} & 0.5896±{\scriptsize 0.0849} & 0.6292±{\scriptsize 0.0773} & \textbf{-} \\
& & EvoDiff (Ours)& 0.5435±{\scriptsize 0.0018} & 0.5652±{\scriptsize 0.0217} & 0.5764±{\scriptsize 0.0543} & 0.5978±{\scriptsize 0.0109} & 0.6196±{\scriptsize 0.0109} & 0.6373±{\scriptsize 0.0543} & \textbf{-} \\
\midrule
& & \cellcolor[HTML]{F8F8D1}FSL (upper bound) & \cellcolor[HTML]{F8F8D1}\textbf{-} & \cellcolor[HTML]{F8F8D1}\textbf{-} & \cellcolor[HTML]{F8F8D1}\textbf{-} & \cellcolor[HTML]{F8F8D1}\textbf{-} & \cellcolor[HTML]{F8F8D1}\textbf{-} & \cellcolor[HTML]{F8F8D1}\textbf{-} & \cellcolor[HTML]{F8F8D1}0.9424±{\scriptsize 0.0261} \\
& & \cellcolor[HTML]{ECF4FF}Random & \cellcolor[HTML]{ECF4FF}0.6670±{\scriptsize 0.0534} & \cellcolor[HTML]{ECF4FF}0.7515±{\scriptsize 0.0340} & \cellcolor[HTML]{ECF4FF}0.7758±{\scriptsize 0.0294} & \cellcolor[HTML]{ECF4FF}0.8273±{\scriptsize 0.0485} & \cellcolor[HTML]{ECF4FF}0.8545±{\scriptsize 0.0312} & \cellcolor[HTML]{ECF4FF}0.8739±{\scriptsize 0.0321} & \cellcolor[HTML]{ECF4FF}\textbf{-} \\
& & VAAL \cite{vaal} & 0.6596±{\scriptsize 0.1755} & 0.6822±{\scriptsize 0.0891} & 0.7282±{\scriptsize 0.0146} & 0.7892±{\scriptsize 0.0797} & 0.8462±{\scriptsize 0.0969} & 0.8315±{\scriptsize 0.1674} & \textbf{-} \\
& & Core-Set \cite{core} & 0.6555±{\scriptsize 0.0201} & 0.7599±{\scriptsize 0.0955} & 0.8044±{\scriptsize 0.0839} & 0.8326±{\scriptsize 0.0378} & 0.8675±{\scriptsize 0.1437} & 0.8798±{\scriptsize 0.0743} & \textbf{-} \\
& & Max-Entropy \cite{entro} & 0.6779±{\scriptsize 0.0121} & 0.7110±{\scriptsize 0.0362} & 0.7567±{\scriptsize 0.0447} & 0.7941±{\scriptsize 0.0965} & 0.8706±{\scriptsize 0.0741} & 0.8731±{\scriptsize 0.0676} & \textbf{-} \\
& \multirow{-6}{*}{BUSI~\cite{BUSI}} & BALD \cite{bald} & 0.6919±{\scriptsize 0.0911} & 0.7449±{\scriptsize 0.0860} & 0.7592±{\scriptsize 0.0363} & 0.8232±{\scriptsize 0.0871} & 0.8531±{\scriptsize 0.0629} & 0.8792±{\scriptsize 0.0829}  & \textbf{-} \\
& & EvoDiff (Ours)& 0.7287±{\scriptsize 0.0109} & 0.8075±{\scriptsize 0.0323} & 0.8318±{\scriptsize 0.0468} & 0.8460±{\scriptsize 0.0266} & 0.8825±{\scriptsize 0.0396} & 0.9010±{\scriptsize 0.0319} & \textbf{-} \\
\cmidrule(lr){2-10}
& & \cellcolor[HTML]{F8F8D1}FSL (upper bound) & \cellcolor[HTML]{F8F8D1}\textbf{-} & \cellcolor[HTML]{F8F8D1}\textbf{-} & \cellcolor[HTML]{F8F8D1}\textbf{-} & \cellcolor[HTML]{F8F8D1}\textbf{-} & \cellcolor[HTML]{F8F8D1}\textbf{-} & \cellcolor[HTML]{F8F8D1}\textbf{-} & \cellcolor[HTML]{F8F8D1}0.9558±{\scriptsize 0.0123} \\
& & \cellcolor[HTML]{ECF4FF}Random & \cellcolor[HTML]{ECF4FF}0.6695±{\scriptsize 0.0787} & \cellcolor[HTML]{ECF4FF}0.7347±{\scriptsize 0.0739} & \cellcolor[HTML]{ECF4FF}0.7663±{\scriptsize 0.0437} & \cellcolor[HTML]{ECF4FF}0.8084±{\scriptsize 0.0344} & \cellcolor[HTML]{ECF4FF}0.8495±{\scriptsize 0.0297} & \cellcolor[HTML]{ECF4FF}0.8632±{\scriptsize 0.0354} & \cellcolor[HTML]{ECF4FF}\textbf{-} \\
& & VAAL \cite{vaal} & 0.6744±{\scriptsize 0.0252} & 0.6905±{\scriptsize 0.0101} & 0.7118±{\scriptsize 0.0375} & 0.7555±{\scriptsize 0.0284} & 0.7946±{\scriptsize 0.0293} & 0.8238±{\scriptsize 0.0695} & \textbf{-} \\
& & Core-Set \cite{core} & 0.6746±{\scriptsize 0.0736} & 0.6959±{\scriptsize 0.0291} & 0.7444±{\scriptsize 0.0907} & 0.7715±{\scriptsize 0.0793} & 0.8290±{\scriptsize 0.1143}  & 0.8578±{\scriptsize 0.0291} & \textbf{-} \\
& & Max-Entropy \cite{entro} & 0.7088±{\scriptsize 0.0317} & 0.7317±{\scriptsize 0.0727} & 0.7455±{\scriptsize 0.0344} & 0.7859±{\scriptsize 0.0461} & 0.8385±{\scriptsize 0.0226} & 0.8528±{\scriptsize 0.0226} & \textbf{-} \\
\multirow{-19}{*}{} & \multirow{-6}{*}{BUS-BRA~\cite{BUS-BRA}} & BALD \cite{bald} & 0.7041±{\scriptsize 0.0329} & 0.7364±{\scriptsize 0.0270} & 0.7807±{\scriptsize 0.0118} & 0.7939±{\scriptsize 0.0686} & 0.8351±{\scriptsize 0.0979} & 0.8681±{\scriptsize 0.0263} & \textbf{-} \\
& & EvoDiff (Ours)& 0.7261±{\scriptsize 0.0519} & 0.7826±{\scriptsize 0.0619} & 0.8327±{\scriptsize 0.0276} & 0.8517±{\scriptsize 0.0365} & 0.8891±{\scriptsize 0.0461} & 0.9136±{\scriptsize 0.0351} & \textbf{-} \\
\cmidrule(lr){2-10}
& & \cellcolor[HTML]{F8F8D1}FSL (upper bound) & \cellcolor[HTML]{F8F8D1}\textbf{-} & \cellcolor[HTML]{F8F8D1}\textbf{-} & \cellcolor[HTML]{F8F8D1}\textbf{-} & \cellcolor[HTML]{F8F8D1}\textbf{-} & \cellcolor[HTML]{F8F8D1}\textbf{-} & \cellcolor[HTML]{F8F8D1}\textbf{-} & \cellcolor[HTML]{F8F8D1}0.9222±{\scriptsize 0.0272} \\
& & \cellcolor[HTML]{ECF4FF}Random & \cellcolor[HTML]{ECF4FF}0.6556±{\scriptsize 0.0737} & \cellcolor[HTML]{ECF4FF}0.6889±{\scriptsize 0.0567} & \cellcolor[HTML]{ECF4FF}0.7333±{\scriptsize 0.0956} & \cellcolor[HTML]{ECF4FF}0.8111±{\scriptsize 0.0667} & \cellcolor[HTML]{ECF4FF}0.8333±{\scriptsize 0.0609} & \cellcolor[HTML]{ECF4FF}0.8556±{\scriptsize 0.0754} & \cellcolor[HTML]{ECF4FF}\textbf{-} \\
& & VAAL \cite{vaal} & 0.6513±{\scriptsize 0.0969} & 0.6772±{\scriptsize 0.0900} & 0.7355±{\scriptsize 0.0587} & 0.7601±{\scriptsize 0.1068} & 0.7856±{\scriptsize 0.0976} & 0.8318±{\scriptsize 0.0691} & \textbf{-} \\
& & Core-Set \cite{core} & 0.6750±{\scriptsize 0.0236} & 0.7186±{\scriptsize 0.0278} & 0.7537±{\scriptsize 0.0119} & 0.7776±{\scriptsize 0.0531} & 0.7810±{\scriptsize 0.0408} & 0.8409±{\scriptsize 0.0302} & \textbf{-} \\
& & Max-Entropy \cite{entro} & 0.6870±{\scriptsize 0.0664} & 0.6856±{\scriptsize 0.0175}  & 0.7579±{\scriptsize 0.0153}  & 0.7786±{\scriptsize 0.0323} & 0.8116±{\scriptsize 0.0657} & 0.8303±{\scriptsize 0.0231} & \textbf{-} \\
\multirow{-19}{*}{Breast} & \multirow{-6}{*}{UDIAT~\cite{UDIAT}} & BALD \cite{bald} & 0.6714±{\scriptsize 0.0278} & 0.7176±{\scriptsize 0.0415} & 0.7930±{\scriptsize 0.0722} & 0.7511±{\scriptsize 0.0762} & 0.7731±{\scriptsize 0.0831} & 0.8334±{\scriptsize 0.0882} & \textbf{-} \\
& & EvoDiff (Ours) & 0.7061±{\scriptsize 0.0761} & 0.7340±{\scriptsize 0.0129} & 0.7851±{\scriptsize 0.0761} & 0.8454±{\scriptsize 0.0718} & 0.8684±{\scriptsize 0.0649} & 0.9074±{\scriptsize 0.0543} & \textbf{-} \\
\midrule
& & \cellcolor[HTML]{F8F8D1}FSL (upper bound) & \cellcolor[HTML]{F8F8D1}\textbf{-} & \cellcolor[HTML]{F8F8D1}\textbf{-} & \cellcolor[HTML]{F8F8D1}\textbf{-} & \cellcolor[HTML]{F8F8D1}\textbf{-} & \cellcolor[HTML]{F8F8D1}\textbf{-} & \cellcolor[HTML]{F8F8D1}\textbf{-} & \cellcolor[HTML]{F8F8D1}0.8906±{\scriptsize 0.0451} \\
& & \cellcolor[HTML]{ECF4FF}Random & \cellcolor[HTML]{ECF4FF}0.6705±{\scriptsize 0.0323} & \cellcolor[HTML]{ECF4FF}0.7122±{\scriptsize 0.0457} & \cellcolor[HTML]{ECF4FF}0.7410±{\scriptsize 0.0561} & \cellcolor[HTML]{ECF4FF}0.7899±{\scriptsize 0.0411} & \cellcolor[HTML]{ECF4FF}0.8029±{\scriptsize 0.0293} & \cellcolor[HTML]{ECF4FF}0.8072±{\scriptsize 0.0537} & \cellcolor[HTML]{ECF4FF}\textbf{-} \\
& & VAAL \cite{vaal} & 0.7048±{\scriptsize 0.0451} & 0.7257±{\scriptsize 0.0528} & 0.7113±{\scriptsize 0.0151} & 0.7657±{\scriptsize 0.0141} & 0.7852±{\scriptsize 0.0955} & 0.7730±{\scriptsize 0.0411} & \textbf{-} \\
& & Core-Set \cite{core} & 0.7241±{\scriptsize 0.0306} & 0.7317±{\scriptsize 0.0204} & 0.7413±{\scriptsize 0.0558} & 0.7799±{\scriptsize 0.0496} & 0.8060±{\scriptsize 0.0125} & 0.8247±{\scriptsize 0.0271} & \textbf{-} \\
& & Max-Entropy \cite{entro} & 0.7212±{\scriptsize 0.0521} & 0.7462±{\scriptsize 0.0321} & 0.7916±{\scriptsize 0.0392} & 0.8186±{\scriptsize 0.0208} & 0.8362±{\scriptsize 0.0924} & 0.8288±{\scriptsize 0.0671} & \textbf{-} \\
\multirow{-6}{*}{Carotid} & \multirow{-6}{*}{CUBS~\cite{CUBS}} & BALD \cite{bald} & 0.7308±{\scriptsize 0.0632} & 0.7403±{\scriptsize 0.0416} & 0.7756±{\scriptsize 0.0797} & 0.8023±{\scriptsize 0.0832} & 0.8262±{\scriptsize 0.0802} & 0.8392±{\scriptsize 0.0389} & \textbf{-} \\
& & EvoDiff (Ours) & 0.7518±{\scriptsize 0.0396} & 0.7951±{\scriptsize 0.0045} & 0.8218±{\scriptsize 0.0184} & 0.8442±{\scriptsize 0.0254} & 0.8593±{\scriptsize 0.0247} & 0.8770±{\scriptsize 0.0450} & \textbf{-} \\
\midrule
& & \cellcolor[HTML]{F8F8D1}FSL (upper bound) & \cellcolor[HTML]{F8F8D1}\textbf{-} & \cellcolor[HTML]{F8F8D1}\textbf{-} & \cellcolor[HTML]{F8F8D1}\textbf{-} & \cellcolor[HTML]{F8F8D1}\textbf{-} & \cellcolor[HTML]{F8F8D1}\textbf{-} & \cellcolor[HTML]{F8F8D1}\textbf{-} & \cellcolor[HTML]{F8F8D1}0.9741±{\scriptsize 0.0361} \\
& & \cellcolor[HTML]{ECF4FF}Random & \cellcolor[HTML]{ECF4FF}0.8251±{\scriptsize 0.0632} & \cellcolor[HTML]{ECF4FF}0.8469±{\scriptsize 0.0766} & \cellcolor[HTML]{ECF4FF}0.8753±{\scriptsize  0.0810} & \cellcolor[HTML]{ECF4FF}0.9157±{\scriptsize 0.0688} & \cellcolor[HTML]{ECF4FF}0.9163±{\scriptsize 0.0641} & \cellcolor[HTML]{ECF4FF}0.9321±{\scriptsize 0.0389} & \cellcolor[HTML]{ECF4FF}\textbf{-} \\
& & VAAL \cite{vaal} & 0.8042±{\scriptsize 0.0713} & 0.8196±{\scriptsize 0.0491} & 0.8417±{\scriptsize 0.0922} & 0.8763±{\scriptsize 0.0859} & 0.8901±{\scriptsize 0.0814} & 0.9036±{\scriptsize 0.0762} & \textbf{-} \\
& & Core-Set \cite{core} & 0.8145±{\scriptsize 0.0449} & 0.8294±{\scriptsize 0.0360} & 0.8490±{\scriptsize 0.0797} & 0.8870±{\scriptsize 0.0831} & 0.8510±{\scriptsize 0.0460}  & 0.8973±{\scriptsize 0.0531} & \textbf{-} \\
& & Max-Entropy \cite{entro} & 0.8258±{\scriptsize 0.0521} & 0.8566±{\scriptsize 0.0720} & 0.8934±{\scriptsize 0.0913} & 0.9011±{\scriptsize 0.0442} & 0.9481±{\scriptsize 0.0770} & 0.9450±{\scriptsize 0.0749} & \textbf{-} \\
\multirow{-6}{*}{Liver} & \multirow{-6}{*}{Fatty-Liver~\cite{liver}} & BALD \cite{bald} & 0.8445±{\scriptsize 0.0429} & 0.8641±{\scriptsize 0.0455} & 0.8987±{\scriptsize 0.0421} &  0.9371±{\scriptsize 0.0870} & 0.9572±{\scriptsize 0.0850} &  0.9620±{\scriptsize 0.0475} & \textbf{-} \\
& & EvoDiff (Ours) & 0.8563±{\scriptsize 0.0208} & 0.8815±{\scriptsize 0.0336} & 0.9240±{\scriptsize 0.0491} & 0.9552±{\scriptsize 0.0767} & 0.9608±{\scriptsize 0.0232} & 0.9724±{\scriptsize 0.0389} & \textbf{-} \\
\midrule
& & \cellcolor[HTML]{F8F8D1}FSL (upper bound) & \cellcolor[HTML]{F8F8D1}\textbf{-} & \cellcolor[HTML]{F8F8D1}\textbf{-} & \cellcolor[HTML]{F8F8D1}\textbf{-} & \cellcolor[HTML]{F8F8D1}\textbf{-} & \cellcolor[HTML]{F8F8D1}\textbf{-} & \cellcolor[HTML]{F8F8D1}\textbf{-} & \cellcolor[HTML]{F8F8D1}0.9410±{\scriptsize 0.0212} \\
& & \cellcolor[HTML]{ECF4FF}Random & \cellcolor[HTML]{ECF4FF}0.7256±{\scriptsize 0.0426} & \cellcolor[HTML]{ECF4FF}0.7564±{\scriptsize 0.0481} & \cellcolor[HTML]{ECF4FF}0.7987±{\scriptsize 0.0119} & \cellcolor[HTML]{ECF4FF}0.8359±{\scriptsize 0.0359} & \cellcolor[HTML]{ECF4FF}0.8679±{\scriptsize 0.0323} & \cellcolor[HTML]{ECF4FF}0.8885±{\scriptsize 0.0188} & \cellcolor[HTML]{ECF4FF}\textbf{-} \\
& & VAAL \cite{vaal} & 0.7273±{\scriptsize 0.0732} & 0.7393±{\scriptsize 0.0866} & 0.7877±{\scriptsize 0.0844} & 0.8195±{\scriptsize 0.0399}  & 0.8442±{\scriptsize 0.0925} & 0.8704±{\scriptsize 0.0737} & \textbf{-} \\
& & Core-Set \cite{core} & 0.7219±{\scriptsize 0.0864} & 0.7630±{\scriptsize 0.0353} & 0.8168±{\scriptsize 0.0850} & 0.8583±{\scriptsize 0.0570} & 0.8687±{\scriptsize 0.0658} & 0.8900±{\scriptsize 0.0803} & \textbf{-} \\
& & Max-Entropy \cite{entro} & 0.7530±{\scriptsize 0.0494} & 0.7747±{\scriptsize 0.0621} & 0.8372±{\scriptsize 0.0693} & 0.8770±{\scriptsize 0.0757} & 0.8867±{\scriptsize 0.0982} & 0.9015±{\scriptsize 0.0888} & \textbf{-} \\
\multirow{-6}{*}{Ovary} & \multirow{-6}{*}{MMOTU~\cite{mmotu}} & BALD \cite{bald} & 0.7450±{\scriptsize 0.0336} & 0.7630±{\scriptsize 0.0861} & 0.8248±{\scriptsize 0.0642} & 0.8475±{\scriptsize 0.0684} & 0.8550±{\scriptsize 0.0212} & 0.8931±{\scriptsize 0.0831} & \textbf{-} \\
& & EvoDiff (Ours)& 0.7513±{\scriptsize 0.0558} & 0.8095±{\scriptsize 0.0496} & 0.8320±{\scriptsize 0.0192} & 0.8682±{\scriptsize 0.0290} & 0.9057±{\scriptsize 0.0528} & 0.9222±{\scriptsize 0.0475} & \textbf{-} \\
\midrule
& & \cellcolor[HTML]{F8F8D1}FSL (upper bound) & \cellcolor[HTML]{F8F8D1}\textbf{-} & \cellcolor[HTML]{F8F8D1}\textbf{-} & \cellcolor[HTML]{F8F8D1}\textbf{-} & \cellcolor[HTML]{F8F8D1}\textbf{-} & \cellcolor[HTML]{F8F8D1}\textbf{-} & \cellcolor[HTML]{F8F8D1}\textbf{-} & \cellcolor[HTML]{F8F8D1}0.8937±{\scriptsize 0.0157} \\
& & \cellcolor[HTML]{ECF4FF}Random & \cellcolor[HTML]{ECF4FF}0.6029±{\scriptsize 0.0261} & \cellcolor[HTML]{ECF4FF}0.6263±{\scriptsize 0.0209} & \cellcolor[HTML]{ECF4FF}0.7109±{\scriptsize 0.0119} & \cellcolor[HTML]{ECF4FF}0.7377±{\scriptsize 0.0314} & \cellcolor[HTML]{ECF4FF}0.7617±{\scriptsize 0.0067} & \cellcolor[HTML]{ECF4FF}0.7749±{\scriptsize 0.0038} & \cellcolor[HTML]{ECF4FF}\textbf{-} \\
& & VAAL \cite{vaal} & 0.5980±{\scriptsize 0.0758} & 0.6018±{\scriptsize 0.2120} & 0.6372±{\scriptsize 0.0840} &  0.7075±{\scriptsize 0.0323} & 0.7573±{\scriptsize 0.0483} & 0.7952±{\scriptsize 0.0694} & \textbf{-} \\
& & Core-Set \cite{core} & 0.6406±{\scriptsize 0.0313} & 0.6625±{\scriptsize 0.2069} & 0.6896±{\scriptsize 0.0484} & 0.7509±{\scriptsize 0.0322} & 0.7670±{\scriptsize 0.0544} & 0.8082±{\scriptsize 0.0576} & \textbf{-} \\
& & Max-Entropy \cite{entro} & 0.6527±{\scriptsize 0.0206} & 0.7023±{\scriptsize 0.0590} & 0.7436±{\scriptsize 0.0509} & 0.7682±{\scriptsize 0.0726} & 0.7729±{\scriptsize 0.0582} & 0.7853±{\scriptsize 0.0514} & \textbf{-} \\
\multirow{-6}{*}{Thyroid} & \multirow{-6}{*}{TN3K~\cite{TN3K}} & BALD \cite{bald} & 0.6334±{\scriptsize 0.0381} & 0.6721±{\scriptsize 0.0839} & 0.7569±{\scriptsize 0.0550} & 0.7877±{\scriptsize 0.0514} & 0.7816±{\scriptsize 0.0710} & 0.8218±{\scriptsize 0.0468} & \textbf{-} \\
& & EvoDiff (Ours) & 0.6730±{\scriptsize 0.0011} & 0.7048±{\scriptsize 0.0451} & 0.7793±{\scriptsize 0.0159} & 0.8126±{\scriptsize 0.0382} & 0.8386±{\scriptsize 0.0208} & 0.8656±{\scriptsize 0.0597} & \textbf{-} \\
\bottomrule
\end{tabular}
}
\label{c1}
\end{table*}

\clearpage

\begin{table*}[!ht]
\centering
\caption{Ablation study on the Appendix dataset \cite{app} under 10\%, 30\%, and 50\% cumulative annotation budgets. Each reduced configuration was compared with the complete EvoDiff framework using two-sided paired $t$-tests, with the five cross-validation folds used as matched pairs. The Upper Bound column reports fully supervised learning on the complete labeled set.}
\resizebox{\textwidth}{!}{
\begin{tabular}{ccc|ccc|ccc|ccc|ccc}
\toprule
\multirow{2}{*}{\textbf{IG Guidance}} & \multirow{2}{*}{\textbf{LOEA}} & \multirow{2}{*}{\textbf{CL}} &
\multicolumn{3}{c|}{\textbf{Annotation Ratio=10\%}} &
\multicolumn{3}{c|}{\textbf{Annotation Ratio=30\%}} &
\multicolumn{3}{c|}{\textbf{Annotation Ratio=50\%}} &
\multicolumn{3}{c}{\textbf{Upper Bound}} \\
\cmidrule{4-15}
& & &
\textbf{ACC} & \textbf{F1-score} & \textbf{AUC} &
\textbf{ACC} & \textbf{F1-score} & \textbf{AUC} &
\textbf{ACC} & \textbf{F1-score} & \textbf{AUC} &
\textbf{ACC} & \textbf{F1-score} & \textbf{AUC} \\
\midrule
$\checkmark$ & $\times$ & $\times$ & 0.5386 & 0.4938 & 0.5305 & 0.5503 & 0.5265 & 0.5627 & 0.5587 & 0.5294 & 0.5846 & -- & -- & -- \\
$\checkmark$ & $\checkmark$ & $\times$ & 0.5383 & 0.4956 & 0.5347 & 0.5479 & 0.5294 & 0.5518 & 0.5592 & 0.5319 & 0.5731 & -- & -- & -- \\
$\checkmark$ & $\checkmark$ & $\checkmark$ & \textbf{0.5652} & \textbf{0.5374} & \textbf{0.5960} & \textbf{0.5978} & \textbf{0.5761} & \textbf{0.6153} & \textbf{0.6373} & \textbf{0.6095} & \textbf{0.6474} & {\cellcolor{gray!15}0.6404} & {\cellcolor{gray!15}0.7032} & {\cellcolor{gray!15}0.6390} \\
\bottomrule
\end{tabular}
}
\label{ab1}
\end{table*}

\clearpage

\begin{table*}[!ht]
\centering
\caption{Ablation study on the TN3K dataset \cite{TN3K} under 10\%, 30\%, and 50\% cumulative annotation budgets. Each reduced configuration was compared with the complete EvoDiff framework using two-sided paired $t$-tests, with the five cross-validation folds used as matched pairs. The Upper Bound column reports fully supervised learning on the complete labeled set.}
\resizebox{\textwidth}{!}{
\begin{tabular}{ccc|ccc|ccc|ccc|ccc}
\toprule
\multirow{2}{*}{\textbf{IG Guidance}} & \multirow{2}{*}{\textbf{LOEA}} & \multirow{2}{*}{\textbf{CL}} &
\multicolumn{3}{c|}{\textbf{Annotation Ratio=10\%}} &
\multicolumn{3}{c|}{\textbf{Annotation Ratio=30\%}} &
\multicolumn{3}{c|}{\textbf{Annotation Ratio=50\%}} &
\multicolumn{3}{c}{\textbf{Upper Bound}} \\
\cmidrule{4-15}
&  &  &
\textbf{ACC} & \textbf{F1-score} & \textbf{AUC} &
\textbf{ACC} & \textbf{F1-score} & \textbf{AUC} &
\textbf{ACC} & \textbf{F1-score} & \textbf{AUC} &
\textbf{ACC} & \textbf{F1-score} & \textbf{AUC} \\
\midrule
$\checkmark$ & $\times$ & $\times$ & 0.6592 & 0.6051 & 0.6560 & 0.7531 & 0.6859 & 0.7803 & 0.8043 & 0.7275 & 0.8092 & -- & -- & -- \\
$\checkmark$ & $\checkmark$ & $\times$ & 0.6631 & 0.6336 & 0.6958 & 0.7894 & 0.7047 & 0.7851 & 0.8383 & 0.7552 & 0.8365 & -- & -- & -- \\
$\checkmark$ & $\checkmark$ & $\checkmark$ & \textbf{0.7048} & \textbf{0.6729} & \textbf{0.7125} & \textbf{0.8126} & \textbf{0.7403} & \textbf{0.8289} & \textbf{0.8656} & \textbf{0.7849} & \textbf{0.8519} & {\cellcolor{gray!15}0.8937} & {\cellcolor{gray!15}0.7621} & {\cellcolor{gray!15}0.8927} \\
\bottomrule
\end{tabular}
}
\label{ab2}
\end{table*}

% ==================== Supplementary Information ====================
% Finish all main-text floats and start the supplement on a new page.
\clearpage
% \nolinenumbers
\setcounter{table}{0}
\setcounter{figure}{0}
\renewcommand{\thetable}{S\arabic{table}}
\renewcommand{\thefigure}{S\arabic{figure}}
% Keep hyperlink destinations distinct after restarting the counters.
\ifdefined\theHtable
  \renewcommand{\theHtable}{supp.\arabic{table}}
\fi
\ifdefined\theHfigure
  \renewcommand{\theHfigure}{supp.\arabic{figure}}
\fi
% Both parts use the single shared bibliography in the main text.

\section*{Supplementary Information}
\subsection*{Generative Active Learning for Label-Efficient Pan-Organ Ultrasound Diagnosis}

%%%%%  S1  ResNet-50 完整指标  %%%%%

\begin{sidewaystable*}[!ht]
\small
\centering
\renewcommand{\arraystretch}{1.2}
\setlength{\tabcolsep}{4pt}
\caption{Comparison of ResNet-50 classification metrics between EvoDiff and other AL methods on the test sets. Values are reported as mean $\pm$ s.d. over five folds. Statistical differences between each active learning strategy and Random sampling were assessed using two-sided paired t-tests on fold-wise AUC, pairing the k-th cross-validation fold of a strategy with the k-th fold of Random. The reported $q$-values are Benjamini--Hochberg-adjusted $p$-values for AUC, with correction applied across the five strategy-versus-Random comparisons that share a given dataset and annotation budget. The appended (+) or ($-$) indicates the sign of the mean fold-wise AUC difference between the strategy and Random. (\textbf{Bold} represents the best result, and \underline{Underline} represents the second best result).}
\textbf{\label{resnet_landscape_metrics}}
\resizebox{\linewidth}{!}{
\begin{tabular}{l|l|c|cccccccc|c}
\toprule
\textbf{Dataset} & \textbf{Method} & \textbf{Annotation Budget} & \textbf{Acc} & \textbf{bAcc} & \textbf{Sens} & \textbf{Spec} & \textbf{PPV} & \textbf{NPV} & \textbf{F1} & \textbf{AUC} & \textbf{$q$-value} \\
\midrule
\multirow{19}{*}{Appendix~\cite{app}} & \cellcolor[HTML]{F8F8D1}FSL (upper bound) & \cellcolor[HTML]{F8F8D1}100\% & \cellcolor[HTML]{F8F8D1}0.6404±{\scriptsize 0.0104} & \cellcolor[HTML]{F8F8D1}0.5024±{\scriptsize 0.0745} & \cellcolor[HTML]{F8F8D1}0.7931±{\scriptsize 0.0617} & \cellcolor[HTML]{F8F8D1}0.2118±{\scriptsize 0.1091} & \cellcolor[HTML]{F8F8D1}0.6323±{\scriptsize 0.0450} & \cellcolor[HTML]{F8F8D1}0.3697±{\scriptsize 0.1543} & \cellcolor[HTML]{F8F8D1}0.7032±{\scriptsize 0.0490} & \cellcolor[HTML]{F8F8D1}0.6390±{\scriptsize 0.0948} & - \\
 & \cellcolor[HTML]{ECF4FF}Random & \cellcolor[HTML]{ECF4FF}10\% & \cellcolor[HTML]{ECF4FF}0.4870±{\scriptsize 0.1163} & \cellcolor[HTML]{ECF4FF}\underline{0.4933±{\scriptsize 0.0327}} & \cellcolor[HTML]{ECF4FF}\underline{0.8690±{\scriptsize 0.0593}} & \cellcolor[HTML]{ECF4FF}0.1176±{\scriptsize 0.0984} & \cellcolor[HTML]{ECF4FF}\underline{0.6277±{\scriptsize 0.0173}} & \cellcolor[HTML]{ECF4FF}0.3033±{\scriptsize 0.1809} & \cellcolor[HTML]{ECF4FF}0.7277±{\scriptsize 0.0218} & \cellcolor[HTML]{ECF4FF}0.5154±{\scriptsize 0.0383} & - \\
 & VAAL~\cite{vaal} & 10\% & 0.4891±{\scriptsize 0.0763} & 0.4856±{\scriptsize 0.0247} & \textbf{0.9241±{\scriptsize 0.0258}} & 0.0471±{\scriptsize 0.0440} & 0.6233±{\scriptsize 0.0122} & 0.2467±{\scriptsize 0.2083} & \textbf{0.7444±{\scriptsize 0.0146}} & 0.5162±{\scriptsize 0.0561} & 0.0940 (+) \\
 & Core-Set~\cite{core} & 10\% & 0.5096±{\scriptsize 0.0149} & 0.4680±{\scriptsize 0.0265} & \textbf{0.9241±{\scriptsize 0.0457}} & 0.0118±{\scriptsize 0.0235} & 0.6144±{\scriptsize 0.0134} & 0.0667±{\scriptsize 0.1333} & \underline{0.7379±{\scriptsize 0.0236}} & 0.5314±{\scriptsize 0.0506} & 0.0940 (+) \\
 & Max-Entropy~\cite{entro} & 10\% & \underline{0.5277±{\scriptsize 0.0645}} & 0.4306±{\scriptsize 0.0521} & 0.6966±{\scriptsize 0.1623} & 0.1647±{\scriptsize 0.1552} & 0.5839±{\scriptsize 0.0443} & 0.1682±{\scriptsize 0.1474} & 0.6288±{\scriptsize 0.0867} & 0.5406±{\scriptsize 0.0297} & 0.0171 (+) \\
 & BALD~\cite{bald} & 10\% & 0.5145±{\scriptsize 0.0692} & 0.4915±{\scriptsize 0.0410} & 0.7241±{\scriptsize 0.2256} & \underline{0.2588±{\scriptsize 0.1769}} & 0.6187±{\scriptsize 0.0268} & \textbf{0.4672±{\scriptsize 0.2726}} & 0.6523±{\scriptsize 0.1244} & \underline{0.5431±{\scriptsize 0.0648}} & 0.0982 (+) \\
 & EvoDiff (Ours) & 10\% & \textbf{0.5652±{\scriptsize 0.0217}} & \textbf{0.5136±{\scriptsize 0.0443}} & 0.7448±{\scriptsize 0.1739} & \textbf{0.2824±{\scriptsize 0.2275}} & \textbf{0.6437±{\scriptsize 0.0277}} & \underline{0.3058±{\scriptsize 0.1832}} & 0.6780±{\scriptsize 0.0783} & \textbf{0.5962±{\scriptsize 0.0798}} & 0.0940 (+) \\
\cmidrule(lr){2-12}
 & \cellcolor[HTML]{ECF4FF}Random & \cellcolor[HTML]{ECF4FF}20\% & \cellcolor[HTML]{ECF4FF}0.4913±{\scriptsize 0.0668} & \cellcolor[HTML]{ECF4FF}\textbf{0.4850±{\scriptsize 0.0198}} & \cellcolor[HTML]{ECF4FF}\underline{0.8759±{\scriptsize 0.1206}} & \cellcolor[HTML]{ECF4FF}\underline{0.0941±{\scriptsize 0.1372}} & \cellcolor[HTML]{ECF4FF}\textbf{0.6231±{\scriptsize 0.0100}} & \cellcolor[HTML]{ECF4FF}0.1417±{\scriptsize 0.1740} & \cellcolor[HTML]{ECF4FF}\underline{0.7242±{\scriptsize 0.0445}} & \cellcolor[HTML]{ECF4FF}0.5232±{\scriptsize 0.0810} & - \\
 & VAAL~\cite{vaal} & 20\% & 0.4936±{\scriptsize 0.0869} & 0.4702±{\scriptsize 0.0460} & 0.8345±{\scriptsize 0.1894} & \textbf{0.1059±{\scriptsize 0.1141}} & 0.6092±{\scriptsize 0.0302} & \textbf{0.3017±{\scriptsize 0.3681}} & 0.6980±{\scriptsize 0.0901} & 0.5128±{\scriptsize 0.0741} & 0.0981 ($-$) \\
 & Core-Set~\cite{core} & 20\% & 0.4858±{\scriptsize 0.0918} & \underline{0.4832±{\scriptsize 0.0148}} & \textbf{0.9310±{\scriptsize 0.0436}} & 0.0353±{\scriptsize 0.0470} & \underline{0.6221±{\scriptsize 0.0071}} & 0.1333±{\scriptsize 0.1633} & \textbf{0.7455±{\scriptsize 0.0161}} & 0.5127±{\scriptsize 0.0732} & 0.0007 ($-$) \\
 & Max-Entropy~\cite{entro} & 20\% & 0.5556±{\scriptsize 0.0794} & 0.4702±{\scriptsize 0.0642} & 0.8345±{\scriptsize 0.0402} & \textbf{0.1059±{\scriptsize 0.1310}} & 0.6161±{\scriptsize 0.0369} & \underline{0.1991±{\scriptsize 0.2009}} & 0.7079±{\scriptsize 0.0289} & 0.5633±{\scriptsize 0.0811} & 0.0981 (+) \\
 & BALD~\cite{bald} & 20\% & \textbf{0.5769±{\scriptsize 0.0870}} & 0.4633±{\scriptsize 0.0309} & 0.8207±{\scriptsize 0.1750} & \textbf{0.1059±{\scriptsize 0.1310}} & 0.6060±{\scriptsize 0.0255} & 0.1705±{\scriptsize 0.1575} & 0.6907±{\scriptsize 0.0885} & \underline{0.5790±{\scriptsize 0.0350}} & 0.0981 (+) \\
 & EvoDiff (Ours) & 20\% & \underline{0.5764±{\scriptsize 0.0543}} & 0.4663±{\scriptsize 0.0302} & 0.8621±{\scriptsize 0.0617} & 0.0706±{\scriptsize 0.0686} & 0.6127±{\scriptsize 0.0158} & 0.1900±{\scriptsize 0.1625} & 0.7155±{\scriptsize 0.0283} & \textbf{0.6067±{\scriptsize 0.0649}} & 0.0981 (+) \\
\cmidrule(lr){2-12}
 & \cellcolor[HTML]{ECF4FF}Random & \cellcolor[HTML]{ECF4FF}30\% & \cellcolor[HTML]{ECF4FF}0.5391±{\scriptsize 0.0422} & \cellcolor[HTML]{ECF4FF}0.4497±{\scriptsize 0.0376} & \cellcolor[HTML]{ECF4FF}0.8759±{\scriptsize 0.0639} & \cellcolor[HTML]{ECF4FF}0.0235±{\scriptsize 0.0470} & \cellcolor[HTML]{ECF4FF}0.6043±{\scriptsize 0.0195} & \cellcolor[HTML]{ECF4FF}0.0667±{\scriptsize 0.1333} & \cellcolor[HTML]{ECF4FF}0.7147±{\scriptsize 0.0326} & \cellcolor[HTML]{ECF4FF}0.5601±{\scriptsize 0.0636} & - \\
 & VAAL~\cite{vaal} & 30\% & 0.5279±{\scriptsize 0.0448} & \textbf{0.5351±{\scriptsize 0.0473}} & \underline{0.9172±{\scriptsize 0.0639}} & \underline{0.1529±{\scriptsize 0.1267}} & \textbf{0.6503±{\scriptsize 0.0250}} & \textbf{0.3600±{\scriptsize 0.3137}} & \textbf{0.7596±{\scriptsize 0.0238}} & 0.5546±{\scriptsize 0.0380} & 0.0678 ($-$) \\
 & Core-Set~\cite{core} & 30\% & 0.5392±{\scriptsize 0.0573} & 0.4797±{\scriptsize 0.0188} & \textbf{0.9241±{\scriptsize 0.0338}} & 0.0353±{\scriptsize 0.0470} & 0.6205±{\scriptsize 0.0094} & 0.1300±{\scriptsize 0.1661} & \underline{0.7422±{\scriptsize 0.0136}} & 0.5586±{\scriptsize 0.0542} & 0.0678 ($-$) \\
 & Max-Entropy~\cite{entro} & 30\% & 0.5804±{\scriptsize 0.0940} & \underline{0.4858±{\scriptsize 0.0340}} & 0.8069±{\scriptsize 0.0941} & \textbf{0.1647±{\scriptsize 0.0440}} & \underline{0.6206±{\scriptsize 0.0223}} & \underline{0.3486±{\scriptsize 0.0688}} & 0.7002±{\scriptsize 0.0512} & 0.5974±{\scriptsize 0.1058} & 0.0852 (+) \\
 & BALD~\cite{bald} & 30\% & \underline{0.5932±{\scriptsize 0.1383}} & 0.4797±{\scriptsize 0.0250} & \textbf{0.9241±{\scriptsize 0.0552}} & 0.0353±{\scriptsize 0.0470} & 0.6202±{\scriptsize 0.0128} & 0.1333±{\scriptsize 0.1633} & 0.7418±{\scriptsize 0.0245} & \underline{0.6018±{\scriptsize 0.0505}} & 0.0852 (+) \\
 & EvoDiff (Ours) & 30\% & \textbf{0.5978±{\scriptsize 0.0109}} & 0.4643±{\scriptsize 0.0322} & 0.8345±{\scriptsize 0.0736} & 0.0941±{\scriptsize 0.0288} & 0.6100±{\scriptsize 0.0201} & 0.2597±{\scriptsize 0.0641} & 0.7042±{\scriptsize 0.0400} & \textbf{0.6152±{\scriptsize 0.0509}} & 0.0678 (+) \\
\midrule
\multirow{19}{*}{BUSI~\cite{BUSI}} & \cellcolor[HTML]{F8F8D1}FSL (upper bound) & \cellcolor[HTML]{F8F8D1}100\% & \cellcolor[HTML]{F8F8D1}0.9424±{\scriptsize 0.0261} & \cellcolor[HTML]{F8F8D1}0.9483±{\scriptsize 0.0517} & \cellcolor[HTML]{F8F8D1}0.9143±{\scriptsize 0.1143} & \cellcolor[HTML]{F8F8D1}0.9822±{\scriptsize 0.0259} & \cellcolor[HTML]{F8F8D1}0.9659±{\scriptsize 0.0487} & \cellcolor[HTML]{F8F8D1}0.9640±{\scriptsize 0.0474} & \cellcolor[HTML]{F8F8D1}0.9333±{\scriptsize 0.0573} & \cellcolor[HTML]{F8F8D1}0.9867±{\scriptsize 0.0239} & - \\
 & \cellcolor[HTML]{ECF4FF}Random & \cellcolor[HTML]{ECF4FF}10\% & \cellcolor[HTML]{ECF4FF}0.7515±{\scriptsize 0.0340} & \cellcolor[HTML]{ECF4FF}0.6441±{\scriptsize 0.0958} & \cellcolor[HTML]{ECF4FF}0.5905±{\scriptsize 0.2670} & \cellcolor[HTML]{ECF4FF}0.6978±{\scriptsize 0.2635} & \cellcolor[HTML]{ECF4FF}0.5599±{\scriptsize 0.1481} & \cellcolor[HTML]{ECF4FF}0.7932±{\scriptsize 0.0636} & \cellcolor[HTML]{ECF4FF}0.5040±{\scriptsize 0.1604} & \cellcolor[HTML]{ECF4FF}0.7967±{\scriptsize 0.0309} & - \\
 & VAAL~\cite{vaal} & 10\% & 0.6822±{\scriptsize 0.0891} & 0.7156±{\scriptsize 0.0692} & 0.6000±{\scriptsize 0.1895} & 0.8311±{\scriptsize 0.1156} & 0.6572±{\scriptsize 0.1144} & 0.8258±{\scriptsize 0.0612} & 0.6006±{\scriptsize 0.1066} & 0.7566±{\scriptsize 0.0292} & 0.0339 ($-$) \\
 & Core-Set~\cite{core} & 10\% & \underline{0.7599±{\scriptsize 0.0955}} & \textbf{0.7584±{\scriptsize 0.0312}} & \textbf{0.7524±{\scriptsize 0.0923}} & 0.7644±{\scriptsize 0.1121} & 0.6209±{\scriptsize 0.0991} & \textbf{0.8729±{\scriptsize 0.0308}} & \textbf{0.6697±{\scriptsize 0.0450}} & 0.7893±{\scriptsize 0.0188} & 0.0245 ($-$) \\
 & Max-Entropy~\cite{entro} & 10\% & 0.7110±{\scriptsize 0.0362} & 0.7181±{\scriptsize 0.0914} & 0.5429±{\scriptsize 0.1966} & \textbf{0.8933±{\scriptsize 0.0295}} & \underline{0.6906±{\scriptsize 0.0799}} & 0.8133±{\scriptsize 0.0636} & 0.5941±{\scriptsize 0.1521} & 0.8083±{\scriptsize 0.0651} & 0.0942 (+) \\
 & BALD~\cite{bald} & 10\% & 0.7449±{\scriptsize 0.0860} & 0.5311±{\scriptsize 0.0352} & 0.2000±{\scriptsize 0.1605} & \underline{0.8622±{\scriptsize 0.0957}} & 0.3273±{\scriptsize 0.1665} & 0.7007±{\scriptsize 0.0231} & 0.2322±{\scriptsize 0.1507} & \underline{0.8125±{\scriptsize 0.0891}} & 0.0051 (+) \\
 & EvoDiff (Ours) & 10\% & \textbf{0.8075±{\scriptsize 0.0323}} & \underline{0.7505±{\scriptsize 0.0556}} & \underline{0.6476±{\scriptsize 0.1494}} & 0.8533±{\scriptsize 0.0959} & \textbf{0.6975±{\scriptsize 0.0880}} & \underline{0.8440±{\scriptsize 0.0479}} & \underline{0.6550±{\scriptsize 0.0785}} & \textbf{0.8355±{\scriptsize 0.0535}} & 0.0229 (+) \\
\cmidrule(lr){2-12}
 & \cellcolor[HTML]{ECF4FF}Random & \cellcolor[HTML]{ECF4FF}20\% & \cellcolor[HTML]{ECF4FF}0.7758±{\scriptsize 0.0294} & \cellcolor[HTML]{ECF4FF}0.6978±{\scriptsize 0.0893} & \cellcolor[HTML]{ECF4FF}0.4667±{\scriptsize 0.2379} & \cellcolor[HTML]{ECF4FF}\textbf{0.9289±{\scriptsize 0.0708}} & \cellcolor[HTML]{ECF4FF}\textbf{0.8252±{\scriptsize 0.1180}} & \cellcolor[HTML]{ECF4FF}0.7981±{\scriptsize 0.0634} & \cellcolor[HTML]{ECF4FF}0.5353±{\scriptsize 0.1987} & \cellcolor[HTML]{ECF4FF}0.8036±{\scriptsize 0.0259} & - \\
 & VAAL~\cite{vaal} & 20\% & 0.7282±{\scriptsize 0.0146} & 0.7175±{\scriptsize 0.0343} & 0.5238±{\scriptsize 0.0602} & \underline{0.9111±{\scriptsize 0.0526}} & 0.7529±{\scriptsize 0.1103} & 0.8041±{\scriptsize 0.0197} & 0.6113±{\scriptsize 0.0522} & 0.7723±{\scriptsize 0.0252} & 0.0869 ($-$) \\
 & Core-Set~\cite{core} & 20\% & \underline{0.8044±{\scriptsize 0.0839}} & 0.7267±{\scriptsize 0.0483} & \textbf{0.8000±{\scriptsize 0.0356}} & 0.6533±{\scriptsize 0.1262} & 0.5359±{\scriptsize 0.0897} & \underline{0.8743±{\scriptsize 0.0123}} & 0.6354±{\scriptsize 0.0530} & 0.7830±{\scriptsize 0.0354} & 0.0709 ($-$) \\
 & Max-Entropy~\cite{entro} & 20\% & 0.7567±{\scriptsize 0.0447} & \underline{0.7657±{\scriptsize 0.0186}} & \underline{0.6381±{\scriptsize 0.0883}} & 0.8933±{\scriptsize 0.0515} & 0.7519±{\scriptsize 0.0686} & 0.8432±{\scriptsize 0.0257} & \underline{0.6813±{\scriptsize 0.0226}} & 0.8034±{\scriptsize 0.0196} & 0.0179 ($-$) \\
 & BALD~\cite{bald} & 20\% & 0.7592±{\scriptsize 0.0363} & 0.7613±{\scriptsize 0.0703} & \underline{0.6381±{\scriptsize 0.2246}} & 0.8844±{\scriptsize 0.1092} & 0.7780±{\scriptsize 0.1216} & 0.8540±{\scriptsize 0.0753} & 0.6583±{\scriptsize 0.1074} & \underline{0.8293±{\scriptsize 0.0226}} & 0.0081 (+) \\
 & EvoDiff (Ours) & 20\% & \textbf{0.8318±{\scriptsize 0.0468}} & \textbf{0.8444±{\scriptsize 0.0448}} & \textbf{0.8000±{\scriptsize 0.1182}} & 0.8889±{\scriptsize 0.0544} & \underline{0.7829±{\scriptsize 0.0707}} & \textbf{0.9094±{\scriptsize 0.0499}} & \textbf{0.7826±{\scriptsize 0.0466}} & \textbf{0.8563±{\scriptsize 0.0331}} & 0.0012 (+) \\
\cmidrule(lr){2-12}
 & \cellcolor[HTML]{ECF4FF}Random & \cellcolor[HTML]{ECF4FF}30\% & \cellcolor[HTML]{ECF4FF}0.8273±{\scriptsize 0.0485} & \cellcolor[HTML]{ECF4FF}0.7549±{\scriptsize 0.0284} & \cellcolor[HTML]{ECF4FF}0.5810±{\scriptsize 0.0555} & \cellcolor[HTML]{ECF4FF}\textbf{0.9289±{\scriptsize 0.0218}} & \cellcolor[HTML]{ECF4FF}\textbf{0.7948±{\scriptsize 0.0527}} & \cellcolor[HTML]{ECF4FF}0.8265±{\scriptsize 0.0191} & \cellcolor[HTML]{ECF4FF}0.6694±{\scriptsize 0.0435} & \cellcolor[HTML]{ECF4FF}\underline{0.8600±{\scriptsize 0.0283}} & - \\
 & VAAL~\cite{vaal} & 30\% & 0.7892±{\scriptsize 0.0797} & 0.7349±{\scriptsize 0.0485} & 0.5810±{\scriptsize 0.1060} & \underline{0.8889±{\scriptsize 0.0445}} & 0.7184±{\scriptsize 0.0801} & 0.8216±{\scriptsize 0.0368} & 0.6351±{\scriptsize 0.0692} & 0.8157±{\scriptsize 0.0310} & 0.0947 ($-$) \\
 & Core-Set~\cite{core} & 30\% & \underline{0.8326±{\scriptsize 0.0378}} & \underline{0.7784±{\scriptsize 0.0319}} & \underline{0.8190±{\scriptsize 0.0819}} & 0.7378±{\scriptsize 0.1008} & 0.6097±{\scriptsize 0.0836} & \textbf{0.9008±{\scriptsize 0.0299}} & 0.6905±{\scriptsize 0.0388} & 0.8067±{\scriptsize 0.0251} & 0.0947 ($-$) \\
 & Max-Entropy~\cite{entro} & 30\% & 0.7941±{\scriptsize 0.0965} & 0.7781±{\scriptsize 0.0312} & 0.7429±{\scriptsize 0.0883} & 0.8133±{\scriptsize 0.0387} & 0.6528±{\scriptsize 0.0333} & 0.8742±{\scriptsize 0.0377} & \underline{0.6915±{\scriptsize 0.0348}} & 0.8376±{\scriptsize 0.0328} & 0.0947 ($-$) \\
 & BALD~\cite{bald} & 30\% & 0.8232±{\scriptsize 0.0871} & 0.7289±{\scriptsize 0.1101} & \textbf{0.8667±{\scriptsize 0.0819}} & 0.5911±{\scriptsize 0.2030} & 0.5275±{\scriptsize 0.1269} & \underline{0.8959±{\scriptsize 0.0680}} & 0.6463±{\scriptsize 0.1053} & 0.8522±{\scriptsize 0.1078} & 0.0947 ($-$) \\
 & EvoDiff (Ours) & 30\% & \textbf{0.8460±{\scriptsize 0.0266}} & \textbf{0.8571±{\scriptsize 0.0302}} & 0.8024±{\scriptsize 0.0128} & 0.8731±{\scriptsize 0.0244} & \underline{0.7792±{\scriptsize 0.0269}} & 0.8919±{\scriptsize 0.0221} & \textbf{0.8053±{\scriptsize 0.0421}} & \textbf{0.8746±{\scriptsize 0.0245}} & 0.0947 (+) \\
\bottomrule
\end{tabular}
}
\end{sidewaystable*}

\begin{sidewaystable*}[!ht]
\small
\centering
\renewcommand{\arraystretch}{1.2}
\setlength{\tabcolsep}{4pt}
\ContinuedFloat
\caption{Comparison of ResNet-50 classification metrics between EvoDiff and other AL methods on test sets. Values are reported as mean $\pm$ s.d. over five folds. Differences between each active learning strategy and random sampling were assessed by two-sided paired t-tests, and all reported $q$-values are Benjamini--Hochberg adjusted $p$-values computed across the five strategy-versus-random comparisons within each annotation budget and metric. (\textbf{Bold} represents the best result, \underline{Underline} represents the second best result).}
\resizebox{\linewidth}{!}{
\begin{tabular}{l|l|c|cccccccc|c}
\toprule
\textbf{Dataset} & \textbf{Method} & \textbf{Annotation Budget} & \textbf{Acc} & \textbf{bAcc} & \textbf{Sens} & \textbf{Spec} & \textbf{PPV} & \textbf{NPV} & \textbf{F1} & \textbf{AUC} & \textbf{$q$-value} \\
\midrule
\multirow{19}{*}{BUS-BRA~\cite{BUS-BRA}} & \cellcolor[HTML]{F8F8D1}FSL (upper bound) & \cellcolor[HTML]{F8F8D1}100\% & \cellcolor[HTML]{F8F8D1}0.9558±{\scriptsize 0.0123} & \cellcolor[HTML]{F8F8D1}0.9583±{\scriptsize 0.0054} & \cellcolor[HTML]{F8F8D1}0.9323±{\scriptsize 0.0158} & \cellcolor[HTML]{F8F8D1}0.9844±{\scriptsize 0.0148} & \cellcolor[HTML]{F8F8D1}0.9679±{\scriptsize 0.0299} & \cellcolor[HTML]{F8F8D1}0.9679±{\scriptsize 0.0069} & \cellcolor[HTML]{F8F8D1}0.9492±{\scriptsize 0.0101} & \cellcolor[HTML]{F8F8D1}0.9875±{\scriptsize 0.0041} & - \\
 & \cellcolor[HTML]{ECF4FF}Random & \cellcolor[HTML]{ECF4FF}10\% & \cellcolor[HTML]{ECF4FF}0.7347±{\scriptsize 0.0739} & \cellcolor[HTML]{ECF4FF}0.5507±{\scriptsize 0.0520} & \cellcolor[HTML]{ECF4FF}0.1968±{\scriptsize 0.1774} & \cellcolor[HTML]{ECF4FF}\textbf{0.9047±{\scriptsize 0.0796}} & \cellcolor[HTML]{ECF4FF}0.3817±{\scriptsize 0.1982} & \cellcolor[HTML]{ECF4FF}0.7030±{\scriptsize 0.0323} & \cellcolor[HTML]{ECF4FF}0.2374±{\scriptsize 0.1882} & \cellcolor[HTML]{ECF4FF}0.7523±{\scriptsize 0.0439} & - \\
 & VAAL~\cite{vaal} & 10\% & 0.6905±{\scriptsize 0.0101} & 0.5995±{\scriptsize 0.0553} & 0.3194±{\scriptsize 0.2284} & 0.8797±{\scriptsize 0.1306} & \textbf{0.5998±{\scriptsize 0.1011}} & 0.7369±{\scriptsize 0.0491} & 0.3631±{\scriptsize 0.1662} & 0.7309±{\scriptsize 0.0273} & 0.0613 ($-$) \\
 & Core-Set~\cite{core} & 10\% & 0.6959±{\scriptsize 0.0291} & 0.5877±{\scriptsize 0.0290} & 0.4097±{\scriptsize 0.0416} & 0.7656±{\scriptsize 0.0320} & 0.4598±{\scriptsize 0.0465} & 0.7280±{\scriptsize 0.0178} & 0.4327±{\scriptsize 0.0408} & 0.7343±{\scriptsize 0.0487} & 0.0627 ($-$) \\
 & Max-Entropy~\cite{entro} & 10\% & 0.7317±{\scriptsize 0.0727} & \underline{0.6193±{\scriptsize 0.0215}} & \textbf{0.5323±{\scriptsize 0.1619}} & 0.7063±{\scriptsize 0.1399} & 0.4929±{\scriptsize 0.0649} & \underline{0.7626±{\scriptsize 0.0251}} & \underline{0.4851±{\scriptsize 0.0707}} & 0.7554±{\scriptsize 0.0264} & 0.0660 (+) \\
 & BALD~\cite{bald} & 10\% & \underline{0.7364±{\scriptsize 0.0270}} & 0.5435±{\scriptsize 0.0592} & 0.1871±{\scriptsize 0.1589} & \underline{0.9000±{\scriptsize 0.0761}} & 0.3758±{\scriptsize 0.2157} & 0.6981±{\scriptsize 0.0343} & 0.2348±{\scriptsize 0.1759} & \underline{0.7625±{\scriptsize 0.0884}} & 0.0613 (+) \\
 & EvoDiff (Ours) & 10\% & \textbf{0.7826±{\scriptsize 0.0619}} & \textbf{0.6367±{\scriptsize 0.0492}} & \underline{0.5000±{\scriptsize 0.1530}} & 0.7734±{\scriptsize 0.0725} & \underline{0.5154±{\scriptsize 0.0553}} & \textbf{0.7666±{\scriptsize 0.0402}} & \textbf{0.4960±{\scriptsize 0.0970}} & \textbf{0.8358±{\scriptsize 0.0498}} & 0.0627 (+) \\
\cmidrule(lr){2-12}
 & \cellcolor[HTML]{ECF4FF}Random & \cellcolor[HTML]{ECF4FF}20\% & \cellcolor[HTML]{ECF4FF}0.7663±{\scriptsize 0.0437} & \cellcolor[HTML]{ECF4FF}0.5762±{\scriptsize 0.0514} & \cellcolor[HTML]{ECF4FF}0.2742±{\scriptsize 0.2051} & \cellcolor[HTML]{ECF4FF}0.8781±{\scriptsize 0.1052} & \cellcolor[HTML]{ECF4FF}0.5573±{\scriptsize 0.0572} & \cellcolor[HTML]{ECF4FF}0.7203±{\scriptsize 0.0391} & \cellcolor[HTML]{ECF4FF}0.3153±{\scriptsize 0.1610} & \cellcolor[HTML]{ECF4FF}0.8014±{\scriptsize 0.0156} & - \\
 & VAAL~\cite{vaal} & 20\% & 0.7118±{\scriptsize 0.0375} & 0.5901±{\scriptsize 0.0625} & 0.2645±{\scriptsize 0.2067} & \underline{0.9156±{\scriptsize 0.0849}} & 0.6075±{\scriptsize 0.0753} & 0.7266±{\scriptsize 0.0463} & 0.3239±{\scriptsize 0.1804} & 0.7661±{\scriptsize 0.0473} & 0.0543 ($-$) \\
 & Core-Set~\cite{core} & 20\% & 0.7444±{\scriptsize 0.0907} & \underline{0.6310±{\scriptsize 0.0351}} & 0.3839±{\scriptsize 0.1374} & 0.8781±{\scriptsize 0.1104} & \textbf{0.6978±{\scriptsize 0.1813}} & \underline{0.7489±{\scriptsize 0.0263}} & 0.4566±{\scriptsize 0.0824} & 0.7834±{\scriptsize 0.0503} & 0.0425 ($-$) \\
 & Max-Entropy~\cite{entro} & 20\% & 0.7455±{\scriptsize 0.0344} & 0.6010±{\scriptsize 0.0616} & \underline{0.4613±{\scriptsize 0.1300}} & 0.7406±{\scriptsize 0.1537} & 0.4852±{\scriptsize 0.0892} & 0.7378±{\scriptsize 0.0378} & \underline{0.4585±{\scriptsize 0.0853}} & 0.7979±{\scriptsize 0.0560} & 0.0425 ($-$) \\
 & BALD~\cite{bald} & 20\% & \underline{0.7807±{\scriptsize 0.0118}} & 0.5957±{\scriptsize 0.0494} & 0.2742±{\scriptsize 0.1326} & \textbf{0.9172±{\scriptsize 0.0693}} & \underline{0.6690±{\scriptsize 0.1528}} & 0.7247±{\scriptsize 0.0287} & 0.3599±{\scriptsize 0.1379} & \underline{0.8184±{\scriptsize 0.0739}} & 0.0425 (+) \\
 & EvoDiff (Ours) & 20\% & \textbf{0.8327±{\scriptsize 0.0276}} & \textbf{0.6571±{\scriptsize 0.0551}} & \textbf{0.5065±{\scriptsize 0.1768}} & 0.8078±{\scriptsize 0.1137} & 0.6017±{\scriptsize 0.1241} & \textbf{0.7783±{\scriptsize 0.0515}} & \textbf{0.5183±{\scriptsize 0.0938}} & \textbf{0.8640±{\scriptsize 0.0465}} & 0.0454 (+) \\
\cmidrule(lr){2-12}
 & \cellcolor[HTML]{ECF4FF}Random & \cellcolor[HTML]{ECF4FF}30\% & \cellcolor[HTML]{ECF4FF}\underline{0.8084±{\scriptsize 0.0344}} & \cellcolor[HTML]{ECF4FF}0.6419±{\scriptsize 0.0586} & \cellcolor[HTML]{ECF4FF}0.4323±{\scriptsize 0.1437} & \cellcolor[HTML]{ECF4FF}0.8516±{\scriptsize 0.1071} & \cellcolor[HTML]{ECF4FF}0.6253±{\scriptsize 0.1097} & \cellcolor[HTML]{ECF4FF}0.7579±{\scriptsize 0.0377} & \cellcolor[HTML]{ECF4FF}0.4859±{\scriptsize 0.1085} & \cellcolor[HTML]{ECF4FF}\underline{0.8441±{\scriptsize 0.0678}} & - \\
 & VAAL~\cite{vaal} & 30\% & 0.7555±{\scriptsize 0.0284} & \underline{0.6565±{\scriptsize 0.0648}} & \underline{0.5161±{\scriptsize 0.1954}} & 0.7969±{\scriptsize 0.1265} & 0.6213±{\scriptsize 0.1569} & \underline{0.7787±{\scriptsize 0.0477}} & \underline{0.5093±{\scriptsize 0.1403}} & 0.8144±{\scriptsize 0.0518} & 0.0995 ($-$) \\
 & Core-Set~\cite{core} & 30\% & 0.7715±{\scriptsize 0.0793} & 0.5991±{\scriptsize 0.0939} & 0.2419±{\scriptsize 0.2302} & \textbf{0.9563±{\scriptsize 0.0460}} & 0.6350±{\scriptsize 0.3395} & 0.7291±{\scriptsize 0.0564} & 0.3030±{\scriptsize 0.2581} & 0.8080±{\scriptsize 0.0486} & 0.0995 ($-$) \\
 & Max-Entropy~\cite{entro} & 30\% & 0.7859±{\scriptsize 0.0461} & 0.6299±{\scriptsize 0.0673} & 0.4645±{\scriptsize 0.2518} & 0.7953±{\scriptsize 0.1562} & 0.6313±{\scriptsize 0.2047} & 0.7694±{\scriptsize 0.0719} & 0.4526±{\scriptsize 0.1657} & 0.8180±{\scriptsize 0.0690} & 0.0995 ($-$) \\
 & BALD~\cite{bald} & 30\% & 0.7939±{\scriptsize 0.0686} & 0.6443±{\scriptsize 0.0549} & 0.4323±{\scriptsize 0.1871} & \underline{0.8562±{\scriptsize 0.0920}} & \textbf{0.6509±{\scriptsize 0.1199}} & 0.7628±{\scriptsize 0.0407} & 0.4709±{\scriptsize 0.1526} & 0.8238±{\scriptsize 0.0413} & 0.0995 ($-$) \\
 & EvoDiff (Ours) & 30\% & \textbf{0.8517±{\scriptsize 0.0365}} & \textbf{0.6895±{\scriptsize 0.0453}} & \textbf{0.5258±{\scriptsize 0.1183}} & 0.8531±{\scriptsize 0.0676} & \underline{0.6473±{\scriptsize 0.0959}} & \textbf{0.7909±{\scriptsize 0.0367}} & \textbf{0.5703±{\scriptsize 0.0742}} & \textbf{0.8914±{\scriptsize 0.0662}} & 0.0995 (+) \\
\midrule
\multirow{19}{*}{UDIAT~\cite{UDIAT}} & \cellcolor[HTML]{F8F8D1}FSL (upper bound) & \cellcolor[HTML]{F8F8D1}100\% & \cellcolor[HTML]{F8F8D1}0.9222±{\scriptsize 0.0272} & \cellcolor[HTML]{F8F8D1}0.8583±{\scriptsize 0.0624} & \cellcolor[HTML]{F8F8D1}0.7333±{\scriptsize 0.1333} & \cellcolor[HTML]{F8F8D1}0.9833±{\scriptsize 0.0333} & \cellcolor[HTML]{F8F8D1}0.9667±{\scriptsize 0.0667} & \cellcolor[HTML]{F8F8D1}0.8840±{\scriptsize 0.0488} & \cellcolor[HTML]{F8F8D1}0.8236±{\scriptsize 0.0893} & \cellcolor[HTML]{F8F8D1}0.9420±{\scriptsize 0.0136} & - \\
 & \cellcolor[HTML]{ECF4FF}Random & \cellcolor[HTML]{ECF4FF}10\% & \cellcolor[HTML]{ECF4FF}0.6889±{\scriptsize 0.0567} & \cellcolor[HTML]{ECF4FF}0.4417±{\scriptsize 0.1333} & \cellcolor[HTML]{ECF4FF}0.1667±{\scriptsize 0.2582} & \cellcolor[HTML]{ECF4FF}0.7167±{\scriptsize 0.2561} & \cellcolor[HTML]{ECF4FF}0.1111±{\scriptsize 0.1721} & \cellcolor[HTML]{ECF4FF}0.6254±{\scriptsize 0.1116} & \cellcolor[HTML]{ECF4FF}0.1333±{\scriptsize 0.2065} & \cellcolor[HTML]{ECF4FF}0.7510±{\scriptsize 0.1281} & - \\
 & VAAL~\cite{vaal} & 10\% & 0.6772±{\scriptsize 0.0900} & 0.4750±{\scriptsize 0.0677} & 0.2667±{\scriptsize 0.3742} & 0.6833±{\scriptsize 0.3551} & \underline{0.3167±{\scriptsize 0.3667}} & 0.5192±{\scriptsize 0.2618} & 0.1971±{\scriptsize 0.1883} & 0.7554±{\scriptsize 0.0721} & 0.0420 (+) \\
 & Core-Set~\cite{core} & 10\% & \underline{0.7186±{\scriptsize 0.0278}} & \underline{0.5417±{\scriptsize 0.0527}} & \underline{0.3333±{\scriptsize 0.3651}} & \underline{0.7500±{\scriptsize 0.2981}} & 0.2750±{\scriptsize 0.2291} & \underline{0.7485±{\scriptsize 0.1285}} & \underline{0.2691±{\scriptsize 0.2261}} & 0.7598±{\scriptsize 0.1408} & 0.0258 (+) \\
 & Max-Entropy~\cite{entro} & 10\% & 0.6856±{\scriptsize 0.0175} & 0.4000±{\scriptsize 0.0677} & 0.2333±{\scriptsize 0.2261} & 0.5667±{\scriptsize 0.2759} & 0.1395±{\scriptsize 0.1215} & 0.5693±{\scriptsize 0.0911} & 0.1715±{\scriptsize 0.1556} & 0.7620±{\scriptsize 0.1067} & 0.0769 (+) \\
 & BALD~\cite{bald} & 10\% & 0.7176±{\scriptsize 0.0415} & 0.5000±{\scriptsize 0.1149} & 0.2333±{\scriptsize 0.2495} & \textbf{0.7667±{\scriptsize 0.0623}} & 0.2476±{\scriptsize 0.2201} & 0.6753±{\scriptsize 0.0759} & 0.2342±{\scriptsize 0.2302} & \underline{0.7903±{\scriptsize 0.1671}} & 0.0769 (+) \\
 & EvoDiff (Ours) & 10\% & \textbf{0.7340±{\scriptsize 0.0129}} & \textbf{0.7167±{\scriptsize 0.0717}} & \textbf{0.7667±{\scriptsize 0.1333}} & 0.6667±{\scriptsize 0.1900} & \textbf{0.5651±{\scriptsize 0.1147}} & \textbf{0.8677±{\scriptsize 0.0787}} & \textbf{0.6369±{\scriptsize 0.0787}} & \textbf{0.8037±{\scriptsize 0.0763}} & 0.0420 (+) \\
\cmidrule(lr){2-12}
 & \cellcolor[HTML]{ECF4FF}Random & \cellcolor[HTML]{ECF4FF}20\% & \cellcolor[HTML]{ECF4FF}0.7333±{\scriptsize 0.0956} & \cellcolor[HTML]{ECF4FF}0.4667±{\scriptsize 0.1161} & \cellcolor[HTML]{ECF4FF}\textbf{0.3333±{\scriptsize 0.3801}} & \cellcolor[HTML]{ECF4FF}0.6000±{\scriptsize 0.2759} & \cellcolor[HTML]{ECF4FF}0.1857±{\scriptsize 0.1651} & \cellcolor[HTML]{ECF4FF}0.6762±{\scriptsize 0.1721} & \cellcolor[HTML]{ECF4FF}0.2267±{\scriptsize 0.2255} & \cellcolor[HTML]{ECF4FF}0.7724±{\scriptsize 0.1754} & - \\
 & VAAL~\cite{vaal} & 20\% & 0.7355±{\scriptsize 0.0587} & \textbf{0.5667±{\scriptsize 0.0624}} & 0.1333±{\scriptsize 0.1247} & \textbf{1.0000±{\scriptsize 0.0000}} & \textbf{0.6000±{\scriptsize 0.4899}} & 0.6990±{\scriptsize 0.0309} & 0.2143±{\scriptsize 0.1917} & 0.7690±{\scriptsize 0.1267} & 0.0767 ($-$) \\
 & Core-Set~\cite{core} & 20\% & 0.7537±{\scriptsize 0.0119} & 0.5083±{\scriptsize 0.0408} & 0.0333±{\scriptsize 0.0667} & \underline{0.9833±{\scriptsize 0.0333}} & 0.2000±{\scriptsize 0.4000} & 0.6706±{\scriptsize 0.0192} & 0.0571±{\scriptsize 0.1143} & 0.7923±{\scriptsize 0.1767} & 0.0896 (+) \\
 & Max-Entropy~\cite{entro} & 20\% & 0.7579±{\scriptsize 0.0153} & 0.5083±{\scriptsize 0.0167} & 0.0333±{\scriptsize 0.0667} & \underline{0.9833±{\scriptsize 0.0333}} & 0.1000±{\scriptsize 0.2000} & 0.6708±{\scriptsize 0.0083} & 0.0500±{\scriptsize 0.1000} & 0.7840±{\scriptsize 0.1045} & 0.0285 (+) \\
 & BALD~\cite{bald} & 20\% & \textbf{0.7930±{\scriptsize 0.0722}} & \underline{0.5583±{\scriptsize 0.0972}} & 0.2000±{\scriptsize 0.1944} & 0.9167±{\scriptsize 0.0746} & 0.4200±{\scriptsize 0.3816} & \underline{0.6997±{\scriptsize 0.0533}} & \underline{0.2591±{\scriptsize 0.2343}} & \underline{0.8431±{\scriptsize 0.1070}} & 0.0820 (+) \\
 & EvoDiff (Ours) & 20\% & \underline{0.7851±{\scriptsize 0.0761}} & \textbf{0.5667±{\scriptsize 0.0773}} & \underline{0.3000±{\scriptsize 0.1247}} & 0.8333±{\scriptsize 0.0913} & \underline{0.4833±{\scriptsize 0.1528}} & \textbf{0.7042±{\scriptsize 0.0444}} & \textbf{0.3611±{\scriptsize 0.1349}} & \textbf{0.8590±{\scriptsize 0.1305}} & 0.0767 (+) \\
\cmidrule(lr){2-12}
 & \cellcolor[HTML]{ECF4FF}Random & \cellcolor[HTML]{ECF4FF}30\% & \cellcolor[HTML]{ECF4FF}\underline{0.8111±{\scriptsize 0.0667}} & \cellcolor[HTML]{ECF4FF}\textbf{0.5833±{\scriptsize 0.0950}} & \cellcolor[HTML]{ECF4FF}\underline{0.3333±{\scriptsize 0.1491}} & \cellcolor[HTML]{ECF4FF}0.8333±{\scriptsize 0.1394} & \cellcolor[HTML]{ECF4FF}\underline{0.6157±{\scriptsize 0.3196}} & \cellcolor[HTML]{ECF4FF}0.7137±{\scriptsize 0.0513} & \cellcolor[HTML]{ECF4FF}\textbf{0.3955±{\scriptsize 0.1607}} & \cellcolor[HTML]{ECF4FF}\underline{0.8544±{\scriptsize 0.1305}} & - \\
 & VAAL~\cite{vaal} & 30\% & 0.7601±{\scriptsize 0.1068} & \underline{0.5500±{\scriptsize 0.0667}} & 0.2000±{\scriptsize 0.1247} & 0.9000±{\scriptsize 0.0334} & 0.4333±{\scriptsize 0.2495} & 0.6936±{\scriptsize 0.0349} & 0.2722±{\scriptsize 0.1652} & 0.8353±{\scriptsize 0.0722} & 0.0878 ($-$) \\
 & Core-Set~\cite{core} & 30\% & 0.7776±{\scriptsize 0.0531} & \textbf{0.5833±{\scriptsize 0.0746}} & \textbf{0.4000±{\scriptsize 0.3091}} & 0.7667±{\scriptsize 0.2495} & \textbf{0.6190±{\scriptsize 0.3130}} & \textbf{0.7578±{\scriptsize 0.1249}} & \underline{0.3883±{\scriptsize 0.1403}} & 0.8401±{\scriptsize 0.1447} & 0.0482 ($-$) \\
 & Max-Entropy~\cite{entro} & 30\% & 0.7786±{\scriptsize 0.0323} & 0.5417±{\scriptsize 0.0456} & 0.1333±{\scriptsize 0.1247} & \underline{0.9500±{\scriptsize 0.0408}} & 0.3333±{\scriptsize 0.2789} & 0.6883±{\scriptsize 0.0243} & 0.1889±{\scriptsize 0.1698} & 0.8446±{\scriptsize 0.1463} & 0.0482 ($-$) \\
 & BALD~\cite{bald} & 30\% & 0.7511±{\scriptsize 0.0762} & \textbf{0.5833±{\scriptsize 0.0950}} & \underline{0.3333±{\scriptsize 0.1491}} & 0.8333±{\scriptsize 0.1178} & 0.5600±{\scriptsize 0.2577} & \underline{0.7144±{\scriptsize 0.0567}} & \textbf{0.3955±{\scriptsize 0.1491}} & 0.8494±{\scriptsize 0.0858} & 0.0878 ($-$) \\
 & EvoDiff (Ours) & 30\% & \textbf{0.8454±{\scriptsize 0.0718}} & 0.5333±{\scriptsize 0.0667} & 0.1000±{\scriptsize 0.2000} & \textbf{0.9667±{\scriptsize 0.0667}} & 0.1200±{\scriptsize 0.2400} & 0.6872±{\scriptsize 0.0410} & 0.1091±{\scriptsize 0.2182} & \textbf{0.9039±{\scriptsize 0.0686}} & 0.0482 (+) \\
\bottomrule
\end{tabular}
}
\end{sidewaystable*}

\begin{sidewaystable*}[!ht]
\small
\centering
\renewcommand{\arraystretch}{1.2}
\setlength{\tabcolsep}{4pt}
\ContinuedFloat
\caption{Comparison of ResNet-50 classification metrics between EvoDiff and other AL methods on test sets. Values are reported as mean $\pm$ s.d. over five folds. Differences between each active learning strategy and random sampling were assessed by two-sided paired t-tests, and all reported $q$-values are Benjamini--Hochberg adjusted $p$-values computed across the five strategy-versus-random comparisons within each annotation budget and metric. (\textbf{Bold} represents the best result, \underline{Underline} represents the second best result).}
\resizebox{\linewidth}{!}{
\begin{tabular}{l|l|c|cccccccc|c}
\toprule
\textbf{Dataset} & \textbf{Method} & \textbf{Annotation Budget} & \textbf{Acc} & \textbf{bAcc} & \textbf{Sens} & \textbf{Spec} & \textbf{PPV} & \textbf{NPV} & \textbf{F1} & \textbf{AUC} & \textbf{$q$-value} \\
\midrule
\multirow{19}{*}{CUBS~\cite{CUBS}} & \cellcolor[HTML]{F8F8D1}FSL (upper bound) & \cellcolor[HTML]{F8F8D1}100\% & \cellcolor[HTML]{F8F8D1}0.8906±{\scriptsize 0.0451} & \cellcolor[HTML]{F8F8D1}0.8648±{\scriptsize 0.0312} & \cellcolor[HTML]{F8F8D1}0.7615±{\scriptsize 0.0632} & \cellcolor[HTML]{F8F8D1}0.9681±{\scriptsize 0.0184} & \cellcolor[HTML]{F8F8D1}0.8515±{\scriptsize 0.0762} & \cellcolor[HTML]{F8F8D1}0.9465±{\scriptsize 0.0135} & \cellcolor[HTML]{F8F8D1}0.8016±{\scriptsize 0.0475} & \cellcolor[HTML]{F8F8D1}0.9368±{\scriptsize 0.0136} & - \\
 & \cellcolor[HTML]{ECF4FF}Random & \cellcolor[HTML]{ECF4FF}10\% & \cellcolor[HTML]{ECF4FF}0.7122±{\scriptsize 0.0457} & \cellcolor[HTML]{ECF4FF}0.5351±{\scriptsize 0.0229} & \cellcolor[HTML]{ECF4FF}0.1923±{\scriptsize 0.0902} & \cellcolor[HTML]{ECF4FF}\textbf{0.8779±{\scriptsize 0.0840}} & \cellcolor[HTML]{ECF4FF}\underline{0.2901±{\scriptsize 0.0944}} & \cellcolor[HTML]{ECF4FF}0.8254±{\scriptsize 0.0079} & \cellcolor[HTML]{ECF4FF}0.2176±{\scriptsize 0.0527} & \cellcolor[HTML]{ECF4FF}0.7290±{\scriptsize 0.0287} & - \\
 & VAAL~\cite{vaal} & 10\% & 0.7257±{\scriptsize 0.0528} & 0.5567±{\scriptsize 0.0317} & 0.2692±{\scriptsize 0.0860} & \underline{0.8442±{\scriptsize 0.0323}} & 0.2805±{\scriptsize 0.0374} & 0.8343±{\scriptsize 0.0126} & 0.2718±{\scriptsize 0.0631} & 0.7421±{\scriptsize 0.0418} & 0.0029 (+) \\
 & Core-Set~\cite{core} & 10\% & 0.7317±{\scriptsize 0.0204} & 0.5601±{\scriptsize 0.0412} & \underline{0.3769±{\scriptsize 0.0688}} & 0.7434±{\scriptsize 0.0570} & 0.2556±{\scriptsize 0.0487} & \underline{0.8380±{\scriptsize 0.0168}} & \underline{0.3030±{\scriptsize 0.0534}} & 0.7423±{\scriptsize 0.0458} & 0.0029 (+) \\
 & Max-Entropy~\cite{entro} & 10\% & \underline{0.7462±{\scriptsize 0.0321}} & 0.5097±{\scriptsize 0.0124} & 0.1769±{\scriptsize 0.0211} & 0.8425±{\scriptsize 0.0383} & 0.2091±{\scriptsize 0.0247} & 0.8163±{\scriptsize 0.0045} & 0.1900±{\scriptsize 0.0123} & 0.7674±{\scriptsize 0.0368} & 0.0003 (+) \\
 & BALD~\cite{bald} & 10\% & 0.7403±{\scriptsize 0.0416} & \underline{0.5610±{\scriptsize 0.0468}} & 0.3308±{\scriptsize 0.0750} & 0.7912±{\scriptsize 0.0698} & 0.2805±{\scriptsize 0.0830} & 0.8367±{\scriptsize 0.0186} & 0.2980±{\scriptsize 0.0606} & \underline{0.7850±{\scriptsize 0.0451}} & 0.0038 (+) \\
 & EvoDiff (Ours) & 10\% & \textbf{0.7951±{\scriptsize 0.0045}} & \textbf{0.6165±{\scriptsize 0.0336}} & \textbf{0.4154±{\scriptsize 0.0632}} & 0.8177±{\scriptsize 0.0445} & \textbf{0.3488±{\scriptsize 0.0547}} & \textbf{0.8588±{\scriptsize 0.0133}} & \textbf{0.3769±{\scriptsize 0.0484}} & \textbf{0.8242±{\scriptsize 0.0274}} & 0.0001 (+) \\
\cmidrule(lr){2-12}
 & \cellcolor[HTML]{ECF4FF}Random & \cellcolor[HTML]{ECF4FF}20\% & \cellcolor[HTML]{ECF4FF}0.7410±{\scriptsize 0.0561} & \cellcolor[HTML]{ECF4FF}0.5586±{\scriptsize 0.0364} & \cellcolor[HTML]{ECF4FF}0.3154±{\scriptsize 0.0688} & \cellcolor[HTML]{ECF4FF}0.8018±{\scriptsize 0.0503} & \cellcolor[HTML]{ECF4FF}0.2735±{\scriptsize 0.0622} & \cellcolor[HTML]{ECF4FF}0.8358±{\scriptsize 0.0142} & \cellcolor[HTML]{ECF4FF}0.2899±{\scriptsize 0.0531} & \cellcolor[HTML]{ECF4FF}0.7753±{\scriptsize 0.0367} & - \\
 & VAAL~\cite{vaal} & 20\% & 0.7113±{\scriptsize 0.0151} & \underline{0.6334±{\scriptsize 0.0597}} & \underline{0.4154±{\scriptsize 0.1500}} & 0.8513±{\scriptsize 0.0630} & \textbf{0.4067±{\scriptsize 0.0818}} & \underline{0.8647±{\scriptsize 0.0251}} & \underline{0.3934±{\scriptsize 0.0958}} & 0.7693±{\scriptsize 0.0474} & 0.0089 ($-$) \\
 & Core-Set~\cite{core} & 20\% & 0.7413±{\scriptsize 0.0558} & 0.5653±{\scriptsize 0.0232} & 0.2615±{\scriptsize 0.0322} & 0.8690±{\scriptsize 0.0202} & 0.3175±{\scriptsize 0.0536} & 0.8364±{\scriptsize 0.0081} & 0.2864±{\scriptsize 0.0396} & 0.7912±{\scriptsize 0.0319} & 0.0116 (+) \\
 & Max-Entropy~\cite{entro} & 20\% & \underline{0.7916±{\scriptsize 0.0392}} & 0.5717±{\scriptsize 0.0265} & 0.2462±{\scriptsize 0.0439} & \textbf{0.8973±{\scriptsize 0.0346}} & 0.3722±{\scriptsize 0.1089} & 0.8380±{\scriptsize 0.0091} & 0.2918±{\scriptsize 0.0482} & 0.7855±{\scriptsize 0.0502} & 0.0485 (+) \\
 & BALD~\cite{bald} & 20\% & 0.7756±{\scriptsize 0.0797} & 0.5671±{\scriptsize 0.0211} & 0.2615±{\scriptsize 0.0570} & \underline{0.8726±{\scriptsize 0.0284}} & 0.3215±{\scriptsize 0.0405} & 0.8372±{\scriptsize 0.0081} & 0.2863±{\scriptsize 0.0422} & \underline{0.8246±{\scriptsize 0.0505}} & 0.0089 (+) \\
 & EvoDiff (Ours) & 20\% & \textbf{0.8218±{\scriptsize 0.0184}} & \textbf{0.6360±{\scriptsize 0.0259}} & \textbf{0.4385±{\scriptsize 0.0583}} & 0.8336±{\scriptsize 0.0296} & \underline{0.3792±{\scriptsize 0.0385}} & \textbf{0.8660±{\scriptsize 0.0106}} & \textbf{0.4051±{\scriptsize 0.0395}} & \textbf{0.8671±{\scriptsize 0.0493}} & 0.0993 (+) \\
\cmidrule(lr){2-12}
 & \cellcolor[HTML]{ECF4FF}Random & \cellcolor[HTML]{ECF4FF}30\% & \cellcolor[HTML]{ECF4FF}0.7899±{\scriptsize 0.0411} & \cellcolor[HTML]{ECF4FF}0.5809±{\scriptsize 0.0325} & \cellcolor[HTML]{ECF4FF}0.2769±{\scriptsize 0.0570} & \cellcolor[HTML]{ECF4FF}0.8850±{\scriptsize 0.0108} & \cellcolor[HTML]{ECF4FF}0.3554±{\scriptsize 0.0663} & \cellcolor[HTML]{ECF4FF}0.8418±{\scriptsize 0.0116} & \cellcolor[HTML]{ECF4FF}0.3111±{\scriptsize 0.0611} & \cellcolor[HTML]{ECF4FF}0.8219±{\scriptsize 0.0237} & - \\
 & VAAL~\cite{vaal} & 30\% & 0.7657±{\scriptsize 0.0141} & 0.6043±{\scriptsize 0.0249} & 0.3308±{\scriptsize 0.1003} & 0.8779±{\scriptsize 0.0639} & 0.3979±{\scriptsize 0.0474} & 0.8514±{\scriptsize 0.0112} & 0.3501±{\scriptsize 0.0500} & 0.8094±{\scriptsize 0.0190} & 0.0137 ($-$) \\
 & Core-Set~\cite{core} & 30\% & 0.7799±{\scriptsize 0.0496} & \textbf{0.6631±{\scriptsize 0.0664}} & \underline{0.4077±{\scriptsize 0.1141}} & 0.9186±{\scriptsize 0.0211} & 0.5325±{\scriptsize 0.1287} & \textbf{0.8709±{\scriptsize 0.0240}} & \textbf{0.4613±{\scriptsize 0.1203}} & 0.8228±{\scriptsize 0.0452} & 0.0538 (+) \\
 & Max-Entropy~\cite{entro} & 30\% & \underline{0.8186±{\scriptsize 0.0208}} & \underline{0.6602±{\scriptsize 0.0402}} & 0.4000±{\scriptsize 0.1039} & \underline{0.9204±{\scriptsize 0.0531}} & \underline{0.5758±{\scriptsize 0.1478}} & 0.8701±{\scriptsize 0.0159} & \underline{0.4550±{\scriptsize 0.0723}} & 0.8450±{\scriptsize 0.0623} & 0.0974 (+) \\
 & BALD~\cite{bald} & 30\% & 0.8023±{\scriptsize 0.0802} & 0.6590±{\scriptsize 0.0394} & \textbf{0.4154±{\scriptsize 0.0688}} & 0.9027±{\scriptsize 0.0313} & 0.5037±{\scriptsize 0.1059} & \underline{0.8704±{\scriptsize 0.0143}} & 0.4526±{\scriptsize 0.0724} & \underline{0.8587±{\scriptsize 0.0410}} & 0.0546 (+) \\
 & EvoDiff (Ours) & 30\% & \textbf{0.8442±{\scriptsize 0.0254}} & 0.6291±{\scriptsize 0.0301} & 0.3077±{\scriptsize 0.0544} & \textbf{0.9504±{\scriptsize 0.0184}} & \textbf{0.5983±{\scriptsize 0.1275}} & 0.8565±{\scriptsize 0.0104} & 0.4037±{\scriptsize 0.0639} & \textbf{0.8962±{\scriptsize 0.0718}} & 0.0538 (+) \\
\midrule
\multirow{19}{*}{Fatty-Liver~\cite{liver}} & \cellcolor[HTML]{F8F8D1}FSL (upper bound) & \cellcolor[HTML]{F8F8D1}100\% & \cellcolor[HTML]{F8F8D1}0.9741±{\scriptsize 0.0361} & \cellcolor[HTML]{F8F8D1}0.9118±{\scriptsize 0.0340} & \cellcolor[HTML]{F8F8D1}0.8486±{\scriptsize 0.0961} & \cellcolor[HTML]{F8F8D1}0.9750±{\scriptsize 0.0306} & \cellcolor[HTML]{F8F8D1}0.9890±{\scriptsize 0.0135} & \cellcolor[HTML]{F8F8D1}0.7618±{\scriptsize 0.1385} & \cellcolor[HTML]{F8F8D1}0.9099±{\scriptsize 0.0489} & \cellcolor[HTML]{F8F8D1}0.9879±{\scriptsize 0.0282} & - \\
 & \cellcolor[HTML]{ECF4FF}Random & \cellcolor[HTML]{ECF4FF}10\% & \cellcolor[HTML]{ECF4FF}0.8469±{\scriptsize 0.0766} & \cellcolor[HTML]{ECF4FF}0.8492±{\scriptsize 0.0666} & \cellcolor[HTML]{ECF4FF}0.8108±{\scriptsize 0.1134} & \cellcolor[HTML]{ECF4FF}0.8875±{\scriptsize 0.0468} & \cellcolor[HTML]{ECF4FF}0.9419±{\scriptsize 0.0255} & \cellcolor[HTML]{ECF4FF}0.6913±{\scriptsize 0.1179} & \cellcolor[HTML]{ECF4FF}0.8677±{\scriptsize 0.0774} & \cellcolor[HTML]{ECF4FF}0.9258±{\scriptsize 0.0769} & - \\
 & VAAL~\cite{vaal} & 10\% & 0.8196±{\scriptsize 0.0491} & 0.5615±{\scriptsize 0.1298} & \textbf{0.9730±{\scriptsize 0.0419}} & 0.1500±{\scriptsize 0.3000} & 0.7357±{\scriptsize 0.0781} & 0.1500±{\scriptsize 0.3000} & 0.8335±{\scriptsize 0.0296} & 0.9145±{\scriptsize 0.1845} & 0.0434 ($-$) \\
 & Core-Set~\cite{core} & 10\% & 0.8294±{\scriptsize 0.0360} & 0.8480±{\scriptsize 0.0415} & 0.7459±{\scriptsize 0.1089} & \underline{0.9500±{\scriptsize 0.0468}} & \underline{0.9743±{\scriptsize 0.0222}} & 0.6374±{\scriptsize 0.1068} & 0.8396±{\scriptsize 0.0648} & 0.9163±{\scriptsize 0.0111} & 0.0709 ($-$) \\
 & Max-Entropy~\cite{entro} & 10\% & 0.8566±{\scriptsize 0.0720} & 0.7921±{\scriptsize 0.0807} & 0.6216±{\scriptsize 0.1519} & \textbf{0.9625±{\scriptsize 0.0306}} & 0.9703±{\scriptsize 0.0285} & 0.5396±{\scriptsize 0.0843} & 0.7470±{\scriptsize 0.1345} & 0.9222±{\scriptsize 0.0657} & 0.0417 ($-$) \\
 & BALD~\cite{bald} & 10\% & \underline{0.8641±{\scriptsize 0.0455}} & \textbf{0.8858±{\scriptsize 0.0650}} & 0.8216±{\scriptsize 0.1363} & \underline{0.9500±{\scriptsize 0.0250}} & \textbf{0.9745±{\scriptsize 0.0136}} & \textbf{0.7417±{\scriptsize 0.1875}} & \textbf{0.8856±{\scriptsize 0.0788}} & \underline{0.9376±{\scriptsize 0.0206}} & 0.0434 (+) \\
 & EvoDiff (Ours) & 10\% & \textbf{0.8815±{\scriptsize 0.0336}} & \underline{0.8537±{\scriptsize 0.0221}} & \underline{0.8324±{\scriptsize 0.0733}} & 0.8750±{\scriptsize 0.0685} & 0.9416±{\scriptsize 0.0258} & \underline{0.7065±{\scriptsize 0.0748}} & \underline{0.8809±{\scriptsize 0.0352}} & \textbf{0.9534±{\scriptsize 0.0280}} & 0.0709 (+) \\
\cmidrule(lr){2-12}
 & \cellcolor[HTML]{ECF4FF}Random & \cellcolor[HTML]{ECF4FF}20\% & \cellcolor[HTML]{ECF4FF}0.8753±{\scriptsize 0.0810} & \cellcolor[HTML]{ECF4FF}0.9328±{\scriptsize 0.0211} & \cellcolor[HTML]{ECF4FF}0.9405±{\scriptsize 0.0359} & \cellcolor[HTML]{ECF4FF}0.9250±{\scriptsize 0.0250} & \cellcolor[HTML]{ECF4FF}0.9668±{\scriptsize 0.0105} & \cellcolor[HTML]{ECF4FF}0.8760±{\scriptsize 0.0703} & \cellcolor[HTML]{ECF4FF}0.9531±{\scriptsize 0.0192} & \cellcolor[HTML]{ECF4FF}\underline{0.9431±{\scriptsize 0.0134}} & - \\
 & VAAL~\cite{vaal} & 20\% & 0.8417±{\scriptsize 0.0922} & \textbf{0.9534±{\scriptsize 0.0285}} & \underline{0.9568±{\scriptsize 0.0471}} & 0.9500±{\scriptsize 0.0250} & 0.9779±{\scriptsize 0.0111} & \underline{0.9132±{\scriptsize 0.0850}} & \textbf{0.9666±{\scriptsize 0.0269}} & 0.9195±{\scriptsize 0.0087} & 0.0019 ($-$) \\
 & Core-Set~\cite{core} & 20\% & 0.8490±{\scriptsize 0.0797} & 0.9218±{\scriptsize 0.0550} & 0.8811±{\scriptsize 0.1062} & \underline{0.9625±{\scriptsize 0.0306}} & \underline{0.9819±{\scriptsize 0.0151}} & 0.8082±{\scriptsize 0.1539} & 0.9254±{\scriptsize 0.0616} & 0.9240±{\scriptsize 0.0187} & 0.0019 ($-$) \\
 & Max-Entropy~\cite{entro} & 20\% & 0.8934±{\scriptsize 0.0913} & 0.9432±{\scriptsize 0.0216} & 0.8865±{\scriptsize 0.0432} & \textbf{1.0000±{\scriptsize 0.0000}} & \textbf{1.0000±{\scriptsize 0.0000}} & 0.7968±{\scriptsize 0.0593} & 0.9393±{\scriptsize 0.0248} & 0.9380±{\scriptsize 0.0106} & 0.0019 ($-$) \\
 & BALD~\cite{bald} & 20\% & \underline{0.8987±{\scriptsize 0.0421}} & 0.9182±{\scriptsize 0.0205} & 0.8865±{\scriptsize 0.0602} & 0.9500±{\scriptsize 0.0250} & 0.9769±{\scriptsize 0.0116} & 0.7942±{\scriptsize 0.0822} & 0.9281±{\scriptsize 0.0301} & 0.9401±{\scriptsize 0.0047} & 0.0021 ($-$) \\
 & EvoDiff (Ours) & 20\% & \textbf{0.9240±{\scriptsize 0.0491}} & \underline{0.9436±{\scriptsize 0.0218}} & \textbf{0.9622±{\scriptsize 0.0471}} & 0.9250±{\scriptsize 0.0250} & 0.9676±{\scriptsize 0.0095} & \textbf{0.9246±{\scriptsize 0.0934}} & \underline{0.9642±{\scriptsize 0.0232}} & \textbf{0.9723±{\scriptsize 0.0039}} & 0.0380 (+) \\
\cmidrule(lr){2-12}
 & \cellcolor[HTML]{ECF4FF}Random & \cellcolor[HTML]{ECF4FF}30\% & \cellcolor[HTML]{ECF4FF}0.9157±{\scriptsize 0.0688} & \cellcolor[HTML]{ECF4FF}0.9392±{\scriptsize 0.0316} & \cellcolor[HTML]{ECF4FF}\textbf{0.9784±{\scriptsize 0.0202}} & \cellcolor[HTML]{ECF4FF}0.9000±{\scriptsize 0.0500} & \cellcolor[HTML]{ECF4FF}0.9579±{\scriptsize 0.0204} & \cellcolor[HTML]{ECF4FF}\textbf{0.9482±{\scriptsize 0.0474}} & \cellcolor[HTML]{ECF4FF}0.9679±{\scriptsize 0.0181} & \cellcolor[HTML]{ECF4FF}\underline{0.9669±{\scriptsize 0.0107}} & - \\
 & VAAL~\cite{vaal} & 30\% & 0.8763±{\scriptsize 0.0859} & 0.9507±{\scriptsize 0.0346} & 0.9514±{\scriptsize 0.0524} & \underline{0.9500±{\scriptsize 0.0468}} & \underline{0.9781±{\scriptsize 0.0200}} & 0.9051±{\scriptsize 0.0968} & 0.9637±{\scriptsize 0.0295} & 0.9333±{\scriptsize 0.0042} & 0.0011 ($-$) \\
 & Core-Set~\cite{core} & 30\% & 0.8870±{\scriptsize 0.0831} & \textbf{0.9784±{\scriptsize 0.0108}} & \underline{0.9568±{\scriptsize 0.0216}} & \textbf{1.0000±{\scriptsize 0.0000}} & \textbf{1.0000±{\scriptsize 0.0000}} & \underline{0.9109±{\scriptsize 0.0399}} & \textbf{0.9778±{\scriptsize 0.0114}} & 0.9296±{\scriptsize 0.0037} & 0.0011 ($-$) \\
 & Max-Entropy~\cite{entro} & 30\% & 0.9011±{\scriptsize 0.0442} & 0.9328±{\scriptsize 0.0263} & 0.9405±{\scriptsize 0.0602} & 0.9250±{\scriptsize 0.0250} & 0.9670±{\scriptsize 0.0093} & 0.8860±{\scriptsize 0.1079} & 0.9525±{\scriptsize 0.0304} & 0.9424±{\scriptsize 0.0117} & 0.0011 ($-$) \\
 & BALD~\cite{bald} & 30\% & \underline{0.9371±{\scriptsize 0.0870}} & \underline{0.9730±{\scriptsize 0.0191}} & 0.9459±{\scriptsize 0.0382} & \textbf{1.0000±{\scriptsize 0.0000}} & \textbf{1.0000±{\scriptsize 0.0000}} & 0.8944±{\scriptsize 0.0707} & \underline{0.9718±{\scriptsize 0.0202}} & 0.9543±{\scriptsize 0.0028} & 0.0011 ($-$) \\
 & EvoDiff (Ours) & 30\% & \textbf{0.9552±{\scriptsize 0.0767}} & \underline{0.9730±{\scriptsize 0.0191}} & 0.9459±{\scriptsize 0.0382} & \textbf{1.0000±{\scriptsize 0.0000}} & \textbf{1.0000±{\scriptsize 0.0000}} & 0.8944±{\scriptsize 0.0707} & \underline{0.9718±{\scriptsize 0.0202}} & \textbf{0.9793±{\scriptsize 0.0014}} & 0.0839 (+) \\
\bottomrule
\end{tabular}
}
\end{sidewaystable*}

\begin{sidewaystable*}[!ht]
\small
\centering
\renewcommand{\arraystretch}{1.2}
\setlength{\tabcolsep}{4pt}
\ContinuedFloat
\caption{Comparison of ResNet-50 classification metrics between EvoDiff and other AL methods on test sets. Values are reported as mean $\pm$ s.d. over five folds. Differences between each active learning strategy and random sampling were assessed by two-sided paired t-tests, and all reported $q$-values are Benjamini--Hochberg adjusted $p$-values computed across the five strategy-versus-random comparisons within each annotation budget and metric. (\textbf{Bold} represents the best result, \underline{Underline} represents the second best result).}
\resizebox{\linewidth}{!}{
\begin{tabular}{l|l|c|cccccccc|c}
\toprule
\textbf{Dataset} & \textbf{Method} & \textbf{Annotation Budget} & \textbf{Acc} & \textbf{bAcc} & \textbf{Sens} & \textbf{Spec} & \textbf{PPV} & \textbf{NPV} & \textbf{F1} & \textbf{AUC} & \textbf{$q$-value} \\
\midrule
\multirow{19}{*}{MMOTU~\cite{mmotu}} & \cellcolor[HTML]{F8F8D1}FSL (upper bound) & \cellcolor[HTML]{F8F8D1}100\% & \cellcolor[HTML]{F8F8D1}0.9410±{\scriptsize 0.0212} & \cellcolor[HTML]{F8F8D1}0.7282±{\scriptsize 0.0275} & \cellcolor[HTML]{F8F8D1}0.9422±{\scriptsize 0.0238} & \cellcolor[HTML]{F8F8D1}0.5143±{\scriptsize 0.0407} & \cellcolor[HTML]{F8F8D1}0.8986±{\scriptsize 0.0089} & \cellcolor[HTML]{F8F8D1}0.6707±{\scriptsize 0.1089} & \cellcolor[HTML]{F8F8D1}0.9198±{\scriptsize 0.0147} & \cellcolor[HTML]{F8F8D1}0.9897±{\scriptsize 0.0100} & - \\
 & \cellcolor[HTML]{ECF4FF}Random & \cellcolor[HTML]{ECF4FF}10\% & \cellcolor[HTML]{ECF4FF}0.7564±{\scriptsize 0.0481} & \cellcolor[HTML]{ECF4FF}0.5586±{\scriptsize 0.0577} & \cellcolor[HTML]{ECF4FF}0.8172±{\scriptsize 0.0782} & \cellcolor[HTML]{ECF4FF}0.3000±{\scriptsize 0.1029} & \cellcolor[HTML]{ECF4FF}0.8418±{\scriptsize 0.0212} & \cellcolor[HTML]{ECF4FF}0.2711±{\scriptsize 0.0865} & \cellcolor[HTML]{ECF4FF}0.8280±{\scriptsize 0.0480} & \cellcolor[HTML]{ECF4FF}0.8370±{\scriptsize 0.0539} & - \\
 & VAAL~\cite{vaal} & 10\% & 0.7393±{\scriptsize 0.0866} & \textbf{0.6266±{\scriptsize 0.0386}} & 0.8531±{\scriptsize 0.0480} & \textbf{0.4000±{\scriptsize 0.0299}} & \underline{0.8663±{\scriptsize 0.0122}} & \textbf{0.3872±{\scriptsize 0.0980}} & 0.8593±{\scriptsize 0.0303} & 0.8361±{\scriptsize 0.0186} & 0.0103 ($-$) \\
 & Core-Set~\cite{core} & 10\% & 0.7630±{\scriptsize 0.0353} & 0.5323±{\scriptsize 0.0287} & \textbf{0.9359±{\scriptsize 0.0178}} & 0.1286±{\scriptsize 0.0696} & 0.8310±{\scriptsize 0.0097} & 0.2951±{\scriptsize 0.0651} & \textbf{0.8802±{\scriptsize 0.0062}} & 0.8409±{\scriptsize 0.0326} & 0.0204 (+) \\
 & Max-Entropy~\cite{entro} & 10\% & \underline{0.7747±{\scriptsize 0.0621}} & 0.5792±{\scriptsize 0.0474} & \underline{0.9156±{\scriptsize 0.0347}} & 0.2429±{\scriptsize 0.1247} & 0.8475±{\scriptsize 0.0170} & \underline{0.3715±{\scriptsize 0.0899}} & \underline{0.8797±{\scriptsize 0.0098}} & \underline{0.8661±{\scriptsize 0.0248}} & 0.0042 (+) \\
 & BALD~\cite{bald} & 10\% & 0.7630±{\scriptsize 0.0861} & 0.5934±{\scriptsize 0.0251} & 0.8797±{\scriptsize 0.0389} & \underline{0.3071±{\scriptsize 0.0860}} & 0.8535±{\scriptsize 0.0103} & 0.3615±{\scriptsize 0.0209} & 0.8658±{\scriptsize 0.0142} & 0.8509±{\scriptsize 0.0202} & 0.0204 (+) \\
 & EvoDiff (Ours) & 10\% & \textbf{0.8095±{\scriptsize 0.0496}} & \underline{0.6242±{\scriptsize 0.0474}} & 0.8484±{\scriptsize 0.0461} & \textbf{0.4000±{\scriptsize 0.1112}} & \textbf{0.8666±{\scriptsize 0.0192}} & 0.3703±{\scriptsize 0.0614} & 0.8567±{\scriptsize 0.0227} & \textbf{0.8748±{\scriptsize 0.0448}} & 0.0270 (+) \\
\cmidrule(lr){2-12}
 & \cellcolor[HTML]{ECF4FF}Random & \cellcolor[HTML]{ECF4FF}20\% & \cellcolor[HTML]{ECF4FF}0.7987±{\scriptsize 0.0119} & \cellcolor[HTML]{ECF4FF}0.6294±{\scriptsize 0.0360} & \cellcolor[HTML]{ECF4FF}0.9016±{\scriptsize 0.0196} & \cellcolor[HTML]{ECF4FF}0.3571±{\scriptsize 0.0911} & \cellcolor[HTML]{ECF4FF}0.8655±{\scriptsize 0.0137} & \cellcolor[HTML]{ECF4FF}0.4392±{\scriptsize 0.0233} & \cellcolor[HTML]{ECF4FF}0.8829±{\scriptsize 0.0030} & \cellcolor[HTML]{ECF4FF}0.8672±{\scriptsize 0.0420} & - \\
 & VAAL~\cite{vaal} & 20\% & 0.7877±{\scriptsize 0.0844} & \underline{0.6631±{\scriptsize 0.0615}} & 0.8547±{\scriptsize 0.0295} & \underline{0.4714±{\scriptsize 0.1112}} & \underline{0.8811±{\scriptsize 0.0238}} & 0.4161±{\scriptsize 0.0810} & 0.8675±{\scriptsize 0.0227} & 0.8412±{\scriptsize 0.0563} & 0.0897 ($-$) \\
 & Core-Set~\cite{core} & 20\% & 0.8168±{\scriptsize 0.0850} & 0.6475±{\scriptsize 0.0320} & 0.9094±{\scriptsize 0.0393} & 0.3857±{\scriptsize 0.0687} & 0.8714±{\scriptsize 0.0117} & 0.5042±{\scriptsize 0.1202} & 0.8896±{\scriptsize 0.0196} & 0.8903±{\scriptsize 0.0304} & 0.0256 (+) \\
 & Max-Entropy~\cite{entro} & 20\% & \textbf{0.8372±{\scriptsize 0.0693}} & 0.6304±{\scriptsize 0.0285} & \textbf{0.9250±{\scriptsize 0.0408}} & 0.3357±{\scriptsize 0.0649} & 0.8644±{\scriptsize 0.0100} & \underline{0.5172±{\scriptsize 0.1202}} & \underline{0.8933±{\scriptsize 0.0198}} & 0.8842±{\scriptsize 0.0449} & 0.0505 (+) \\
 & BALD~\cite{bald} & 20\% & 0.8248±{\scriptsize 0.0642} & 0.6221±{\scriptsize 0.0263} & 0.9156±{\scriptsize 0.0338} & 0.3286±{\scriptsize 0.0687} & 0.8620±{\scriptsize 0.0095} & 0.4708±{\scriptsize 0.0714} & 0.8877±{\scriptsize 0.0148} & \underline{0.8911±{\scriptsize 0.0269}} & 0.0294 (+) \\
 & EvoDiff (Ours) & 20\% & \underline{0.8320±{\scriptsize 0.0192}} & \textbf{0.6987±{\scriptsize 0.0225}} & \underline{0.9187±{\scriptsize 0.0225}} & \textbf{0.4786±{\scriptsize 0.0542}} & \textbf{0.8897±{\scriptsize 0.0089}} & \textbf{0.5682±{\scriptsize 0.0508}} & \textbf{0.9038±{\scriptsize 0.0102}} & \textbf{0.9018±{\scriptsize 0.0334}} & 0.0112 (+) \\
\cmidrule(lr){2-12}
 & \cellcolor[HTML]{ECF4FF}Random & \cellcolor[HTML]{ECF4FF}30\% & \cellcolor[HTML]{ECF4FF}0.8359±{\scriptsize 0.0359} & \cellcolor[HTML]{ECF4FF}\underline{0.7056±{\scriptsize 0.0211}} & \cellcolor[HTML]{ECF4FF}\textbf{0.9469±{\scriptsize 0.0261}} & \cellcolor[HTML]{ECF4FF}0.4643±{\scriptsize 0.0565} & \cellcolor[HTML]{ECF4FF}0.8900±{\scriptsize 0.0086} & \cellcolor[HTML]{ECF4FF}\textbf{0.6724±{\scriptsize 0.0901}} & \cellcolor[HTML]{ECF4FF}\textbf{0.9174±{\scriptsize 0.0107}} & \cellcolor[HTML]{ECF4FF}0.8966±{\scriptsize 0.0131} & - \\
 & VAAL~\cite{vaal} & 30\% & 0.8195±{\scriptsize 0.0399} & 0.6646±{\scriptsize 0.0329} & 0.9078±{\scriptsize 0.0486} & 0.4214±{\scriptsize 0.0774} & 0.8779±{\scriptsize 0.0124} & 0.5299±{\scriptsize 0.1250} & 0.8920±{\scriptsize 0.0235} & 0.8720±{\scriptsize 0.0419} & 0.0629 ($-$) \\
 & Core-Set~\cite{core} & 30\% & 0.8583±{\scriptsize 0.0570} & 0.6942±{\scriptsize 0.0304} & \underline{0.9313±{\scriptsize 0.0187}} & 0.4571±{\scriptsize 0.0687} & 0.8871±{\scriptsize 0.0121} & \underline{0.5967±{\scriptsize 0.0598}} & \underline{0.9085±{\scriptsize 0.0085}} & \underline{0.9065±{\scriptsize 0.0118}} & 0.0821 (+) \\
 & Max-Entropy~\cite{entro} & 30\% & \textbf{0.8770±{\scriptsize 0.0757}} & 0.7038±{\scriptsize 0.0360} & 0.9219±{\scriptsize 0.0259} & \underline{0.4857±{\scriptsize 0.0741}} & \underline{0.8914±{\scriptsize 0.0137}} & 0.5849±{\scriptsize 0.0902} & 0.9062±{\scriptsize 0.0139} & 0.8988±{\scriptsize 0.0361} & 0.0629 (+) \\
 & BALD~\cite{bald} & 30\% & 0.8475±{\scriptsize 0.0684} & 0.6831±{\scriptsize 0.0466} & 0.9234±{\scriptsize 0.0224} & 0.4429±{\scriptsize 0.1089} & 0.8840±{\scriptsize 0.0182} & 0.5604±{\scriptsize 0.0579} & 0.9030±{\scriptsize 0.0079} & 0.9031±{\scriptsize 0.0222} & 0.0629 (+) \\
 & EvoDiff (Ours) & 30\% & \underline{0.8682±{\scriptsize 0.0290}} & \textbf{0.7087±{\scriptsize 0.0536}} & 0.9031±{\scriptsize 0.0437} & \textbf{0.5143±{\scriptsize 0.1089}} & \textbf{0.8951±{\scriptsize 0.0203}} & 0.5527±{\scriptsize 0.1123} & 0.8986±{\scriptsize 0.0239} & \textbf{0.9370±{\scriptsize 0.0514}} & 0.0629 (+) \\
\midrule
\multirow{19}{*}{TN3K~\cite{TN3K}} & \cellcolor[HTML]{F8F8D1}FSL (upper bound) & \cellcolor[HTML]{F8F8D1}100\% & \cellcolor[HTML]{F8F8D1}0.8937±{\scriptsize 0.0157} & \cellcolor[HTML]{F8F8D1}0.8203±{\scriptsize 0.0241} & \cellcolor[HTML]{F8F8D1}0.7934±{\scriptsize 0.0370} & \cellcolor[HTML]{F8F8D1}0.8472±{\scriptsize 0.0304} & \cellcolor[HTML]{F8F8D1}0.7346±{\scriptsize 0.0416} & \cellcolor[HTML]{F8F8D1}0.8861±{\scriptsize 0.0184} & \cellcolor[HTML]{F8F8D1}0.7621±{\scriptsize 0.0310} & \cellcolor[HTML]{F8F8D1}0.8927±{\scriptsize 0.0279} & - \\
 & \cellcolor[HTML]{ECF4FF}Random & \cellcolor[HTML]{ECF4FF}10\% & \cellcolor[HTML]{ECF4FF}0.6263±{\scriptsize 0.0209} & \cellcolor[HTML]{ECF4FF}0.5747±{\scriptsize 0.0185} & \cellcolor[HTML]{ECF4FF}0.3669±{\scriptsize 0.1091} & \cellcolor[HTML]{ECF4FF}0.7825±{\scriptsize 0.0829} & \cellcolor[HTML]{ECF4FF}0.4784±{\scriptsize 0.0327} & \cellcolor[HTML]{ECF4FF}0.7024±{\scriptsize 0.0172} & \cellcolor[HTML]{ECF4FF}0.4040±{\scriptsize 0.0626} & \cellcolor[HTML]{ECF4FF}0.6180±{\scriptsize 0.0403} & - \\
 & VAAL~\cite{vaal} & 10\% & 0.6018±{\scriptsize 0.2120} & 0.5123±{\scriptsize 0.0262} & 0.1041±{\scriptsize 0.1919} & \textbf{0.9205±{\scriptsize 0.1396}} & 0.2004±{\scriptsize 0.1720} & 0.6638±{\scriptsize 0.0197} & 0.1025±{\scriptsize 0.1748} & 0.6044±{\scriptsize 0.0335} & 0.0329 ($-$) \\
 & Core-Set~\cite{core} & 10\% & 0.6625±{\scriptsize 0.2069} & \textbf{0.5805±{\scriptsize 0.0233}} & \underline{0.3802±{\scriptsize 0.1672}} & 0.7808±{\scriptsize 0.1366} & \underline{0.5044±{\scriptsize 0.0682}} & \textbf{0.7096±{\scriptsize 0.0280}} & \underline{0.4052±{\scriptsize 0.0855}} & 0.6125±{\scriptsize 0.0243} & 0.0673 ($-$) \\
 & Max-Entropy~\cite{entro} & 10\% & \underline{0.7023±{\scriptsize 0.0590}} & 0.5427±{\scriptsize 0.0287} & 0.1719±{\scriptsize 0.1091} & \underline{0.9135±{\scriptsize 0.0558}} & 0.4113±{\scriptsize 0.2098} & 0.6774±{\scriptsize 0.0165} & 0.2366±{\scriptsize 0.1350} & \underline{0.6582±{\scriptsize 0.0361}} & 0.0673 (+) \\
 & BALD~\cite{bald} & 10\% & 0.6721±{\scriptsize 0.0839} & \underline{0.5778±{\scriptsize 0.0377}} & 0.3041±{\scriptsize 0.1420} & 0.8515±{\scriptsize 0.0725} & \textbf{0.5342±{\scriptsize 0.0482}} & 0.7012±{\scriptsize 0.0250} & 0.3603±{\scriptsize 0.1367} & 0.6312±{\scriptsize 0.0451} & 0.0747 (+) \\
 & EvoDiff (Ours) & 10\% & \textbf{0.7048±{\scriptsize 0.0451}} & 0.5639±{\scriptsize 0.0136} & \textbf{0.4579±{\scriptsize 0.1113}} & 0.6699±{\scriptsize 0.1002} & 0.4268±{\scriptsize 0.0216} & \underline{0.7027±{\scriptsize 0.0168}} & \textbf{0.4339±{\scriptsize 0.0441}} & \textbf{0.6863±{\scriptsize 0.0195}} & 0.0329 (+) \\
\cmidrule(lr){2-12}
 & \cellcolor[HTML]{ECF4FF}Random & \cellcolor[HTML]{ECF4FF}20\% & \cellcolor[HTML]{ECF4FF}0.7109±{\scriptsize 0.0119} & \cellcolor[HTML]{ECF4FF}0.5530±{\scriptsize 0.0309} & \cellcolor[HTML]{ECF4FF}0.2562±{\scriptsize 0.1591} & \cellcolor[HTML]{ECF4FF}\underline{0.8498±{\scriptsize 0.1002}} & \cellcolor[HTML]{ECF4FF}0.4714±{\scriptsize 0.0460} & \cellcolor[HTML]{ECF4FF}0.6869±{\scriptsize 0.0219} & \cellcolor[HTML]{ECF4FF}0.3008±{\scriptsize 0.1467} & \cellcolor[HTML]{ECF4FF}0.6666±{\scriptsize 0.0126} & - \\
 & VAAL~\cite{vaal} & 20\% & 0.6372±{\scriptsize 0.0840} & 0.5486±{\scriptsize 0.0309} & 0.2264±{\scriptsize 0.0899} & \textbf{0.8707±{\scriptsize 0.0405}} & 0.4744±{\scriptsize 0.0709} & 0.6816±{\scriptsize 0.0186} & 0.2992±{\scriptsize 0.0878} & 0.6284±{\scriptsize 0.0388} & 0.0862 ($-$) \\
 & Core-Set~\cite{core} & 20\% & 0.6896±{\scriptsize 0.0484} & 0.5619±{\scriptsize 0.0269} & 0.2810±{\scriptsize 0.1318} & 0.8428±{\scriptsize 0.0832} & \underline{0.4922±{\scriptsize 0.0318}} & 0.6915±{\scriptsize 0.0191} & 0.3360±{\scriptsize 0.1142} & 0.6410±{\scriptsize 0.0263} & 0.0862 ($-$) \\
 & Max-Entropy~\cite{entro} & 20\% & 0.7436±{\scriptsize 0.0509} & \underline{0.5896±{\scriptsize 0.0458}} & 0.3653±{\scriptsize 0.2379} & 0.8140±{\scriptsize 0.1484} & \textbf{0.5520±{\scriptsize 0.0665}} & 0.7191±{\scriptsize 0.0459} & 0.3792±{\scriptsize 0.1578} & 0.6960±{\scriptsize 0.0163} & 0.0862 (+) \\
 & BALD~\cite{bald} & 20\% & \underline{0.7569±{\scriptsize 0.0550}} & 0.5867±{\scriptsize 0.0372} & \underline{0.4364±{\scriptsize 0.2035}} & 0.7371±{\scriptsize 0.1428} & 0.4778±{\scriptsize 0.0359} & \underline{0.7203±{\scriptsize 0.0361}} & \underline{0.4239±{\scriptsize 0.1196}} & \underline{0.7032±{\scriptsize 0.0111}} & 0.0862 (+) \\
 & EvoDiff (Ours) & 20\% & \textbf{0.7793±{\scriptsize 0.0159}} & \textbf{0.5980±{\scriptsize 0.0152}} & \textbf{0.4694±{\scriptsize 0.0698}} & 0.7266±{\scriptsize 0.0716} & 0.4822±{\scriptsize 0.0401} & \textbf{0.7223±{\scriptsize 0.0124}} & \textbf{0.4707±{\scriptsize 0.0268}} & \textbf{0.7605±{\scriptsize 0.0116}} & 0.0862 (+) \\
\cmidrule(lr){2-12}
 & \cellcolor[HTML]{ECF4FF}Random & \cellcolor[HTML]{ECF4FF}30\% & \cellcolor[HTML]{ECF4FF}0.7377±{\scriptsize 0.0314} & \cellcolor[HTML]{ECF4FF}0.5797±{\scriptsize 0.0356} & \cellcolor[HTML]{ECF4FF}0.3289±{\scriptsize 0.1212} & \cellcolor[HTML]{ECF4FF}0.8306±{\scriptsize 0.0584} & \cellcolor[HTML]{ECF4FF}0.5056±{\scriptsize 0.0291} & \cellcolor[HTML]{ECF4FF}0.7032±{\scriptsize 0.0269} & \cellcolor[HTML]{ECF4FF}0.3867±{\scriptsize 0.0903} & \cellcolor[HTML]{ECF4FF}0.6742±{\scriptsize 0.0240} & - \\
 & VAAL~\cite{vaal} & 30\% & 0.7075±{\scriptsize 0.0323} & \underline{0.5918±{\scriptsize 0.0554}} & \textbf{0.3653±{\scriptsize 0.1734}} & 0.8183±{\scriptsize 0.0752} & 0.5112±{\scriptsize 0.0460} & \textbf{0.7151±{\scriptsize 0.0461}} & \textbf{0.4061±{\scriptsize 0.1231}} & 0.6463±{\scriptsize 0.0466} & 0.0882 ($-$) \\
 & Core-Set~\cite{core} & 30\% & 0.7509±{\scriptsize 0.0322} & 0.5858±{\scriptsize 0.0260} & 0.3174±{\scriptsize 0.0912} & \underline{0.8541±{\scriptsize 0.0441}} & 0.5364±{\scriptsize 0.0232} & 0.7043±{\scriptsize 0.0173} & 0.3897±{\scriptsize 0.0826} & 0.6740±{\scriptsize 0.0139} & 0.0882 ($-$) \\
 & Max-Entropy~\cite{entro} & 30\% & 0.7682±{\scriptsize 0.0726} & 0.5751±{\scriptsize 0.0452} & \underline{0.3388±{\scriptsize 0.1744}} & 0.8114±{\scriptsize 0.1247} & \underline{0.5375±{\scriptsize 0.1106}} & 0.7025±{\scriptsize 0.0294} & 0.3684±{\scriptsize 0.1572} & \underline{0.7428±{\scriptsize 0.0415}} & 0.0797 (+) \\
 & BALD~\cite{bald} & 30\% & \underline{0.7877±{\scriptsize 0.0514}} & 0.5755±{\scriptsize 0.0307} & 0.3256±{\scriptsize 0.1591} & 0.8253±{\scriptsize 0.1050} & 0.5228±{\scriptsize 0.0542} & 0.7020±{\scriptsize 0.0241} & 0.3675±{\scriptsize 0.1255} & 0.7329±{\scriptsize 0.0152} & 0.0797 (+) \\
 & EvoDiff (Ours) & 30\% & \textbf{0.8126±{\scriptsize 0.0382}} & \textbf{0.5930±{\scriptsize 0.0232}} & 0.3240±{\scriptsize 0.0907} & \textbf{0.8620±{\scriptsize 0.0486}} & \textbf{0.5591±{\scriptsize 0.0335}} & \underline{0.7084±{\scriptsize 0.0193}} & \underline{0.4021±{\scriptsize 0.0597}} & \textbf{0.8389±{\scriptsize 0.0200}} & 0.0974 (+) \\
\bottomrule
\end{tabular}
}
\end{sidewaystable*}

\clearpage

%%%%%  S2  USFM 完整指标  %%%%%

\begin{sidewaystable*}[!ht]
\small
\centering
\renewcommand{\arraystretch}{1.2}
\setlength{\tabcolsep}{4pt}
\caption{Comparison of USFM classification metrics between EvoDiff and other AL methods on the test sets. Values are reported as mean $\pm$ s.d. over five folds. Statistical differences between each active learning strategy and Random sampling were assessed using two-sided paired $t$-tests based on fold-wise AUC. The reported $q$-values are the Benjamini--Hochberg-calibrated $p$-values across the five strategy-versus-Random comparisons within each dataset and annotation budget; the appended $(+)$ or $(-)$ indicates the sign of \texttt{mean\_diff(method-Random)} for AUC. (\textbf{Bold} represents the best result, and \underline{Underline} represents the second best result).}
\textbf{\label{usfm_landscape_metrics}}
\resizebox{\linewidth}{!}{
\begin{tabular}{l|l|c|cccccccc|c}
\toprule
\textbf{Dataset} & \textbf{Method} & \textbf{Annotation Budget} & \textbf{Acc} & \textbf{bAcc} & \textbf{Sens} & \textbf{Spec} & \textbf{PPV} & \textbf{NPV} & \textbf{F1} & \textbf{AUC} & \textbf{$q$-value} \\
\midrule
\multirow{19}{*}{Appendix~\cite{app}} & \cellcolor[HTML]{F8F8D1}FSL (upper bound) & \cellcolor[HTML]{F8F8D1}100\% & \cellcolor[HTML]{F8F8D1}0.6435±{\scriptsize 0.0248} & \cellcolor[HTML]{F8F8D1}0.5590±{\scriptsize 0.0259} & \cellcolor[HTML]{F8F8D1}0.8828±{\scriptsize 0.0308} & \cellcolor[HTML]{F8F8D1}0.2353±{\scriptsize 0.0416} & \cellcolor[HTML]{F8F8D1}0.6633±{\scriptsize 0.0144} & \cellcolor[HTML]{F8F8D1}0.5445±{\scriptsize 0.0883} & \cellcolor[HTML]{F8F8D1}0.7573±{\scriptsize 0.0180} & \cellcolor[HTML]{F8F8D1}0.6471±{\scriptsize 0.0349} & - \\
 & \cellcolor[HTML]{ECF4FF}Random & \cellcolor[HTML]{ECF4FF}10\% & \cellcolor[HTML]{ECF4FF}0.4652±{\scriptsize 0.0763} & \cellcolor[HTML]{ECF4FF}\underline{0.5150±{\scriptsize 0.0549}} & \cellcolor[HTML]{ECF4FF}0.3241±{\scriptsize 0.1592} & \cellcolor[HTML]{ECF4FF}0.7059±{\scriptsize 0.1101} & \cellcolor[HTML]{ECF4FF}\underline{0.6291±{\scriptsize 0.1195}} & \cellcolor[HTML]{ECF4FF}\underline{0.3828±{\scriptsize 0.0354}} & \cellcolor[HTML]{ECF4FF}0.4154±{\scriptsize 0.1675} & \cellcolor[HTML]{ECF4FF}\underline{0.5509±{\scriptsize 0.0271}} & - \\
 & VAAL~\cite{vaal} & 10\% & \underline{0.4870±{\scriptsize 0.0763}} & 0.5031±{\scriptsize 0.0646} & \underline{0.4414±{\scriptsize 0.1154}} & 0.5647±{\scriptsize 0.0526} & 0.6270±{\scriptsize 0.0705} & 0.3771±{\scriptsize 0.0563} & \underline{0.5153±{\scriptsize 0.0987}} & 0.5270±{\scriptsize 0.0949} & 0.0679 (\ensuremath{-}) \\
 & Core-Set~\cite{core} & 10\% & 0.4261±{\scriptsize 0.0808} & \textbf{0.5351±{\scriptsize 0.0541}} & 0.1172±{\scriptsize 0.1628} & \textbf{0.9529±{\scriptsize 0.0767}} & 0.4571±{\scriptsize 0.4534} & \textbf{0.3910±{\scriptsize 0.0331}} & 0.1740±{\scriptsize 0.2241} & 0.4769±{\scriptsize 0.0666} & 0.0317 (\ensuremath{-}) \\
 & Max-Entropy~\cite{entro} & 10\% & 0.3652±{\scriptsize 0.0097} & 0.4673±{\scriptsize 0.0155} & 0.0759±{\scriptsize 0.0288} & 0.8588±{\scriptsize 0.0526} & 0.4800±{\scriptsize 0.0447} & 0.3524±{\scriptsize 0.0083} & 0.1292±{\scriptsize 0.0446} & 0.4428±{\scriptsize 0.0466} & 0.0003 (\ensuremath{-}) \\
 & BALD~\cite{bald} & 10\% & 0.4130±{\scriptsize 0.1109} & 0.5053±{\scriptsize 0.0708} & 0.1517±{\scriptsize 0.2346} & \underline{0.8588±{\scriptsize 0.1220}} & 0.3302±{\scriptsize 0.3374} & 0.3811±{\scriptsize 0.0559} & 0.1931±{\scriptsize 0.2722} & 0.4714±{\scriptsize 0.0968} & 0.0333 (\ensuremath{-}) \\
 & EvoDiff (Ours) & 10\% & \textbf{0.5261±{\scriptsize 0.0563}} & 0.5000±{\scriptsize 0.0665} & \textbf{0.6000±{\scriptsize 0.1413}} & 0.4000±{\scriptsize 0.2096} & \textbf{0.6370±{\scriptsize 0.0535}} & 0.3630±{\scriptsize 0.0930} & \textbf{0.6091±{\scriptsize 0.0677}} & \textbf{0.5748±{\scriptsize 0.1012}} & 0.0679 (+) \\
\cmidrule(lr){2-12}
 & \cellcolor[HTML]{ECF4FF}Random & \cellcolor[HTML]{ECF4FF}20\% & \cellcolor[HTML]{ECF4FF}0.5609±{\scriptsize 0.0238} & \cellcolor[HTML]{ECF4FF}0.5592±{\scriptsize 0.0376} & \cellcolor[HTML]{ECF4FF}0.5655±{\scriptsize 0.0463} & \cellcolor[HTML]{ECF4FF}\textbf{0.5530±{\scriptsize 0.1069}} & \cellcolor[HTML]{ECF4FF}\underline{0.6866±{\scriptsize 0.0376}} & \cellcolor[HTML]{ECF4FF}0.4248±{\scriptsize 0.0337} & \cellcolor[HTML]{ECF4FF}0.6183±{\scriptsize 0.0220} & \cellcolor[HTML]{ECF4FF}0.5655±{\scriptsize 0.0566} & - \\
 & VAAL~\cite{vaal} & 20\% & 0.5609±{\scriptsize 0.0418} & 0.5105±{\scriptsize 0.0388} & \underline{0.7034±{\scriptsize 0.1133}} & 0.3176±{\scriptsize 0.1354} & 0.6388±{\scriptsize 0.0279} & 0.3873±{\scriptsize 0.0596} & 0.6656±{\scriptsize 0.0528} & 0.5448±{\scriptsize 0.0330} & 0.0750 (\ensuremath{-}) \\
 & Core-Set~\cite{core} & 20\% & \textbf{0.6174±{\scriptsize 0.0291}} & \underline{0.5748±{\scriptsize 0.0368}} & \textbf{0.7379±{\scriptsize 0.0393}} & 0.4117±{\scriptsize 0.0832} & 0.6826±{\scriptsize 0.0282} & \underline{0.4781±{\scriptsize 0.0472}} & \textbf{0.7085±{\scriptsize 0.0220}} & \underline{0.5951±{\scriptsize 0.0699}} & 0.0750 (+) \\
 & Max-Entropy~\cite{entro} & 20\% & \underline{0.5652±{\scriptsize 0.0266}} & 0.5408±{\scriptsize 0.0497} & 0.6345±{\scriptsize 0.1106} & 0.4471±{\scriptsize 0.1933} & 0.6696±{\scriptsize 0.0470} & 0.4057±{\scriptsize 0.0656} & 0.6443±{\scriptsize 0.0455} & 0.5586±{\scriptsize 0.0837} & 0.0913 (\ensuremath{-}) \\
 & BALD~\cite{bald} & 20\% & 0.5000±{\scriptsize 0.0461} & 0.4793±{\scriptsize 0.0516} & 0.5586±{\scriptsize 0.0859} & 0.4000±{\scriptsize 0.1275} & 0.6155±{\scriptsize 0.0392} & 0.3440±{\scriptsize 0.0646} & 0.5825±{\scriptsize 0.0518} & 0.4836±{\scriptsize 0.0438} & 0.0015 (\ensuremath{-}) \\
 & EvoDiff (Ours) & 20\% & \textbf{0.6174±{\scriptsize 0.0663}} & \textbf{0.5992±{\scriptsize 0.0673}} & 0.6690±{\scriptsize 0.1107} & \underline{0.5294±{\scriptsize 0.1380}} & \textbf{0.7101±{\scriptsize 0.0493}} & \textbf{0.4881±{\scriptsize 0.0999}} & \underline{0.6850±{\scriptsize 0.0662}} & \textbf{0.6381±{\scriptsize 0.0491}} & 0.0235 (+) \\
\cmidrule(lr){2-12}
 & \cellcolor[HTML]{ECF4FF}Random & \cellcolor[HTML]{ECF4FF}30\% & \cellcolor[HTML]{ECF4FF}\underline{0.6478±{\scriptsize 0.0283}} & \cellcolor[HTML]{ECF4FF}\underline{0.6136±{\scriptsize 0.0320}} & \cellcolor[HTML]{ECF4FF}0.7448±{\scriptsize 0.1079} & \cellcolor[HTML]{ECF4FF}0.4823±{\scriptsize 0.1465} & \cellcolor[HTML]{ECF4FF}0.7139±{\scriptsize 0.0310} & \cellcolor[HTML]{ECF4FF}\underline{0.5445±{\scriptsize 0.0748}} & \cellcolor[HTML]{ECF4FF}\underline{0.7248±{\scriptsize 0.0384}} & \cellcolor[HTML]{ECF4FF}\underline{0.6783±{\scriptsize 0.0607}} & - \\
 & VAAL~\cite{vaal} & 30\% & 0.5870±{\scriptsize 0.0407} & 0.5434±{\scriptsize 0.0525} & 0.7104±{\scriptsize 0.0462} & 0.3765±{\scriptsize 0.1147} & 0.6619±{\scriptsize 0.0352} & 0.4240±{\scriptsize 0.0834} & 0.6843±{\scriptsize 0.0291} & 0.5574±{\scriptsize 0.0639} & 0.0852 (\ensuremath{-}) \\
 & Core-Set~\cite{core} & 30\% & 0.5869±{\scriptsize 0.0407} & 0.5240±{\scriptsize 0.0338} & \underline{0.7655±{\scriptsize 0.0707}} & 0.2823±{\scriptsize 0.0492} & 0.6448±{\scriptsize 0.0201} & 0.4204±{\scriptsize 0.0728} & 0.6992±{\scriptsize 0.0400} & 0.5562±{\scriptsize 0.0506} & 0.0029 (\ensuremath{-}) \\
 & Max-Entropy~\cite{entro} & 30\% & 0.6391±{\scriptsize 0.0195} & 0.6116±{\scriptsize 0.0307} & 0.7172±{\scriptsize 0.0511} & \underline{0.5059±{\scriptsize 0.0984}} & \underline{0.7146±{\scriptsize 0.0310}} & 0.5117±{\scriptsize 0.0289} & 0.7143±{\scriptsize 0.0182} & 0.6385±{\scriptsize 0.0501} & 0.0166 (\ensuremath{-}) \\
 & BALD~\cite{bald} & 30\% & 0.5261±{\scriptsize 0.1164} & 0.5243±{\scriptsize 0.0813} & 0.5310±{\scriptsize 0.2421} & \textbf{0.5176±{\scriptsize 0.1523}} & 0.6285±{\scriptsize 0.0887} & 0.4126±{\scriptsize 0.1039} & 0.5577±{\scriptsize 0.2048} & 0.5229±{\scriptsize 0.0675} & 0.0029 (\ensuremath{-}) \\
 & EvoDiff (Ours) & 30\% & \textbf{0.7000±{\scriptsize 0.0519}} & \textbf{0.6574±{\scriptsize 0.0494}} & \textbf{0.8207±{\scriptsize 0.0787}} & 0.4941±{\scriptsize 0.0789} & \textbf{0.7347±{\scriptsize 0.0313}} & \textbf{0.6289±{\scriptsize 0.1014}} & \textbf{0.7741±{\scriptsize 0.0455}} & \textbf{0.6868±{\scriptsize 0.0976}} & 0.0123 (+) \\
\midrule
\multirow{19}{*}{BUSI~\cite{BUSI}} & \cellcolor[HTML]{F8F8D1}FSL (upper bound) & \cellcolor[HTML]{F8F8D1}100\% & \cellcolor[HTML]{F8F8D1}0.8273±{\scriptsize 0.2852} & \cellcolor[HTML]{F8F8D1}0.8505±{\scriptsize 0.1988} & \cellcolor[HTML]{F8F8D1}0.9143±{\scriptsize 0.0916} & \cellcolor[HTML]{F8F8D1}0.7867±{\scriptsize 0.4402} & \cellcolor[HTML]{F8F8D1}0.8359±{\scriptsize 0.2919} & \cellcolor[HTML]{F8F8D1}0.7625±{\scriptsize 0.4274} & \cellcolor[HTML]{F8F8D1}0.8361±{\scriptsize 0.2006} & \cellcolor[HTML]{F8F8D1}0.9131±{\scriptsize 0.1423} & - \\
 & \cellcolor[HTML]{ECF4FF}Random & \cellcolor[HTML]{ECF4FF}10\% & \cellcolor[HTML]{ECF4FF}\textbf{0.8424±{\scriptsize 0.0395}} & \cellcolor[HTML]{ECF4FF}\textbf{0.7981±{\scriptsize 0.0480}} & \cellcolor[HTML]{ECF4FF}\underline{0.6762±{\scriptsize 0.0916}} & \cellcolor[HTML]{ECF4FF}0.9200±{\scriptsize 0.0461} & \cellcolor[HTML]{ECF4FF}\textbf{0.8051±{\scriptsize 0.0978}} & \cellcolor[HTML]{ECF4FF}\underline{0.8601±{\scriptsize 0.0353}} & \cellcolor[HTML]{ECF4FF}\textbf{0.7312±{\scriptsize 0.0697}} & \cellcolor[HTML]{ECF4FF}\textbf{0.8873±{\scriptsize 0.0336}} & - \\
 & VAAL~\cite{vaal} & 10\% & 0.7485±{\scriptsize 0.0395} & 0.6378±{\scriptsize 0.0573} & 0.3333±{\scriptsize 0.1117} & \textbf{0.9422±{\scriptsize 0.0254}} & 0.7271±{\scriptsize 0.1145} & 0.7528±{\scriptsize 0.0333} & 0.4514±{\scriptsize 0.1110} & 0.7874±{\scriptsize 0.0482} & 0.0088 (\ensuremath{-}) \\
 & Core-Set~\cite{core} & 10\% & 0.7879±{\scriptsize 0.0214} & 0.7378±{\scriptsize 0.0297} & 0.6000±{\scriptsize 0.0543} & 0.8756±{\scriptsize 0.0122} & 0.6916±{\scriptsize 0.0329} & 0.8246±{\scriptsize 0.0203} & 0.6421±{\scriptsize 0.0435} & 0.8493±{\scriptsize 0.0102} & 0.0664 (\ensuremath{-}) \\
 & Max-Entropy~\cite{entro} & 10\% & 0.7636±{\scriptsize 0.0747} & 0.6641±{\scriptsize 0.1505} & 0.3905±{\scriptsize 0.3602} & \underline{0.9378±{\scriptsize 0.0617}} & 0.4502±{\scriptsize 0.4122} & 0.7810±{\scriptsize 0.0919} & 0.4162±{\scriptsize 0.3803} & 0.7893±{\scriptsize 0.0625} & 0.0095 (\ensuremath{-}) \\
 & BALD~\cite{bald} & 10\% & 0.7091±{\scriptsize 0.2197} & 0.6749±{\scriptsize 0.1028} & 0.5810±{\scriptsize 0.2414} & 0.7689±{\scriptsize 0.4299} & \underline{0.7429±{\scriptsize 0.2412}} & 0.6381±{\scriptsize 0.3572} & 0.5838±{\scriptsize 0.0809} & 0.8329±{\scriptsize 0.0848} & 0.0356 (\ensuremath{-}) \\
 & EvoDiff (Ours) & 10\% & \underline{0.8182±{\scriptsize 0.0214}} & \underline{0.7956±{\scriptsize 0.0307}} & \textbf{0.7333±{\scriptsize 0.0987}} & 0.8578±{\scriptsize 0.0558} & 0.7149±{\scriptsize 0.0600} & \textbf{0.8757±{\scriptsize 0.0332}} & \underline{0.7179±{\scriptsize 0.0406}} & \underline{0.8788±{\scriptsize 0.0259}} & 0.0154 (\ensuremath{-}) \\
\cmidrule(lr){2-12}
 & \cellcolor[HTML]{ECF4FF}Random & \cellcolor[HTML]{ECF4FF}20\% & \cellcolor[HTML]{ECF4FF}0.8485±{\scriptsize 0.0303} & \cellcolor[HTML]{ECF4FF}0.7975±{\scriptsize 0.0358} & \cellcolor[HTML]{ECF4FF}0.6572±{\scriptsize 0.0783} & \cellcolor[HTML]{ECF4FF}0.9378±{\scriptsize 0.0456} & \cellcolor[HTML]{ECF4FF}0.8432±{\scriptsize 0.0940} & \cellcolor[HTML]{ECF4FF}0.8551±{\scriptsize 0.0250} & \cellcolor[HTML]{ECF4FF}0.7332±{\scriptsize 0.0550} & \cellcolor[HTML]{ECF4FF}0.8534±{\scriptsize 0.0290} & - \\
 & VAAL~\cite{vaal} & 20\% & 0.7758±{\scriptsize 0.0249} & 0.6603±{\scriptsize 0.0362} & 0.3429±{\scriptsize 0.0706} & \textbf{0.9778±{\scriptsize 0.0157}} & \underline{0.8800±{\scriptsize 0.0829}} & 0.7616±{\scriptsize 0.0196} & 0.4902±{\scriptsize 0.0791} & 0.8456±{\scriptsize 0.0222} & 0.0535 (\ensuremath{-}) \\
 & Core-Set~\cite{core} & 20\% & \underline{0.8848±{\scriptsize 0.0203}} & \underline{0.8571±{\scriptsize 0.0270}} & \underline{0.7809±{\scriptsize 0.0543}} & 0.9333±{\scriptsize 0.0223} & 0.8470±{\scriptsize 0.0426} & \underline{0.9018±{\scriptsize 0.0220}} & \underline{0.8115±{\scriptsize 0.0353}} & \underline{0.9089±{\scriptsize 0.0249}} & 0.0030 (+) \\
 & Max-Entropy~\cite{entro} & 20\% & 0.7818±{\scriptsize 0.0854} & 0.7003±{\scriptsize 0.1728} & 0.4762±{\scriptsize 0.4138} & 0.9245±{\scriptsize 0.0696} & 0.6455±{\scriptsize 0.3781} & 0.8112±{\scriptsize 0.1138} & 0.4740±{\scriptsize 0.3930} & 0.8852±{\scriptsize 0.0204} & 0.0162 (+) \\
 & BALD~\cite{bald} & 20\% & 0.8636±{\scriptsize 0.0355} & 0.8238±{\scriptsize 0.0345} & 0.7143±{\scriptsize 0.0476} & 0.9334±{\scriptsize 0.0445} & 0.8409±{\scriptsize 0.0923} & 0.8750±{\scriptsize 0.0200} & 0.7702±{\scriptsize 0.0517} & 0.8913±{\scriptsize 0.0081} & 0.0017 (+) \\
 & EvoDiff (Ours) & 20\% & \textbf{0.9121±{\scriptsize 0.0311}} & \textbf{0.8924±{\scriptsize 0.0495}} & \textbf{0.8381±{\scriptsize 0.1043}} & \underline{0.9467±{\scriptsize 0.0199}} & \textbf{0.8811±{\scriptsize 0.0331}} & \textbf{0.9279±{\scriptsize 0.0417}} & \textbf{0.8560±{\scriptsize 0.0606}} & \textbf{0.9302±{\scriptsize 0.0236}} & 0.0030 (+) \\
\cmidrule(lr){2-12}
 & \cellcolor[HTML]{ECF4FF}Random & \cellcolor[HTML]{ECF4FF}30\% & \cellcolor[HTML]{ECF4FF}\textbf{0.9060±{\scriptsize 0.0249}} & \cellcolor[HTML]{ECF4FF}\underline{0.8600±{\scriptsize 0.0327}} & \cellcolor[HTML]{ECF4FF}\underline{0.7333±{\scriptsize 0.0543}} & \cellcolor[HTML]{ECF4FF}\textbf{0.9867±{\scriptsize 0.0122}} & \cellcolor[HTML]{ECF4FF}\textbf{0.9617±{\scriptsize 0.0350}} & \cellcolor[HTML]{ECF4FF}0.8882±{\scriptsize 0.0214} & \cellcolor[HTML]{ECF4FF}\underline{0.8318±{\scriptsize 0.0470}} & \cellcolor[HTML]{ECF4FF}0.8959±{\scriptsize 0.0104} & - \\
 & VAAL~\cite{vaal} & 30\% & 0.8636±{\scriptsize 0.0152} & 0.8060±{\scriptsize 0.0215} & 0.6476±{\scriptsize 0.0426} & 0.9645±{\scriptsize 0.0122} & \underline{0.8955±{\scriptsize 0.0330}} & 0.8545±{\scriptsize 0.0152} & 0.7509±{\scriptsize 0.0313} & \underline{0.9146±{\scriptsize 0.0094}} & 0.0046 (+) \\
 & Core-Set~\cite{core} & 30\% & \underline{0.8939±{\scriptsize 0.1193}} & 0.8461±{\scriptsize 0.1939} & 0.7143±{\scriptsize 0.3998} & \underline{0.9778±{\scriptsize 0.0222}} & 0.7514±{\scriptsize 0.4219} & \underline{0.8973±{\scriptsize 0.1208}} & 0.7319±{\scriptsize 0.4097} & 0.9107±{\scriptsize 0.0685} & 0.0852 (+) \\
 & Max-Entropy~\cite{entro} & 30\% & 0.8485±{\scriptsize 0.0240} & 0.8152±{\scriptsize 0.0170} & 0.7238±{\scriptsize 0.0521} & 0.9067±{\scriptsize 0.0507} & 0.7926±{\scriptsize 0.0752} & 0.8763±{\scriptsize 0.0172} & 0.7531±{\scriptsize 0.0275} & 0.8684±{\scriptsize 0.0178} & 0.0072 (\ensuremath{-}) \\
 & BALD~\cite{bald} & 30\% & 0.8515±{\scriptsize 0.0990} & 0.7895±{\scriptsize 0.1640} & 0.6190±{\scriptsize 0.3515} & 0.9600±{\scriptsize 0.0553} & 0.7164±{\scriptsize 0.4141} & 0.8572±{\scriptsize 0.1002} & 0.6590±{\scriptsize 0.3707} & 0.9007±{\scriptsize 0.0526} & 0.0033 (+) \\
 & EvoDiff (Ours) & 30\% & 0.8849±{\scriptsize 0.0449} & \textbf{0.8927±{\scriptsize 0.0273}} & \textbf{0.9143±{\scriptsize 0.0399}} & 0.8711±{\scriptsize 0.0792} & 0.7816±{\scriptsize 0.1020} & \textbf{0.9572±{\scriptsize 0.0166}} & \textbf{0.8381±{\scriptsize 0.0493}} & \textbf{0.9533±{\scriptsize 0.0189}} & 0.0861 (+) \\
\bottomrule
\end{tabular}
}
\end{sidewaystable*}

\begin{sidewaystable*}[!ht]
\small
\centering
\renewcommand{\arraystretch}{1.2}
\setlength{\tabcolsep}{4pt}
\ContinuedFloat
\caption{Comparison of USFM classification metrics between EvoDiff and other AL methods on the test sets. Values are reported as mean $\pm$ s.d. over five folds. Statistical differences between each active learning strategy and Random sampling were assessed using two-sided paired $t$-tests based on fold-wise AUC. The reported $q$-values are the Benjamini--Hochberg-calibrated $p$-values across the five strategy-versus-Random comparisons within each dataset and annotation budget; the appended $(+)$ or $(-)$ indicates the sign of \texttt{mean\_diff(method-Random)} for AUC. (\textbf{Bold} represents the best result, and \underline{Underline} represents the second best result).}
\resizebox{\linewidth}{!}{
\begin{tabular}{l|l|c|cccccccc|c}
\toprule
\textbf{Dataset} & \textbf{Method} & \textbf{Annotation Budget} & \textbf{Acc} & \textbf{bAcc} & \textbf{Sens} & \textbf{Spec} & \textbf{PPV} & \textbf{NPV} & \textbf{F1} & \textbf{AUC} & \textbf{$q$-value} \\
\midrule
\multirow{19}{*}{BUS-BRA~\cite{BUS-BRA}} & \cellcolor[HTML]{F8F8D1}FSL (upper bound) & \cellcolor[HTML]{F8F8D1}100\% & \cellcolor[HTML]{F8F8D1}0.9263±{\scriptsize 0.0378} & \cellcolor[HTML]{F8F8D1}0.8937±{\scriptsize 0.0537} & \cellcolor[HTML]{F8F8D1}0.8000±{\scriptsize 0.0998} & \cellcolor[HTML]{F8F8D1}0.9875±{\scriptsize 0.0089} & \cellcolor[HTML]{F8F8D1}0.9669±{\scriptsize 0.0262} & \cellcolor[HTML]{F8F8D1}0.9119±{\scriptsize 0.0403} & \cellcolor[HTML]{F8F8D1}0.8737±{\scriptsize 0.0715} & \cellcolor[HTML]{F8F8D1}0.9459±{\scriptsize 0.0214} & - \\
 & \cellcolor[HTML]{ECF4FF}Random & \cellcolor[HTML]{ECF4FF}10\% & \cellcolor[HTML]{ECF4FF}\textbf{0.7810±{\scriptsize 0.0246}} & \cellcolor[HTML]{ECF4FF}\underline{0.7294±{\scriptsize 0.0358}} & \cellcolor[HTML]{ECF4FF}\underline{0.5807±{\scriptsize 0.0782}} & \cellcolor[HTML]{ECF4FF}0.8781±{\scriptsize 0.0268} & \cellcolor[HTML]{ECF4FF}\underline{0.6990±{\scriptsize 0.0441}} & \cellcolor[HTML]{ECF4FF}\underline{0.8130±{\scriptsize 0.0267}} & \cellcolor[HTML]{ECF4FF}\underline{0.6319±{\scriptsize 0.0541}} & \cellcolor[HTML]{ECF4FF}\underline{0.7931±{\scriptsize 0.0518}} & - \\
 & VAAL~\cite{vaal} & 10\% & 0.7084±{\scriptsize 0.0231} & 0.6123±{\scriptsize 0.0248} & 0.3355±{\scriptsize 0.0350} & 0.8891±{\scriptsize 0.0237} & 0.5962±{\scriptsize 0.0675} & 0.7342±{\scriptsize 0.0132} & 0.4288±{\scriptsize 0.0431} & 0.7164±{\scriptsize 0.0168} & 0.0002 (\ensuremath{-}) \\
 & Core-Set~\cite{core} & 10\% & \underline{0.7758±{\scriptsize 0.0466}} & 0.7213±{\scriptsize 0.0602} & 0.5645±{\scriptsize 0.1076} & 0.8781±{\scriptsize 0.0357} & 0.6916±{\scriptsize 0.0747} & 0.8075±{\scriptsize 0.0415} & 0.6189±{\scriptsize 0.0875} & 0.7279±{\scriptsize 0.0311} & 0.0249 (\ensuremath{-}) \\
 & Max-Entropy~\cite{entro} & 10\% & 0.7179±{\scriptsize 0.0151} & 0.6135±{\scriptsize 0.0251} & 0.3129±{\scriptsize 0.0708} & \underline{0.9141±{\scriptsize 0.0322}} & 0.6441±{\scriptsize 0.0544} & 0.7336±{\scriptsize 0.0154} & 0.4162±{\scriptsize 0.0629} & 0.7864±{\scriptsize 0.0443} & 0.0175 (\ensuremath{-}) \\
 & BALD~\cite{bald} & 10\% & 0.7590±{\scriptsize 0.0126} & 0.6756±{\scriptsize 0.0247} & 0.4355±{\scriptsize 0.0730} & \textbf{0.9156±{\scriptsize 0.0309}} & \textbf{0.7215±{\scriptsize 0.0584}} & 0.7707±{\scriptsize 0.0183} & 0.5383±{\scriptsize 0.0493} & 0.7430±{\scriptsize 0.0363} & 0.0012 (\ensuremath{-}) \\
 & EvoDiff (Ours) & 10\% & 0.7705±{\scriptsize 0.0259} & \textbf{0.7324±{\scriptsize 0.0395}} & \textbf{0.6226±{\scriptsize 0.0827}} & 0.8422±{\scriptsize 0.0178} & 0.6554±{\scriptsize 0.0316} & \textbf{0.8227±{\scriptsize 0.0318}} & \textbf{0.6372±{\scriptsize 0.0546}} & \textbf{0.8277±{\scriptsize 0.0146}} & 0.0012 (+) \\
\cmidrule(lr){2-12}
 & \cellcolor[HTML]{ECF4FF}Random & \cellcolor[HTML]{ECF4FF}20\% & \cellcolor[HTML]{ECF4FF}0.8231±{\scriptsize 0.0704} & \cellcolor[HTML]{ECF4FF}0.7681±{\scriptsize 0.1283} & \cellcolor[HTML]{ECF4FF}0.6097±{\scriptsize 0.2964} & \cellcolor[HTML]{ECF4FF}\textbf{0.9266±{\scriptsize 0.0437}} & \cellcolor[HTML]{ECF4FF}\textbf{0.8377±{\scriptsize 0.0954}} & \cellcolor[HTML]{ECF4FF}0.8416±{\scriptsize 0.0842} & \cellcolor[HTML]{ECF4FF}0.6446±{\scriptsize 0.2777} & \cellcolor[HTML]{ECF4FF}0.8684±{\scriptsize 0.0850} & - \\
 & VAAL~\cite{vaal} & 20\% & 0.7484±{\scriptsize 0.0176} & 0.6902±{\scriptsize 0.0234} & 0.5226±{\scriptsize 0.0762} & 0.8578±{\scriptsize 0.0457} & 0.6471±{\scriptsize 0.0558} & 0.7887±{\scriptsize 0.0206} & 0.5736±{\scriptsize 0.0383} & 0.7491±{\scriptsize 0.0252} & 0.0233 (\ensuremath{-}) \\
 & Core-Set~\cite{core} & 20\% & 0.8116±{\scriptsize 0.0198} & 0.7953±{\scriptsize 0.0132} & \underline{0.7484±{\scriptsize 0.0577}} & 0.8422±{\scriptsize 0.0514} & 0.7031±{\scriptsize 0.0583} & 0.8747±{\scriptsize 0.0201} & 0.7218±{\scriptsize 0.0180} & 0.8681±{\scriptsize 0.0126} & 0.0992 (\ensuremath{-}) \\
 & Max-Entropy~\cite{entro} & 20\% & 0.8379±{\scriptsize 0.0233} & 0.8015±{\scriptsize 0.0269} & 0.6968±{\scriptsize 0.0385} & 0.9063±{\scriptsize 0.0183} & 0.7828±{\scriptsize 0.0403} & 0.8606±{\scriptsize 0.0173} & 0.7371±{\scriptsize 0.0378} & 0.8668±{\scriptsize 0.0298} & 0.0992 (\ensuremath{-}) \\
 & BALD~\cite{bald} & 20\% & \textbf{0.8526±{\scriptsize 0.0118}} & \underline{0.8224±{\scriptsize 0.0147}} & 0.7355±{\scriptsize 0.0315} & \underline{0.9094±{\scriptsize 0.0162}} & \underline{0.7981±{\scriptsize 0.0273}} & \underline{0.8767±{\scriptsize 0.0122}} & \underline{0.7650±{\scriptsize 0.0194}} & \underline{0.8821±{\scriptsize 0.0201}} & 0.0992 (+) \\
 & EvoDiff (Ours) & 20\% & \underline{0.8495±{\scriptsize 0.0184}} & \textbf{0.8259±{\scriptsize 0.0217}} & \textbf{0.7581±{\scriptsize 0.0655}} & 0.8938±{\scriptsize 0.0419} & 0.7819±{\scriptsize 0.0601} & \textbf{0.8852±{\scriptsize 0.0245}} & \textbf{0.7664±{\scriptsize 0.0262}} & \textbf{0.8987±{\scriptsize 0.0137}} & 0.0992 (+) \\
\cmidrule(lr){2-12}
 & \cellcolor[HTML]{ECF4FF}Random & \cellcolor[HTML]{ECF4FF}30\% & \cellcolor[HTML]{ECF4FF}\underline{0.8452±{\scriptsize 0.0765}} & \cellcolor[HTML]{ECF4FF}0.7845±{\scriptsize 0.1261} & \cellcolor[HTML]{ECF4FF}0.6097±{\scriptsize 0.2702} & \cellcolor[HTML]{ECF4FF}\underline{0.9594±{\scriptsize 0.0256}} & \cellcolor[HTML]{ECF4FF}\textbf{0.8822±{\scriptsize 0.0373}} & \cellcolor[HTML]{ECF4FF}0.8437±{\scriptsize 0.0802} & \cellcolor[HTML]{ECF4FF}0.6830±{\scriptsize 0.2567} & \cellcolor[HTML]{ECF4FF}\underline{0.8699±{\scriptsize 0.0896}} & - \\
 & VAAL~\cite{vaal} & 30\% & 0.7990±{\scriptsize 0.0180} & 0.7726±{\scriptsize 0.0374} & \underline{0.6968±{\scriptsize 0.0957}} & 0.8484±{\scriptsize 0.0238} & 0.6904±{\scriptsize 0.0134} & 0.8543±{\scriptsize 0.0341} & \underline{0.6906±{\scriptsize 0.0517}} & 0.8441±{\scriptsize 0.0236} & 0.0715 (\ensuremath{-}) \\
 & Core-Set~\cite{core} & 30\% & \textbf{0.8811±{\scriptsize 0.0192}} & \textbf{0.8760±{\scriptsize 0.0183}} & \textbf{0.8613±{\scriptsize 0.0184}} & 0.8906±{\scriptsize 0.0221} & 0.7931±{\scriptsize 0.0354} & \textbf{0.9298±{\scriptsize 0.0098}} & \textbf{0.8256±{\scriptsize 0.0256}} & \textbf{0.9237±{\scriptsize 0.0123}} & 0.0647 (+) \\
 & Max-Entropy~\cite{entro} & 30\% & 0.7590±{\scriptsize 0.1113} & 0.6481±{\scriptsize 0.1932} & 0.3290±{\scriptsize 0.4299} & \textbf{0.9672±{\scriptsize 0.0470}} & 0.5300±{\scriptsize 0.4900} & 0.7679±{\scriptsize 0.1249} & 0.3431±{\scriptsize 0.4303} & 0.8208±{\scriptsize 0.0926} & 0.0647 (\ensuremath{-}) \\
 & BALD~\cite{bald} & 30\% & 0.8242±{\scriptsize 0.0847} & \underline{0.7847±{\scriptsize 0.1595}} & 0.6710±{\scriptsize 0.3760} & 0.8984±{\scriptsize 0.0597} & 0.6104±{\scriptsize 0.3421} & \underline{0.8692±{\scriptsize 0.1099}} & 0.6388±{\scriptsize 0.3573} & 0.8596±{\scriptsize 0.1242} & 0.0903 (\ensuremath{-}) \\
 & EvoDiff (Ours) & 30\% & 0.8316±{\scriptsize 0.0770} & 0.7744±{\scriptsize 0.1335} & 0.6097±{\scriptsize 0.2986} & 0.9391±{\scriptsize 0.0396} & \underline{0.8405±{\scriptsize 0.0680}} & 0.8449±{\scriptsize 0.1000} & 0.6662±{\scriptsize 0.2173} & 0.8489±{\scriptsize 0.0960} & 0.0715 (\ensuremath{-}) \\
\midrule
\multirow{19}{*}{UDIAT~\cite{UDIAT}} & \cellcolor[HTML]{F8F8D1}FSL (upper bound) & \cellcolor[HTML]{F8F8D1}100\% & \cellcolor[HTML]{F8F8D1}0.8555±{\scriptsize 0.1501} & \cellcolor[HTML]{F8F8D1}0.7917±{\scriptsize 0.2320} & \cellcolor[HTML]{F8F8D1}0.6000±{\scriptsize 0.4802} & \cellcolor[HTML]{F8F8D1}0.9833±{\scriptsize 0.0373} & \cellcolor[HTML]{F8F8D1}0.7714±{\scriptsize 0.4357} & \cellcolor[HTML]{F8F8D1}0.8591±{\scriptsize 0.1615} & \cellcolor[HTML]{F8F8D1}0.6236±{\scriptsize 0.4516} & \cellcolor[HTML]{F8F8D1}0.9736±{\scriptsize 0.0400} & - \\
 & \cellcolor[HTML]{ECF4FF}Random & \cellcolor[HTML]{ECF4FF}10\% & \cellcolor[HTML]{ECF4FF}\underline{0.7222±{\scriptsize 0.0393}} & \cellcolor[HTML]{ECF4FF}0.5833±{\scriptsize 0.0589} & \cellcolor[HTML]{ECF4FF}0.1667±{\scriptsize 0.1178} & \cellcolor[HTML]{ECF4FF}\textbf{1.0000±{\scriptsize 0.0000}} & \cellcolor[HTML]{ECF4FF}\underline{0.8000±{\scriptsize 0.4472}} & \cellcolor[HTML]{ECF4FF}0.7069±{\scriptsize 0.0295} & \cellcolor[HTML]{ECF4FF}0.2714±{\scriptsize 0.1779} & \cellcolor[HTML]{ECF4FF}0.7139±{\scriptsize 0.0848} & - \\
 & VAAL~\cite{vaal} & 10\% & \underline{0.7222±{\scriptsize 0.0393}} & 0.6000±{\scriptsize 0.0632} & 0.2333±{\scriptsize 0.1491} & \underline{0.9667±{\scriptsize 0.0456}} & 0.6667±{\scriptsize 0.4082} & 0.7178±{\scriptsize 0.0327} & 0.3349±{\scriptsize 0.2036} & 0.7139±{\scriptsize 0.1014} & 0.1000 \\
 & Core-Set~\cite{core} & 10\% & 0.7000±{\scriptsize 0.1601} & \underline{0.6417±{\scriptsize 0.1003}} & \underline{0.4667±{\scriptsize 0.2472}} & 0.8167±{\scriptsize 0.3249} & 0.7714±{\scriptsize 0.3170} & \underline{0.7540±{\scriptsize 0.0451}} & \underline{0.5038±{\scriptsize 0.1669}} & \textbf{0.7417±{\scriptsize 0.0578}} & 0.1000 (+) \\
 & Max-Entropy~\cite{entro} & 10\% & 0.6778±{\scriptsize 0.0248} & 0.5167±{\scriptsize 0.0373} & 0.0333±{\scriptsize 0.0746} & \textbf{1.0000±{\scriptsize 0.0000}} & 0.2000±{\scriptsize 0.4472} & 0.6745±{\scriptsize 0.0175} & 0.0571±{\scriptsize 0.1278} & 0.5917±{\scriptsize 0.0837} & 0.0555 (\ensuremath{-}) \\
 & BALD~\cite{bald} & 10\% & \underline{0.7222±{\scriptsize 0.0680}} & 0.5833±{\scriptsize 0.1021} & 0.1667±{\scriptsize 0.2041} & \textbf{1.0000±{\scriptsize 0.0000}} & 0.6000±{\scriptsize 0.5477} & 0.7090±{\scriptsize 0.0545} & 0.2476±{\scriptsize 0.2744} & \underline{0.7278±{\scriptsize 0.0854}} & 0.1000 (+) \\
 & EvoDiff (Ours) & 10\% & \textbf{0.7667±{\scriptsize 0.0913}} & \textbf{0.7000±{\scriptsize 0.0685}} & \textbf{0.5000±{\scriptsize 0.1667}} & 0.9000±{\scriptsize 0.1807} & \textbf{0.8489±{\scriptsize 0.2421}} & \textbf{0.7848±{\scriptsize 0.0402}} & \textbf{0.5855±{\scriptsize 0.1049}} & 0.7056±{\scriptsize 0.0690} & 0.1000 (\ensuremath{-}) \\
\cmidrule(lr){2-12}
 & \cellcolor[HTML]{ECF4FF}Random & \cellcolor[HTML]{ECF4FF}20\% & \cellcolor[HTML]{ECF4FF}\textbf{0.7889±{\scriptsize 0.0465}} & \cellcolor[HTML]{ECF4FF}\underline{0.6917±{\scriptsize 0.0812}} & \cellcolor[HTML]{ECF4FF}0.4000±{\scriptsize 0.1900} & \cellcolor[HTML]{ECF4FF}\textbf{0.9833±{\scriptsize 0.0373}} & \cellcolor[HTML]{ECF4FF}\textbf{0.9600±{\scriptsize 0.0894}} & \cellcolor[HTML]{ECF4FF}0.7704±{\scriptsize 0.0539} & \cellcolor[HTML]{ECF4FF}0.5359±{\scriptsize 0.1724} & \cellcolor[HTML]{ECF4FF}\underline{0.8667±{\scriptsize 0.0656}} & - \\
 & VAAL~\cite{vaal} & 20\% & \underline{0.7778±{\scriptsize 0.0393}} & \textbf{0.7417±{\scriptsize 0.0186}} & \underline{0.6334±{\scriptsize 0.0746}} & 0.8500±{\scriptsize 0.0913} & 0.7143±{\scriptsize 0.1650} & \underline{0.8236±{\scriptsize 0.0147}} & \textbf{0.6564±{\scriptsize 0.0229}} & 0.8222±{\scriptsize 0.0465} & 0.0654 (\ensuremath{-}) \\
 & Core-Set~\cite{core} & 20\% & 0.7667±{\scriptsize 0.0465} & 0.6833±{\scriptsize 0.0864} & 0.4333±{\scriptsize 0.2236} & 0.9333±{\scriptsize 0.0697} & 0.8267±{\scriptsize 0.1673} & 0.7737±{\scriptsize 0.0624} & 0.5248±{\scriptsize 0.1771} & 0.8000±{\scriptsize 0.0304} & 0.0254 (\ensuremath{-}) \\
 & Max-Entropy~\cite{entro} & 20\% & 0.6444±{\scriptsize 0.1338} & 0.6167±{\scriptsize 0.1514} & 0.5333±{\scriptsize 0.2173} & 0.7000±{\scriptsize 0.1118} & 0.4693±{\scriptsize 0.1716} & 0.7512±{\scriptsize 0.1103} & 0.4969±{\scriptsize 0.1913} & 0.6778±{\scriptsize 0.0870} & 0.0191 (\ensuremath{-}) \\
 & BALD~\cite{bald} & 20\% & 0.7667±{\scriptsize 0.0608} & 0.6667±{\scriptsize 0.0779} & 0.3667±{\scriptsize 0.1394} & \underline{0.9667±{\scriptsize 0.0456}} & \underline{0.8500±{\scriptsize 0.2236}} & 0.7546±{\scriptsize 0.0435} & 0.5033±{\scriptsize 0.1583} & 0.8556±{\scriptsize 0.1131} & 0.0893 (\ensuremath{-}) \\
 & EvoDiff (Ours) & 20\% & 0.7667±{\scriptsize 0.0249} & \textbf{0.7417±{\scriptsize 0.0745}} & \textbf{0.6667±{\scriptsize 0.2357}} & 0.8167±{\scriptsize 0.0913} & 0.6534±{\scriptsize 0.0298} & \textbf{0.8466±{\scriptsize 0.0960}} & \underline{0.6389±{\scriptsize 0.1146}} & \textbf{0.8750±{\scriptsize 0.0428}} & 0.0893 (+) \\
\cmidrule(lr){2-12}
 & \cellcolor[HTML]{ECF4FF}Random & \cellcolor[HTML]{ECF4FF}30\% & \cellcolor[HTML]{ECF4FF}0.7667±{\scriptsize 0.0465} & \cellcolor[HTML]{ECF4FF}\textbf{0.7083±{\scriptsize 0.0659}} & \cellcolor[HTML]{ECF4FF}\textbf{0.5333±{\scriptsize 0.1395}} & \cellcolor[HTML]{ECF4FF}0.8833±{\scriptsize 0.0457} & \cellcolor[HTML]{ECF4FF}0.6967±{\scriptsize 0.0785} & \cellcolor[HTML]{ECF4FF}\textbf{0.7935±{\scriptsize 0.0465}} & \cellcolor[HTML]{ECF4FF}\textbf{0.5968±{\scriptsize 0.1093}} & \cellcolor[HTML]{ECF4FF}\underline{0.8472±{\scriptsize 0.0581}} & - \\
 & VAAL~\cite{vaal} & 30\% & 0.7222±{\scriptsize 0.0878} & 0.6083±{\scriptsize 0.1003} & 0.2667±{\scriptsize 0.1491} & 0.9500±{\scriptsize 0.0745} & 0.7667±{\scriptsize 0.3249} & 0.7220±{\scriptsize 0.0533} & 0.3849±{\scriptsize 0.1918} & 0.7556±{\scriptsize 0.0975} & 0.0245 (\ensuremath{-}) \\
 & Core-Set~\cite{core} & 30\% & \underline{0.7778±{\scriptsize 0.0393}} & \textbf{0.7083±{\scriptsize 0.0510}} & \underline{0.5000±{\scriptsize 0.1179}} & 0.9167±{\scriptsize 0.0589} & 0.7800±{\scriptsize 0.1440} & \underline{0.7874±{\scriptsize 0.0360}} & \underline{0.5946±{\scriptsize 0.0852}} & \textbf{0.8583±{\scriptsize 0.0386}} & 0.0777 (+) \\
 & Max-Entropy~\cite{entro} & 30\% & 0.7000±{\scriptsize 0.0304} & 0.5500±{\scriptsize 0.0456} & 0.1000±{\scriptsize 0.0913} & \textbf{1.0000±{\scriptsize 0.0000}} & 0.6000±{\scriptsize 0.5477} & 0.6902±{\scriptsize 0.0215} & 0.1714±{\scriptsize 0.1565} & 0.5500±{\scriptsize 0.1545} & 0.0056 (\ensuremath{-}) \\
 & BALD~\cite{bald} & 30\% & 0.7556±{\scriptsize 0.0305} & 0.6333±{\scriptsize 0.0457} & 0.2667±{\scriptsize 0.0913} & \textbf{1.0000±{\scriptsize 0.0000}} & \textbf{1.0000±{\scriptsize 0.0000}} & 0.7324±{\scriptsize 0.0242} & 0.4143±{\scriptsize 0.1174} & 0.8139±{\scriptsize 0.0732} & 0.0540 (\ensuremath{-}) \\
 & EvoDiff (Ours) & 30\% & \textbf{0.7889±{\scriptsize 0.0465}} & \underline{0.7000±{\scriptsize 0.0456}} & 0.4333±{\scriptsize 0.0913} & \underline{0.9667±{\scriptsize 0.0746}} & \underline{0.9200±{\scriptsize 0.1789}} & 0.7738±{\scriptsize 0.0251} & 0.5758±{\scriptsize 0.0851} & 0.7986±{\scriptsize 0.0505} & 0.0245 (\ensuremath{-}) \\
\bottomrule
\end{tabular}
}
\end{sidewaystable*}

\begin{sidewaystable*}[!ht]
\small
\centering
\renewcommand{\arraystretch}{1.2}
\setlength{\tabcolsep}{4pt}
\ContinuedFloat
\caption{Comparison of USFM classification metrics between EvoDiff and other AL methods on the test sets. Values are reported as mean $\pm$ s.d. over five folds. Statistical differences between each active learning strategy and Random sampling were assessed using two-sided paired $t$-tests based on fold-wise AUC. The reported $q$-values are the Benjamini--Hochberg-calibrated $p$-values across the five strategy-versus-Random comparisons within each dataset and annotation budget; the appended $(+)$ or $(-)$ indicates the sign of \texttt{mean\_diff(method-Random)} for AUC. (\textbf{Bold} represents the best result, and \underline{Underline} represents the second best result).}
\resizebox{\linewidth}{!}{
\begin{tabular}{l|l|c|cccccccc|c}
\toprule
\textbf{Dataset} & \textbf{Method} & \textbf{Annotation Budget} & \textbf{Acc} & \textbf{bAcc} & \textbf{Sens} & \textbf{Spec} & \textbf{PPV} & \textbf{NPV} & \textbf{F1} & \textbf{AUC} & \textbf{$q$-value} \\
\midrule
\multirow{19}{*}{CUBS~\cite{CUBS}} & \cellcolor[HTML]{F8F8D1}FSL (upper bound) & \cellcolor[HTML]{F8F8D1}100\% & \cellcolor[HTML]{F8F8D1}0.8259±{\scriptsize 0.0094} & \cellcolor[HTML]{F8F8D1}0.5346±{\scriptsize 0.0251} & \cellcolor[HTML]{F8F8D1}0.0692±{\scriptsize 0.0501} & \cellcolor[HTML]{F8F8D1}1.0000±{\scriptsize 0.0000} & \cellcolor[HTML]{F8F8D1}0.8000±{\scriptsize 0.4472} & \cellcolor[HTML]{F8F8D1}0.8237±{\scriptsize 0.0078} & \cellcolor[HTML]{F8F8D1}0.1262±{\scriptsize 0.0894} & \cellcolor[HTML]{F8F8D1}0.7455±{\scriptsize 0.0266} & - \\
 & \cellcolor[HTML]{ECF4FF}Random & \cellcolor[HTML]{ECF4FF}10\% & \cellcolor[HTML]{ECF4FF}\textbf{0.8129±{\scriptsize 0.0144}} & \cellcolor[HTML]{ECF4FF}\underline{0.5770±{\scriptsize 0.0298}} & \cellcolor[HTML]{ECF4FF}0.2000±{\scriptsize 0.0740} & \cellcolor[HTML]{ECF4FF}\underline{0.9540±{\scriptsize 0.0245}} & \cellcolor[HTML]{ECF4FF}\underline{0.5180±{\scriptsize 0.1012}} & \cellcolor[HTML]{ECF4FF}\underline{0.8385±{\scriptsize 0.0106}} & \cellcolor[HTML]{ECF4FF}\underline{0.2802±{\scriptsize 0.0757}} & \cellcolor[HTML]{ECF4FF}\underline{0.6170±{\scriptsize 0.0658}} & - \\
 & VAAL~\cite{vaal} & 10\% & 0.7856±{\scriptsize 0.0307} & 0.5246±{\scriptsize 0.0285} & 0.1077±{\scriptsize 0.0322} & 0.9416±{\scriptsize 0.0334} & 0.3283±{\scriptsize 0.1328} & 0.8208±{\scriptsize 0.0093} & 0.1602±{\scriptsize 0.0527} & 0.5302±{\scriptsize 0.0567} & 0.0115 (\ensuremath{-}) \\
 & Core-Set~\cite{core} & 10\% & 0.7813±{\scriptsize 0.0181} & 0.5397±{\scriptsize 0.0220} & 0.1538±{\scriptsize 0.0471} & 0.9257±{\scriptsize 0.0239} & 0.3318±{\scriptsize 0.0800} & 0.8263±{\scriptsize 0.0076} & 0.2066±{\scriptsize 0.0502} & 0.5243±{\scriptsize 0.0228} & 0.0475 (\ensuremath{-}) \\
 & Max-Entropy~\cite{entro} & 10\% & \underline{0.8101±{\scriptsize 0.0213}} & 0.5397±{\scriptsize 0.0157} & 0.1077±{\scriptsize 0.0322} & \textbf{0.9717±{\scriptsize 0.0290}} & \textbf{0.5771±{\scriptsize 0.2880}} & 0.8256±{\scriptsize 0.0049} & 0.1740±{\scriptsize 0.0415} & 0.5244±{\scriptsize 0.0596} & 0.0097 (\ensuremath{-}) \\
 & BALD~\cite{bald} & 10\% & 0.7612±{\scriptsize 0.0320} & 0.5481±{\scriptsize 0.0329} & \underline{0.2077±{\scriptsize 0.0644}} & 0.8885±{\scriptsize 0.0409} & 0.3119±{\scriptsize 0.0964} & 0.8298±{\scriptsize 0.0116} & 0.2447±{\scriptsize 0.0614} & 0.5321±{\scriptsize 0.0278} & 0.0097 (\ensuremath{-}) \\
 & EvoDiff (Ours) & 10\% & 0.7971±{\scriptsize 0.0256} & \textbf{0.5850±{\scriptsize 0.0387}} & \textbf{0.2462±{\scriptsize 0.0798}} & 0.9239±{\scriptsize 0.0311} & 0.4389±{\scriptsize 0.1139} & \textbf{0.8421±{\scriptsize 0.0138}} & \textbf{0.3095±{\scriptsize 0.0786}} & \textbf{0.6555±{\scriptsize 0.0770}} & 0.0097 (+) \\
\cmidrule(lr){2-12}
 & \cellcolor[HTML]{ECF4FF}Random & \cellcolor[HTML]{ECF4FF}20\% & \cellcolor[HTML]{ECF4FF}0.7727±{\scriptsize 0.0277} & \cellcolor[HTML]{ECF4FF}0.5433±{\scriptsize 0.0310} & \cellcolor[HTML]{ECF4FF}0.1769±{\scriptsize 0.0439} & \cellcolor[HTML]{ECF4FF}0.9097±{\scriptsize 0.0283} & \cellcolor[HTML]{ECF4FF}0.3231±{\scriptsize 0.1070} & \cellcolor[HTML]{ECF4FF}0.8276±{\scriptsize 0.0104} & \cellcolor[HTML]{ECF4FF}0.2268±{\scriptsize 0.0581} & \cellcolor[HTML]{ECF4FF}0.5361±{\scriptsize 0.0262} & - \\
 & VAAL~\cite{vaal} & 20\% & 0.7856±{\scriptsize 0.0212} & 0.5276±{\scriptsize 0.0255} & 0.1154±{\scriptsize 0.0385} & 0.9398±{\scriptsize 0.0211} & 0.3177±{\scriptsize 0.1237} & 0.8219±{\scriptsize 0.0083} & 0.1680±{\scriptsize 0.0570} & 0.5662±{\scriptsize 0.0284} & 0.0355 (+) \\
 & Core-Set~\cite{core} & 20\% & \underline{0.8058±{\scriptsize 0.0113}} & 0.5489±{\scriptsize 0.0150} & 0.1384±{\scriptsize 0.0344} & \underline{0.9593±{\scriptsize 0.0161}} & 0.4470±{\scriptsize 0.0861} & 0.8288±{\scriptsize 0.0049} & 0.2090±{\scriptsize 0.0459} & \underline{0.6040±{\scriptsize 0.0352}} & 0.0039 (+) \\
 & Max-Entropy~\cite{entro} & 20\% & \textbf{0.8144±{\scriptsize 0.0156}} & 0.5483±{\scriptsize 0.0274} & 0.1231±{\scriptsize 0.0501} & \textbf{0.9735±{\scriptsize 0.0125}} & \textbf{0.5264±{\scriptsize 0.1842}} & 0.8283±{\scriptsize 0.0089} & 0.1973±{\scriptsize 0.0739} & 0.5754±{\scriptsize 0.0313} & 0.0122 (+) \\
 & BALD~\cite{bald} & 20\% & 0.7612±{\scriptsize 0.0186} & \underline{0.5511±{\scriptsize 0.0303}} & \underline{0.2154±{\scriptsize 0.0583}} & 0.8867±{\scriptsize 0.0192} & 0.3035±{\scriptsize 0.0607} & \underline{0.8309±{\scriptsize 0.0108}} & \underline{0.2509±{\scriptsize 0.0597}} & 0.5645±{\scriptsize 0.0412} & 0.0392 (+) \\
 & EvoDiff (Ours) & 20\% & 0.8043±{\scriptsize 0.0280} & \textbf{0.5865±{\scriptsize 0.0262}} & \textbf{0.2385±{\scriptsize 0.0322}} & 0.9345±{\scriptsize 0.0310} & \underline{0.4765±{\scriptsize 0.1276}} & \textbf{0.8420±{\scriptsize 0.0084}} & \textbf{0.3152±{\scriptsize 0.0527}} & \textbf{0.6637±{\scriptsize 0.0349}} & 0.0039 (+) \\
\cmidrule(lr){2-12}
 & \cellcolor[HTML]{ECF4FF}Random & \cellcolor[HTML]{ECF4FF}30\% & \cellcolor[HTML]{ECF4FF}0.7957±{\scriptsize 0.0219} & \cellcolor[HTML]{ECF4FF}0.5753±{\scriptsize 0.0228} & \cellcolor[HTML]{ECF4FF}0.2231±{\scriptsize 0.0322} & \cellcolor[HTML]{ECF4FF}0.9274±{\scriptsize 0.0237} & \cellcolor[HTML]{ECF4FF}0.4278±{\scriptsize 0.1108} & \cellcolor[HTML]{ECF4FF}0.8384±{\scriptsize 0.0075} & \cellcolor[HTML]{ECF4FF}0.2910±{\scriptsize 0.0457} & \cellcolor[HTML]{ECF4FF}0.6395±{\scriptsize 0.0740} & - \\
 & VAAL~\cite{vaal} & 30\% & 0.8029±{\scriptsize 0.0219} & 0.5975±{\scriptsize 0.0317} & 0.2692±{\scriptsize 0.0666} & 0.9257±{\scriptsize 0.0277} & 0.4644±{\scriptsize 0.0850} & 0.8464±{\scriptsize 0.0115} & 0.3363±{\scriptsize 0.0616} & 0.6939±{\scriptsize 0.0154} & 0.0405 (+) \\
 & Core-Set~\cite{core} & 30\% & \underline{0.8403±{\scriptsize 0.0186}} & \underline{0.6264±{\scriptsize 0.0387}} & 0.2846±{\scriptsize 0.0750} & \textbf{0.9682±{\scriptsize 0.0134}} & \underline{0.6755±{\scriptsize 0.1047}} & \underline{0.8548±{\scriptsize 0.0133}} & \underline{0.3970±{\scriptsize 0.0876}} & 0.6401±{\scriptsize 0.0362} & 0.0347 (+) \\
 & Max-Entropy~\cite{entro} & 30\% & 0.8187±{\scriptsize 0.0171} & 0.6161±{\scriptsize 0.0227} & \underline{0.2923±{\scriptsize 0.0439}} & 0.9399±{\scriptsize 0.0202} & 0.5369±{\scriptsize 0.0859} & 0.8524±{\scriptsize 0.0078} & 0.3758±{\scriptsize 0.0483} & \underline{0.6988±{\scriptsize 0.0532}} & 0.0405 (+) \\
 & BALD~\cite{bald} & 30\% & 0.7885±{\scriptsize 0.0286} & 0.5975±{\scriptsize 0.0247} & 0.2923±{\scriptsize 0.0211} & 0.9027±{\scriptsize 0.0313} & 0.4191±{\scriptsize 0.0821} & 0.8470±{\scriptsize 0.0080} & 0.3430±{\scriptsize 0.0427} & 0.6030±{\scriptsize 0.0313} & 0.0527 (\ensuremath{-}) \\
 & EvoDiff (Ours) & 30\% & \textbf{0.8489±{\scriptsize 0.0283}} & \textbf{0.6583±{\scriptsize 0.0394}} & \textbf{0.3538±{\scriptsize 0.0632}} & \underline{0.9628±{\scriptsize 0.0254}} & \textbf{0.7046±{\scriptsize 0.1730}} & \textbf{0.8662±{\scriptsize 0.0132}} & \textbf{0.4682±{\scriptsize 0.0848}} & \textbf{0.7061±{\scriptsize 0.0263}} & 0.0986 (+) \\
\midrule
\multirow{19}{*}{Fatty-Liver~\cite{liver}} & \cellcolor[HTML]{F8F8D1}FSL (upper bound) & \cellcolor[HTML]{F8F8D1}100\% & \cellcolor[HTML]{F8F8D1}0.8868±{\scriptsize 0.1092} & \cellcolor[HTML]{F8F8D1}0.8444±{\scriptsize 0.1931} & \cellcolor[HTML]{F8F8D1}0.9514±{\scriptsize 0.0645} & \cellcolor[HTML]{F8F8D1}0.7375±{\scriptsize 0.4155} & \cellcolor[HTML]{F8F8D1}0.9132±{\scriptsize 0.1222} & \cellcolor[HTML]{F8F8D1}0.7063±{\scriptsize 0.4049} & \cellcolor[HTML]{F8F8D1}0.9254±{\scriptsize 0.0622} & \cellcolor[HTML]{F8F8D1}0.9299±{\scriptsize 0.1068} & - \\
 & \cellcolor[HTML]{ECF4FF}Random & \cellcolor[HTML]{ECF4FF}10\% & \cellcolor[HTML]{ECF4FF}\textbf{0.9547±{\scriptsize 0.0215}} & \cellcolor[HTML]{ECF4FF}\textbf{0.9285±{\scriptsize 0.0299}} & \cellcolor[HTML]{ECF4FF}\textbf{0.9946±{\scriptsize 0.0121}} & \cellcolor[HTML]{ECF4FF}0.8625±{\scriptsize 0.0523} & \cellcolor[HTML]{ECF4FF}0.9438±{\scriptsize 0.0208} & \cellcolor[HTML]{ECF4FF}\textbf{0.9857±{\scriptsize 0.0319}} & \cellcolor[HTML]{ECF4FF}\textbf{0.9685±{\scriptsize 0.0149}} & \cellcolor[HTML]{ECF4FF}\textbf{0.9785±{\scriptsize 0.0246}} & - \\
 & VAAL~\cite{vaal} & 10\% & 0.7472±{\scriptsize 0.0620} & 0.5919±{\scriptsize 0.1099} & \underline{0.9838±{\scriptsize 0.0148}} & 0.2000±{\scriptsize 0.2313} & 0.7436±{\scriptsize 0.0571} & 0.4900±{\scriptsize 0.4506} & 0.8457±{\scriptsize 0.0321} & 0.7648±{\scriptsize 0.0906} & 0.0028 (\ensuremath{-}) \\
 & Core-Set~\cite{core} & 10\% & \underline{0.8604±{\scriptsize 0.1022}} & 0.8042±{\scriptsize 0.1737} & 0.9459±{\scriptsize 0.0936} & 0.6625±{\scriptsize 0.3817} & 0.8831±{\scriptsize 0.1084} & \underline{0.7038±{\scriptsize 0.4183}} & \underline{0.9063±{\scriptsize 0.0611}} & 0.8987±{\scriptsize 0.1071} & 0.0128 (\ensuremath{-}) \\
 & Max-Entropy~\cite{entro} & 10\% & 0.7962±{\scriptsize 0.0338} & 0.8505±{\scriptsize 0.0281} & 0.7135±{\scriptsize 0.0453} & \underline{0.9875±{\scriptsize 0.0280}} & \underline{0.9926±{\scriptsize 0.0165}} & 0.6002±{\scriptsize 0.0391} & 0.8296±{\scriptsize 0.0323} & 0.8427±{\scriptsize 0.0739} & 0.0028 (\ensuremath{-}) \\
 & BALD~\cite{bald} & 10\% & 0.8302±{\scriptsize 0.0766} & 0.7649±{\scriptsize 0.1488} & 0.9297±{\scriptsize 0.0491} & 0.6000±{\scriptsize 0.3354} & 0.8548±{\scriptsize 0.0876} & 0.6326±{\scriptsize 0.3578} & 0.8867±{\scriptsize 0.0395} & 0.8588±{\scriptsize 0.1071} & 0.0174 (\ensuremath{-}) \\
 & EvoDiff (Ours) & 10\% & 0.8415±{\scriptsize 0.0315} & \underline{0.8865±{\scriptsize 0.0226}} & 0.7730±{\scriptsize 0.0452} & \textbf{1.0000±{\scriptsize 0.0000}} & \textbf{1.0000±{\scriptsize 0.0000}} & 0.6581±{\scriptsize 0.0434} & 0.8713±{\scriptsize 0.0293} & \underline{0.9375±{\scriptsize 0.0415}} & 0.0128 (\ensuremath{-}) \\
\cmidrule(lr){2-12}
 & \cellcolor[HTML]{ECF4FF}Random & \cellcolor[HTML]{ECF4FF}20\% & \cellcolor[HTML]{ECF4FF}\textbf{0.9510±{\scriptsize 0.0169}} & \cellcolor[HTML]{ECF4FF}\textbf{0.9365±{\scriptsize 0.0279}} & \cellcolor[HTML]{ECF4FF}\textbf{0.9730±{\scriptsize 0.0000}} & \cellcolor[HTML]{ECF4FF}0.9000±{\scriptsize 0.0559} & \cellcolor[HTML]{ECF4FF}\underline{0.9579±{\scriptsize 0.0224}} & \cellcolor[HTML]{ECF4FF}\textbf{0.9349±{\scriptsize 0.0040}} & \cellcolor[HTML]{ECF4FF}\textbf{0.9653±{\scriptsize 0.0115}} & \cellcolor[HTML]{ECF4FF}\textbf{0.9814±{\scriptsize 0.0266}} & - \\
 & VAAL~\cite{vaal} & 20\% & 0.8642±{\scriptsize 0.0938} & 0.8034±{\scriptsize 0.1710} & 0.9567±{\scriptsize 0.0242} & 0.6500±{\scriptsize 0.3660} & 0.8767±{\scriptsize 0.1013} & 0.6931±{\scriptsize 0.3875} & 0.9112±{\scriptsize 0.0505} & 0.9399±{\scriptsize 0.0641} & 0.0373 (\ensuremath{-}) \\
 & Core-Set~\cite{core} & 20\% & 0.8038±{\scriptsize 0.2866} & 0.8488±{\scriptsize 0.2004} & 0.7351±{\scriptsize 0.4184} & \underline{0.9625±{\scriptsize 0.0342}} & 0.7824±{\scriptsize 0.4376} & 0.7430±{\scriptsize 0.2736} & 0.7568±{\scriptsize 0.4257} & 0.9215±{\scriptsize 0.0967} & 0.0373 (\ensuremath{-}) \\
 & Max-Entropy~\cite{entro} & 20\% & 0.6642±{\scriptsize 0.3326} & 0.7488±{\scriptsize 0.2278} & 0.5351±{\scriptsize 0.4930} & \underline{0.9625±{\scriptsize 0.0559}} & 0.5836±{\scriptsize 0.5331} & 0.6056±{\scriptsize 0.2904} & 0.5567±{\scriptsize 0.5091} & \underline{0.9476±{\scriptsize 0.0690}} & 0.0411 (\ensuremath{-}) \\
 & BALD~\cite{bald} & 20\% & \underline{0.8792±{\scriptsize 0.1039}} & 0.8177±{\scriptsize 0.1800} & \underline{0.9730±{\scriptsize 0.0271}} & 0.6625±{\scriptsize 0.3740} & 0.8828±{\scriptsize 0.1052} & 0.7331±{\scriptsize 0.4136} & \underline{0.9220±{\scriptsize 0.0581}} & 0.8965±{\scriptsize 0.3175} & 0.0373 (\ensuremath{-}) \\
 & EvoDiff (Ours) & 20\% & 0.8340±{\scriptsize 0.2564} & \underline{0.8775±{\scriptsize 0.1816}} & 0.7676±{\scriptsize 0.3709} & \textbf{0.9875±{\scriptsize 0.0280}} & \textbf{0.9946±{\scriptsize 0.0121}} & \underline{0.7607±{\scriptsize 0.2513}} & 0.8080±{\scriptsize 0.3432} & 0.9138±{\scriptsize 0.1356} & 0.0373 (\ensuremath{-}) \\
\cmidrule(lr){2-12}
 & \cellcolor[HTML]{ECF4FF}Random & \cellcolor[HTML]{ECF4FF}30\% & \cellcolor[HTML]{ECF4FF}\textbf{0.9472±{\scriptsize 0.0207}} & \cellcolor[HTML]{ECF4FF}\textbf{0.9302±{\scriptsize 0.0324}} & \cellcolor[HTML]{ECF4FF}\underline{0.9730±{\scriptsize 0.0191}} & \cellcolor[HTML]{ECF4FF}\underline{0.8875±{\scriptsize 0.0685}} & \cellcolor[HTML]{ECF4FF}\textbf{0.9531±{\scriptsize 0.0279}} & \cellcolor[HTML]{ECF4FF}\textbf{0.9366±{\scriptsize 0.0443}} & \cellcolor[HTML]{ECF4FF}\textbf{0.9626±{\scriptsize 0.0146}} & \cellcolor[HTML]{ECF4FF}\underline{0.9720±{\scriptsize 0.0230}} & - \\
 & VAAL~\cite{vaal} & 30\% & 0.8491±{\scriptsize 0.1380} & 0.7606±{\scriptsize 0.2380} & \textbf{0.9838±{\scriptsize 0.0242}} & 0.5375±{\scriptsize 0.4913} & 0.8529±{\scriptsize 0.1417} & 0.5631±{\scriptsize 0.5158} & 0.9074±{\scriptsize 0.0780} & 0.8829±{\scriptsize 0.1355} & 0.0381 (\ensuremath{-}) \\
 & Core-Set~\cite{core} & 30\% & 0.9019±{\scriptsize 0.1151} & 0.8552±{\scriptsize 0.1989} & \underline{0.9730±{\scriptsize 0.0331}} & 0.7375±{\scriptsize 0.4132} & 0.9128±{\scriptsize 0.1205} & 0.7417±{\scriptsize 0.4189} & 0.9373±{\scriptsize 0.0655} & 0.8507±{\scriptsize 0.3193} & 0.0518 (\ensuremath{-}) \\
 & Max-Entropy~\cite{entro} & 30\% & 0.8604±{\scriptsize 0.1489} & 0.8326±{\scriptsize 0.2036} & 0.9027±{\scriptsize 0.1878} & 0.7625±{\scriptsize 0.4339} & 0.9246±{\scriptsize 0.1307} & 0.6765±{\scriptsize 0.4282} & 0.8960±{\scriptsize 0.1178} & 0.8831±{\scriptsize 0.1710} & 0.0518 (\ensuremath{-}) \\
 & BALD~\cite{bald} & 30\% & \underline{0.9321±{\scriptsize 0.0286}} & \underline{0.9123±{\scriptsize 0.0203}} & 0.9622±{\scriptsize 0.0452} & 0.8625±{\scriptsize 0.0280} & \underline{0.9420±{\scriptsize 0.0097}} & \underline{0.9172±{\scriptsize 0.0938}} & \underline{0.9515±{\scriptsize 0.0218}} & 0.9301±{\scriptsize 0.0054} & 0.0776 (\ensuremath{-}) \\
 & EvoDiff (Ours) & 30\% & 0.8415±{\scriptsize 0.3018} & 0.8829±{\scriptsize 0.2145} & 0.7784±{\scriptsize 0.4351} & \textbf{0.9875±{\scriptsize 0.0280}} & 0.7946±{\scriptsize 0.4443} & 0.8126±{\scriptsize 0.2855} & 0.7864±{\scriptsize 0.4396} & \textbf{0.9790±{\scriptsize 0.0384}} & 0.0086 (+) \\
\bottomrule
\end{tabular}
}
\end{sidewaystable*}

\begin{sidewaystable*}[!ht]
\small
\centering
\renewcommand{\arraystretch}{1.2}
\setlength{\tabcolsep}{4pt}
\ContinuedFloat
\caption{Comparison of USFM classification metrics between EvoDiff and other AL methods on the test sets. Values are reported as mean $\pm$ s.d. over five folds. Statistical differences between each active learning strategy and Random sampling were assessed using two-sided paired $t$-tests based on fold-wise AUC. The reported $q$-values are the Benjamini--Hochberg-calibrated $p$-values across the five strategy-versus-Random comparisons within each dataset and annotation budget; the appended $(+)$ or $(-)$ indicates the sign of \texttt{mean\_diff(method-Random)} for AUC. (\textbf{Bold} represents the best result, and \underline{Underline} represents the second best result).}
\resizebox{\linewidth}{!}{
\begin{tabular}{l|l|c|cccccccc|c}
\toprule
\textbf{Dataset} & \textbf{Method} & \textbf{Annotation Budget} & \textbf{Acc} & \textbf{bAcc} & \textbf{Sens} & \textbf{Spec} & \textbf{PPV} & \textbf{NPV} & \textbf{F1} & \textbf{AUC} & \textbf{$q$-value} \\
\midrule
\multirow{19}{*}{MMOTU~\cite{mmotu}} & \cellcolor[HTML]{F8F8D1}FSL (upper bound) & \cellcolor[HTML]{F8F8D1}100\% & \cellcolor[HTML]{F8F8D1}0.8692±{\scriptsize 0.0107} & \cellcolor[HTML]{F8F8D1}0.7473±{\scriptsize 0.0248} & \cellcolor[HTML]{F8F8D1}0.9375±{\scriptsize 0.0055} & \cellcolor[HTML]{F8F8D1}0.5571±{\scriptsize 0.0479} & \cellcolor[HTML]{F8F8D1}0.9064±{\scriptsize 0.0093} & \cellcolor[HTML]{F8F8D1}0.6605±{\scriptsize 0.0315} & \cellcolor[HTML]{F8F8D1}0.9217±{\scriptsize 0.0062} & \cellcolor[HTML]{F8F8D1}0.8322±{\scriptsize 0.0126} & - \\
 & \cellcolor[HTML]{ECF4FF}Random & \cellcolor[HTML]{ECF4FF}10\% & \cellcolor[HTML]{ECF4FF}0.8013±{\scriptsize 0.0294} & \cellcolor[HTML]{ECF4FF}0.6808±{\scriptsize 0.0313} & \cellcolor[HTML]{ECF4FF}0.8688±{\scriptsize 0.0319} & \cellcolor[HTML]{ECF4FF}\underline{0.4929±{\scriptsize 0.0465}} & \cellcolor[HTML]{ECF4FF}0.8867±{\scriptsize 0.0106} & \cellcolor[HTML]{ECF4FF}0.4569±{\scriptsize 0.0747} & \cellcolor[HTML]{ECF4FF}0.8774±{\scriptsize 0.0197} & \cellcolor[HTML]{ECF4FF}0.8091±{\scriptsize 0.0192} & - \\
 & VAAL~\cite{vaal} & 10\% & 0.8231±{\scriptsize 0.0154} & 0.5992±{\scriptsize 0.0554} & \underline{0.9485±{\scriptsize 0.0280}} & 0.2500±{\scriptsize 0.1312} & 0.8532±{\scriptsize 0.0195} & 0.5238±{\scriptsize 0.1085} & 0.8979±{\scriptsize 0.0091} & 0.7615±{\scriptsize 0.0338} & 0.0136 (\ensuremath{-}) \\
 & Core-Set~\cite{core} & 10\% & 0.8243±{\scriptsize 0.0147} & 0.5972±{\scriptsize 0.0494} & \textbf{0.9516±{\scriptsize 0.0102}} & 0.2429±{\scriptsize 0.1053} & 0.8521±{\scriptsize 0.0168} & 0.5095±{\scriptsize 0.0776} & 0.8990±{\scriptsize 0.0075} & 0.7369±{\scriptsize 0.0317} & 0.0029 (\ensuremath{-}) \\
 & Max-Entropy~\cite{entro} & 10\% & 0.7974±{\scriptsize 0.0173} & 0.6562±{\scriptsize 0.0458} & 0.8766±{\scriptsize 0.0150} & 0.4357±{\scriptsize 0.0958} & 0.8769±{\scriptsize 0.0175} & 0.4324±{\scriptsize 0.0562} & 0.8766±{\scriptsize 0.0101} & 0.7755±{\scriptsize 0.0325} & 0.0136 (\ensuremath{-}) \\
 & BALD~\cite{bald} & 10\% & \underline{0.8359±{\scriptsize 0.0107}} & \underline{0.7103±{\scriptsize 0.0281}} & 0.9062±{\scriptsize 0.0199} & \textbf{0.5143±{\scriptsize 0.0696}} & \underline{0.8953±{\scriptsize 0.0116}} & \underline{0.5478±{\scriptsize 0.0344}} & \underline{0.9006±{\scriptsize 0.0073}} & \textbf{0.8311±{\scriptsize 0.0222}} & 0.0178 (+) \\
 & EvoDiff (Ours) & 10\% & \textbf{0.8474±{\scriptsize 0.0146}} & \textbf{0.7173±{\scriptsize 0.0305}} & 0.9203±{\scriptsize 0.0169} & \textbf{0.5143±{\scriptsize 0.0649}} & \textbf{0.8967±{\scriptsize 0.0119}} & \textbf{0.5874±{\scriptsize 0.0520}} & \textbf{0.9082±{\scriptsize 0.0090}} & \underline{0.8267±{\scriptsize 0.0240}} & 0.0029 (+) \\
\cmidrule(lr){2-12}
 & \cellcolor[HTML]{ECF4FF}Random & \cellcolor[HTML]{ECF4FF}20\% & \cellcolor[HTML]{ECF4FF}0.8372±{\scriptsize 0.0073} & \cellcolor[HTML]{ECF4FF}0.7138±{\scriptsize 0.0303} & \cellcolor[HTML]{ECF4FF}0.9062±{\scriptsize 0.0166} & \cellcolor[HTML]{ECF4FF}0.5214±{\scriptsize 0.0740} & \cellcolor[HTML]{ECF4FF}0.8968±{\scriptsize 0.0129} & \cellcolor[HTML]{ECF4FF}0.5500±{\scriptsize 0.0264} & \cellcolor[HTML]{ECF4FF}0.9013±{\scriptsize 0.0047} & \cellcolor[HTML]{ECF4FF}0.8382±{\scriptsize 0.0203} & - \\
 & VAAL~\cite{vaal} & 20\% & 0.8166±{\scriptsize 0.0282} & 0.6651±{\scriptsize 0.0482} & 0.9016±{\scriptsize 0.0356} & 0.4286±{\scriptsize 0.1041} & 0.8786±{\scriptsize 0.0176} & 0.4928±{\scriptsize 0.0751} & 0.8895±{\scriptsize 0.0187} & 0.7990±{\scriptsize 0.0192} & 0.0090 (\ensuremath{-}) \\
 & Core-Set~\cite{core} & 20\% & 0.8346±{\scriptsize 0.0229} & 0.6872±{\scriptsize 0.1049} & \underline{0.9172±{\scriptsize 0.0557}} & 0.4571±{\scriptsize 0.2568} & 0.8884±{\scriptsize 0.0381} & 0.4461±{\scriptsize 0.2586} & 0.9008±{\scriptsize 0.0149} & 0.8114±{\scriptsize 0.0404} & 0.0335 (\ensuremath{-}) \\
 & Max-Entropy~\cite{entro} & 20\% & \underline{0.8462±{\scriptsize 0.0227}} & 0.7109±{\scriptsize 0.0387} & \textbf{0.9219±{\scriptsize 0.0166}} & 0.5000±{\scriptsize 0.0668} & 0.8940±{\scriptsize 0.0138} & \textbf{0.5847±{\scriptsize 0.0773}} & \underline{0.9077±{\scriptsize 0.0137}} & 0.8125±{\scriptsize 0.0366} & 0.0857 (\ensuremath{-}) \\
 & BALD~\cite{bald} & 20\% & 0.8436±{\scriptsize 0.0267} & \textbf{0.7484±{\scriptsize 0.0446}} & 0.8969±{\scriptsize 0.0210} & \textbf{0.6000±{\scriptsize 0.0774}} & \textbf{0.9112±{\scriptsize 0.0166}} & 0.5615±{\scriptsize 0.0743} & 0.9039±{\scriptsize 0.0165} & \textbf{0.8581±{\scriptsize 0.0157}} & 0.0090 (+) \\
 & EvoDiff (Ours) & 20\% & \textbf{0.8474±{\scriptsize 0.0124}} & \underline{0.7257±{\scriptsize 0.0263}} & 0.9156±{\scriptsize 0.0086} & \underline{0.5357±{\scriptsize 0.0505}} & \underline{0.9002±{\scriptsize 0.0099}} & \underline{0.5811±{\scriptsize 0.0377}} & \textbf{0.9078±{\scriptsize 0.0073}} & \underline{0.8423±{\scriptsize 0.0318}} & 0.0439 (+) \\
\cmidrule(lr){2-12}
 & \cellcolor[HTML]{ECF4FF}Random & \cellcolor[HTML]{ECF4FF}30\% & \cellcolor[HTML]{ECF4FF}\underline{0.8910±{\scriptsize 0.0091}} & \cellcolor[HTML]{ECF4FF}\underline{0.7745±{\scriptsize 0.0258}} & \cellcolor[HTML]{ECF4FF}0.9562±{\scriptsize 0.0070} & \cellcolor[HTML]{ECF4FF}\underline{0.5928±{\scriptsize 0.0542}} & \cellcolor[HTML]{ECF4FF}\underline{0.9149±{\scriptsize 0.0100}} & \cellcolor[HTML]{ECF4FF}0.7481±{\scriptsize 0.0270} & \cellcolor[HTML]{ECF4FF}\underline{0.9351±{\scriptsize 0.0051}} & \cellcolor[HTML]{ECF4FF}\textbf{0.8935±{\scriptsize 0.0255}} & - \\
 & VAAL~\cite{vaal} & 30\% & 0.8718±{\scriptsize 0.0078} & 0.7545±{\scriptsize 0.0299} & 0.9375±{\scriptsize 0.0096} & 0.5714±{\scriptsize 0.0668} & 0.9093±{\scriptsize 0.0125} & 0.6672±{\scriptsize 0.0230} & 0.9231±{\scriptsize 0.0043} & 0.8240±{\scriptsize 0.0370} & 0.0020 (\ensuremath{-}) \\
 & Core-Set~\cite{core} & 30\% & \textbf{0.9000±{\scriptsize 0.0133}} & \textbf{0.7884±{\scriptsize 0.0097}} & 0.9625±{\scriptsize 0.0170} & \textbf{0.6143±{\scriptsize 0.0160}} & \textbf{0.9194±{\scriptsize 0.0030}} & \textbf{0.7884±{\scriptsize 0.0797}} & \textbf{0.9404±{\scriptsize 0.0084}} & 0.8299±{\scriptsize 0.0316} & 0.0069 (\ensuremath{-}) \\
 & Max-Entropy~\cite{entro} & 30\% & 0.8872±{\scriptsize 0.0402} & 0.7471±{\scriptsize 0.1400} & \underline{0.9656±{\scriptsize 0.0238}} & 0.5286±{\scriptsize 0.2982} & 0.9065±{\scriptsize 0.0488} & 0.6196±{\scriptsize 0.3521} & 0.9341±{\scriptsize 0.0205} & \underline{0.8382±{\scriptsize 0.1347}} & 0.0435 (\ensuremath{-}) \\
 & BALD~\cite{bald} & 30\% & 0.8526±{\scriptsize 0.0290} & 0.7120±{\scriptsize 0.1217} & 0.9312±{\scriptsize 0.0440} & 0.4928±{\scriptsize 0.2782} & 0.8967±{\scriptsize 0.0436} & 0.4922±{\scriptsize 0.2836} & 0.9122±{\scriptsize 0.0158} & 0.8078±{\scriptsize 0.0886} & 0.0172 (\ensuremath{-}) \\
 & EvoDiff (Ours) & 30\% & 0.8859±{\scriptsize 0.0153} & 0.7407±{\scriptsize 0.0332} & \textbf{0.9672±{\scriptsize 0.0116}} & 0.5143±{\scriptsize 0.0649} & 0.9011±{\scriptsize 0.0119} & \underline{0.7760±{\scriptsize 0.0662}} & 0.9330±{\scriptsize 0.0089} & 0.8362±{\scriptsize 0.0398} & 0.0069 (\ensuremath{-}) \\
\midrule
\multirow{19}{*}{TN3K~\cite{TN3K}} & \cellcolor[HTML]{F8F8D1}FSL (upper bound) & \cellcolor[HTML]{F8F8D1}100\% & \cellcolor[HTML]{F8F8D1}0.6651±{\scriptsize 0.0227} & \cellcolor[HTML]{F8F8D1}0.5157±{\scriptsize 0.0329} & \cellcolor[HTML]{F8F8D1}0.0314±{\scriptsize 0.0657} & \cellcolor[HTML]{F8F8D1}1.0000±{\scriptsize 0.0000} & \cellcolor[HTML]{F8F8D1}0.4000±{\scriptsize 0.5477} & \cellcolor[HTML]{F8F8D1}0.6618±{\scriptsize 0.0157} & \cellcolor[HTML]{F8F8D1}0.0551±{\scriptsize 0.1142} & \cellcolor[HTML]{F8F8D1}0.7354±{\scriptsize 0.0324} & - \\
 & \cellcolor[HTML]{ECF4FF}Random & \cellcolor[HTML]{ECF4FF}10\% & \cellcolor[HTML]{ECF4FF}\textbf{0.7103±{\scriptsize 0.0184}} & \cellcolor[HTML]{ECF4FF}\textbf{0.6737±{\scriptsize 0.0156}} & \cellcolor[HTML]{ECF4FF}\textbf{0.5554±{\scriptsize 0.0288}} & \cellcolor[HTML]{ECF4FF}\underline{0.7921±{\scriptsize 0.0319}} & \cellcolor[HTML]{ECF4FF}\textbf{0.5871±{\scriptsize 0.0321}} & \cellcolor[HTML]{ECF4FF}\textbf{0.7713±{\scriptsize 0.0104}} & \cellcolor[HTML]{ECF4FF}\textbf{0.5700±{\scriptsize 0.0198}} & \cellcolor[HTML]{ECF4FF}\textbf{0.7475±{\scriptsize 0.0119}} & - \\
 & VAAL~\cite{vaal} & 10\% & \underline{0.6794±{\scriptsize 0.0187}} & \underline{0.6194±{\scriptsize 0.0231}} & 0.4248±{\scriptsize 0.0683} & \textbf{0.8140±{\scriptsize 0.0424}} & \underline{0.5503±{\scriptsize 0.0435}} & 0.7288±{\scriptsize 0.0177} & 0.4760±{\scriptsize 0.0442} & \underline{0.6870±{\scriptsize 0.0265}} & 0.0007 (\ensuremath{-}) \\
 & Core-Set~\cite{core} & 10\% & 0.6589±{\scriptsize 0.0177} & 0.6048±{\scriptsize 0.0179} & 0.4298±{\scriptsize 0.0383} & 0.7799±{\scriptsize 0.0299} & 0.5090±{\scriptsize 0.0329} & 0.7215±{\scriptsize 0.0122} & 0.4651±{\scriptsize 0.0274} & 0.6845±{\scriptsize 0.0138} & 0.0001 (\ensuremath{-}) \\
 & Max-Entropy~\cite{entro} & 10\% & 0.6669±{\scriptsize 0.0108} & 0.6156±{\scriptsize 0.0065} & 0.4496±{\scriptsize 0.0440} & 0.7817±{\scriptsize 0.0376} & 0.5231±{\scriptsize 0.0218} & 0.7291±{\scriptsize 0.0074} & 0.4819±{\scriptsize 0.0183} & 0.6819±{\scriptsize 0.0175} & 0.0002 (\ensuremath{-}) \\
 & BALD~\cite{bald} & 10\% & 0.6549±{\scriptsize 0.0138} & 0.6162±{\scriptsize 0.0087} & \underline{0.4909±{\scriptsize 0.0150}} & 0.7415±{\scriptsize 0.0268} & 0.5018±{\scriptsize 0.0200} & 0.7337±{\scriptsize 0.0043} & \underline{0.4959±{\scriptsize 0.0077}} & 0.6710±{\scriptsize 0.0093} & 0.0001 (\ensuremath{-}) \\
 & EvoDiff (Ours) & 10\% & 0.6389±{\scriptsize 0.0211} & 0.6008±{\scriptsize 0.0620} & 0.4777±{\scriptsize 0.3294} & 0.7240±{\scriptsize 0.2057} & 0.5266±{\scriptsize 0.1125} & \underline{0.7457±{\scriptsize 0.0724}} & 0.4182±{\scriptsize 0.2096} & 0.6536±{\scriptsize 0.0480} & 0.0008 (\ensuremath{-}) \\
\cmidrule(lr){2-12}
 & \cellcolor[HTML]{ECF4FF}Random & \cellcolor[HTML]{ECF4FF}20\% & \cellcolor[HTML]{ECF4FF}0.7440±{\scriptsize 0.0152} & \cellcolor[HTML]{ECF4FF}0.7112±{\scriptsize 0.0191} & \cellcolor[HTML]{ECF4FF}0.6050±{\scriptsize 0.0343} & \cellcolor[HTML]{ECF4FF}\underline{0.8175±{\scriptsize 0.0113}} & \cellcolor[HTML]{ECF4FF}\underline{0.6364±{\scriptsize 0.0214}} & \cellcolor[HTML]{ECF4FF}0.7968±{\scriptsize 0.0145} & \cellcolor[HTML]{ECF4FF}0.6201±{\scriptsize 0.0267} & \cellcolor[HTML]{ECF4FF}0.8006±{\scriptsize 0.0228} & - \\
 & VAAL~\cite{vaal} & 20\% & 0.7371±{\scriptsize 0.0111} & 0.7181±{\scriptsize 0.0212} & 0.6562±{\scriptsize 0.0550} & 0.7799±{\scriptsize 0.0137} & 0.6114±{\scriptsize 0.0077} & 0.8118±{\scriptsize 0.0219} & 0.6323±{\scriptsize 0.0291} & 0.7902±{\scriptsize 0.0210} & 0.0563 (\ensuremath{-}) \\
 & Core-Set~\cite{core} & 20\% & \underline{0.7486±{\scriptsize 0.0111}} & 0.7147±{\scriptsize 0.0148} & 0.6050±{\scriptsize 0.0358} & \textbf{0.8245±{\scriptsize 0.0173}} & \textbf{0.6459±{\scriptsize 0.0191}} & 0.7982±{\scriptsize 0.0132} & 0.6242±{\scriptsize 0.0209} & 0.7966±{\scriptsize 0.0141} & 0.0816 (\ensuremath{-}) \\
 & Max-Entropy~\cite{entro} & 20\% & 0.7457±{\scriptsize 0.0270} & \underline{0.7324±{\scriptsize 0.0277}} & \underline{0.6892±{\scriptsize 0.0323}} & 0.7755±{\scriptsize 0.0270} & 0.6193±{\scriptsize 0.0366} & \underline{0.8252±{\scriptsize 0.0192}} & \underline{0.6523±{\scriptsize 0.0337}} & \underline{0.8152±{\scriptsize 0.0252}} & 0.0563 (+) \\
 & BALD~\cite{bald} & 20\% & 0.7434±{\scriptsize 0.0169} & 0.7217±{\scriptsize 0.0166} & 0.6513±{\scriptsize 0.0244} & 0.7921±{\scriptsize 0.0226} & 0.6241±{\scriptsize 0.0275} & 0.8113±{\scriptsize 0.0116} & 0.6371±{\scriptsize 0.0209} & 0.7980±{\scriptsize 0.0104} & 0.0816 (\ensuremath{-}) \\
 & EvoDiff (Ours) & 20\% & \textbf{0.7491±{\scriptsize 0.0253}} & \textbf{0.7405±{\scriptsize 0.0230}} & \textbf{0.7124±{\scriptsize 0.0306}} & 0.7686±{\scriptsize 0.0360} & 0.6207±{\scriptsize 0.0365} & \textbf{0.8349±{\scriptsize 0.0155}} & \textbf{0.6629±{\scriptsize 0.0276}} & \textbf{0.8209±{\scriptsize 0.0235}} & 0.0563 (+) \\
\cmidrule(lr){2-12}
 & \cellcolor[HTML]{ECF4FF}Random & \cellcolor[HTML]{ECF4FF}30\% & \cellcolor[HTML]{ECF4FF}\textbf{0.7920±{\scriptsize 0.0098}} & \cellcolor[HTML]{ECF4FF}0.7646±{\scriptsize 0.0118} & \cellcolor[HTML]{ECF4FF}0.6760±{\scriptsize 0.0189} & \cellcolor[HTML]{ECF4FF}\textbf{0.8533±{\scriptsize 0.0059}} & \cellcolor[HTML]{ECF4FF}\textbf{0.7088±{\scriptsize 0.0132}} & \cellcolor[HTML]{ECF4FF}0.8329±{\scriptsize 0.0088} & \cellcolor[HTML]{ECF4FF}0.6920±{\scriptsize 0.0159} & \cellcolor[HTML]{ECF4FF}0.8456±{\scriptsize 0.0143} & - \\
 & VAAL~\cite{vaal} & 30\% & 0.7709±{\scriptsize 0.0104} & 0.7606±{\scriptsize 0.0118} & 0.7273±{\scriptsize 0.0202} & 0.7939±{\scriptsize 0.0113} & 0.6510±{\scriptsize 0.0141} & 0.8465±{\scriptsize 0.0099} & 0.6869±{\scriptsize 0.0145} & 0.8335±{\scriptsize 0.0081} & 0.0216 (\ensuremath{-}) \\
 & Core-Set~\cite{core} & 30\% & 0.7840±{\scriptsize 0.0156} & 0.7640±{\scriptsize 0.0177} & 0.6992±{\scriptsize 0.0265} & \underline{0.8288±{\scriptsize 0.0129}} & \underline{0.6834±{\scriptsize 0.0217}} & 0.8391±{\scriptsize 0.0132} & 0.6911±{\scriptsize 0.0229} & 0.8160±{\scriptsize 0.0140} & 0.0125 (\ensuremath{-}) \\
 & Max-Entropy~\cite{entro} & 30\% & \underline{0.7903±{\scriptsize 0.0210}} & \underline{0.7777±{\scriptsize 0.0249}} & 0.7372±{\scriptsize 0.0385} & 0.8183±{\scriptsize 0.0133} & 0.6817±{\scriptsize 0.0258} & 0.8551±{\scriptsize 0.0198} & \underline{0.7083±{\scriptsize 0.0312}} & 0.8509±{\scriptsize 0.0138} & 0.0624 (+) \\
 & BALD~\cite{bald} & 30\% & \underline{0.7903±{\scriptsize 0.0152}} & \textbf{0.7793±{\scriptsize 0.0142}} & \underline{0.7438±{\scriptsize 0.0143}} & 0.8149±{\scriptsize 0.0184} & 0.6802±{\scriptsize 0.0243} & \textbf{0.8575±{\scriptsize 0.0085}} & \textbf{0.7105±{\scriptsize 0.0181}} & \underline{0.8544±{\scriptsize 0.0120}} & 0.0576 (+) \\
 & EvoDiff (Ours) & 30\% & 0.7617±{\scriptsize 0.0125} & 0.7602±{\scriptsize 0.0191} & \textbf{0.7554±{\scriptsize 0.0407}} & 0.7651±{\scriptsize 0.0057} & 0.6292±{\scriptsize 0.0110} & \underline{0.8559±{\scriptsize 0.0202}} & 0.6863±{\scriptsize 0.0228} & \textbf{0.8616±{\scriptsize 0.0129}} & 0.0169 (+) \\
\bottomrule
\end{tabular}
}
\vspace{1pt}
\begin{minipage}{\linewidth}
\footnotesize
\textit{Note:} Higher is better for every metric. Bold indicates the best result and underline the second-best result within each dataset and annotation budget. The $q$-value is the Benjamini--Hochberg-calibrated $p$-value for the two-sided paired test against Random at the same dataset and annotation budget, and (+) or ($-$) indicates the sign of the mean AUC difference relative to Random.
\end{minipage}
\end{sidewaystable*}

\clearpage

%%%%%  S3  跨域外部验证  %%%%%

\begin{table*}[!ht]
\centering
\caption{\textbf{Cross-domain generalization of USFM across external validation datasets.}
USFM was fine-tuned exclusively on source-domain data using training subsets selected by either random sampling or EvoDiff at annotation budgets of 10\%, 20\%, and 30\%, and was subsequently evaluated without target-domain adaptation on seven unseen target-domain datasets, including four public and three private datasets. Performance is reported as balanced accuracy (bAcc). Statistical significance was assessed using two-sided (Random versus EvoDiff) paired $t$-tests. The Generalization Gain column reports each row's largest relative gain across the three budgets and is therefore a post-hoc selected maximum, it is not corrected for this selection. For each annotation budget, the higher result between random sampling and EvoDiff is shown in bold.}
\label{tab:external_private_landscape_metrics}

\resizebox{\textwidth}{!}{

\begin{tabular}{@{}l|ccc|ccc|ccc|c@{}}

\toprule

\multirow{3}{*}{\makecell{\textbf{Source Domain}\\\textbf{$\rightarrow$ Target Domain}}} & 

\multicolumn{9}{c|}{\textbf{Cross-domain Generalization Performance (bAcc)}} & 

\multirow{3}{*}{\makecell{\textbf{Generalization}\\\textbf{Gain}$^{a}$}} \\

\cmidrule{2-10}

& \multicolumn{3}{c|}{\textbf{10\% Annotation}} & \multicolumn{3}{c|}{\textbf{20\% Annotation}} & \multicolumn{3}{c|}{\textbf{30\% Annotation}} & \\

\cmidrule(lr){2-4} \cmidrule(lr){5-7} \cmidrule(lr){8-10}

& \textbf{Random} & \textbf{EvoDiff (Ours)} & \textbf{$p$-value} & \textbf{Random} & \textbf{EvoDiff (Ours)} & \textbf{$p$-value} & \textbf{Random} & \textbf{EvoDiff (Ours)} & \textbf{$p$-value} & \\

\midrule

\multicolumn{11}{l}{\textit{\textbf{Target Domain: ALN-Ultra \cite{ALN-Ultra}} (n=257; 121 benign, 136 malignant)}} \\

BUS-BRA $\rightarrow$ ALN-Ultra & 0.5858 & \textbf{0.6013} & 0.0566 & 0.5925 & \textbf{0.6376} & 0.0156 & 0.6174 & \textbf{0.6385} & 0.2422 & +7.6\% \\

BUSI $\rightarrow$ ALN-Ultra & 0.5940 & \textbf{0.6288} & 0.0433 & 0.5881 & \textbf{0.6394} & 0.0156 & 0.6197 & \textbf{0.6453} & 0.1563 & +8.7\% \\

UDIAT $\rightarrow$ ALN-Ultra & 0.5325 & \textbf{0.5826} & 0.0420 & 0.5595 & \textbf{0.5713} & 0.1797 & 0.5517 & \textbf{0.5982} & 0.1875 & +9.4\% \\

\midrule

\multicolumn{11}{l}{\textit{\textbf{Target Domain: BUS\_UC \cite{busuc}} (n=811; 358 benign, 453 malignant)}} \\

BUS-BRA $\rightarrow$ BUS\_UC & 0.5885 & \textbf{0.5894} & 0.0779 & 0.5895 & \textbf{0.6113} & 0.2500 & 0.6358 & \textbf{0.6373} & 0.2969 & +3.7\% \\

BUSI $\rightarrow$ BUS\_UC & 0.5670 & \textbf{0.5746} & 0.0825 & 0.5729 & \textbf{0.5895} & 0.0625 & 0.5881 & \textbf{0.6017} & 0.4922 & +2.9\% \\

UDIAT $\rightarrow$ BUS\_UC & \textbf{0.5043} & 0.5019 & 0.0906 & 0.5057 & \textbf{0.5700} & 0.0078 & 0.5172 & \textbf{0.5729} & 0.0656 & +12.7\% \\

\midrule

\multicolumn{11}{l}{\textit{\textbf{Target Domain: QAMEBI \cite{QAMEBI}} (n=232; 109 benign, 123 malignant)}} \\

BUS-BRA $\rightarrow$ QAMEBI & 0.6325 & \textbf{0.6465} & 0.0664 & 0.6359 & \textbf{0.7314} & 0.2109 & 0.7410 & \textbf{0.7552} & 0.3750 & +15.0\% \\

BUSI $\rightarrow$ QAMEBI & 0.5917 & \textbf{0.6342} & 0.0820 & 0.6124 & \textbf{0.6357} & 0.3828 & 0.5714 & \textbf{0.5972} & 0.6406 & +7.2\% \\

UDIAT $\rightarrow$ QAMEBI & 0.5746 & \textbf{0.5857} & 0.0500 & 0.5757 & \textbf{0.6728} & 0.0156 & 0.5863 & \textbf{0.6841} & 0.2813 & +16.9\% \\

\midrule

\multicolumn{11}{l}{\textit{\textbf{Target Domain: TN5000 \cite{tn5000}} (n=5,000; 1,426 benign, 3,574 malignant)}} \\

TN3K $\rightarrow$ TN5000 & 0.6192 & \textbf{0.6302} & 0.0583 & 0.6367 & \textbf{0.6396} & 0.0594 & 0.6428 & \textbf{0.6737} & 0.0156 & +4.8\% \\

\midrule

\multicolumn{11}{l}{\textit{\textbf{Target Domain: PrivAppendix} (n=45; 21 normal, 24 appendicitis)}} \\

Appendix~\cite{app} $\rightarrow$ PrivAppendix & 0.4286 & \textbf{0.5357} & 0.0367 & 0.4286 & \textbf{0.4940} & 0.0320 & 0.5357 & \textbf{0.5714} & 0.0156 & +25.0\% \\

\midrule

\multicolumn{11}{l}{\textit{\textbf{Target Domain: PrivBreast} (n=100; 59 benign, 41 malignant)}} \\

BUS-BRA~\cite{BUS-BRA} $\rightarrow$ PrivBreast & 0.5477 & \textbf{0.6153} & 0.0410 & 0.5759 & \textbf{0.6819} & 0.0078 & 0.7344 & \textbf{0.7679} & 0.0781 & +18.4\% \\

BUSI~\cite{BUSI} $\rightarrow$ PrivBreast & 0.5250 & \textbf{0.5589} & 0.0521 & 0.5473 & \textbf{0.5759} & 0.0727 & 0.5738 & \textbf{0.6077} & 0.0625 & +6.5\% \\

UDIAT~\cite{UDIAT} $\rightarrow$ PrivBreast & \textbf{0.6079} & 0.6077 & 0.0213 & 0.5488 & \textbf{0.5589} & 0.2969 & \textbf{0.5951} & 0.5759 & 0.0156 & +1.8\% \\

\midrule

\multicolumn{11}{l}{\textit{\textbf{Target Domain: PrivLiver} (n=69; 24 normal, 45 fatty-liver)}} \\

Fatty-Liver~\cite{liver} $\rightarrow$ PrivLiver & 0.5611 & \textbf{0.6639} & 0.0078 & 0.6514 & \textbf{0.6639} & 0.2188 & 0.6514 & \textbf{0.6611} & 0.0156 & +18.3\% \\

\bottomrule
\multicolumn{11}{l}{\textsuperscript{a}Generalization Gain: relative improvement in balanced accuracy of EvoDiff over random sampling, and reported at the optimal annotation budget for each source-target pair.} \\

\end{tabular}
}
\end{table*}

\clearpage

%%%%%  S4  HAE 估计  %%%%%

\begin{table*}[!ht]
\centering
\caption{Model-based estimates of Human Annotation Effort (HAE, minutes) required by senior and junior radiologists to achieve matched downstream ResNet-50 performance using random sampling versus EvoDiff-selected annotations. Estimates were recomputed using the expanded six-reader panel. ``-/w EvoDiff'' denotes random sampling/EvoDiff selection, respectively.}
\label{supp.reader}
\renewcommand{\arraystretch}{1.2}
\setlength{\tabcolsep}{3pt}

\resizebox{\textwidth}{!}{
\begin{tabular}{c|cc|cc|cc|cc|cc|cc}
\toprule
\multirow{2}{*}{
\textbf{\begin{tabular}[c]{@{}c@{}}
Budget
\end{tabular}}
}
&
\multicolumn{2}{c|}{\textbf{Appendix}}
&
\multicolumn{2}{c|}{\textbf{Breast}}
&
\multicolumn{2}{c|}{\textbf{Carotid}}
&
\multicolumn{2}{c|}{\textbf{Liver}}
&
\multicolumn{2}{c|}{\textbf{Ovary}}
&
\multicolumn{2}{c}{\textbf{Thyroid}}
\\

\cmidrule{2-13}

&
\begin{tabular}[c]{@{}c@{}}
Senior\\
-/w EvoDiff
\end{tabular}
&
\begin{tabular}[c]{@{}c@{}}
Junior\\
-/w EvoDiff
\end{tabular}

&
\begin{tabular}[c]{@{}c@{}}
Senior\\
-/w EvoDiff
\end{tabular}
&
\begin{tabular}[c]{@{}c@{}}
Junior\\
-/w EvoDiff
\end{tabular}

&
\begin{tabular}[c]{@{}c@{}}
Senior\\
-/w EvoDiff
\end{tabular}
&
\begin{tabular}[c]{@{}c@{}}
Junior\\
-/w EvoDiff
\end{tabular}

&
\begin{tabular}[c]{@{}c@{}}
Senior\\
-/w EvoDiff
\end{tabular}
&
\begin{tabular}[c]{@{}c@{}}
Junior\\
-/w EvoDiff
\end{tabular}

&
\begin{tabular}[c]{@{}c@{}}
Senior\\
-/w EvoDiff
\end{tabular}
&
\begin{tabular}[c]{@{}c@{}}
Junior\\
-/w EvoDiff
\end{tabular}

&
\begin{tabular}[c]{@{}c@{}}
Senior\\
-/w EvoDiff
\end{tabular}
&
\begin{tabular}[c]{@{}c@{}}
Junior\\
-/w EvoDiff
\end{tabular}
\\

\midrule

5\%
& 17.1/2.8
& 29.6/4.8
& 19.3/10.3
& 26.8/14.3
& 23.3/5.2
& 28.7/6.5
& 4.1/1.6
& 6.5/2.5
& 31.6/17.1
& 37.3/20.3
& 53.0/17.1
& 61.5/19.8
\\

10\%
& 19.1/5.5
& 32.9/9.6
& 49.0/20.6
& 68.2/28.6
& 35.6/10.5
& 43.9/12.9
& 6.7/3.1
& 10.5/4.9
& 78.8/34.4
& 93.0/40.6
& 65.8/34.2
& 76.3/39.6
\\

20\%
& 20.1/11.1
& 34.7/19.2
& 73.8/41.2
& 102.6/57.3
& 61.0/20.9
& 75.4/25.8
& 14.0/6.2
& 21.9/9.8
& 99.6/68.9
& 117.6/81.2
& 177.2/68.4
& 205.3/79.2
\\

30\%
& 22.0/16.7
& 38.0/28.8
& 86.6/61.8
& 120.4/85.9
& 75.4/31.4
& 93.1/38.7
& 24.1/9.3
& 38.0/14.7
& 138.1/103.3
& 163.1/121.9
& 225.0/102.5
& 260.8/118.8
\\

40\%
& 29.2/22.2
& 50.4/38.4
& 131.5/82.4
& 182.8/114.6
& 84.9/41.8
& 104.8/51.6
& 26.2/12.5
& 41.2/19.6
& 228.4/137.6
& 273.1/164.5
& 262.4/136.6
& 304.1/158.4
\\

50\%
& 52.9/27.8
& 91.3/48.0
& 158.7/103.0
& 220.6/143.2
& 96.0/52.3
& 118.5/64.5
& 30.5/15.5
& 48.0/24.5
& 282.5/172.0
& 337.9/205.7
& 301.2/170.8
& 349.2/198.0
\\

\bottomrule

\end{tabular}
}
\end{table*}

\clearpage

%%%%%  S5  USFM 微调超参数  %%%%%

\begin{table}[!ht]
\centering
\caption{Hyperparameter settings for different datasets on USFM fine-tuning.}
\resizebox{0.7\textwidth}{!}{
\begin{tabular}{l|cccccc}
\toprule
\multirow{2}{*}{\textbf{Dataset}} & \multicolumn{6}{c}{\textbf{Hyperparameter Setting}} \\
\cmidrule{2-7}
& \textbf{Warmup Epoch} & \textbf{Warmup LR} & \textbf{Base LR} & \textbf{Minimum LR} & \textbf{Weight Decay} & \textbf{Layer Decay} \\
\midrule
Appendix~\cite{app}      & 2 & 1.25e-3 & 1.25e-3 & 1.25e-4 & 0.001 & 1.0  \\
BUSI~\cite{BUSI}         & 2 & 2.5e-4  & 1.25e-2 & 2.5e-7  & 0.05  & 0.65 \\
BUS-BRA~\cite{BUS-BRA}   & 2 & 2.5e-4  & 1.25e-2 & 2.5e-7  & 0.05  & 0.65 \\
UDIAT~\cite{UDIAT}       & 2 & 2.5e-4  & 1.25e-2 & 2.5e-7  & 0.05  & 0.65 \\
Fatty-Liver~\cite{liver} & 1 & 1.25e-3 & 1.25e-3 & 1.25e-4 & 0.001 & 1.0  \\
MMOTU~\cite{mmotu}       & 2 & 1.25e-3 & 1e-3    & 1.25e-4 & 0.001 & 1.0  \\
TN3K~\cite{TN3K}         & 2 & 1.25e-3 & 1e-3    & 1.25e-4 & 0.001 & 1.0  \\
\bottomrule
\end{tabular}
}
\label{canshu}
\end{table}

\clearpage

%%%%%  S6  逐读者标注表现  %%%%%

\begin{table*}[!ht]
\centering
\small
\renewcommand{\arraystretch}{1.05}
\setlength{\tabcolsep}{3.2pt}
\caption{Per-reader annotation performance across the six organ tasks. Six board-certified radiologists (R1--R3 senior, R4--R6 junior) annotated the same image sets under the same protocol. Accuracy is the proportion of correctly annotated images. Normalized counts project each reader's observed rate to a 30-minute period. ``Per image'' and ``Per correct'' denote the mean seconds per annotated image and the mean seconds per correctly annotated image, the latter being the total annotation time in seconds divided by the number of correctly annotated images.}
\label{tab:reader_annotation_performance}
\resizebox{\textwidth}{!}{%
\begin{tabular}{lclcccccccc}
\toprule
\multirow{2}{*}{\textbf{Reader}} &
\multirow{2}{*}{\shortstack{\textbf{Experience}\\\textbf{(years)}}} &
\multirow{2}{*}{\textbf{Organ}} &
\multirow{2}{*}{\shortstack{\textbf{Reading time}\\\textbf{(min)}}} &
\multicolumn{2}{c}{\textbf{Observed count ($n$)}} &
\multirow{2}{*}{\textbf{Accuracy}} &
\multicolumn{2}{c}{\textbf{Normalized count ($n$/30 min)}} &
\multicolumn{2}{c}{\textbf{Mean time (s)}} \\
\cmidrule(lr){5-6}\cmidrule(lr){8-9}\cmidrule(lr){10-11}
& & & &
\textbf{Annotated} & \textbf{Correct} & &
\textbf{Annotated} & \textbf{Correct} &
\textbf{Per image} & \textbf{Per correct} \\
\midrule
\multirow{6}{*}{R1 (Senior)} & \multirow{6}{*}{7} & Appendix & 16.8 & 380 & 239 & 0.629 & 680 & 428 & 2.65 & 4.21 \\
  &  & Breast & 21.1 & 400 & 326 & 0.815 & 569 & 464 & 3.16 & 3.88 \\
  &  & Carotid & 12.5 & 400 & 320 & 0.800 & 959 & 767 & 1.88 & 2.35 \\
  &  & Liver & 20.0 & 400 & 369 & 0.923 & 600 & 554 & 3.00 & 3.25 \\
  &  & Ovary & 21.2 & 500 & 147 & 0.294 & 708 & 208 & 2.54 & 8.65 \\
  &  & Thyroid & 22.7 & 500 & 390 & 0.780 & 661 & 516 & 2.72 & 3.49 \\
\midrule
\multirow{6}{*}{R2 (Senior)} & \multirow{6}{*}{13} & Appendix & 30.0 & 235 & 143 & 0.609 & 235 & 143 & 7.66 & 12.59 \\
  &  & Breast & 30.0 & 300 & 241 & 0.803 & 300 & 241 & 6.00 & 7.47 \\
  &  & Carotid & 30.0 & 335 & 261 & 0.779 & 335 & 261 & 5.37 & 6.90 \\
  &  & Liver & 23.0 & 400 & 318 & 0.795 & 522 & 415 & 3.45 & 4.34 \\
  &  & Ovary & 30.0 & 329 & 76 & 0.231 & 329 & 76 & 5.47 & 23.68 \\
  &  & Thyroid & 30.0 & 253 & 199 & 0.787 & 253 & 199 & 7.11 & 9.05 \\
\midrule
\multirow{6}{*}{R3 (Senior)} & \multirow{6}{*}{8} & Appendix & 30.0 & 300 & 189 & 0.630 & 300 & 189 & 6.00 & 9.52 \\
  &  & Breast & 30.0 & 300 & 217 & 0.723 & 300 & 217 & 6.00 & 8.29 \\
  &  & Carotid & 22.5 & 400 & 303 & 0.758 & 533 & 404 & 3.38 & 4.46 \\
  &  & Liver & 30.0 & 400 & 351 & 0.878 & 400 & 351 & 4.50 & 5.13 \\
  &  & Ovary & 30.0 & 323 & 179 & 0.554 & 323 & 179 & 5.57 & 10.06 \\
  &  & Thyroid & 30.0 & 441 & 354 & 0.803 & 441 & 354 & 4.08 & 5.08 \\
\midrule
\multirow{6}{*}{R4 (Junior)} & \multirow{6}{*}{2} & Appendix & 30.0 & 295 & 175 & 0.593 & 295 & 175 & 6.10 & 10.29 \\
  &  & Breast & 30.0 & 362 & 248 & 0.685 & 362 & 248 & 4.97 & 7.26 \\
  &  & Carotid & 13.0 & 400 & 308 & 0.770 & 923 & 711 & 1.95 & 2.53 \\
  &  & Liver & 12.3 & 400 & 300 & 0.750 & 977 & 733 & 1.84 & 2.46 \\
  &  & Ovary & 25.0 & 500 & 171 & 0.342 & 600 & 205 & 3.00 & 8.77 \\
  &  & Thyroid & 25.0 & 500 & 394 & 0.788 & 600 & 473 & 3.00 & 3.81 \\
\midrule
\multirow{6}{*}{R5 (Junior)} & \multirow{6}{*}{3} & Appendix & 25.0 & 380 & 237 & 0.624 & 456 & 284 & 3.95 & 6.33 \\
  &  & Breast & 27.0 & 400 & 307 & 0.768 & 444 & 341 & 4.05 & 5.28 \\
  &  & Carotid & 23.0 & 400 & 308 & 0.770 & 522 & 402 & 3.45 & 4.48 \\
  &  & Liver & 30.0 & 303 & 186 & 0.614 & 303 & 186 & 5.94 & 9.68 \\
  &  & Ovary & 29.0 & 500 & 218 & 0.436 & 517 & 226 & 3.48 & 7.98 \\
  &  & Thyroid & 25.0 & 500 & 392 & 0.784 & 600 & 470 & 3.00 & 3.83 \\
\midrule
\multirow{6}{*}{R6 (Junior)} & \multirow{6}{*}{3} & Appendix & 30.0 & 119 & 64 & 0.538 & 119 & 64 & 15.13 & 28.13 \\
  &  & Breast & 30.0 & 144 & 119 & 0.826 & 144 & 119 & 12.50 & 15.13 \\
  &  & Carotid & 30.0 & 235 & 183 & 0.779 & 235 & 183 & 7.66 & 9.84 \\
  &  & Liver & 30.0 & 266 & 233 & 0.876 & 266 & 233 & 6.77 & 7.73 \\
  &  & Ovary & 30.0 & 142 & 53 & 0.373 & 142 & 53 & 12.68 & 33.96 \\
  &  & Thyroid & 30.0 & 185 & 141 & 0.762 & 185 & 141 & 9.73 & 12.77 \\
\bottomrule
\end{tabular}%
}
\vspace{2pt}
\begin{minipage}{\textwidth}
\footnotesize
\textit{Note.} Readers are identified as R1--R3 (Senior) and R4--R6 (Junior), with clinical experience reported separately. Accuracy is the proportion of correctly annotated images. Normalized counts are the projected numbers of annotated and correctly annotated images within a 30-minute reading period, calculated from each reader's observed annotation rate. ``Per image'' and ``Per correct'' denote the mean seconds per annotated image and per correctly annotated image, respectively.
\end{minipage}
\end{table*}

\clearpage

%%%%%  S7  Diagnostic-model recognition of synthetic probes  %%%%%

\begin{table}[!ht]
\centering
\caption{\textbf{Diagnostic-model recognition of synthetic probes.}
AUC (mean $\pm$ s.d.\ over five folds) from classifiers trained exclusively
on real images, reported across five guidance strengths and for the pooled
synthetic images. s1, s3, s5, s7, and s9 denote the guidance strengths used in the EvoDiff generation process. The real-image reference is the AUC of the same models on real images.}
\label{TR29}
\renewcommand{\arraystretch}{1.2}
\small
\resizebox{\textwidth}{!}{
\begin{tabular}{l c c c c c c c c}
\toprule
\textbf{Organ} & \textbf{Real-image reference} & \textbf{s1} & \textbf{s3} &
\textbf{s5} & \textbf{s7} & \textbf{s9} &
\textbf{Pooled EvoDiff synthetic} & \textbf{Paired $t$-test vs reference} \\
\midrule
Appendix        & 0.624 $\pm$ 0.063 & 0.741 & 0.786 & 0.688 & 0.703
& 0.744 & 0.733 $\pm$ 0.079 & $0.176$ \\
Breast & 0.929 $\pm$ 0.014 & 0.790 & 0.867 & 0.901 & 0.914
& 0.774 & 0.847 $\pm$ 0.018 & $< 0.001$ \\
Carotid         & 0.657 $\pm$ 0.024 & 0.779 & 0.731 & 0.706 & 0.773
& 0.763 & 0.751 $\pm$ 0.045 & $0.018$ \\
Liver     & 0.970 $\pm$ 0.053 & 0.945 & 0.970 & 0.963 & 0.972
& 0.981 & 0.966 $\pm$ 0.023 & $0.803$ \\
Ovary   & 0.932 $\pm$ 0.031 & 0.849 & 0.766 & 0.852 & 0.898
& 0.859 & 0.844 $\pm$ 0.030 & $0.034$ \\
Thyroid & 0.853 $\pm$ 0.012 & 0.821 & 0.762 & 0.781 & 0.726
& 0.847 & 0.789 $\pm$ 0.016 & $0.007$ \\
\bottomrule
\end{tabular}
}
\end{table}

\clearpage

%%%%%  S8  Blinded real-versus-synthetic discrimination  %%%%%

\begin{table*}[!ht]
\centering
\caption{\textbf{Blinded real-versus-synthetic discrimination.}
Four radiologists (two senior and two junior) each assessed 20 real and
20 synthetic images per organ. Synthetic images were evenly sampled from
five guidance strengths. Accuracy denotes the proportion of correct image-origin judgments.}
\label{TR32}

\scriptsize
\renewcommand{\arraystretch}{0.8}
\setlength{\tabcolsep}{1.pt}

% ============================================================
% Appendix, Breast, and Carotid
% ============================================================
\resizebox{\textwidth}{!}{%
\begin{tabular}{ll|cc|cc|c|cc|cc|c|cc|cc|c}
\toprule

\multirow{3}{*}{\textbf{Reader}}
&
\multirow{3}{*}{\textbf{Experience}}
&
\multicolumn{5}{c|}{\textbf{Appendix}}
&
\multicolumn{5}{c|}{\textbf{Breast}}
&
\multicolumn{5}{c}{\textbf{Carotid}}
\\

\cmidrule(lr){3-7}
\cmidrule(lr){8-12}
\cmidrule(lr){13-17}

&
&
\multicolumn{2}{c|}{\textbf{Synthetic Data}}
&
\multicolumn{2}{c|}{\textbf{Real Data}}
&
\multirow{2}{*}{\textbf{Acc}}
&
\multicolumn{2}{c|}{\textbf{Synthetic Data}}
&
\multicolumn{2}{c|}{\textbf{Real Data}}
&
\multirow{2}{*}{\textbf{Acc}}
&
\multicolumn{2}{c|}{\textbf{Synthetic Data}}
&
\multicolumn{2}{c|}{\textbf{Real Data}}
&
\multirow{2}{*}{\textbf{Acc}}
\\

\cmidrule(lr){3-4}
\cmidrule(lr){5-6}
\cmidrule(lr){8-9}
\cmidrule(lr){10-11}
\cmidrule(lr){13-14}
\cmidrule(lr){15-16}

&
&
\shortstack{Synthetic\\(Correct)}
&
\shortstack{Real\\(Wrong)}
&
\shortstack{Real\\(Correct)}
&
\shortstack{Synthetic\\(Wrong)}
&
&
\shortstack{Synthetic\\(Correct)}
&
\shortstack{Real\\(Wrong)}
&
\shortstack{Real\\(Correct)}
&
\shortstack{Synthetic\\(Wrong)}
&
&
\shortstack{Synthetic\\(Correct)}
&
\shortstack{Real\\(Wrong)}
&
\shortstack{Real\\(Correct)}
&
\shortstack{Synthetic\\(Wrong)}
&
\\

\midrule

Senior 1 & 9 years
& 7 & 13 & 12 & 8 & 0.475
& 8 & 12 & 12 & 8 & 0.500
& 15 & 5 & 10 & 10 & 0.625
\\

Senior 2 & 8 years
& 2 & 18 & 10 & 10 & 0.300
& 2 & 18 & 20 & 0 & 0.550
& 4 & 16 & 17 & 3 & 0.525
\\

Junior 1 & 3 years
& 8 & 12 & 9 & 11 & 0.425
& 8 & 12 & 20 & 0 & 0.700
& 7 & 13 & 12 & 8 & 0.475
\\

Junior 2 & 4 years
& 7 & 13 & 11 & 9 & 0.450
& 5 & 15 & 13 & 7 & 0.450
& 4 & 16 & 12 & 8 & 0.400
\\

\midrule

\textbf{Average} & 
& \multicolumn{4}{c|}{} & \textbf{0.413}
& \multicolumn{4}{c|}{} & \textbf{0.550}
& \multicolumn{4}{c|}{} & \textbf{0.506}
\\

\bottomrule
\end{tabular}%
}

\vspace{0.8em}

% ============================================================
% Liver, Ovary, and Thyroid
% ============================================================
\resizebox{\textwidth}{!}{%
\begin{tabular}{ll|cc|cc|c|cc|cc|c|cc|cc|c}
\toprule

\multirow{3}{*}{\textbf{Reader}}
&
\multirow{3}{*}{\textbf{Experience}}
&
\multicolumn{5}{c|}{\textbf{Liver}}
&
\multicolumn{5}{c|}{\textbf{Ovary}}
&
\multicolumn{5}{c}{\textbf{Thyroid}}
\\

\cmidrule(lr){3-7}
\cmidrule(lr){8-12}
\cmidrule(lr){13-17}

&
&
\multicolumn{2}{c|}{\textbf{Synthetic Data}}
&
\multicolumn{2}{c|}{\textbf{Real Data}}
&
\multirow{2}{*}{\textbf{Acc}}
&
\multicolumn{2}{c|}{\textbf{Synthetic Data}}
&
\multicolumn{2}{c|}{\textbf{Real Data}}
&
\multirow{2}{*}{\textbf{Acc}}
&
\multicolumn{2}{c|}{\textbf{Synthetic Data}}
&
\multicolumn{2}{c|}{\textbf{Real Data}}
&
\multirow{2}{*}{\textbf{Acc}}
\\

\cmidrule(lr){3-4}
\cmidrule(lr){5-6}
\cmidrule(lr){8-9}
\cmidrule(lr){10-11}
\cmidrule(lr){13-14}
\cmidrule(lr){15-16}

&
&
\shortstack{Synthetic\\(Correct)}
&
\shortstack{Real\\(Wrong)}
&
\shortstack{Real\\(Correct)}
&
\shortstack{Synthetic\\(Wrong)}
&
&
\shortstack{Synthetic\\(Correct)}
&
\shortstack{Real\\(Wrong)}
&
\shortstack{Real\\(Correct)}
&
\shortstack{Synthetic\\(Wrong)}
&
&
\shortstack{Synthetic\\(Correct)}
&
\shortstack{Real\\(Wrong)}
&
\shortstack{Real\\(Correct)}
&
\shortstack{Synthetic\\(Wrong)}
&
\\

\midrule

Senior 1 & 9 years
& 8 & 12 & 8 & 12 & 0.400
& 5 & 15 & 6 & 14 & 0.275
& 5 & 15 & 15 & 5 & 0.500
\\

Senior 2 & 8 years
& 6 & 14 & 10 & 10 & 0.400
& 0 & 20 & 16 & 4 & 0.400
& 1 & 19 & 19 & 1 & 0.500
\\

Junior 1 & 3 years
& 7 & 13 & 3 & 17 & 0.250
& 7 & 13 & 16 & 4 & 0.575
& 8 & 12 & 12 & 8 & 0.500
\\

Junior 2 & 4 years
& 4 & 16 & 6 & 14 & 0.250
& 1 & 19 & 16 & 4 & 0.425
& 4 & 16 & 13 & 7 & 0.425
\\

\midrule

\textbf{Average} &
& \multicolumn{4}{c|}{} & \textbf{0.325}
& \multicolumn{4}{c|}{} & \textbf{0.419}
& \multicolumn{4}{c|}{} & \textbf{0.481}
\\

\bottomrule
\end{tabular}%
}
\end{table*}

\clearpage

%%%%%  S9  Blinded reader diagnostic scores  %%%%%

\begin{table*}[!ht]
\centering
\caption{\textbf{Blinded reader diagnostic scores by organ and image category.}
Pooled 1--5 scores from four readers are reported for real and synthetic
non-diseased and diseased images, where 1 denotes a confidently non-diseased
appearance and 5 a confidently diseased appearance. For each category, $n$,
mean $\pm$ s.d., median, and the proportion of scores $\geq 4$ are shown;
bold marks the category rated most disease-like within each organ.}
\label{tab:R34}

\small
\renewcommand{\arraystretch}{1.1}
\setlength{\tabcolsep}{6pt}

\resizebox{0.7\textwidth}{!}{%
\begin{tabular}{llccccccccc}

\toprule

\textbf{Organ}
&
\textbf{Category}
&
\textbf{\textit{n}}
&
\textbf{1}
&
\textbf{2}
&
\textbf{3}
&
\textbf{4}
&
\textbf{5}
&
\textbf{Mean $\pm$ s.d.}
&
\textbf{Median}
&
\textbf{Score $\geq$ 4}
\\

\midrule

\multirow{4}{*}{Appendix}
& Real normal
& 40 & 16 & 11 & 5 & 2 & 6
& 2.28 $\pm$ 1.43
& 2
& 20\%
\\

& Synthetic normal
& 40 & 17 & 8 & 6 & 5 & 4
& 2.28 $\pm$ 1.40
& 2
& 22\%
\\

& Real diseased
& 40 & 4 & 1 & 4 & 14 & 17
& \textbf{3.98 $\pm$ 1.25}
& 4
& 78\%
\\

& Synthetic diseased
& 40 & 9 & 8 & 2 & 10 & 11
& 3.15 $\pm$ 1.58
& 4
& 52\%
\\

\midrule

\multirow{4}{*}{Breast}
& Real normal
& 40 & 17 & 13 & 2 & 7 & 1
& 2.05 $\pm$ 1.20
& 2
& 20\%
\\

& Synthetic normal
& 40 & 25 & 5 & 5 & 4 & 1
& 1.78 $\pm$ 1.17
& 1
& 12\%
\\

& Real diseased
& 40 & 7 & 9 & 8 & 7 & 9
& 3.05 $\pm$ 1.43
& 3
& 40\%
\\

& Synthetic diseased
& 40 & 1 & 2 & 10 & 16 & 11
& \textbf{3.85 $\pm$ 0.98}
& 4
& 68\%
\\

\midrule

\multirow{4}{*}{Carotid}
& Real normal
& 40 & 30 & 7 & 2 & 1 & 0
& 1.35 $\pm$ 0.70
& 1
& 2\%
\\

& Synthetic normal
& 40 & 27 & 9 & 4 & 0 & 0
& 1.42 $\pm$ 0.68
& 1
& 0\%
\\

& Real diseased
& 40 & 17 & 5 & 7 & 9 & 2
& 2.35 $\pm$ 1.37
& 2
& 28\%
\\

& Synthetic diseased
& 40 & 10 & 15 & 5 & 8 & 2
& \textbf{2.42 $\pm$ 1.22}
& 2
& 25\%
\\

\midrule

\multirow{4}{*}{Liver}
& Real normal
& 40 & 25 & 11 & 1 & 2 & 1
& 1.58 $\pm$ 0.96
& 1
& 8\%
\\

& Synthetic normal
& 40 & 14 & 13 & 5 & 3 & 5
& 2.30 $\pm$ 1.36
& 2
& 20\%
\\

& Real diseased
& 40 & 0 & 3 & 2 & 5 & 30
& 4.55 $\pm$ 0.90
& 5
& 88\%
\\

& Synthetic diseased
& 40 & 0 & 0 & 2 & 7 & 31
& \textbf{4.72 $\pm$ 0.55}
& 5
& 95\%
\\

\midrule

\multirow{4}{*}{Ovary}
& Real normal
& 40 & 15 & 7 & 14 & 2 & 2
& 2.22 $\pm$ 1.17
& 2
& 10\%
\\

& Synthetic normal
& 40 & 16 & 5 & 9 & 5 & 5
& 2.45 $\pm$ 1.45
& 2
& 25\%
\\

& Real diseased
& 40 & 8 & 7 & 10 & 10 & 5
& \textbf{2.92 $\pm$ 1.33}
& 3
& 38\%
\\

& Synthetic diseased
& 40 & 13 & 7 & 8 & 8 & 4
& 2.58 $\pm$ 1.39
& 2.5
& 30\%
\\

\midrule

\multirow{4}{*}{Thyroid}
& Real normal
& 40 & 22 & 6 & 7 & 4 & 1
& 1.90 $\pm$ 1.17
& 1
& 12\%
\\

& Synthetic normal
& 40 & 21 & 8 & 4 & 7 & 0
& 1.92 $\pm$ 1.16
& 1
& 18\%
\\

& Real diseased
& 40 & 7 & 9 & 5 & 10 & 9
& \textbf{3.12 $\pm$ 1.45}
& 3
& 48\%
\\

& Synthetic diseased
& 40 & 10 & 4 & 8 & 10 & 8
& 3.05 $\pm$ 1.48
& 3
& 45\%
\\

\bottomrule
\end{tabular}%
}

\end{table*}

\clearpage

%%%%%  S10  Acquisition cost per round  %%%%%

\begin{table*}[!ht]
\centering

\caption{Acquisition cost per round, measured on a single NVIDIA A40 for a representative round (BUS-BRA at a 10\% budget). EvoDiff is resolved into its pipeline stages; Stage~1 pretraining is a one-time cost amortized across organs, datasets, and rounds, whereas Stages~2 and~3 recur at each round.}
\label{TR35}

\small
\renewcommand{\arraystretch}{1.15}
\setlength{\tabcolsep}{6pt}

\resizebox{\textwidth}{!}{%
\begin{tabular}{llccc}

\toprule

\textbf{Method}
&
\textbf{Selection model(s)}
&
\textbf{Params (M)}
&
\textbf{Cost per round}
&
\textbf{Peak GPU memory}
\\

\midrule

Random
& none
& 0
& 0
& 0
\\

Max-Entropy
& ResNet-18 proxy
& 11.7
& 19.4 s
& 66 MB
\\

Core-Set
& ResNet-18 proxy
& 11.7
& 19.4 s
& 66 MB
\\

BALD
& ResNet-18 proxy, 25 MC passes
& 11.7
& 86.5 s
& 66 MB
\\

VAAL
& VAE and discriminator
& 23.6
& 104.1 s
& 1,997 MB
\\

EvoDiff Stage~1 (pretraining)
& diffusion UNet and guidance classifier
& 462
& one-time, amortized
& N/A
\\

EvoDiff Stage~2 (guided generation)
& diffusion UNet and guidance classifier
& 462
& 6.8 GPU-hours
& 1,915 MB
\\

EvoDiff Stage~3 (retrieval selection)
& ResNet-50 retrieval encoder
& 25.6
& $\approx$34 s
& 123 MB
\\

\textbf{EvoDiff (all stages)}
& UNet, guidance classifier, retrieval encoder
& 487.6
& 6.8 GPU-hours
& 1,915 MB
\\

\bottomrule

\end{tabular}%
}
\end{table*}

\clearpage

%%%%%%  Supplementary Figure S1  accuracy  %%%%%%
\begin{figure}[p]
\centering
\includegraphics[width=\textwidth]{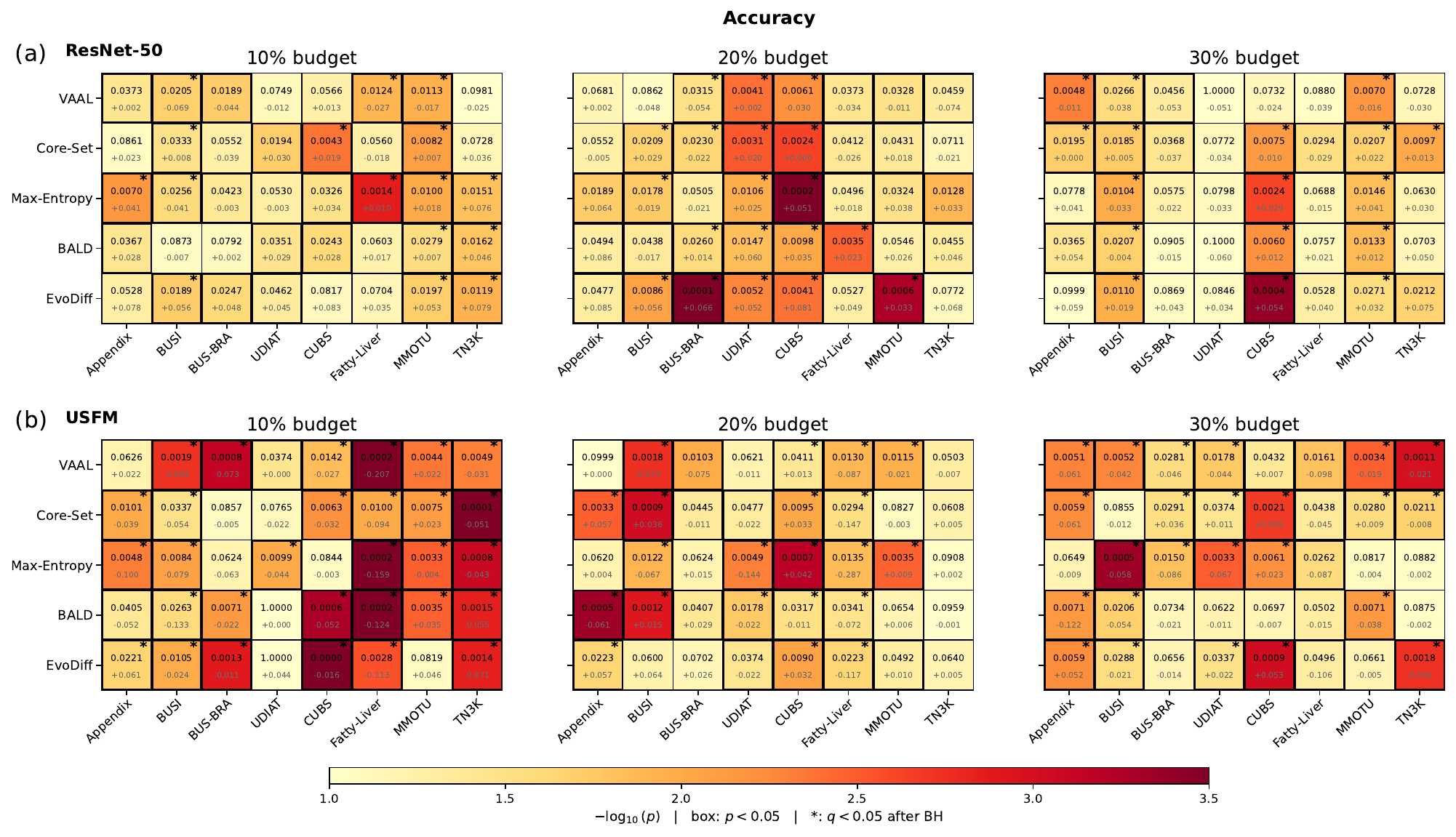}
\caption{\textbf{Statistical comparison of sample-selection strategies against Random for accuracy across eight datasets and three cumulative annotation budgets.} \textbf{a)} Results for the ResNet-50 downstream model. \textbf{b)} Results for the ultrasound foundation model (USFM). In each panel, the three heat maps correspond to cumulative annotation budgets of 10\%, 20\%, and 30\%, rows list the five sample-selection strategies compared against Random (VAAL, Core-Set, Max-Entropy, BALD, and EvoDiff), and columns list the eight internal datasets. In each cell, the upper value is the unadjusted $p$-value from a two-sided paired $t$-test on the five matched cross-validation folds, and the lower value is the mean difference in accuracy relative to Random, with positive values indicating better performance than Random. The cell colour encodes $-\log_{10}(p)$ from light yellow for values near 1 to dark red for values of 3.5 and above, as shown by the colour bar. A black cell border marks an unadjusted $p<0.05$, and an asterisk marks significance after Benjamini--Hochberg correction, with $q<0.05$ within the family of the five strategy-versus-Random comparisons that share a dataset and annotation budget. The testing procedure is identical for both downstream models.}
\label{fig:bh_heatmap_acc}
\end{figure}

\clearpage

%%%%%%  Supplementary Figure S2  balanced accuracy  %%%%%%
\begin{figure}[p]
\centering
\includegraphics[width=\textwidth]{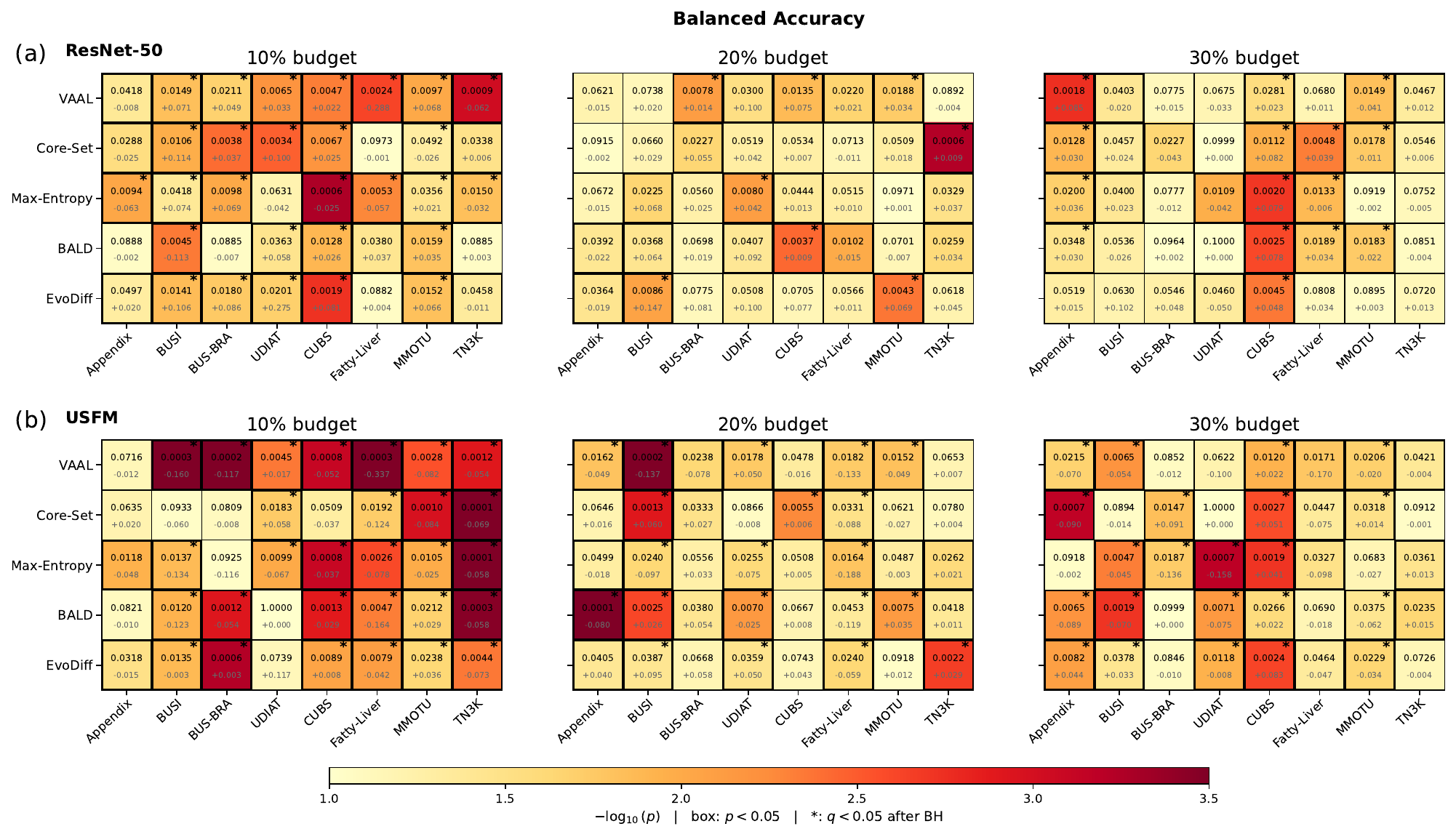}
\caption{\textbf{Statistical comparison of sample-selection strategies against Random for balanced accuracy across eight datasets and three cumulative annotation budgets.} \textbf{a)} Results for the ResNet-50 downstream model. \textbf{b)} Results for the ultrasound foundation model (USFM). In each panel, the three heat maps correspond to cumulative annotation budgets of 10\%, 20\%, and 30\%, rows list the five sample-selection strategies compared against Random (VAAL, Core-Set, Max-Entropy, BALD, and EvoDiff), and columns list the eight internal datasets. In each cell, the upper value is the unadjusted $p$-value from a two-sided paired $t$-test on the five matched cross-validation folds, and the lower value is the mean difference in balanced accuracy (bAcc) relative to Random, with positive values indicating better performance than Random. The cell colour encodes $-\log_{10}(p)$ from light yellow for values near 1 to dark red for values of 3.5 and above, as shown by the colour bar. A black cell border marks an unadjusted $p<0.05$, and an asterisk marks significance after Benjamini--Hochberg correction, with $q<0.05$ within the family of the five strategy-versus-Random comparisons that share a dataset and annotation budget. The testing procedure is identical for both downstream models.}
\label{fig:bh_heatmap_bacc}
\end{figure}

\clearpage

%%%%%%  Supplementary Figure S3  sensitivity  %%%%%%
\begin{figure}[p]
\centering
\includegraphics[width=\textwidth]{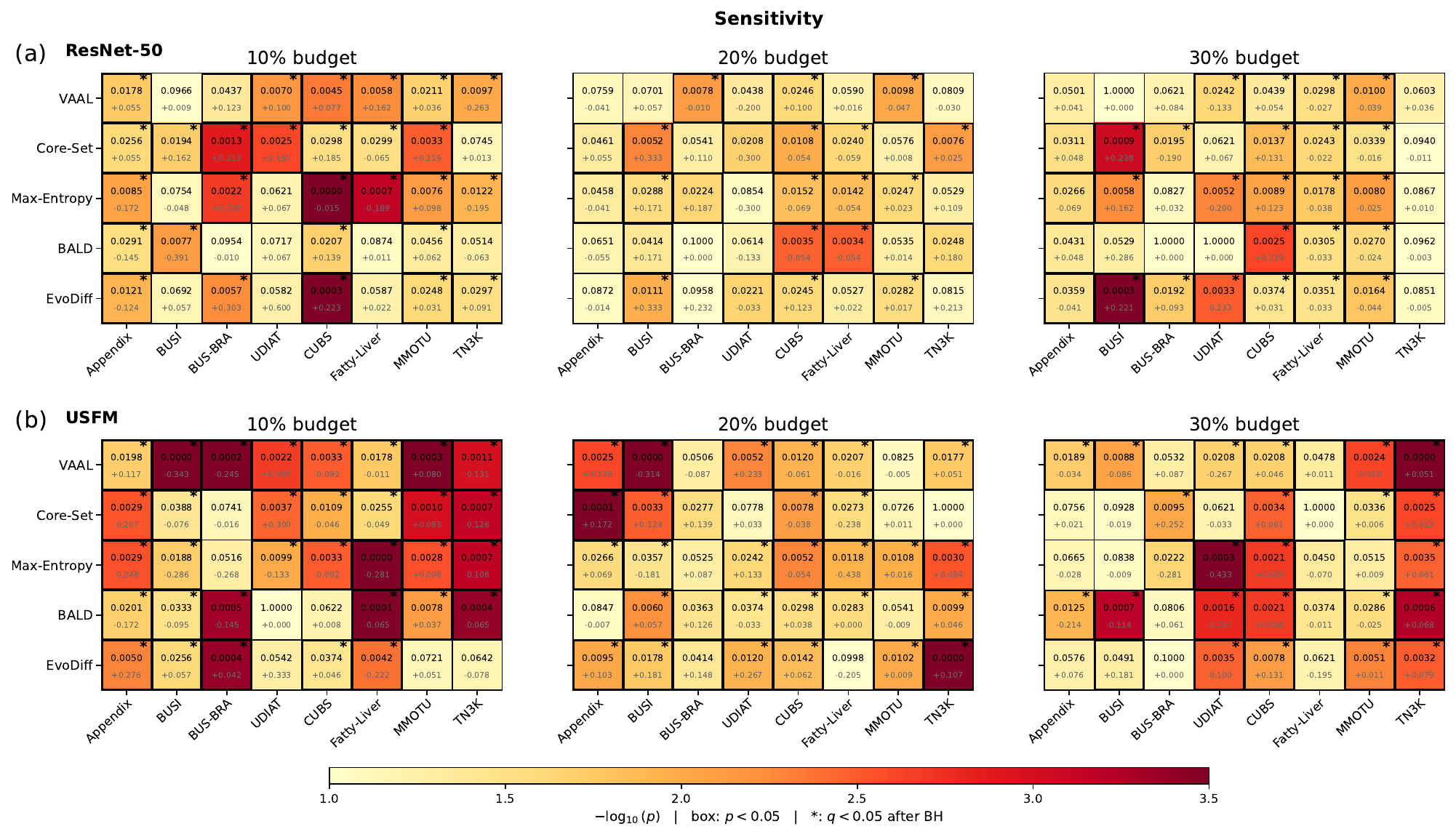}
\caption{\textbf{Statistical comparison of sample-selection strategies against Random for sensitivity across eight datasets and three cumulative annotation budgets.} \textbf{a)} Results for the ResNet-50 downstream model. \textbf{b)} Results for the ultrasound foundation model (USFM). In each panel, the three heat maps correspond to cumulative annotation budgets of 10\%, 20\%, and 30\%, rows list the five sample-selection strategies compared against Random (VAAL, Core-Set, Max-Entropy, BALD, and EvoDiff), and columns list the eight internal datasets. In each cell, the upper value is the unadjusted $p$-value from a two-sided paired $t$-test on the five matched cross-validation folds, and the lower value is the mean difference in sensitivity relative to Random, with positive values indicating better performance than Random. The cell colour encodes $-\log_{10}(p)$ from light yellow for values near 1 to dark red for values of 3.5 and above, as shown by the colour bar. A black cell border marks an unadjusted $p<0.05$, and an asterisk marks significance after Benjamini--Hochberg correction, with $q<0.05$ within the family of the five strategy-versus-Random comparisons that share a dataset and annotation budget. The testing procedure is identical for both downstream models.}
\label{fig:bh_heatmap_sens}
\end{figure}

\clearpage

%%%%%%  Supplementary Figure S4  specificity  %%%%%%
\begin{figure}[p]
\centering
\includegraphics[width=\textwidth]{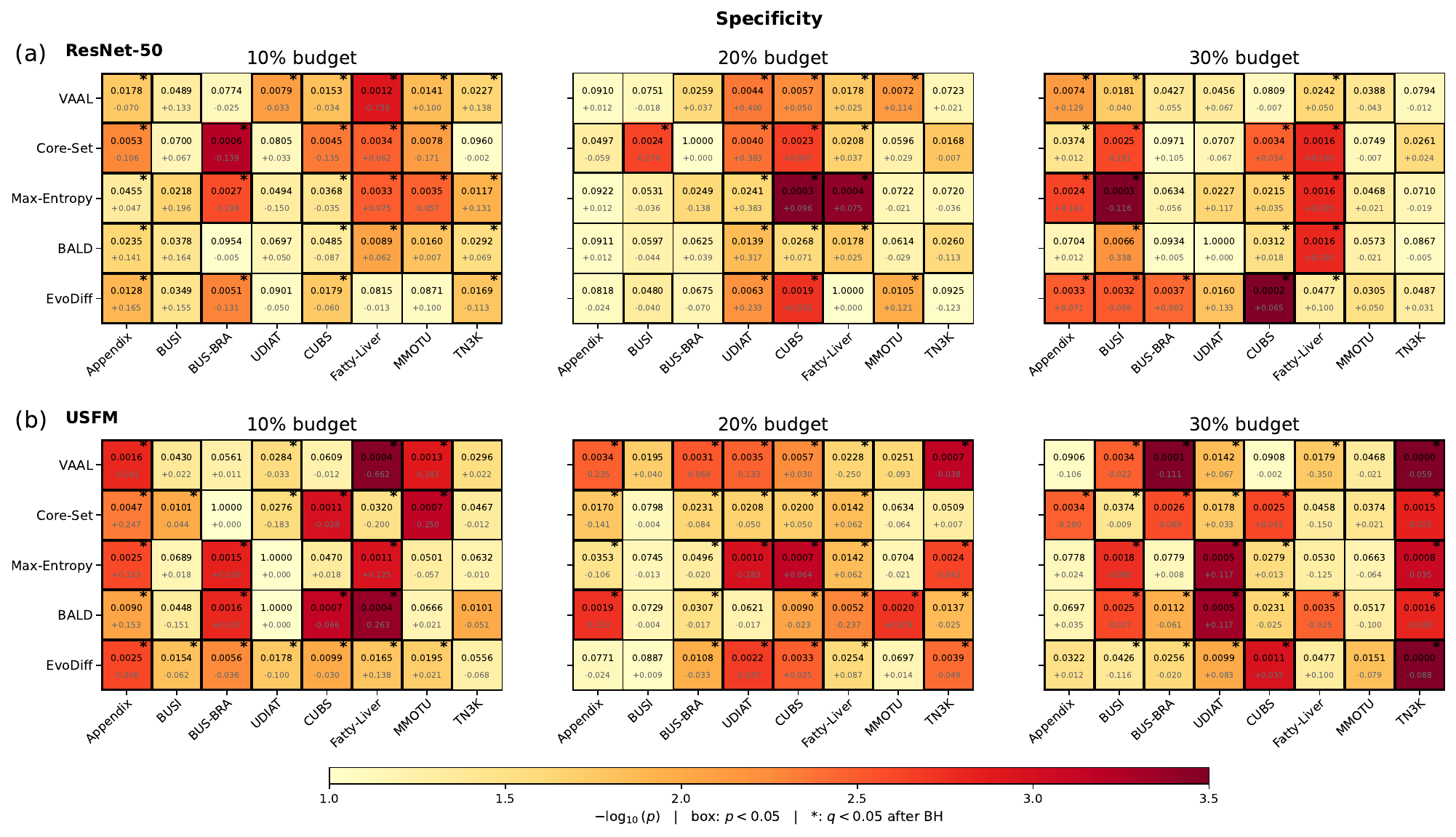}
\caption{\textbf{Statistical comparison of sample-selection strategies against Random for specificity across eight datasets and three cumulative annotation budgets.} \textbf{a)} Results for the ResNet-50 downstream model. \textbf{b)} Results for the ultrasound foundation model (USFM). In each panel, the three heat maps correspond to cumulative annotation budgets of 10\%, 20\%, and 30\%, rows list the five sample-selection strategies compared against Random (VAAL, Core-Set, Max-Entropy, BALD, and EvoDiff), and columns list the eight internal datasets. In each cell, the upper value is the unadjusted $p$-value from a two-sided paired $t$-test on the five matched cross-validation folds, and the lower value is the mean difference in specificity relative to Random, with positive values indicating better performance than Random. The cell colour encodes $-\log_{10}(p)$ from light yellow for values near 1 to dark red for values of 3.5 and above, as shown by the colour bar. A black cell border marks an unadjusted $p<0.05$, and an asterisk marks significance after Benjamini--Hochberg correction, with $q<0.05$ within the family of the five strategy-versus-Random comparisons that share a dataset and annotation budget. The testing procedure is identical for both downstream models.}
\label{fig:bh_heatmap_spec}
\end{figure}

\clearpage

%%%%%%  Supplementary Figure S5  positive predictive value (PPV)  %%%%%%
\begin{figure}[p]
\centering
\includegraphics[width=\textwidth]{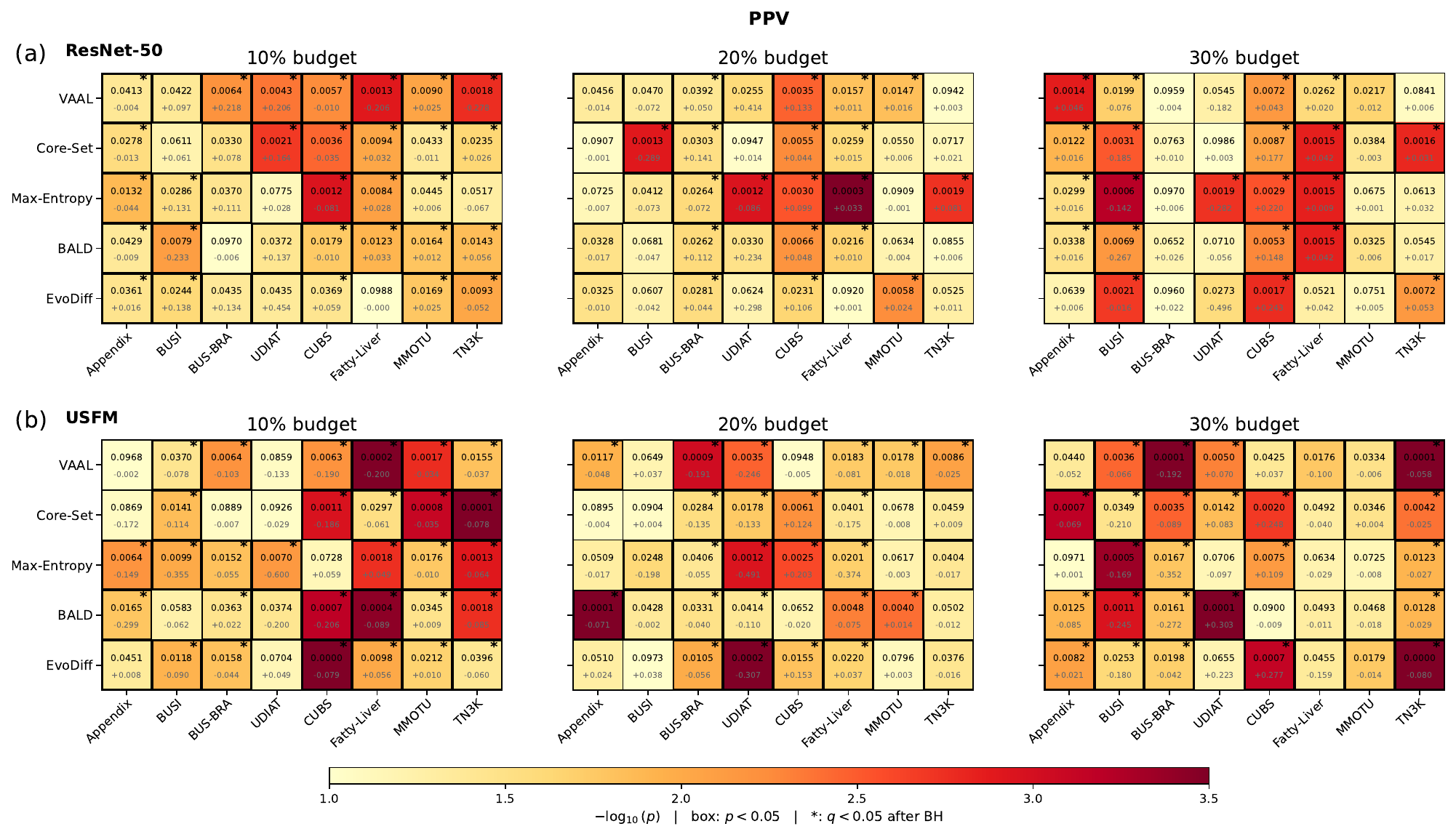}
\caption{\textbf{Statistical comparison of sample-selection strategies against Random for positive predictive value (PPV) across eight datasets and three cumulative annotation budgets.} \textbf{a)} Results for the ResNet-50 downstream model. \textbf{b)} Results for the ultrasound foundation model (USFM). In each panel, the three heat maps correspond to cumulative annotation budgets of 10\%, 20\%, and 30\%, rows list the five sample-selection strategies compared against Random (VAAL, Core-Set, Max-Entropy, BALD, and EvoDiff), and columns list the eight internal datasets. In each cell, the upper value is the unadjusted $p$-value from a two-sided paired $t$-test on the five matched cross-validation folds, and the lower value is the mean difference in PPV relative to Random, with positive values indicating better performance than Random. The cell colour encodes $-\log_{10}(p)$ from light yellow for values near 1 to dark red for values of 3.5 and above, as shown by the colour bar. A black cell border marks an unadjusted $p<0.05$, and an asterisk marks significance after Benjamini--Hochberg correction, with $q<0.05$ within the family of the five strategy-versus-Random comparisons that share a dataset and annotation budget. The testing procedure is identical for both downstream models.}
\label{fig:bh_heatmap_ppv}
\end{figure}

\clearpage

%%%%%%  Supplementary Figure S6  negative predictive value (NPV)  %%%%%%
\begin{figure}[p]
\centering
\includegraphics[width=\textwidth]{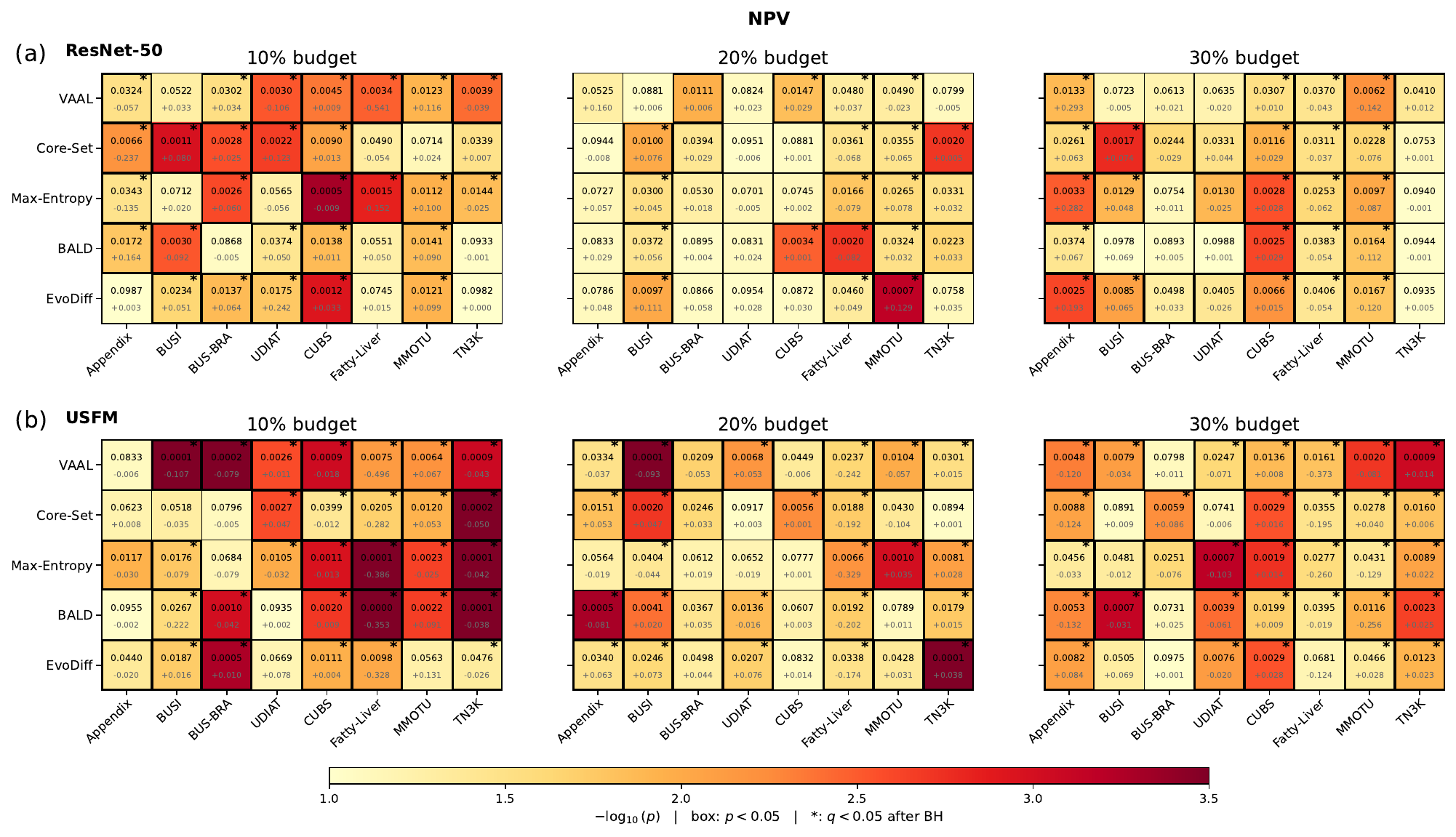}
\caption{\textbf{Statistical comparison of sample-selection strategies against Random for negative predictive value (NPV) across eight datasets and three cumulative annotation budgets.} \textbf{a)} Results for the ResNet-50 downstream model. \textbf{b)} Results for the ultrasound foundation model (USFM). In each panel, the three heat maps correspond to cumulative annotation budgets of 10\%, 20\%, and 30\%, rows list the five sample-selection strategies compared against Random (VAAL, Core-Set, Max-Entropy, BALD, and EvoDiff), and columns list the eight internal datasets. In each cell, the upper value is the unadjusted $p$-value from a two-sided paired $t$-test on the five matched cross-validation folds, and the lower value is the mean difference in NPV relative to Random, with positive values indicating better performance than Random. The cell colour encodes $-\log_{10}(p)$ from light yellow for values near 1 to dark red for values of 3.5 and above, as shown by the colour bar. A black cell border marks an unadjusted $p<0.05$, and an asterisk marks significance after Benjamini--Hochberg correction, with $q<0.05$ within the family of the five strategy-versus-Random comparisons that share a dataset and annotation budget. The testing procedure is identical for both downstream models.}
\label{fig:bh_heatmap_npv}
\end{figure}

\clearpage

%%%%%%  Supplementary Figure S7  F1-score  %%%%%%
\begin{figure}[p]
\centering
\includegraphics[width=\textwidth]{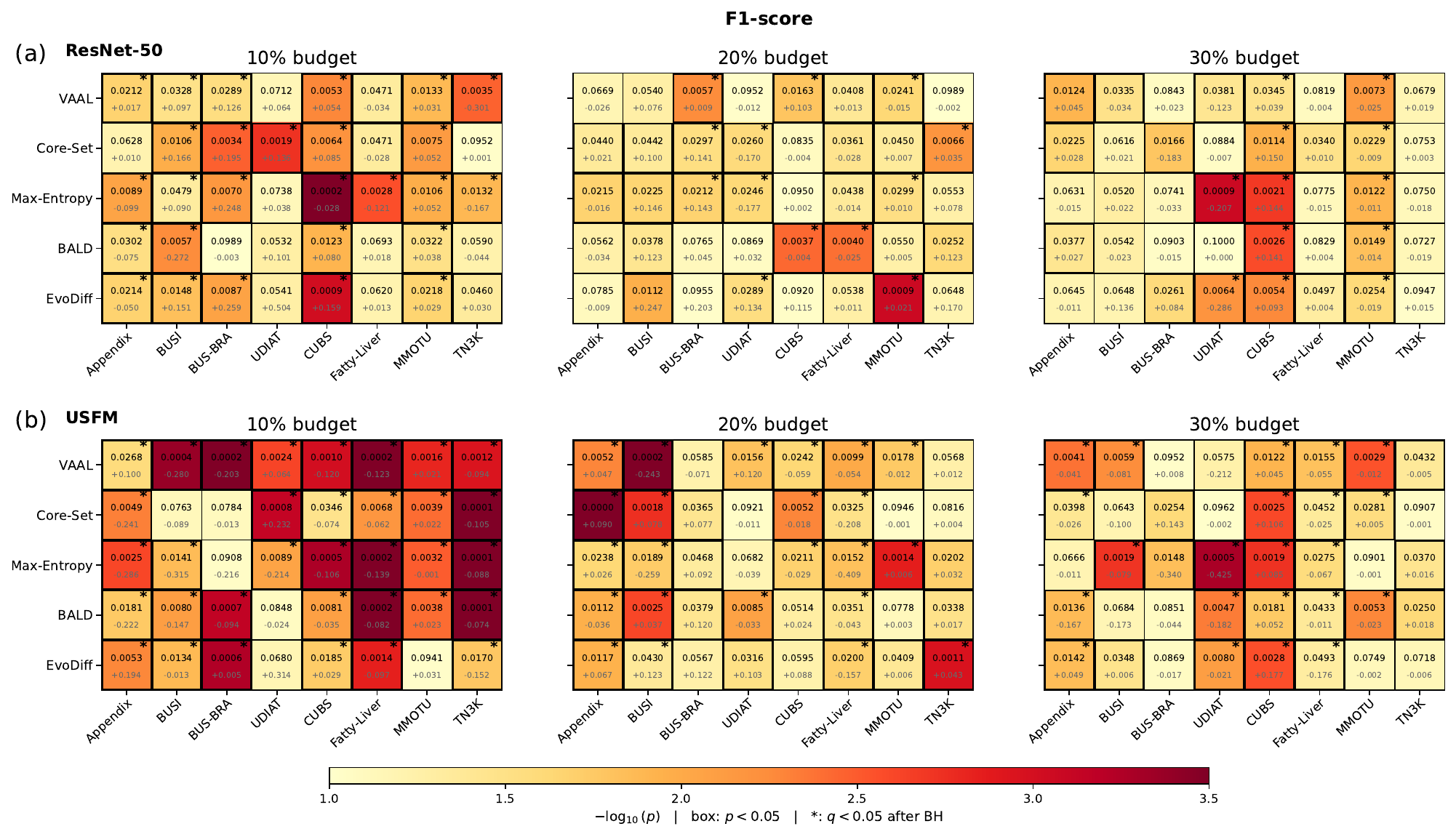}
\caption{\textbf{Statistical comparison of sample-selection strategies against Random for F1-score across eight datasets and three cumulative annotation budgets.} \textbf{a)} Results for the ResNet-50 downstream model. \textbf{b)} Results for the ultrasound foundation model (USFM). In each panel, the three heat maps correspond to cumulative annotation budgets of 10\%, 20\%, and 30\%, rows list the five sample-selection strategies compared against Random (VAAL, Core-Set, Max-Entropy, BALD, and EvoDiff), and columns list the eight internal datasets. In each cell, the upper value is the unadjusted $p$-value from a two-sided paired $t$-test on the five matched cross-validation folds, and the lower value is the mean difference in F1-score relative to Random, with positive values indicating better performance than Random. The cell colour encodes $-\log_{10}(p)$ from light yellow for values near 1 to dark red for values of 3.5 and above, as shown by the colour bar. A black cell border marks an unadjusted $p<0.05$, and an asterisk marks significance after Benjamini--Hochberg correction, with $q<0.05$ within the family of the five strategy-versus-Random comparisons that share a dataset and annotation budget. The testing procedure is identical for both downstream models.}
\label{fig:bh_heatmap_f1}
\end{figure}

\clearpage

%%%%%%  Supplementary Figure S8  area under the receiver operating characteristic curve (AUC)  %%%%%%
\begin{figure}[p]
\centering
\includegraphics[width=\textwidth]{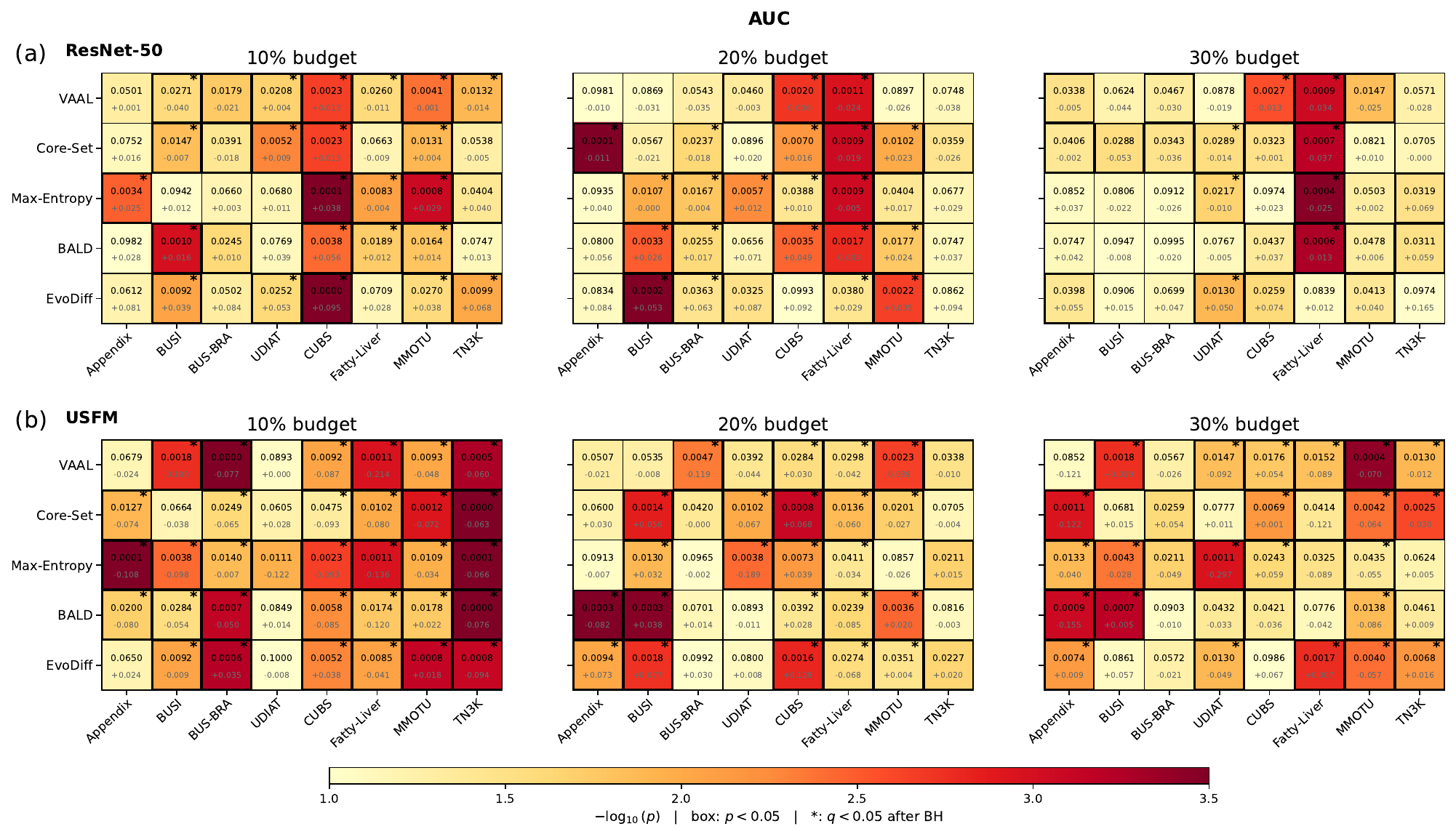}
\caption{\textbf{Statistical comparison of sample-selection strategies against Random for area under the receiver operating characteristic curve (AUC) across eight datasets and three cumulative annotation budgets.} \textbf{a)} Results for the ResNet-50 downstream model. \textbf{b)} Results for the ultrasound foundation model (USFM). In each panel, the three heat maps correspond to cumulative annotation budgets of 10\%, 20\%, and 30\%, rows list the five sample-selection strategies compared against Random (VAAL, Core-Set, Max-Entropy, BALD, and EvoDiff), and columns list the eight internal datasets. In each cell, the upper value is the unadjusted $p$-value from a two-sided paired $t$-test on the five matched cross-validation folds, and the lower value is the mean difference in AUC relative to Random, with positive values indicating better performance than Random. The cell colour encodes $-\log_{10}(p)$ from light yellow for values near 1 to dark red for values of 3.5 and above, as shown by the colour bar. A black cell border marks an unadjusted $p<0.05$, and an asterisk marks significance after Benjamini--Hochberg correction, with $q<0.05$ within the family of the five strategy-versus-Random comparisons that share a dataset and annotation budget. The testing procedure is identical for both downstream models.}
\label{fig:bh_heatmap_auc}
\end{figure}

\clearpage

%%%%%%  Supplementary Figure S9  reader scores for non-diseased images  %%%%%%
\begin{figure*}[!ht]
    \centering
    \includegraphics[width=0.74\linewidth]{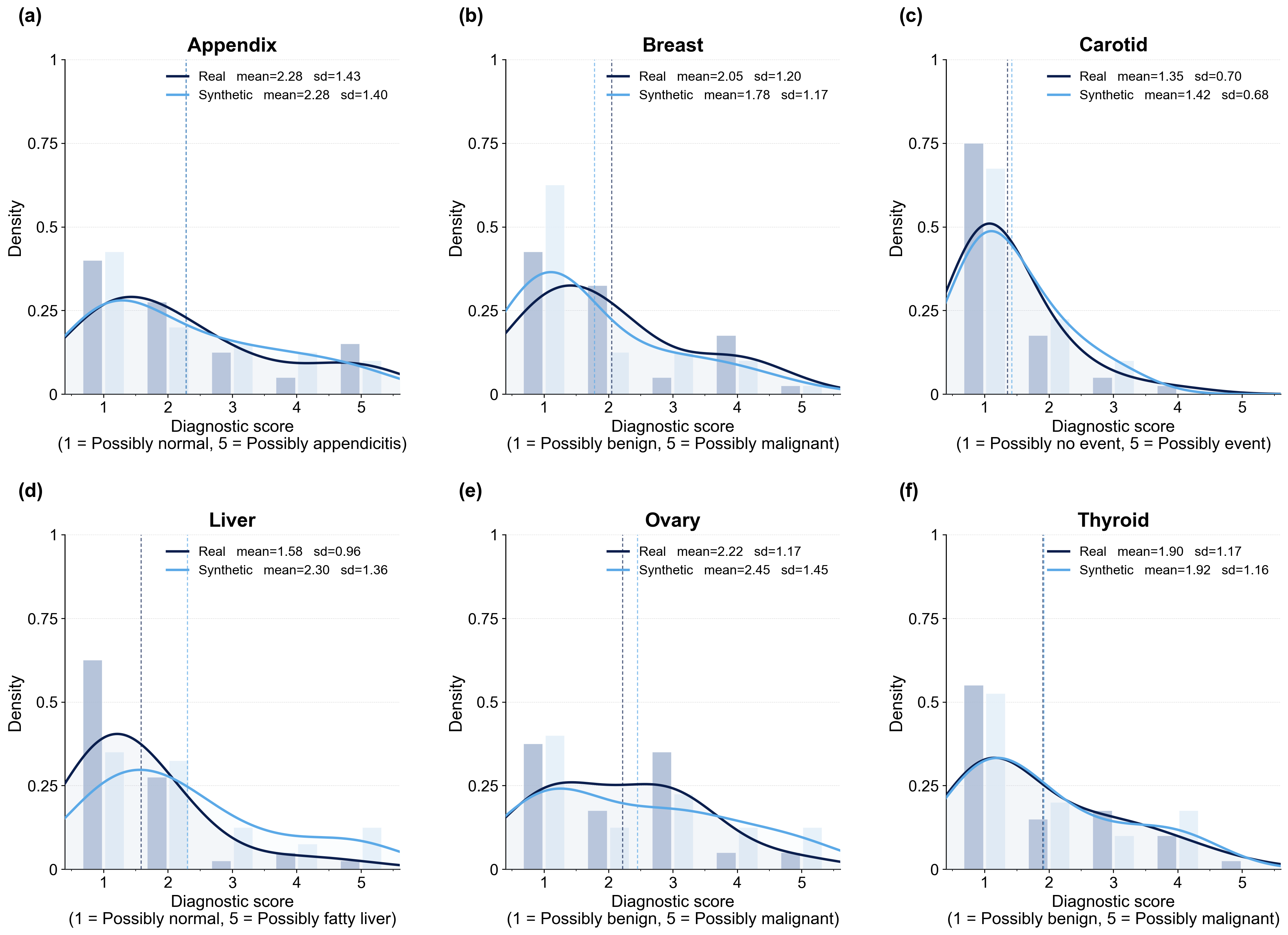}
    \caption{\textbf{Reader scores for non-diseased images.}
    Four blinded readers rated 10 real and 10 synthetic images per organ on a
    1--5 scale (1, confidently non-diseased; 5, confidently diseased).
    Panels show the appendix, breast, carotid, liver, ovary, and thyroid tasks,
    respectively. Bars show empirical probabilities, solid curves Gaussian
    kernel-density estimates, and dashed lines group means; legends report
    mean $\pm$ s.d.}
    \label{nor}
\end{figure*}

\clearpage

%%%%%%  Supplementary Figure S10  reader scores for diseased images  %%%%%%
\begin{figure*}[!ht]
    \centering
    \includegraphics[width=0.74\linewidth]{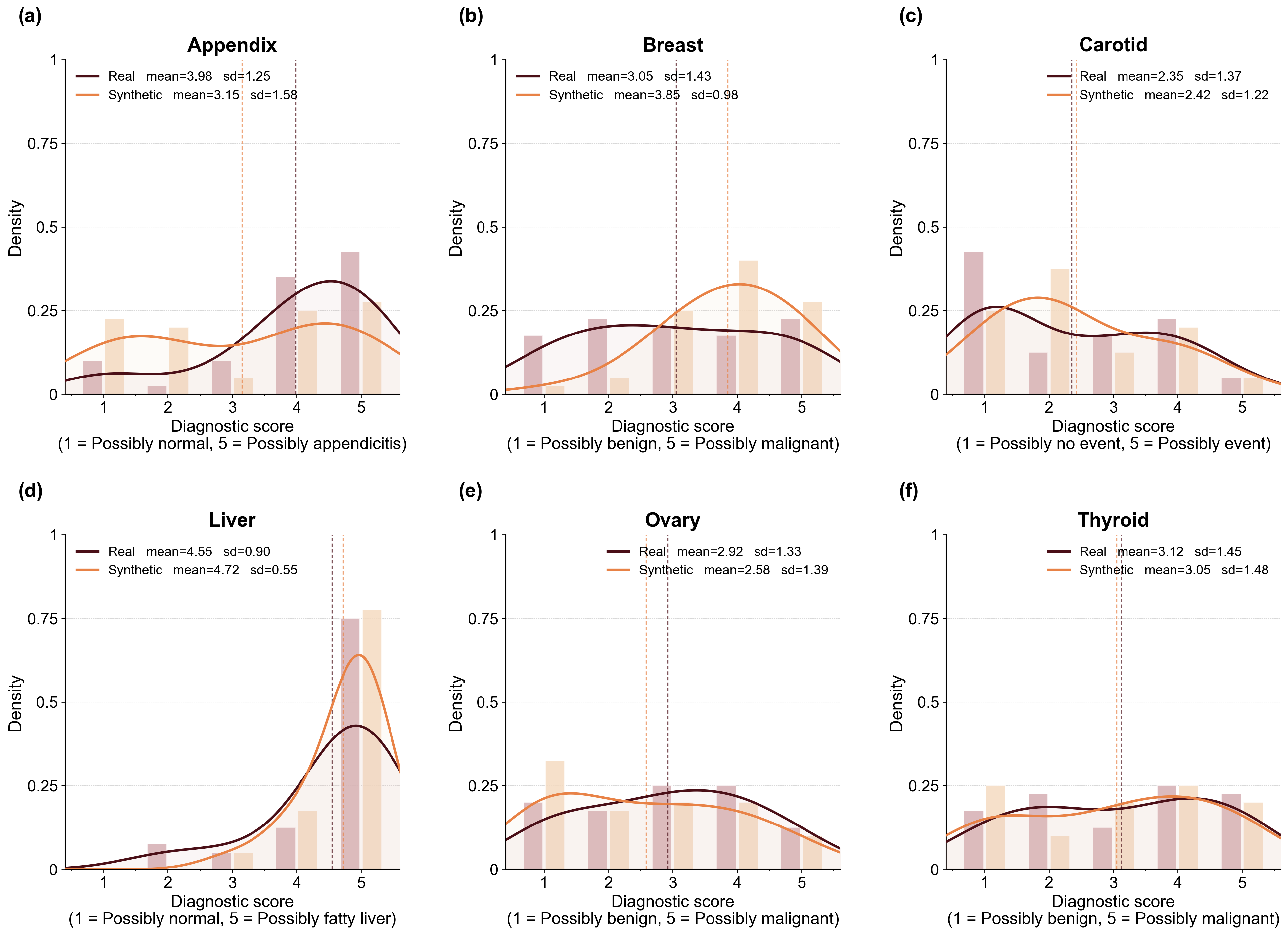}
    \caption{\textbf{Reader scores for diseased images.}
    The reader study used the same settings as Supplementary Figure~\ref{nor}.}
    \label{dis}
\end{figure*}

\clearpage

\end{document}